\documentclass[showinstructions,faculty=firw,department=elt,phddegree=elt,final]{adsphd}
\title{Bridging Modalities and Transferring Knowledge: Enhanced Multimodal Understanding and Recognition} % TODO
\author{Gorjan}{Radevski}

\supervisor{Prof.~dr.~ir.~Tinne~Tuytelaars}{}
\supervisor{Prof.~dr.~Marie-Francine~Moens}{}
\president{Prof.~dr.~ir.~Yves Willems}{}
\jurymember{Prof.~dr.~ir.~Hugo~Van~hamme}{}
\jurymember{Prof.~dr.~ir.~Hendrik~Blockeel}{}
\externaljurymember{Prof.~dr.~Dima~Damen}{University of Bristol}
\externaljurymember{Prof.~dr.~Goran~Glavaš}{University of Würzburg}

\researchgroup{Processing Speech and Images (PSI)}
\website{https://www.esat.kuleuven.be/} % Leave empty to hide
\email{gorjan.radevski@esat.kuleuven.be} % Leave empty to hide

\address{Kasteelpark Arenberg 10 box 2441, B-3001 Leuven}
\date{\today}
\copyyear{2024}
\usepackage{nomencl}   % For nomenclature
\renewcommand{\nomname}{List of Symbols}

\makenomenclature%
\usepackage{glossaries} % For list of abbreviations
\newcommand{\glossname}{List of Abbreviations}
\newcommand{\myprintglossary}{%
  \renewcommand{\glossaryname}{\glossname}
  \cleardoublepage%
  \printglossary[title=\glossname]
  \chaptermark{\glossname}
  \addcontentsline{toc}{chapter}{\glossname} %% comment to exclude from TOC
}
\newcommand{\etal}{\textit{et al.}} % and others...
\newcommand{\NA}{---} % Not applicable
\newcommand{\error}[1]{\small{$\pm$ #1}}
\newcommand*\circled[1]{\tikz[baseline=(char.base)]{
            \node[shape=circle,draw,inner sep=.6pt] (char) {#1};}}
\newcommand{\tikzxmark}{%
\tikz[scale=0.23] {
    \draw[line width=0.7,line cap=round] (0,0) to [bend left=6] (1,1);
    \draw[line width=0.7,line cap=round] (0.2,0.95) to [bend right=3] (0.8,0.05);
}}
\newcommand{\tikzcmark}{%
\tikz[scale=0.23] {
    \draw[line width=0.7,line cap=round] (0.25,0) to [bend left=10] (1,1);
    \draw[line width=0.8,line cap=round] (0,0.35) to [bend right=1] (0.23,0);
}}

\newcommand{\preranker}{
OIE$^{\text{pre}}_{\text{ranker}}$
}
\newcommand{\reranker}{
Fact$^{\text{re}}_{\text{ranker}}$
}

\newcommand{\researchquestion}[2]{
    \begin{tcolorbox}[
        title=#1,
        fonttitle=\bfseries,
        colframe=KULeuvenBlue, % Frame color
        colback=LightBlue,     % Background color
        boxrule=1pt,           % Border thickness
        titlerule=0pt,         % No additional line below the title
        toprule=2pt,           % Thicker top border
        bottomrule=2pt,         % Thicker bottom border
        left=2mm,             % Reduced left padding
        right=2mm            % Reduced right padding
    ]
    #2
    \end{tcolorbox}
}

\newcommand{\subresearchquestion}[2]{
    \begin{tcolorbox}[
        title=#1,
        fonttitle=\bfseries\small,
        colframe=SubtleKULeuvenBlue, % More subtle frame color
        colback=SubtleLightBlue,   % More opaque background color
        coltext=black,        % Text color
        boxrule=0.5pt,        % Thinner border thickness
        titlerule=0pt,        % No additional line below the title
        toprule=1pt,          % Thinner top border
        bottomrule=1pt,       % Thinner bottom border
        left=2mm,             % Reduced left padding
        right=2mm,            % Reduced right padding
        top=1mm,              % Reduced top padding
        bottom=1mm            % Reduced bottom padding
    ]
    \small #2 % Smaller text size
    \end{tcolorbox}
}

\newcommand{\universalresearchquestion}[2]{
    \begin{tcolorbox}[
        title=#1,
        fonttitle=\bfseries\small,
        colframe=UniversalRq, % More subtle frame color
        colback=SubtleUniversalRq,   % More opaque background color
        coltext=black,        % Text color
        boxrule=0.5pt,        % Thinner border thickness
        titlerule=0pt,        % No additional line below the title
        toprule=1pt,          % Thinner top border
        bottomrule=1pt,       % Thinner bottom border
        left=2mm,             % Reduced left padding
        right=2mm,            % Reduced right padding
        top=1mm,              % Reduced top padding
        bottom=1mm            % Reduced bottom padding
    ]
    \small #2 % Smaller text size
    \end{tcolorbox}
}

\newcommand{\inlineresearchquestion}[1]{%
\tikz[baseline=(X.base)] 
    \node (X) [draw=KULeuvenBlue, thick, shape=rectangle, rounded corners, inner sep=2pt, fill=LightBlue] {\textbf{#1}};%
}

\newcommand{\inlinesubresearchquestion}[1]{%
\tikz[baseline=(X.base)] 
    \node (X) [draw=SubtleKULeuvenBlue, thick, shape=rectangle, rounded corners, inner sep=2pt, fill=SubtleLightBlue] {\textbf{#1}};%
}

\definecolor{shadecolor}{RGB}{255,203,203}
\definecolor{shadecolor}{RGB}{255,203,203}
\definecolor{lightgreen}{RGB}{39,179,118}
\definecolor{student}{RGB}{76,115,176}
\definecolor{baseline}{RGB}{221,133,83}
\definecolor{teacher}{RGB}{85,168,105}
\definecolor{redred}{RGB}{196,23,63}
\definecolor{pinkpink}{RGB}{243,157,153}
\definecolor{greengreen}{RGB}{200,228,206}
\definecolor{blueblue}{RGB}{168,218,224}
\definecolor{orangeorange}{RGB}{231,96,83}
\definecolor{graygray}{RGB}{229,229,229}
\definecolor{yellowyellow}{RGB}{255,229,153}
\definecolor{rowrow}{RGB}{164, 188, 219}
\definecolor{row}{RGB}{236,244,233}
\definecolor{rowrow}{RGB}{208,219,203}
\definecolor{transductive}{HTML}{F57251}
\definecolor{inductive}{HTML}{67AB9F}
\definecolor{polysemous}{HTML}{A4BCDB}
\definecolor{outofkg}{HTML}{C4AD9D}
\makeglossaries%

\usepackage{fancyhdr} % Signature in acknowledgements

\usepackage{textcomp} % nice greek alphabet
\usepackage{pifont}   % Dingbats
\usepackage{booktabs}
\usepackage{amssymb,amsthm}
\usepackage{amsmath}
\usepackage{paralist} % Inline lists
\usepackage{caption} 
\usepackage{subcaption}
\usepackage{xcolor,colortbl}
\usepackage{mdframed}
\usepackage{xspace}
\usepackage[export]{adjustbox}% http://ctan.org/pkg/adjustbox
\usepackage{multirow}
\usepackage[most]{tcolorbox}
\usepackage{tikz}
\usepackage{epigraph} % Quotes for preface
\usepackage{enumitem} % 1) 2) 3) .. lists
\usepackage{graphicx} % Emoji
\usepackage[utf8]{inputenc} % Macedonian
\usepackage{hyperref}
\hypersetup{
    colorlinks,
    linkcolor={red!50!black},
    citecolor={blue!50!black},
    urlcolor={blue!80!black}
}
\definecolor{KULeuvenBlue}{RGB}{29,141,176} % Adjust the RGB values to match the exact blue used by KU Leuven
\definecolor{LightBlue}{RGB}{214,234,248}   % Lighter blue for the box background
\colorlet{SubtleKULeuvenBlue}{KULeuvenBlue!50} % Mix the blue with 50% white
\colorlet{SubtleLightBlue}{LightBlue!50} % Mix the blue with 50% white
\definecolor{UniversalRq}{RGB}{210, 92, 111} % Adjust the RGB values to match the exact blue used by KU Leuven
\colorlet{SubtleUniversalRq}{UniversalRq!10}   % Lighter blue for the box background

\begin{document}

%%%%%%%%%%%%%%%%%%%%%%%%%%%%%%%%%%%%%%%%%%%%%%%%%%%%%%%%%%%%%%%%%%%%%%

\makefrontcoverXII

\maketitle

\frontmatter % to get \pagenumbering{roman}

\includepreface{preface}
\includeabstract{abstract}
\includeabstractnl{abstractnl}

% To create a list of abbreviations, there are 2 options
% 1. manual creation of list of abbreviations and inclusion as a chapter
%    \includeabbreviations{abbreviations}
% 2. automatic generation via the glossary package
%    \glossary{name=MD,description=molecular dynamics}

% --------------- ABBREVIATIONS START ---------------------
\newcommand{\blackgls}[1]{%
  \glsadd{#1}% Mark the term as used
  \textcolor{black}{\glsentrytext{#1}}% Display the term's text in black
}
\newcommand{\insertemoji}[1]{\includegraphics[width=1em]{#1}}

% Macedonian
% \newcommand{\mktext}[1]{\textnormal{\fontfamily{cmr}\selectfont #1}}

% Background
\newglossaryentry{ood}{name={OOD},description={Out-of-Distribution}}
\newglossaryentry{id}{name={ID},description={In-Distribution}}
\newglossaryentry{nlp}{name={NLP},description={Natural Language Processing}}
\newglossaryentry{mlm}{name={MLM},description={Masked Language Modelling}}
\newglossaryentry{cnn}{name={CNN},description={Convolutional Neural Network}}
\newglossaryentry{rnn}{name={RNN},description={Recurrent Neural Network}}
\newglossaryentry{3d-cnn}{name={3D-CNN},description={3D Convolutional Neural Network}}
\newglossaryentry{videomae}{name={VideoMAE},description={Video Masked Autoencoders}}

% EMNLP 2020
\newglossaryentry{bert}{name={\textsc{Bert}},description={Bidirectional Encoder Representations from Transformers}}
\newglossaryentry{sr-bert}{name={\textsc{Sr-Bert}},description={Spatial-Reasoning Bert Model}}
\newglossaryentry{mpm}{name={MPM},description={Masked Position Modelling}}
\newglossaryentry{gelu}{name={GELU},description={Gaussian Error Linear Unit}}
\newglossaryentry{ss}{name={SS},description={Single step decoding}}
\newglossaryentry{ro}{name={RO},description={Random order decoding}}
\newglossaryentry{ho}{name={HO},description={Human order decoding}}
\newglossaryentry{hc}{name={HC},description={Highest-confidence decoding}}
\newglossaryentry{le}{name={LE},description={Lowest-entropy decoding}}
\newglossaryentry{gru}{name={GRU},description={Gated Recurrent Unit}}
% CoNLL 2020
\newglossaryentry{sod}{name={SOD},description={Soft Organ Distance loss function}}
\newglossaryentry{ior}{name={IOR},description={Inside Organ Ration}}
\newglossaryentry{nvd}{name={NVD},description={Nearest Voxel Distance}}
\newglossaryentry{nvdo}{name={NVD-O},description={Nearest Voxel Distance Outside of the organ}}
\newglossaryentry{mesh}{name={MeSH},description={Medical Subject Headings}}
\newglossaryentry{igr}{name={IGR},description={Inside Group Ratio}}
% EMNLP 2023
\newglossaryentry{roberta}{name={RoBERTa},description={A Robustly Optimized BERT Pretraining Approach}}
\newglossaryentry{oie}{name={OIE},description={Open Information Extraction}}
\newglossaryentry{kg}{name={KG},description={Knowledge Graph or Knowledge Graphs}}
\newglossaryentry{falb}{name={FaLB},description={Fact Linking Benchmark}}
\newglossaryentry{llm}{name={LLM},description={Large Language Model}}
\newglossaryentry{out-of-kg}{name={Out-of-KG},description={Outside-of-Knowledge Graph}}
\newglossaryentry{brkg}{name={BRKG},description={Benchmark-Restricted Knowledge Graph}}
% BMVC 2021
\newglossaryentry{stlt}{name={STLT},description={Spatial-Temporal Layout Transformer}}
\newglossaryentry{pff}{name={PFF},description={Per-Frame (Multimodal) Fusion}}
\newglossaryentry{pbf}{name={PBF},description={Per-Box (Multimodal) Fusion}}
\newglossaryentry{ef}{name={EF},description={Early (Multimodal) Fusion}}
\newglossaryentry{vatf}{name={VATF},description={Video Action Transformer (Multimodal) Fusion}}
\newglossaryentry{lcf}{name={LCF},description={Late Concatenation (Multimodal) Fusion}}
\newglossaryentry{caf}{name={CAF},description={Cross-Attention (Multimodal) Fusion}}
\newglossaryentry{cacnf}{name={CACNF},description={Cross-Attention CentralNet (Multimodal) Fusion model}}
\newglossaryentry{roi}{name={RoI},description={Region of Interest}}
\newglossaryentry{roialign}{name={RoIAlign},description={Region of Interest Alignment}}
\newglossaryentry{s_and_tlt}{name={S\&TLT},description={Spatial and Temporal Layout Transformer}}
\newglossaryentry{map}{name={mAP},description={Mean Average Precision}}
\newglossaryentry{r3d}{name={R3D},description={3D ResNet, usually 50 layers deep}}
\newglossaryentry{i3d}{name={I3D},description={Inflated 3D Convolutional Neural Network}}
% ICCV 2023
\newglossaryentry{ece}{name={ECE},description={Expected Calibration Error}}
\newglossaryentry{of}{name={OF},description={Optical Flow}}
\newglossaryentry{audio}{name={A},description={Audio, when used as an input modality}}
\newglossaryentry{rgb}{name={RGB},description={Regular RGB video frames, when used as an input modality}}
\newglossaryentry{swin_t}{name={Swin-T},description={Swin-Tiny Transformer Video Encoder}}
\newglossaryentry{obj}{name={OBJ},description={Object detections}}
\newglossaryentry{imu}{name={IMU},description={Inertial Measurement Unit}}
\myprintglossary

% --------------- ABBREVIATIONS END ---------------------

% To create a list of symbols, there are 2 options
% 1. include a manually created nomenclature as a chapter
%    \includenomenclature{nomenclaturechapter}
% 2. automatic generation via the nomencl package
%    \nomenclature[cB]{$c_B(\vec{x})$}{Characteristic function of $B$}

% ------------------------ SYMBOLS START ------------------------
% \nomenclature{$\tau$}{Temperature parameter used for normalization.}
% \myprintnomenclature
% ------------------------ SYMBOLS END ------------------------

\tableofcontents
\listoffigures
\listoftables
%%%%%%%%%%%%%%%%%%%%%%%%%%%%%%%%%%%%%%%%%%%%%%%%%%%%%%%%%%%%%%%%%%%%%%

\mainmatter % to get \pagenumbering{arabic}

% Show instructions on a separate page
\instructionschapters\cleardoublepage

  \graphicspath{{\chaptersdir/introduction/image/}}%
  %\cleardoublepage%
  % !TeX root = ../../manuscript.tex
\chapter{Introduction}\label{ch:introduction}
The world we live in is inherently \textit{multimodal}; our experiences and understanding of the surrounding environment are shaped by inputs from various senses: we see objects, people, and scenes; we hear sounds and speech; we smell odors; we feel textures. More concretely, a \textit{modality} can refer to the representation of a natural phenomenon, or an event \cite{baltruvsaitis2018multimodal}. For example, the event of a tree falling can be represented via: the visual depiction of the event (the process of falling), the static observation of the event (the aftermath: i.e., the fallen tree), the sound made by the tree falling, or a piece of text describing the event itself. Overall, all depictions represent the same event, albeit in their unique ways, and share the same semantics.

In the context of Machine Learning, a \textit{modality} often refers to the form of data: text, images, videos, or audio \cite{baltruvsaitis2018multimodal, liang2022foundations}. These forms of data are often obtained from sensors, for example, images and videos captured from cameras or audio recorded through a microphone. In the literature \cite{liang2022foundations}, the modalities are placed along a spectrum from \textit{raw} to \textit{abstract}. Raw modalities are the ones that come from the sensor itself; e.g., camera-captured images or videos, or audio recorded from a microphone. On the other hand, an abstract modality is typically derived from a raw modality and offers a higher-level, more refined representation of the data. For example, object detections \cite{ren2015faster} obtained from a video represent such abstract modality. The object detections abstract away the video appearance-based properties while highlighting object-centric spatial and temporal-based properties: the spatial arrangements and temporal continuity of the objects that appear in the video. Other abstract modalities can be, for example, optical flow estimation from a video \cite{perez2013tv}, or a knowledge graph representation of the facts present in natural language text \cite{vrandevcic2012wikidata}.

A machine learning problem (or task) is therefore considered \textit{multimodal} if multiple such modalities are involved, regardless of their type (raw or abstract) \cite{liang2022foundations}. For example, such tasks could be generating captions about images, generating images from text, or recognizing human actions from videos using the video frames and the corresponding audio accompanying the video. In order to build machine learning models capable of understanding the world around us, and hence aid people in everyday tasks, such models need to relate information and knowledge encoded in diverse multimodal data. Multimodal data exhibits three (main) properties which make it both suitable and challenging for machines to learn from \cite{baltruvsaitis2018multimodal, liang2022foundations}:

\begin{itemize}
    \item \textit{Heterogeneity}: The information present in each modality manifests itself in a distinct structure. For example, text represents information as a sequence of words, images represent information as spatially arranged pixels along a 2D grid, while videos represent information in both spatial structure and temporal structure, via pixels and frames respectively. Furthermore, different modalities convey information at a different semantic level, i.e., ``A picture is worth a thousand words''. Therefore, the property of heterogeneity plays a vital role in the context of multimodal machine learning tasks. 
    
    \item \textit{Connectivity}: Despite being heterogeneous, the modalities are often connected, namely, they represent the same information. To be specific, certain elements of one modality (e.g., a noun phrase of a sentence) can be connected (i.e., share information), with elements of another modality (e.g., a set of pixels in the image representing a particular object). Arising from the property of connectivity are a multitude of skills machine learning models can learn from multimodal data: e.g., generating images from text.
    
    \item \textit{Interactivity}: The modalities can interact such that new information emerges. Lian~\etal~\cite{liang2022foundations} note an important difference between connectivity and interactivity: ``Connections inherently exist within the multimodal data itself, whereas interactions only arise when modalities are integrated and processed together to bring a new response.'' Consequently, modeling such interaction between two or more distinct modalities is fundamental to performing fusion of multimodal data: the task of combining diverse modalities to solve a specific task.
\end{itemize}

Liang~\etal~\cite{liang2022foundations} and Baltruvsaitis~\etal~\cite{baltruvsaitis2018multimodal} outline multiple challenges and subchallenges that represent the cornerstone of multimodal machine learning research. This manuscript aims to investigate four key challenges:
\begin{inparaenum}[(i)]
    \item Alignment;
    \item Translation;
    \item Fusion; and
    \item Transference.
\end{inparaenum}

\section*{Chapter Overview}
% The rest of this chapter is organized as follows: In \S\ref{chapter1:alignment-translation} we introduce the problems of multimodal alignment and multimodal translation, and in \S\ref{chapter1:fusion-colearning} we introduce the problems of multimodal fusion and multimodal transference. Then, in a bottom-up manner, we give an overview of this manuscript, such that we first introduce the problem-specific research questions for each chapter (\S\ref{chapter1:overview-problem-specific}), and then we cluster them together and introduce the universal research questions of this manuscript (\S\ref{chapter1:epilogue-universal}).
The rest of this chapter is organized as follows. First, we delve into the challenges of multimodal alignment and translation in \S\ref{chapter1:alignment-translation}. Subsequently, we examine the challenges associated with multimodal fusion and transference in \S\ref{chapter1:fusion-colearning}. Then, in a bottom-up manner, we first detail the specific research questions addressed in each chapter in \S\ref{chapter1:overview-chapter-specific}, and lastly, we synthesize these chapter-specific research questions into two transversal research questions that encapsulate the manuscript's core objectives in \S\ref{chapter1:epilogue-universal}.

\section{Multimodal Alignment and Translation}\label{chapter1:alignment-translation}
\textit{Multimodal alignment} aims to establish cross-modal connections between elements of multiple distinct modalities. For example, in a typical multimodal alignment scenario, we would align the representations of videos and captions, a task known as cross-modal retrieval \cite{frome2013devise}. Multimodal alignment is challenging as it is often ambiguous (i.e., the caption might match with multiple videos simultaneously, and only partially describe each of the videos), requires modeling long-range dependencies (i.e., modeling a large context window), or the alignment itself might not exist at all.

The alignment can further be (i) discrete, and (ii) continuous. In a discrete scenario, we learn an alignment between the representation spaces of two modalities which can be segmented into atomic units for the task at hand. A typical case of discrete alignment would be linking a real-world entity represented in text (e.g., ``Michael Jordan'') with the corresponding canonical representation in a knowledge graph \cite{vrandevcic2012wikidata, radevski-etal-2023-linking}. In a continuous setting, on the other hand, at least one of the modalities cannot be segmented into such atomic units. If we were to arrange a set of objects on a 2D canvas based on a description of the scene, the 2D canvas renders the alignment continuous. That is, while the atomic unit of the text is a word, the 2D canvas exhibits an ambiguous segmentation.

The challenge of multimodal alignment is directly related to the property of heterogeneity between modalities. The inherent structural and semantic differences between modalities---e.g., the sequential structure of natural language text and the spatial and temporal structure of videos---render multimodal alignment as a complex task. Addressing the challenge of multimodal alignment therefore involves developing models that can cope with the heterogeneity of data representations, ensuring that, despite their structural and semantic differences, the modalities are effectively aligned.

On the other hand, the goal of \textit{multimodal translation} is to transform (i.e., translate) the representation of one modality into another while preserving the information content.  Within this manuscript, we want to establish an important distinction between alignment and translation. That is, the multimodal alignment is always bi-directional, where each modality can be mapped to the other. Multimodal translation, however, can be considered a form of one-directional alignment, where the representation from one modality can be converted to another, but not necessarily vice versa. For instance, in standard multimodal alignment tasks, a video description can be aligned to the corresponding video and vice versa. On the other hand, in translation tasks, it is possible to find a corresponding video given the description, but the reverse may not be feasible.

Connectivity, the property indicating that different modalities can represent the same concept or event, makes the challenge of \textit{translation} between modalities feasible. Multimodal translation relies on the inter-connectedness of modalities, where elements from one modality can enable the generation or interpretation in another. To overcome the challenge of multimodal translation, models must leverage the connectivity property, understanding the complex relationships that allow for coherent cross-modal translation.

\subsubsection{Is multimodal alignment a precondition for multimodal translation?}

While multimodal alignment and translation are closely related, they are not \textit{always} interdependent. Multimodal alignment involves creating bidirectional mappings between modalities, ensuring that elements from one modality can be directly associated with elements from another. This bidirectional mapping is the backbone for tasks that require mutual understanding and retrieval between modalities: e.g., cross-modal retrieval or grounding language in visual contexts.

Multimodal translation, on the other hand, can sometimes be achieved without explicit alignment. Translation focuses on converting information from one modality to another, typically in a unidirectional manner. For example, generating a textual description from an image (image captioning) is an instance where the end goal is not necessarily to align elements across modalities, but rather, to faithfully reproduce the semantic content in a different form. In these cases, while \textit{implicit} alignment might emerge as part of the translation,
% (e.g., identifying key objects in an image to describe them in text),
\textit{explicit} alignment of every element between the modalities is not necessary.

In conclusion, while multimodal alignment might enhance multimodal translation, it is not a strict precondition. Namely, effective multimodal translation can, in certain scenarios, be achieved by models designed to understand and map the high-level semantics across modalities without relying on explicit alignment.

\subsubsection{Chapters Overview}

In \textbf{Part~\ref{part:alignment-translation}} of this manuscript, which consists of \textbf{Chapter~\ref{ch:emnlp2020}}, \textbf{Chapter~\ref{ch:conll2020}}, and \textbf{Chapter~\ref{ch:emnlp2023}}, we investigate the problems of multimodal alignment and multimodal translation.

In \textbf{Chapter~\ref{ch:emnlp2020}}, we translate a set of spatial relations expressed in language to a set of 2D spatial arrangements \cite{radevski2020decoding}. Given a set of sentences (i.e., a scene description) describing the spatial relations between a set of objects, we arrange the set of objects in a 2D space, such that the generated spatial arrangements satisfy the imposed spatial relations. Importantly, we learn a one-directional mapping: we seek to find the 2D spatial arrangements between the objects that correspond best with the language-expressed spatial relations.

In \textbf{Chapter~\ref{ch:conll2020}}, we translate medical text into a 3D location in an anatomical model of the human body (i.e., a human atlas) \cite{grujicic-etal-2020-learning}. Given a segment of medical text, which typically relates to a specific location in the human body (e.g., a particular organ), we learn a one-directional mapping to a 3D location in the human atlas. Conversely, the reverse process is not feasible: we cannot identify the medical document most closely associated with a given 3D location in the human atlas.
% is not possible.

In \textbf{Chapter~\ref{ch:emnlp2023}}, we focus on linking, or translating, general-purpose text expressed in a structured format to knowledge graph facts, which represent canonical entities and relations \cite{radevski-etal-2023-linking}. Similarly, we learn a one-directional mapping: we translate text to a knowledge graph but not the other way around.

\section{Multimodal Fusion and Transference}\label{chapter1:fusion-colearning} 
The problem of \textit{multimodal fusion} is concerned with how we can combine the representations of two modalities, usually heterogeneous, such that we capture and model their cross-modal interactions. The goal is to reduce the number of representations or to come up with a single representation that emphasizes only the task-relevant semantics from each modality. Ultimately, what emerges from the fusion process is a representation that is more task-relevant and informative than the individual, modality-specific representations. The fusion can be \textit{raw}, if it involves the raw modalities themselves without any processing, or \textit{abstract}, if it involves abstract modalities (e.g., object detections) and their representations.

Even though multimodal fusion is effective for numerous downstream applications (e.g., recognizing human actions \cite{carreira2017quo, radevski2021revisiting, herzig2022object}), it inherently assumes that all modalities are available throughout the whole lifespan of the application. The goal of \textit{multimodal transference}, on the other hand, is to transfer knowledge between the representations of two or more distinct modalities during training of the machine learning model \cite{girdhar2022omnivore, radevski2023multimodal}. Usually, such knowledge transfer is one-directional: i.e., we transfer knowledge from a secondary modality---or a set of modalities of a secondary nature---to a primary modality. For example, the primary modality could be the representation of the video frames, while the secondary modality could be the audio recording. Finally, what comes out of multimodal knowledge transference is a model that can induce multimodal knowledge while relying \textit{only} on the primary modality at inference time.

Interactivity, the emergence of new information through the integration of multiple modalities, is closely related to the challenges of \textit{multimodal fusion} and \textit{multimodal transference}. Multimodal fusion, in particular, is predicated on the ability of models to interactively combine two or more modalities, harnessing their combined strengths to achieve superior performance or generate novel insights. The interactivity property emphasizes the need for machine learning models to not just co-exist alongside data types of different modalities, but also to integrate them, creating a cohesive understanding that transcends the sum of its parts. Furthermore, interactivity plays a crucial role in multimodal transference; it involves the models' capacity to not only integrate but also transfer the insights gained from such integrations across different modalities.

\subsubsection{Chapters Overview}

In \textbf{Part~\ref{part:fusion-transference}} of this manuscript, which consists of \textbf{Chapter~\ref{ch:bmvc2021}} and \textbf{Chapter~\ref{ch:iccv2023}}, we investigate the problems of multimodal fusion and multimodal knowledge transference.

In \textbf{Chapter~\ref{ch:bmvc2021}}, we investigate the optimal way to perform multimodal fusion of two abstract modalities: video representations and object detections (bounding boxes and object categories) \cite{radevski2021revisiting}. To perform the analysis we focus on the task of action recognition, that is, a task where a machine learning model should recognize (i.e., classify) the main action taking place in a video. Importantly, such models make use of \textit{all} modalities participating in the fusion at inference time.

In \textbf{Chapter~\ref{ch:iccv2023}}, we ease the aforementioned requirement. For the context of human action recognition from videos, we perform multimodal knowledge transference between several diverse modalities \cite{radevski2023multimodal}. As the multimodal knowledge transference takes place during training, at inference time the model relies only on the primary modality (i.e., the video itself) to recognize the human actions.

\section{Overview and Chapter-specific Research Questions}\label{chapter1:overview-chapter-specific}
The main topics (i.e., broad challenges) this manuscript covers are multimodal alignment and translation (in \textbf{Part~\ref{part:alignment-translation}}) and multimodal fusion and transference (in \textbf{Part~\ref{part:fusion-transference}}). We dedicate three chapters to \textbf{Part~\ref{part:alignment-translation}}, and two chapters to \textbf{Part~\ref{part:fusion-transference}}. In each chapter, we investigate a specific problem related to a practical real-world task, where each problem is derived from one of the fundamental challenges in multimodal machine learning that are the focus of this manuscript. Before we dive into all of this, in \textbf{Chapter~\ref{ch:background}} we provide a short introduction of the background concepts necessary for understanding the content of this manuscript.

\subsection{Part I: Multimodal Alignment and Translation}
In \textbf{Part~\ref{part:alignment-translation}} of this manuscript, we delve into the intricacies of multimodal alignment and translation, and we seek to understand how we can effectively align and translate different data forms (e.g., text and visual inputs). Our research has implications on, and contributes to, applications such as automated scene generation, tools for enhancing medical education, and automated fact-checking.

The goal of \textbf{Chapter~\ref{ch:emnlp2020}} is to build models that can translate a set of language-expressed spatial relations to a set of 2D spatial arrangements represented as objects on a 2D canvas. Such a topic is at the intersection of natural language processing and visual data interpretation, an area central to multimodal machine learning. By decoding language-expressed spatial relations to 2D spatial arrangements of objects, we address the challenge of translating abstract, linguistic descriptions into concrete, visual representations---crucial for applications like automated scene generation and robotics. The main research question this chapter seeks to answer is:

\researchquestion{Research Question 1}{How effectively can we translate language-expressed spatial relations into 2D spatial arrangements between objects without constraints on their number and type?}

We further expand on this research question by exploring the nuances of spatial language and distinguishing between explicit spatial relations (such as ``above'' or ``left-of'') and implicit spatial relations (such as ``wearing'' or ``holding''). The motivation stems from the complexity and variability of the natural language used to describe spatial arrangements between objects. Namely, spatial language encompasses a wide range of expressions: from clear, direction-based descriptions to more contextual, interpretive phrases that imply spatial relationships. Understanding how these different types of spatial descriptions impact the performance of translating them into visual representations affects numerous real-world applications. Models that accurately interpret both explicit and implicit spatial descriptions are more versatile and applicable to a broader range of scenarios. In robotics, for example, commands given to a robot might vary from explicit (``place the book on the table'') to more implicit instructions (``put the book back where it belongs''). Consequently, we pose the following research question:

\subresearchquestion{Research Question 1.1}{Does the nature of spatial language relations (explicit like ``above'', ``left-of'' or implicit like ``wearing'', ``holding'') significantly influence the correctness of the translation between the language and the 2D spatial arrangements?}

Furthermore, we investigate the models' robustness, particularly their ability to handle spatial relations that are out-of-distribution (OOD)---unseen during training. The models' ability to generalize to unseen data is particularly important in practical applications. Specifically, in real-world scenarios, a model frequently encounters descriptions it was not explicitly trained on. Therefore, measuring the robustness to OOD samples is a crucial step toward deploying such a method in diverse real-world environments. Consequently, we ask:

\subresearchquestion{Research Question 1.2}{How robust is the translation model we learn, in dealing with spatial relations which are out-of-distribution (i.e., not encountered during training)?}

Both secondary research questions we pose are connected to the main one. With each, we delve deeper into specific aspects of the challenge of translating language-expressed spatial relations into visual representations. Together, by answering these questions we aim to advance the field of multimodal machine learning by tackling the nuanced challenges of interpreting and understanding spatial language.

The content of this chapter is based on the following publication:

\begin{mdframed}[backgroundcolor=gray!30,linewidth=0pt]
\underline{Radevski, G.}, Collell, G., Moens, M-F., and Tuytelaars, T., (2020). Decoding Language Spatial Relations to 2D Spatial Arrangements. Findings of The Conference on Empirical Methods in Natural Language Processing (EMNLP). \cite{radevski2020decoding}
\end{mdframed}

The goal of \textbf{Chapter~\ref{ch:conll2020}} is to build models that ground (i.e., translate) medical texts to 3D locations in an anatomical model of the human body. In that way, we contribute to the vital task of interpreting, visualizing, and exploring complex medical texts. By grounding textual descriptions in a 3D anatomical model, such models can enhance medical education and improve diagnostic tools, which helps medical professionals be more time-efficient. The main research question this chapter seeks to answer is:

\researchquestion{Research Question 2}{Can we train models that translate medical texts to 3D locations within a human anatomical atlas, leveraging the spatial co-occurrence of medical terms to create meaningful grounding?}

Next, we also measure the generalization ability of the model beyond its training data. In real-world medical applications, the variability in medical texts and descriptions is immense, and therefore, the ability to handle out-of-distribution data could significantly enhance their utility. Similarly, 
% of educational resource tools.
performance on unseen data is directly related to the continuous applicability of the models, which are meant to support medical professionals without the need for constant retraining. Consequently, we ask:

\subresearchquestion{Research Question 2.1}{How does the model perform in grounding medical texts to locations within the human body that were not part of the training dataset (i.e., human organs that are never referred to in the training data)?}

To reduce the reliance on extensively annotated datasets, we further investigate the feasibility of using self-supervised learning methods to train such models. Namely, annotated medical datasets are expensive and time-consuming to create. If models can be trained effectively in a self-supervised manner, it would significantly lower the barriers to developing advanced medical visualization tools, making them more accessible and scalable. Consequently, we ask:

\subresearchquestion{Research Question 2.2}{Is it feasible to train medical text grounding models in a self-supervised manner, thereby eliminating the dependency on human-annotated data?}

While the main research question establishes the foundational goal of translating medical text into a 3D anatomical context, Research Question 2.1 tests the limits of this translation in terms of generalization to unseen data. Meanwhile, Research Question 2.2 addresses a critical aspect of the model's training process, exploring whether we could obtain such models in a more label-efficient---self-supervised---manner. Overall, we address both the performance and practical aspects of building models capable of grounding medical text in an anatomical atlas, attempting to address practical implications in the medical and machine learning fields.

The content of this chapter is based on the following publication:

\begin{mdframed}[backgroundcolor=gray!30,linewidth=0pt]
Grujicic*, D., \underline{Radevski*, G.}, Tuytelaars, T., Blaschko, M., (2020). Learning to ground medical text in a 3D human atlas. The Conference on Computational Natural Language Learning (CoNLL). \cite{grujicic-etal-2020-learning} (*authors contributed equally)
\end{mdframed}

The goal of \textbf{Chapter~\ref{ch:emnlp2023}} is to build models that translate general purpose text, represented as structured triplets, e.g., \emph{(``Michael Jordan''; ``played for''; ``Chicago Bulls'')}, to canonical facts in a knowledge graph. The ability to accurately link textual information to structured knowledge graphs is crucial for the development of systems that can perform fact-checking given natural language text, or populating knowledge graphs with novel information. The main research question this chapter seeks to answer is:

\researchquestion{Research Question 3}{How effectively can we translate general-purpose textual data, represented as structured triplets, to corresponding facts in a knowledge graph?}

Furthermore, we complement the main research question by diving into the nuanced scenario where such translation might be infeasible due to the absence of the corresponding facts in the knowledge graph. Different from \textbf{Chapter~\ref{ch:emnlp2020}} and \textbf{Chapter~\ref{ch:conll2020}}---where a translation between the modalities always exists---within the scope of this chapter, we also consider the scenario where a translation between the text and the knowledge graph might not exist at all. Importantly, by identifying instances where text triplets cannot be linked to existing knowledge graph facts, we can identify novel or incorrect information that is outside of the knowledge graph. This functionality is essential both for knowledge graph population and fact-checking. Consequently, we ask:

\subresearchquestion{Research Question 3.1}{Can we accurately identify instances where a valid translation between the text triplets and knowledge graph facts does not exist? That is, the text represents knowledge that is not in the knowledge graph: i.e., factually incorrect or novel information.}

Our motivation for posing this secondary research question is expanding the model capabilities beyond mere translation/linking tasks. Together, both research questions contribute to insights that are essential for understanding the limitations and challenges in automating the knowledge graph population and detection of novel/incorrect information.

The content of this chapter is based on the following publication:

\begin{mdframed}[backgroundcolor=gray!30,linewidth=0pt]
\underline{Radevski, G.}, Gashteovski, K., Hung, C-C., Lawrence, C., Glavaš, G., (2023). Linking Surface Facts to Large-Scale Knowledge Graphs. The Conference on Empirical Methods in Natural Language Processing (EMNLP). \cite{radevski-etal-2023-linking}
\end{mdframed}

\subsection{Part II: Multimodal Fusion and Transference}
While in \textbf{Chapter~\ref{ch:emnlp2020}}, \textbf{Chapter~\ref{ch:conll2020}}, and \textbf{Chapter~\ref{ch:emnlp2023}}, we analyze the problem of multimodal alignment and translation between two distinct modalities, in \textbf{Part~\ref{part:fusion-transference}}, we shift our focus on multimodal fusion and transference. 

We first explore the benefits of emergent multimodal representations, arising from the act of combining (i.e., fusing) different modalities and the knowledge transfer that occurs between them. On a similar note, we then focus on multimodal knowledge transference to address the practical limitations of multimodal fusion, thus decreasing the gap between research and real-time processing applications. The goal of this research is to contribute towards more capable and more efficient video recognition systems in resource-constrained environments.

The goal of \textbf{Chapter~\ref{ch:bmvc2021}} is to explore---in the context of human action recognition---how two modalities can be fused so that the task-specific performance of the model increases. More concretely, in this chapter, we analyze the complex challenge of understanding and interpreting actions from videos, a key area in computer vision with wide-ranging applications. The main research question this chapter seeks to answer is:

\researchquestion{Research Question 4}{What is the most effective way to utilize multimodal fusion of two abstract modalities---representations of video frames and object detections---in order to enhance task-specific performance in video-based action recognition?}

The core of multimodal fusion lies in effectively integrating diverse modalities to leverage their unique strengths. For that reason, next, we focus on the essence of what makes multimodal fusion powerful: the preservation of distinct, complementary information from each modality. Understanding how to maintain this complementarity, up to the point of the multimodal fusion, is crucial for ensuring that the integrated representation is richer and more informative than any single modality alone. This is particularly important in video action recognition, where different modalities (like video frames and object detections) contribute unique information. Consequently, we ask:

\subresearchquestion{Research Question 4.1}{How do we preserve the complementarity of the modality representations prior to their fusion, which is the most crucial property that needs to be maintained to achieve optimal task performance?}

While the main question establishes the goal of utilizing multimodal fusion to improve performance, with the secondary research question, we analyze a critical detail in achieving this goal: ensuring that each modality's unique and complementary information is fully leveraged in the fusion process.
% The focus on the complementarity before the fusion is essential for realizing the potential benefits of integrating multiple modalities.
Specifically, our goal is to underline the importance of not just the fusion itself, but also to offer insights into the foundation of an effective multimodal fusion.

The content of this chapter is based on the following publication (Oral presentation):

\begin{mdframed}[backgroundcolor=gray!30,linewidth=0pt]
\underline{Radevski, G.}, Moens, M-F., and Tuytelaars, T. (2021). Revisiting Spatio-Temporal Layouts for Compositional Action Recognition. The British Machine Vision Conference (BMVC).
\end{mdframed}

The main benefit of multimodal fusion is leveraging the emergent knowledge stemming from the interaction between the modalities at inference time. However, this inherently assumes the availability of both modalities, as well as a sufficient compute budget to process all of them during inference. Such assumption is often violated in practical settings: some of the modalities might be unavailable, or, the compute budget may not allow joint processing of the multimodal data.

The goal of \textbf{Chapter~\ref{ch:iccv2023}} is to make use of multimodal knowledge transference to overcome the aforementioned limitation. In particular, we investigate the transference of multimodal knowledge, via knowledge distillation, with a special focus on resource-constrained environments featuring mainly egocentric video data. This renders our research particularly relevant for developing efficient models capable of understanding and interpreting human actions, essential for technologies such as wearable computing and augmented reality. The main research question this chapter seeks to answer is:

\researchquestion{Research Question 5}{Can we come up with a method which enables successful multimodal knowledge transference during training, to emulate the capabilities of a multimodal fusion-based model during inference?}

Importantly, in real-world applications, the quality of data from different modalities can vary significantly due to factors like sensor quality, environmental conditions, or operational constraints. To address this requirement, we explore how to optimize the knowledge transference process, and attempt to determine how lower-quality modalities contribute to the models' learning and performance. Consequently, we ask:
% Practical Applicability: By addressing the challenges posed by noisier modalities, this research could significantly broaden the applicability of multimodal learning models, making them more versatile and useful across a wider range of conditions and applications.

\subresearchquestion{Research Question 5.1}{What is the impact of incorporating a new modality that is less reliable, or noisier, compared to other modalities in the multimodal transference process?}

We finally want to measure the computational efficiency vs. task effectiveness trade-offs between multimodal fusion and multimodal transference, which is a key consideration in the deployment of machine learning systems in resource-constrained environments (e.g., wearable devices). The ultimate goal is to offer insights that can guide practitioners in choosing between multimodal fusion and multimodal transference strategies based on their specific requirements for performance and computational resources. To do so, we pose the following secondary research question:

\subresearchquestion{Research Question 5.2}{How does multimodal knowledge transference compare to multimodal fusion in terms of downstream task performance and computational complexity?}

With these questions, we attempt to comprehensively explore the challenge of multimodal knowledge transference in the context of (egocentric) video understanding, from its feasibility and optimization to its practical implementation and impact. Overall, with the research in \textbf{Chapter \ref{ch:iccv2023}} we seek to advance our understanding of multimodal knowledge transference in a way that is practically valuable for emerging technologies that operate within real-world constraints.

The content of this chapter is based on the following publications:

\begin{mdframed}[backgroundcolor=gray!30,linewidth=0pt]
\underline{Radevski*, G.}, Grujicic*, D., Blaschko, M., Moens, M-F., Tuytelaars, T., (2022). Students taught by multimodal teachers are superior action recognizers. The 2nd International Ego4D Workshop @ European Conference on Computer Vision (ECCV). \cite{radevski2022students} (*authors contributed equally)
\end{mdframed}

\begin{mdframed}[backgroundcolor=gray!30,linewidth=0pt]
\underline{Radevski*, G.}, Grujicic*, D., Blaschko, M., M-F., Moens, Tuytelaars, T., (2023). Multimodal Distillation for Egocentric Action Recognition. The International Conference on Computer Vision (ICCV). \cite{radevski2023multimodal} (*authors contributed equally)
\end{mdframed}

\section{Chapter Epilogue: Transversal Research Questions}\label{chapter1:epilogue-universal}
All in all, this manuscript investigates the challenges and possibilities within multimodal machine learning, which we can broadly group into two transversal research questions. These research questions are not chapter-specific, but rather transverse across several chapters belonging to one part. With the transversal research questions, we address issues that cannot be fully explored or answered within the boundaries of a single chapter. Therefore, our goal with these questions is to synthesize the diverse research goals across individual chapters, while highlighting the unified objectives of this work. In \textbf{Chapter~\ref{ch:conclusion}}, we revisit these questions to summarize the main contributions of the manuscript.

\subsection*{Part I: Multimodal Alignment and Translation}

\universalresearchquestion{Transversal Research Question I}{How effectively can we align and translate across different heterogeneous modalities, while accounting for the modality-specific nuances and challenges of out-of-distribution data?}

With this question, we unify our research goals of developing multimodal methods that not only bridge the semantic differences between various types of data, but also consider the complexities of spatial descriptions, medical texts, or diverse surface-form facts. We highlight the importance of creating models that can adapt to new, unseen data, which is a key requirement for real-world applications such as automated scene generation, medical education tools, and the detection of incorrect or novel information based on knowledge graphs.

\subsection*{Part II: Multimodal Fusion and Transference}

\universalresearchquestion{Transversal Research Question II}{How can we harness the power of multimodal fusion and knowledge transference to create more capable and efficient video understanding machine learning models?}

With the second transversal research question, we consolidate our research on the integration of multimodal data, to enhance the capabilities of video understanding machine learning models beyond what models trained on unimodal data can achieve. In doing so, we contribute towards the development of models that can harness the complementarity of multimodal data---both during training and inference---to ultimately lessen the gap between research and industry.

% vim: tw=70 nocindent expandtab foldmethod=marker foldmarker={{{}{,}{}}}
%

% \includechapter{manual} % Remove this chapter
%
  \graphicspath{{\chaptersdir/background/image/}}%
  %\cleardoublepage%
  \chapter{Background and Fundamentals}\label{ch:background}
Within this chapter, we briefly describe the key concepts necessary to follow the remainder of the manuscript. First, in \S\ref{background:sec:raw} we introduce how, in the context of machine learning, we represent different raw modalities: text, images, videos, and audio. Further, in \S\ref{background:sec:abstract}, we discuss the abstract modalities that we use as well as how we obtain representations from them. Finally, in \S\ref{background:sec:derived-independent}, we attempt to clarify the differences between derived and independent modalities. In \S\ref{background:sec:neural-networks}, we concisely introduce what neural networks are, how they can be trained to solve a task of interest (\S\ref{background:sec:training-nns}), and discuss in more detail why different types of neural networks are better suited for certain modalities (\S\ref{background:sec:modality-specific-nns}).
% We discuss in more detail why different types of neural networks are better suited for certain modalities,
% and 
In \S\ref{background:sec:transformers}, we introduce a specific neural network model---the Transformer \cite{vaswani2017attention}---which we make use of in all remaining chapters, and discuss why it is particularly well suited for transfer learning (\S\ref{background:sec:transfer}).
% , and discuss the potential of a single transformer neural network to encode \textit{all} modalities jointly (\S\ref{background:sec:any-encoder}).
Next, in \S\ref{background:sec:alignment-translation} and \S\ref{background:sec:fusion-transference}, we provide necessary background material which is specific to \textbf{Part~\ref{part:alignment-translation}} and \textbf{Part~\ref{part:fusion-transference}}, respectively. Lastly, in \S\ref{background:sec:videodata}, we provide a consolidated description of the action recognition datasets we use in \textbf{Part~\ref{part:fusion-transference}}, and discuss the properties of egocentric vision.

\section{Modality Representations}\label{background:sec:unimodal-representations}
In \textbf{Chapter~\ref{ch:introduction}}, we introduced the concept of a modality as a representation of a natural phenomenon. In the context of machine learning, however, and in this manuscript too, we refer to a modality as the form of data: text, images, audio, etc. Importantly, different modalities exhibit a specific structure and properties.

\subsection{Raw Modalities}\label{background:sec:raw}
The most common raw modalities, and the ones used in this manuscript are text, images, videos, and audio. Importantly, each modality inherently aligns with machine learning models that are predisposed with particular inductive biases\footnote{An inductive bias refers to the set of assumptions a learning algorithm makes to generalize beyond the training data.
% Importantly, each of these modalities requires a machine learning model to employ a specific inductive bias
These biases influence the types of patterns and solutions a machine learning algorithm is more likely to learn \cite{mitchell1980need}.},
% ---stemming from how the data is structured---
which influences how the respective modality is processed and interpreted.

\textbf{Images.} In image processing and computer vision, the primary inductive biases are spatial locality and translation invariance. Spatial locality refers to the assumption that pixels adjacent to each other are more likely to be related or be a part of the same object \cite{lecun1989backpropagation}. Translation invariance, on the other hand, ensures that the feature map generated in response to an object remains consistent, regardless of the object's location in the image. It is important to differentiate between ``translation invariance'' and ``translation equivariance'', which are terms often used interchangeably. Translation equivariance refers to the property of Convolutional Neural Networks (CNNs), where the feature map adjusts in direct correspondence to any shift of the object within the image. Consequently, even though the CNN's ability to detect the object remains unaffected by the object's location, the internal representation of the features related to the object moves in sync with the movement of the object, maintaining a coherent spatial relationship.

Images are represented as matrices of pixel values, or, to be specific, a regular 3-channel image is represented as a three-dimensional tensor, denoted as $\mathbf{I} \in \mathbb{R}^{c \times h \times w}$, where $c$ represents the number of color channels (such as RGB), and $h$ and $w$ represent the height and width of the image, respectively.

\textbf{Text.} In natural language processing the primary inductive bias is sequentiality: emphasizing the significance of the order of words, characters, or tokens. This bias is inherent due to the sequential structure of language. In the past, text was represented via order-invariant methods such as bag-of-words or TF-IDF \cite{sparck2004statistical} approaches. In the deep learning era, however, text is most commonly represented via sequences of word or token embeddings \cite{mikolov2013distributed}. For instance, a text sequence can be denoted as a matrix $\mathbf{T} \in \mathbb{R}^{n \times d}$, with $n$ indicating the number of words or tokens, and $d$ representing the embedding space dimensionality.

\textbf{Videos.} In video processing and analysis, the primary inductive bias is the combination of spatial locality and temporal continuity. This bias arises from the nature of video as a sequence of images (frames): spatially adjacent pixels are likely to be related, while there is also a strong semantic and structural relationship between adjacent frames over time \cite{ji20123d}. The spatial-temporal aspect of the videos indicates that video understanding requires not only recognizing objects and patterns within each frame, but also the interaction between these elements over time.

Videos are typically represented as four-dimensional tensors. A video can be denoted as $\mathbf{V} \in \mathbb{R}^{c \times t \times h \times w}$, where $c$ represents the number of channels (such as RGB), $t$ is the number frames in the video, and $h$ and $w$ represent the height and width of the video frame, respectively. This representation effectively captures both the spatial information within each frame and the temporal information across the sequence of frames.

\textbf{Audio.} In audio processing and analysis, the primary inductive bias is the temporal and frequency-based nature of the sound. Such bias reflects the inherent structure of audio as a time-based sequence which evolves over time and is defined by varying frequencies. Audio signals are commonly represented as waveforms, which are discrete amplitudes measured at regular intervals (sampling rate). This representation is commonly denoted as $\mathbf{A} \in \mathbb{R}^{t}$, where $t$ is determined by the audio's duration and the sampling rate. For example, a 1-second audio sampled at 44,100 Hz (samples per second) would be represented with $t = 44,100$. In machine learning applications, however, audio is often transformed into an abstract modality---a spectrogram---which we introduce later in \S\ref{background:sec:abstract}.

\subsection{Abstract Modalities}\label{background:sec:abstract}

% \begin{figure}[t]
% \centering
% \includegraphics[width=1.0\textwidth]{chapters/background/images/modalities.pdf}
% \caption[Abstract modalities used in Part I]{Abstract modalities used in \textbf{Part~\ref{part:alignment-translation}} of this manuscript. Left to right: Abstract scenes (used in \textbf{Chapter~\ref{ch:emnlp2020}}); Human body atlas (used in \textbf{Chapter~\ref{ch:conll2020}}); Knowledge graph (used in \textbf{Chapter~\ref{ch:emnlp2023}}).}
% \label{background:fig:part-i-modalities}
% \end{figure}

\begin{figure}[t]
\centering
\begin{subfigure}[t]{0.25\textwidth}
  \centering
  \includegraphics[width=\linewidth]{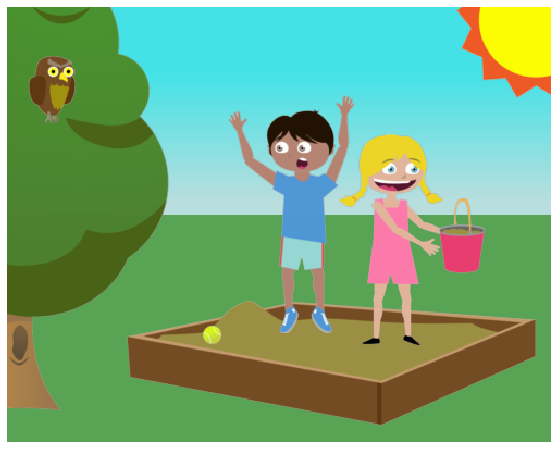}
  \captionsetup{font=scriptsize}
  \caption{Abstract scenes}
\end{subfigure}%
\begin{subfigure}[t]{0.25\textwidth}
  \centering
  \includegraphics[width=0.5\linewidth]{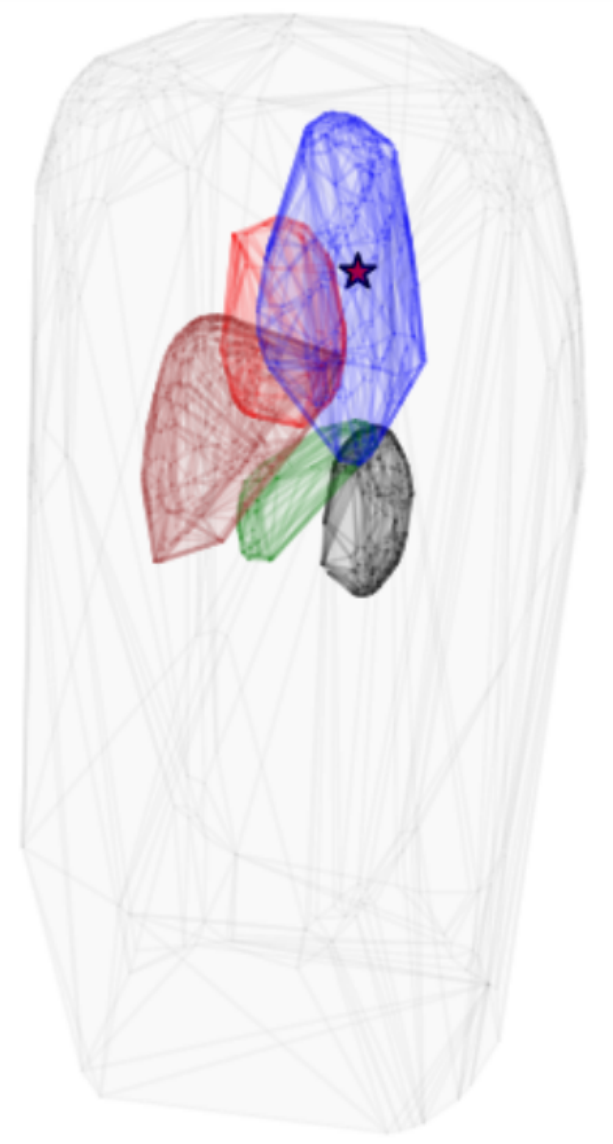}
  \captionsetup{font=scriptsize}
  \caption{Human atlas}
\end{subfigure}
\begin{subfigure}[t]{0.45\textwidth}
  \centering
  \includegraphics[width=\linewidth]{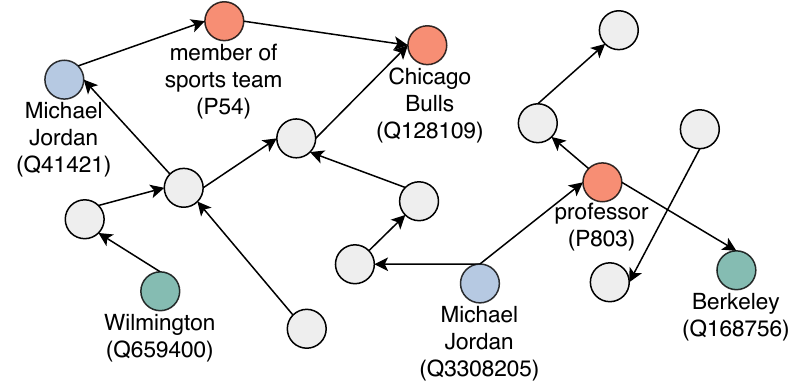}
  \captionsetup{font=scriptsize}
  \caption{Knowledge graph}
\end{subfigure}
\caption[Abstract modalities used in Part I]{Abstract modalities used in \textbf{Part~\ref{part:alignment-translation}} of this manuscript. (a) is used in \textbf{Chapter~\ref{ch:emnlp2020}}; (b) is used in \textbf{Chapter~\ref{ch:conll2020}}; (c) is used in \textbf{Chapter~\ref{ch:emnlp2023}}.}
\label{background:fig:part-i-modalities}
\end{figure}

While the raw modalities capture the essential characteristics inherent to each input type, many insights and data patterns are neither apparent nor trivial to study in the raw form. In this section, we discuss how an abstract modality, which can often, but not necessarily, be derived from a raw modality, highlights specific data properties and allows studying specific phenomena in a less complex manner.
% Such abstract modalities can provide a more organized and compact structure of the data, which in turn allows us to study and leverage specific data properties.
While there is an abundance of such abstract modalities, in this section we only introduce the ones that (re)occur in the scope of this manuscript.

\textbf{Abstract scenes} are proposed by Zitnick~\etal~\cite{zitnick2013bringing}, featuring a finite set of clip-arts (e.g., \insertemoji{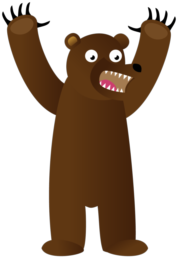}, \insertemoji{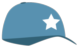}, \insertemoji{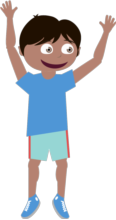}, \insertemoji{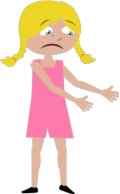}, \insertemoji{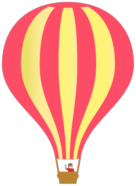}, etc.) which are spatially arranged on a 2D canvas; see Figure~\ref{background:fig:part-i-modalities} (a). The abstract scenes enable us to directly study the high-level semantic and spatial knowledge present in the text (given by scene descriptions), without dealing with low-level information such as image pixels. In \textbf{Chapter \ref{ch:emnlp2020}}, we use the abstract scenes as the translation target modality: we translate a set of scene descriptions---entailing a set of spatial relations---to abstract scenes.

\textbf{Human body atlas} refers to a comprehensive representation of the human body that details its various (sub)structures; see Figure~\ref{background:fig:part-i-modalities} (b). In this manuscript, we use the Segmented Inner Organs (SIO) \cite{pommert2001creating}. The value of each voxel in the atlas represents the label of the corresponding organ or tissue. An organ can be described as the set of indices of voxels which contain the value corresponding to the organ’s segmentation label. In \textbf{Chapter \ref{ch:conll2020}}, we use the human atlas as the translation target modality: we translate medical text to a 3D location in the human atlas.

\textbf{Knowledge graph} is a structured way to represent knowledge of entities: e.g., objects, people, concepts, and the relationships between them \cite{vrandevcic2012wikidata}; see Figure~\ref{background:fig:part-i-modalities} (c). Knowledge graphs excel at representing complex interconnections and dependencies between different entities, particularly effective for illustrating relationships that are difficult to convey through text alone, such as the connections between various people, places, or organizations. In \textbf{Chapter \ref{ch:emnlp2023}}, the knowledge graph is the translation target modality: we translate text to canonical facts in a knowledge graph.

\begin{figure}[t]
\centering
\begin{subfigure}[t]{0.2\textwidth}
  \centering
  \includegraphics[width=\linewidth]{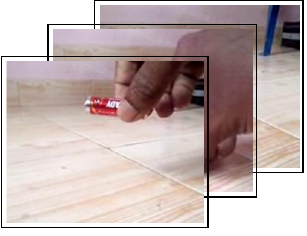}
  \captionsetup{font=scriptsize}
  \caption{Video frames}
  % \label{fig:sub1}
\end{subfigure}%
\hfill
\begin{subfigure}[t]{0.2\textwidth}
  \centering
  \includegraphics[width=\linewidth]{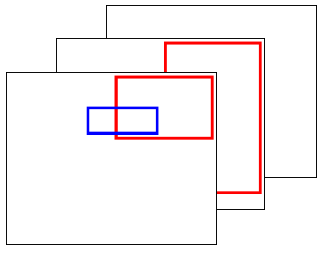}
  \captionsetup{font=scriptsize}
  \caption{Object\\detections}
  % \label{fig:sub2}
\end{subfigure}
\hfill
\begin{subfigure}[t]{0.2\textwidth}
  \centering
  \includegraphics[width=\linewidth]{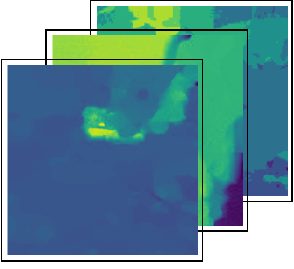}
  \captionsetup{font=scriptsize}
  \caption{Optical flow}
  % \label{fig:sub3}
\end{subfigure}
\hfill
\begin{subfigure}[t]{0.2\textwidth}
  \centering
  \includegraphics[width=\linewidth]{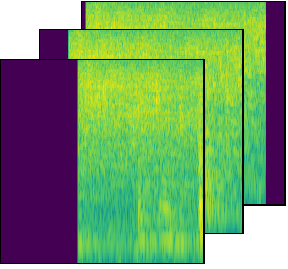}
  \captionsetup{font=scriptsize}
  \caption{Audio\\Spectrogram}
  % \label{fig:sub4}
\end{subfigure}
\caption[Abstract modalities used in Part II]{Video frames (represented as a raw modality), and the abstract modalities used in \textbf{Part~\ref{part:fusion-transference}} of this manuscript.}
\label{background:fig:part-ii-modalities}
\end{figure}

% \begin{figure}[t]
% \centering
% \includegraphics[width=1.0\textwidth]{chapters/background/images/abstract-part-ii.pdf}
% \caption[Abstract modalities used in Part II]{Video frames | Abstract modalities used in \textbf{Part~\ref{part:fusion-transference}} of this manuscript. Left to right: Video frames | Object detections, Optical Flow, and Spectogram obtained from the audio.}
% \label{background:fig:part-ii-modalities}
% \end{figure}

\textbf{Object detections} serve as a refined representation of images or videos, where the raw data is transformed to precise locations (bounding boxes) and categories of the objects of interest \cite{radevski2021revisiting, herzig2022object, materzynska2020something}; see Figure~\ref{background:fig:part-ii-modalities} (b). This abstraction (hence, an abstract modality) is useful in video understanding, where videos may contain extraneous elements such as varying illumination and motion blur. High-quality and accurate object detections can effectively isolate and emphasize the spatial arrangement and temporal dynamics of objects. This abstraction is beneficial for tasks like object-agnostic action recognition, where the focus is on identifying actions (e.g., pushing, moving, etc.) irrespective of the specific objects involved. In \textbf{Chapter \ref{ch:bmvc2021}}, we refer to the object detections as spatio-temporal layouts, integrating them with video data through multimodal fusion. Conversely, in \textbf{Chapter \ref{ch:iccv2023}}, object detections are utilized to facilitate multimodal transference.

\textbf{Optical flow} refers to the pattern of apparent motion of objects, surfaces, and edges in a visual scene, caused by the relative motion between the observer and the scene \cite{horn1981determining}, as well as the motion of scene itself; see Figure~\ref{background:fig:part-ii-modalities} (c). Optical flow methods calculate the motion between two frames, taken at times $t$ and $t + \delta t$ at every pixel position. The resulting abstract modality is valuable in understanding the motion characteristics in videos: e.g., the direction and speed of moving objects. We use optical flow as an abstract modality in \textbf{Chapter \ref{ch:iccv2023}} in the process of multimodal knowledge transference.

\textbf{Spectrogram} is a 2D representation of the audio, showing the evolution of the frequency spectrum of the signal over time, denoted as $\mathbf{S} \in \mathbb{R}^{f \times t}$, where $f$ represents the frequency bins and $t$ represents time steps; see Figure~\ref{background:fig:part-ii-modalities} (d). Due to the nature of sound and the human perception of it,
% as well as phenomena associated with the processing of a digital signal (such as spectral leakage), 
the values at adjacent frequencies in the spectrum tend to be associated with the same or other semantically related components of the signal. This results in an additional inductive bias in audio processing, whereby the bins at nearby frequencies tend to be related. These, coupled with the aforementioned temporal dependencies, make image processing models (e.g., Vision Transformers \cite{dosovitskiy2020image}) suitable for spectrograms as well \cite{gong2021ast}.

In \textbf{Chapter \ref{ch:iccv2023}}, we use spectrograms to obtain an audio representation which we then use in the multimodal transference process.

\subsection{Clarifying Derived and Independent modalities}\label{background:sec:derived-independent}
We also distinguish between derived and independent modalities, intending to clarify the relationships and dependencies between different types of data used in machine learning pipelines.

\textbf{Derived Modalities.} Derived modalities are computed or extracted from other primary or raw modalities. These modalities often highlight specific features or properties that are not easily discernible in the raw form, enabling a more nuanced analysis of certain data aspects. For example, a spectrogram is a derived modality from raw audio.
% While the raw audio signal is a one-dimensional waveform representing amplitude over time, the spectrogram provides a two-dimensional representation of the frequency content of the audio signal over time.
The spectrogram reveals important frequency patterns and temporal dynamics relevant to certain audio processing tasks: e.g., speech recognition or event detection.

Similarly, optical flow is a derived modality from video data. It represents the apparent motion of objects between consecutive video frames, highlighting movement patterns and dynamics not directly observable from raw video frames.
% Optical flow captures the direction and velocity of motion, making it a valuable modality for
This makes it valuable for certain video processing tasks, e.g. motion detection or action recognition.

\textbf{Independent Modalities.} Independent modalities, on the other hand, are primary forms of data that are not derived from other modalities but rather exist as distinct sources of information. These modalities contain their inherent structure and properties, and provide unique insights not dependent on other data forms.
% Audio, when considered as raw waveforms, is an independent modality. It inherently contains temporal and frequency-based information that is crucial for tasks related to sound processing and analysis.

Comparing audio and optical flow illustrates the difference between independent and derived modalities. Raw audio signals are independent because they represent original sensory data captured directly from the environment. They are not transformed from another modality, but are a primary source of information with unique temporal characteristics. In contrast, optical flow is a derived modality computed from the sequence of raw video frames. The raw video captures spatial and temporal information about the scene, but optical flow specifically extracts motion information between frames, providing a different perspective important for understanding dynamic activities in the video.

\textbf{Is every abstract modality a derived modality?} No, not every abstract modality is necessarily a derived modality. Abstract modalities are indeed often derived from raw data, however, some can be considered independent. For example, spectrograms are derived from audio, optical flow and object detections are derived from video. These represent derived abstract modalities. On the other hand, abstract scenes, knowledge graphs, and the human body atlas are abstract modalities that are not necessarily derived from other raw modalities and serve as primary sources of information.

\section{Neural Networks}
Inspired by the structure of the human brain, neural networks consist of layers of interconnected artificial neurons. They are used for modeling complex relationships in the data, and their versatility has led to their use on a wide range of tasks across various domains. The parameters associated with the artificial neurons and their interconnections are adjusted during the course of an iterative training process, which typically involves large numbers of training data samples.

In the context of multimodal machine learning, a neural network can be conceptualized as a function $f: \mathbb{R}^n \rightarrow \mathbb{R}^k$, that transforms any modality with an arbitrary dimension ($n$) into a high-dimensional ($k$) representation \cite{goodfellow2016deep}.

\subsection{Feed-Forward Neural Networks}\label{background:sec:neural-networks}
In their simplest form, neural networks (specifically, feed-forward neural networks) perform a weighted sum of their inputs in each layer, followed by a non-linear activation function: $\mathbf{y} = \sigma(\mathbf{W}\mathbf{x} + \mathbf{b})$, where $\mathbf{x}$ is the input modality's vectorized representation, $\mathbf{W}$ represents the weight matrix, $\mathbf{b}$ is the bias vector, and $\sigma$ is the non-linear activation function (e.g., ReLU \cite{nair2010rectified}, GeLU \cite{hendrycks2016gaussian}, etc.). This structure allows the neural network to learn complex, non-linear mappings from any modality (e.g., text, images, audio, etc.) into a compact, latent representation.

\subsection{Training of Neural Networks}\label{background:sec:training-nns}
Training a neural network requires updating the weights $\mathbf{W}$ and biases $\mathbf{b}$ to minimize the discrepancy (i.e., the loss) between the neural network predictions and the actual target values (i.e., the ground truth). The choice of the loss function, which quantifies this discrepancy, mainly depends on the task at hand, which usually is classification, regression, or distribution approximation. In the content below, $N$ represents the number of samples in the dataset, $\mathbf{x}_i$ denotes a vector representation of an input sample (which can technically be of any modality), $\mathbf{y}_i$ is the ground truth target, and $\mathbf{\hat{y}}_i$ is the neural network's prediction for $\mathbf{x}_i$.

\textbf{Classification and Cross-Entropy:}  In classification tasks, where the objective is to predict discrete values, neural networks are usually trained with cross-entropy loss: $\mathcal{L}_{\text{CE}} = - \frac{1}{N}\sum_{i=1}^{N} \mathbf{y}_i \log \mathbf{\hat{y}}_i$, where $\mathbf{y}_i$ is a one-hot vector indicating the position of the true label, and $\mathbf{\hat{y}}_i$ represents probability distribution over the classes.

\textbf{Regression and Mean Squared Error:} In tasks where the objective is to predict continuous values, the Mean Squared Error (MSE) is the standard loss function being used: $\mathcal{L}_{\text{MSE}} = \frac{1}{N} \sum_{i=1}^{N} (\mathbf{y}_i - \mathbf{\hat{y}}_i)^2$, where $\mathbf{y}_i$ and $\hat{\mathbf{y}}_i$ are the vectors of ground truth and predicted continuous values, respectively.

\textbf{Distribution approximation and Kullback-Leibler Divergence:} When the target is a probability distribution, the Kullback-Leibler (KL) divergence is used to measure the discrepancy between the predicted distribution $\mathbf{\hat{y}_i}$ and the ground truth distribution $\mathbf{y}_i$, defined as: $\mathcal{L}_{\text{KL}} = \sum_{i=1}^{N} \mathbf{y}_i \log (\frac{\mathbf{y}_i}{\mathbf{\hat{y}}_i})$. The goal is to minimize $\mathcal{L}_{\text{KL}}$, such that $\mathbf{\hat{y}}_i$ is as close as possible to $\mathbf{y}_i$.

In most cases, optimization algorithms like Stochastic Gradient Descent (SGD) or its variants (e.g., Adam \cite{kingma2014adam}, AdamW \cite{loshchilov2017decoupled}, etc.) are used to update the weights of the neural network in order to minimize the respective loss functions.

\subsection{Modality-specific Neural Networks}\label{background:sec:modality-specific-nns}
Different neural network architectures are designed to exhibit specific inductive biases in order to exploit the unique properties of different data modalities. Here, we discuss how Convolutional Neural Networks (\blackgls{cnn}s), Recurrent Neural Networks (RNNs), and their combinations, are particularly well-suited for specific modality types like images, text, and videos. Finally, we introduce the Transformer as a general-purpose neural network, which exhibits \textit{no} inherent inductive bias, making it perfectly suitable to be adapted to encode \textit{any} modality. These modality representations (i.e., embeddings\footnote{Note that in this manuscript, we interchangeably use the terms ``embedding'', ``feature'', ``feature map'', or ``hidden state'' to refer to a multidimensional modality representation obtained from a neural network.}) can then be used to facilitate various multimodal tasks: e.g., alignment, translation, fusion, and transference among others.

\textbf{Convolutional Neural Networks (\blackgls{cnn}s)} are tailored for images due to their inductive bias towards spatial hierarchy \cite{lecun1989backpropagation}. Each layer in a CNN uses filters to capture local spatial correlations between image pixels, thus making them effective for tasks like image understanding. The spatial inductive bias aligns with the nature of image data---where pixel proximity (typically) implies semantic similarity or relatedness. A convolution operation in a CNN can be denoted as: $\mathbf{h}_i = \sigma(\text{Conv}(\mathbf{x}_i, \mathbf{W}) + \mathbf{b})$, where $\mathbf{x}_i$ is the input image or feature map from the previous layer, $\mathbf{h}_i$ represents the hidden state (i.e., output feature map), $\text{Conv}$ denotes the convolution operation, $\mathbf{W}$ is a kernel (i.e., filter applied to the input via the convolution), $\mathbf{b}$ is the bias term, and $\sigma$ is a non-linear activation function.

\textbf{Recurrent Neural Networks (\blackgls{rnn}s)} are effective for text due to their sequential inductive bias. RNNs process sequential data while maintaining a ``memory'' (hidden state) of what has been processed so far, making them suitable for handling the sequential nature of text, where the understanding of each word (or character) depends on its predecessors \cite{jozefowicz2016exploring, bengio2000neural}. An RNN's operation can be denoted as: $\mathbf{h}^t_i = \sigma(\mathbf{W}[\mathbf{h}^{t-1}_i, \mathbf{x}^t_i] + \mathbf{b})$, where $\mathbf{h}^t_i$ is the hidden state for the current time step $t$, $\mathbf{x}^t_i$ is the input for the current time step $t$, and $\mathbf{h}^{t-1}_i$ is the hidden state from the previous time step $t-1$.

\textbf{Spatial and Sequential Neural Networks.} While images and text have the inductive bias of spatial hierarchy and temporal connectivity respectively, videos present a more complex scenario, blending both the spatial and the temporal dimensions. Therefore, to process videos, neural network models which can encode both the spatial and temporal properties of the videos---such as \blackgls{cnn}s combined with \blackgls{rnn}s, or \blackgls{3d-cnn}s---are effective. In this case, the CNN extracts spatial features from the individual frames, while the RNN models the temporal dynamics between them. Alternatively, \blackgls{3d-cnn}s extend the convolution operation to three dimensions (height, width, and time), directly capturing both spatial and temporal correlations.

\section{Transformers: General-Purpose Encoders}\label{background:sec:transformers}
\begin{figure}[t]
\centering
\includegraphics[width=0.5\textwidth]{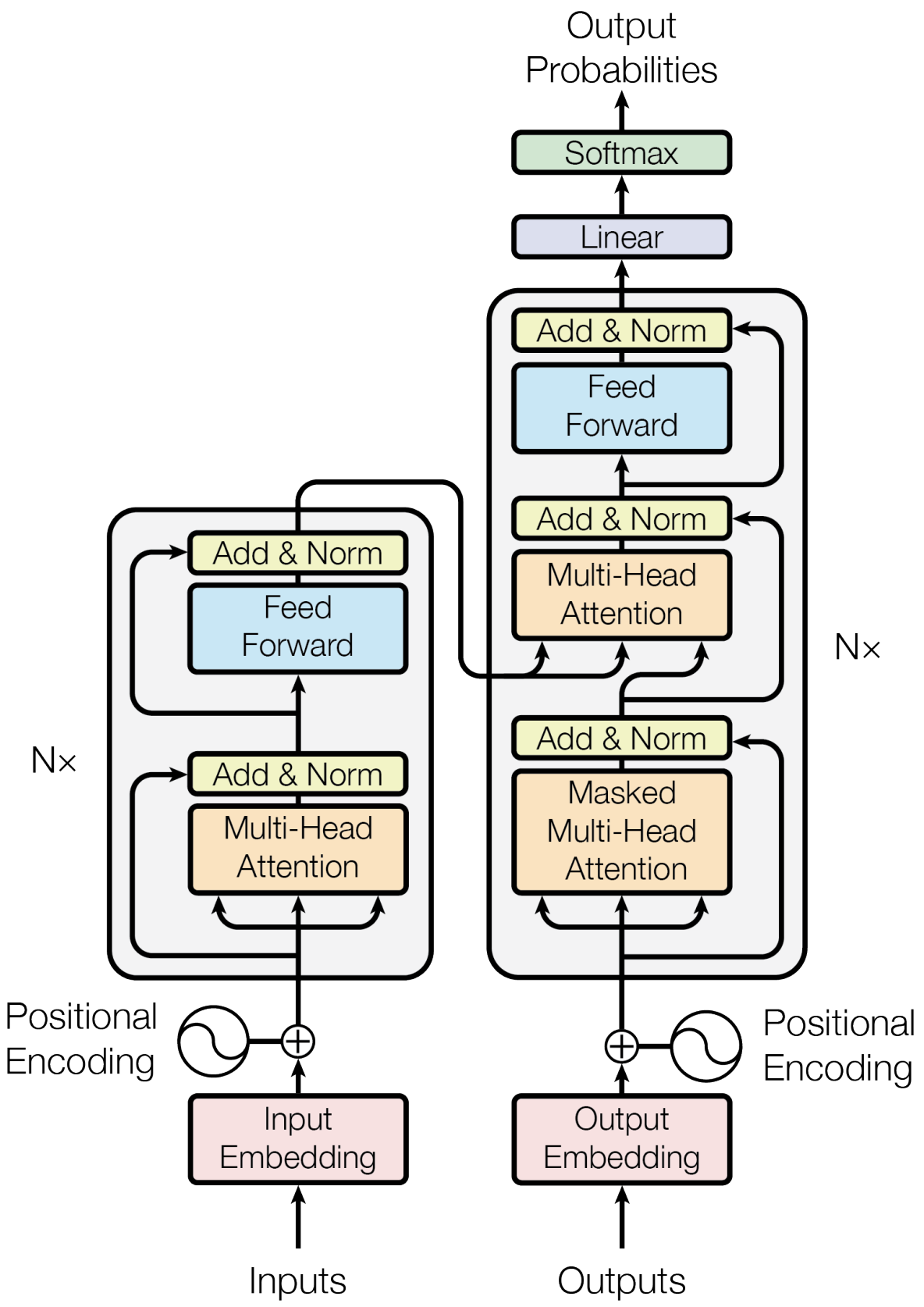}
\caption[The Transformer model architecture]{The Transformer encoder-decoder model architecture. Image source: Vaswani~\etal~\cite{vaswani2017attention}. It is important to note that within the context of this manuscript, and the broader literature, references to the Transformer model typically pertain to the Transformer encoder only.}
\label{background:fig:transformer}
\end{figure}

% Unlike CNNs and RNNs, which are tailored to specific inductive biases like spatial and sequential processing, the Transformers (see Figure~\ref{background:fig:transformer}) are inherently free of any inductive bias \cite{vaswani2017attention}. This makes them adaptable for encoding a wide range of data modalities.
Unlike \blackgls{cnn}s and \blackgls{rnn}s, which are tailored to specific inductive biases for spatial and sequential processing, respectively, the Transformer (see Figure~\ref{background:fig:transformer}), as introduced by Vaswani~\etal~\cite{vaswani2017attention}, exhibits a broader and more adaptable set of biases. Specifically, through the self-attention mechanism and the positional encoding (details follow in the next paragraph), the Transformer is capable of modeling direct relationships between the input elements regardless of their position. This adaptability allows Transformers to encode a wide range of data modalities effectively, without being inherently restricted to any particular data structure.

% The main building block of the Transformer architecture is the self-attention mechanism: $\mathbf{h} = \text{Attention}(Q, K, V) = \text{softmax}\left(\frac{QK^T}{\sqrt{d_k}}\right)V$, where $\mathbf{Q}$, $\mathbf{K}$, and $\mathbf{V}$ represent the queries, keys, and values matrices, respectively, which are all derived from the input data. The term $d_k$ denotes the data dimensionality, and $\mathbf{h}$ signifies the hidden representation post-self-attention. Each of these matrices $\mathbf{Q}$, $\mathbf{K}$, $\mathbf{V}$ are learned transformations of the input elements, for instance, $\mathbf{Q} = \text{Linear}(\mathbf{x}_j)$, where $\mathbf{x}_j$ is the vector representation of the $j$-th element in the input modality (e.g., a token or patch embedding). This mechanism enables dynamic contextualization of each input element relative to the whole sequence.
The main building block of the Transformer architecture is the self-attention mechanism, defined as $\mathbf{h}^j_i = \text{Attention}(\mathbf{Q}, \mathbf{K}, \mathbf{V}) = \text{softmax}\left(\frac{\mathbf{Q}\mathbf{K}^T}{\sqrt{d_k}}\right)\mathbf{V}$, where $\mathbf{Q}$, $\mathbf{K}$, and $\mathbf{V}$ denote the query, key, and value matrices, respectively, derived from the input. Here, $d_k$ represents the dimensionality of the keys, ensuring that the softmax function's argument is scaled appropriately, and $\mathbf{h}$ is the resulting hidden representation after applying self-attention. The matrices $\mathbf{Q}$, $\mathbf{K}$, and $\mathbf{V}$ are produced through learned linear transformations of the input: $\mathbf{Q} = \text{Linear}(\mathbf{x}^j_i)$, with $\mathbf{x}^i_j$ being the vector representation of the $j$-th element of the input $\mathbf{x}_i$ (e.g., token or an image patch embedding). This mechanism enables dynamic contextualization of each input element in relation to the entire sequence, enabling the model to adjust the attention focus based on the context.

% In the purest form, the most important aspect of the Transformer is its positional invariance, an attribute indicating the absence of an inherent inductive bias. In practical terms, this indicates that the output of the self-attention mechanism does not change even if the positions of input elements are altered. To adapt the Transformer to a specific modality, we inject the appropriate inductive bias. For example, for sequential data such as text, sequential positional information is injected to the inputs of the Transformer model. For images, spatial positional encoding is injected to each image patch, forcing the Transformer to learn local features. For video data, a combination of spatial encoding for individual frame patches and temporal encoding across frames is utilized.
A key characteristic of the vanilla Transformer\footnote{Vanilla Transformer: Indicating that the input embeddings are provided to the Transformer as is, without any information of their position and structure (e.g., sequential, spatial, etc.).} is its ability to handle inputs without a predefined notion of positional relationships. This attribute highlights the lack of inherent spatial or sequential inductive bias, different from prior model architectures like \blackgls{cnn}s and \blackgls{rnn}s. In practice, the Transformer’s self-attention mechanism, by itself, treats all inputs equally regardless of how they are structured. To adapt the Transformer for a specific modality, it is necessary to introduce an appropriate form of inductive bias. For example, in the case of textual data, sequential positional information is added to the Transformer’s inputs, enabling it to respect the order of words (or tokens). Similarly, for processing images, spatial positional encodings are added to the individual image patches.
% while for videos, both spatial and temporal positional encoding are added to each of the video frame patches.
This encourages the Transformer to consider the spatial arrangement of patches, thereby learning spatially relevant features. For video data, a combination of spatial encoding for individual frame patches and sequential (i.e., temporal) encoding across frames is utilized.

This inherent flexibility and adaptability make the Transformer a versatile choice for \textit{any} modality. In this manuscript, we leverage the Transformer architecture as our primary model across various modalities, demonstrating its efficacy and general applicability in diverse contexts.

\subsubsection{The Role of Transformers in Machine Learning}
The question of whether Transformers alone suffice in Machine Learning hinges on several factors. With sufficient (pre-)training data available, a general-purpose learning algorithm like the Transformer can capture relevant inductive biases from the data itself. In such scenarios, where the data is both sufficient in quantity and diverse, relying on learned features often leads to superior generalization. However, in scenarios where (pre-)training data is scarce, the flexibility of Transformers may become a double-edged sword, potentially leading to overfitting. In such cases, methods that explicitly encode the data properties---e.g., Convolutional Neural Networks (\blackgls{cnn}s)---can offer more robust generalization.
% \subsubsection{The Role of Transformers in Machine Learning?}
% Depends. Should we have enough (pre-)training data, a general-purpose learning algorithm (e.g., a Transformer) would pick up the (relevant) inductive biases from the data, which if sufficient, is more likely that the algorithm generalizes better than imposing hand-crafted inductive biases. On the other hand, such flexibility could be undesirable if we have a lack of (pre-)training data; i.e., it is more likely that we overfit the data we have. In those cases, a method that explicitly encodes the properties of the data we consider relevant (e.g., \blackgls{cnn}) is more likely to generalize better.

\section{Transfer Learning}\label{background:sec:transfer}
Transfer learning entails applying knowledge acquired from solving one task to a different, albeit related, downstream task. Such a method becomes especially significant when labeled data for the downstream task is limited, enabling the utilization of models pre-trained on large datasets to enhance performance on comparatively smaller datasets \cite{bengio2000neural}. The transfer learning process usually consists of two phases: a pre-training phase subsequently followed by a fine-tuning phase. During pre-training, the model is trained using a large and diverse dataset, facilitating the learning of a wide spectrum of features or patterns. In the fine-tuning phase, this pre-trained model is tailored to a specific, often narrower downstream task or dataset.

\subsection{Transformers in Transfer Learning}\label{background:sec:transformers-transfer}
Transformers are exceptionally suited for transfer learning, which can be attributed to their:
\begin{inparaenum}[(i)]
    \item high parallelizability, allowing efficient processing of large-scale data;
    \item capability to model long-range dependencies, essential for understanding complex patterns; and
    \item inherent versatility to adapt to different data modalities.
\end{inparaenum}
Due to these attributes, the different variants of the Transformer model have demonstrated remarkable success when pre-trained on large-scale and general-purpose datasets \cite{devlin-etal-2019-bert, liu2021swin, dosovitskiy2020image}. After pre-training, these models are fine-tuned for various specific tasks with minimal task-specific data, effectively adapting their learned representations to new contexts.

\textbf{Transfer Learning in \blackgls{nlp} with Transformers.} In the domain of NLP, transformers are typically pre-trained using self-supervised methods like Masked Language Modeling (\blackgls{mlm}) \cite{devlin-etal-2019-bert}. In \blackgls{mlm}, the input token sequence is partially corrupted, with certain tokens replaced by a special \texttt{[MASK]} token. The model then predicts the original, uncorrupted tokens. This pre-training task enables the model to learn the underlying language structure to a large extent. In \textbf{Part \ref{part:alignment-translation}} of this manuscript, we leverage such transformers pre-trained with \blackgls{mlm}---\blackgls{bert} \cite{devlin-etal-2019-bert} or \blackgls{roberta} \cite{liu2019roberta}---across our experiments.

\textbf{Transfer Learning in Computer Vision with Transformers.} 
In computer vision, transformers have shown promising results through supervised pre-training \cite{he2016deep} on large-scale datasets like ImageNet \cite{deng2009imagenet} or via self-supervised methods \cite{chen2021exploring, lebailly2023adaptive, lebailly2023cribo, lebailly2023global, stegmuller2023croc}. In the context of video processing, transformers are often pre-trained in a fully-supervised way on large-scale video datasets like Kinetics \cite{kay2017kinetics}, or through adaptations of \blackgls{mlm} tailored for videos---\blackgls{videomae} \cite{tong2022videomae}. In most of the experiments we conduct in \textbf{Part \ref{part:fusion-transference}} of this manuscript, we utilize a Transformer pre-trained in a fully-supervised fashion on Kinetics---Video Swin \cite{liu2021swin}.

\section{Multimodal Alignment and Translation}\label{background:sec:alignment-translation}
In this section, we lay out the fundamentals on which we build in \textbf{Part \ref{part:alignment-translation}}, which addresses the problems of multimodal alignment and translation. While the literature on these two problems is extensive, here we only focus on the parts relevant for this manuscript.

\subsection{Alignment via Contrastive Learning}\label{background:sec:contrastive}
Multimodal alignment involves establishing connections between the elements of two or more different modalities. This process is often facilitated by contrastive learning: a method used to learn representations by contrasting positive pairs (semantically similar samples, such as those belonging to the same class) against negative pairs (semantically unrelated samples, such as those belonging to different classes) \cite{oord2018representation, hermans2017defense}.

A common approach in contrastive learning for multimodal alignment is the use of the InfoNCE loss \cite{oord2018representation}. This loss function is designed to bring representations of similar (positive) pairs closer while pushing dissimilar (negative) pairs further apart in the embedding (i.e., representation) space. Formally, the InfoNCE loss for a batch of $N$ positive pairs, $i$ and $j$, is defined as:
% $\mathcal{L}_{\text{InfoNCE}} = -\sum_{i=1}^{N} \log \frac{\exp(\text{sim}(\mathbf{x}_i, \mathbf{y}_i)/\tau)}{\exp(\text{sim}(\mathbf{x}_i, \mathbf{y}_i)/\tau) + \sum_{j=1}^{N-1} \exp(\text{sim}(\mathbf{x}_i, \mathbf{y}_j)/\tau)}$,

\begin{equation*}
    \mathcal{L}_{\text{InfoNCE}} = -\sum_{i=1}^{N} \log \frac{\exp(\text{sim}(\mathbf{x}_i, \mathbf{y}_i)/\tau)}{\exp(\text{sim}(\mathbf{x}_i, \mathbf{y}_i)/\tau) + \sum_{j=1, j \neq i}^{N} \exp(\text{sim}(\mathbf{x}_i, \mathbf{y}_j)/\tau)},
\end{equation*}

where $\text{sim}(\mathbf{x}, \mathbf{y})$ denotes the similarity (e.g., cosine similarity) between the representations of the modalities $\mathbf{x}$ and $\mathbf{y}$, and $\tau$ is a temperature parameter that scales the similarity scores.

\subsection{Translation via Grounding}
Multimodal translation
% , distinct from multimodal alignment,
involves the transformation of the representation from one modality to another modality, ensuring the preservation of information content. Grounding is a crucial process in this context, where abstract concepts from one modality (e.g., text) are associated with specific, identifiable elements in another: e.g., 2D scenes \cite{radevski2020decoding} (\textbf{Chapter~\ref{ch:emnlp2020}}), 3D locations in a human atlas \cite{grujicic-etal-2020-learning} (\textbf{Chapter~\ref{ch:conll2020}}), or facts in a knowledge graph \cite{radevski-etal-2023-linking} (\textbf{Chapter~\ref{ch:emnlp2023}}).

% In this manuscript, we treat translation as a special case of one-directional alignment. Unlike bi-directional alignment, which establishes a mutual connection between modalities, translation focuses on a unidirectional transformation of the representations.

When translating a source modality into a target modality that can be discretely represented (i.e., each node in the knowledge graph is a canonical instance), we train models using the standard cross-entropy loss (\S\ref{background:sec:training-nns}). However, when representing the entire target modality is impractical (e.g., grounding to large-scale knowledge graphs with millions of entities and predicates), we adopt a contrastive training setup. In this setup, we sample a fixed set of negative elements, as employed in \textbf{Chapter \ref{ch:emnlp2023}}.

If the target modality cannot be segmented unambiguously, we use a variant of the regression loss (\S\ref{background:sec:training-nns}) to train the models. This approach is particularly relevant for modalities with ambiguous segmentation, such as 2D scenes of objects or locations, or a 3D human atlas, as discussed in \textbf{Chapter~\ref{ch:emnlp2020}} and \textbf{Chapter~\ref{ch:conll2020}}.

\section{Multimodal Fusion and Transference}\label{background:sec:fusion-transference}
In this section, we provide the fundamentals on which we build in \textbf{Part \ref{part:fusion-transference}}. First, we briefly introduce cross-attention as a method that facilitates modeling cross-modal interactions between a pair of modalities. Next, we discuss how we can achieve knowledge transfer between a pair of neural networks which may differ in complexity or encode different modalities.
% We discuss three basic strategies for integrating information from different modalities (i.e., multimodal fusion strategies that we employ in our methods, or as baselines), as well as how knowledge can be transferred across them during training.

\subsection{Cross-Attention for Cross-Modal Interactions}
Cross-attention, as a method, can enable fine-grained interaction between two modality representations exhibiting different structures.
% Assuming we are given two modality representations, initially processing each modality independently, the representations are then combined using cross-attention \cite{tan2019lxmert, lu2019vilbert, chen2019uniter}.
Assuming we are given with hidden representations from modality-specific encoders for two modalities $\mathbf{H}_1$ and $\mathbf{H}_2$, the cross-attention mechanism can be represented as: $\mathbf{C}_{1 \rightarrow 2} = \text{Attention}(Q(\mathbf{H}_1), K(\mathbf{H}_2), V(\mathbf{H}_2))$ and $\mathbf{C}_{2 \rightarrow 1} = \text{Attention}(Q(\mathbf{H}_2), K(\mathbf{H}_1), V(\mathbf{H}_1))$. Here, $\mathbf{C}_{1 \rightarrow 2}$ represents the multimodal representation where the first modality attends to the second, and $\mathbf{C}_{2 \rightarrow 1}$ represents the multimodal representation for the reverse process. The attention function, as defined in \S\ref{background:sec:transformers}, allows each element in the first modality to interact with and be contextualized by all other elements in the second modality.

The cross-attention-based allows for context-sensitive integration of modalities, leading to a more nuanced multimodal representation, especially when the interplay between the modalities is complex (e.g., text and image representations \cite{tan2019lxmert, lu2019vilbert, chen2019uniter}).

\subsection{Transfering Knowledge Across Modalities}
Multimodal knowledge transference is the process of transferring knowledge across different modality-specific neural networks and their respective output representations. Within this manuscript, we focus on a subset of transference known as multimodal co-learning \cite{baltruvsaitis2018multimodal, liang2022foundations}. This approach aims to transfer knowledge from secondary modalities to a primary modality, which is later used during inference, to enhance the model's performance beyond what would be achieved by training solely on the primary modality. In \textbf{Chapter~\ref{ch:bmvc2021}} and \textbf{Chapter~\ref{ch:iccv2023}}, we explore the significant benefits of transferring knowledge across multiple modalities in comparison to the process of multimodal fusion.
% A key advantage is that, since transference occurs during training, multimodal data is not required at inference time, alleviating the constraints seen in multimodal fusion.

\textbf{Knowledge Distillation} is a prevalent method for transferring knowledge between two neural networks, typically from a larger and complex model (teacher), to a smaller and simpler model (student). The student model is typically trained to approximate the output probabilities of the teacher model, employing the KL-divergence loss. The KL-divergence loss for knowledge distillation is given by: $\mathcal{L}_{\text{KD}} = \sum_{i} \text{KL}\left(P_{\text{teacher}}(\mathbf{y}|\mathbf{x}_i) || P_{\text{student}}(\mathbf{y}|\mathbf{x}_i)\right)$, where for input $\mathbf{x}_i$, $P_{\text{teacher}}(\mathbf{y}|\mathbf{x}_i)$ and $P_{\text{student}}(\mathbf{y}|\mathbf{x}_i)$ are the output probability distributions of the teacher and student models, respectively.

% \textbf{Omnivorous Models} are jointly trained on multiple modalities, such as images, videos, and 3D data, leveraging the diverse nature of these data types for comprehensive learning \cite{girdhar2022omnimae, girdhar2022omnivore}. The fundamental principle behind omnivorous models is to facilitate multimodal transference, resulting in a model that not only understands multiple modalities but also exhibits enhanced performance on the primary modality. Unlike traditional approaches, omnivorous training does not necessitate parallel data; modalities can originate from different datasets and involve different tasks. The training objective for omnivorous models can be represented as:
% $\mathcal{L}_{\text{Omnivorous}} = \sum_{\text{modality } m} \mathcal{L}_m(\text{data}_m)$ where $\mathcal{L}_m$ is the loss function specific to modality $m$, and $\text{data}_m$ represents the data corresponding to that modality.

\section{(Egocentric) Video Action Recognition Datasets}\label{background:sec:videodata}
In this section, we lay out the background information about the video action recognition datasets we use in \textbf{Part~\ref{part:fusion-transference}} of this manuscript. Lastly, in \S\ref{background:sec:egovision-properties}, we discuss the distinctive properties of egocentric vision.

\subsubsection{Something-Something} 
The Something-Something \cite{goyal2017something} dataset consists of
% (mainly) egocentric
videos of people performing 174 unique object-agnostic actions with their hands: e.g. ``pushing [something] left'', ``taking [something] out of [something]''. 
% While the dataset also contains exocentric videos, most of the sequences describe similar hand-object interactions, with the main distinction being the orientation of the hands.
Notice that the action classes do not account for specific objects, but rather, only for the activity itself. Therefore, on Something-Something, there is an increased focus on capturing temporal relationships that characterize the actions. The training and validation set contains $\sim$169k and $\sim$26k videos respectively. Furthermore, to deal with environmental bias---models relying on unrelated environmental cues to discriminate between the actions---videos recorded by the same participant can be in either the training or validation set. Nevertheless, the objects the participants interact with---even though unrelated to the action label---can appear in both the training and the testing data, indicating that models that observe the videos through the RGB modality (i.e., the video appearance) can overfit on the visual appearance of the objects.

% \textbf{Something-Else} \cite{materzynska2020something} proposes two data splits according to the distribution of the objects at training and test time. In the compositional split, to validate the compositional generalization, the data is divided such that the models encounter a novel set of objects during training and testing. The training and validation sets contain $\sim$55k and $\sim$58k videos respectively, with 174 actions. In the few-shot split, there are $\sim$112k pre-training videos (with 88 base actions), 5 $\times$ 86 and 10 $\times$ 86 videos in the 5-shot and 10-shot setup respectively for fine-tuning, and $\sim$49k testing videos (with 86 novel actions). On the compositional and few-shot splits, we perform experiments with Faster R-CNN \cite{ren2015faster} object detections (object predictions setting), and with ground truth object detections (oracle setting), both released by \cite{materzynska2020something}. The input object categories in STLT are either ``hand'' or ``object''. We measure performance using top-1 and top-5 accuracy and train all models with the cross-entropy loss.

\subsubsection{Something-Else}
In Something-Something, the objects present in the scene (i.e., where the action takes place) may appear in both the training and the testing data. The goal of Something-Else \cite{materzynska2020something} is to deal with the issue of models exploiting visual cues related to the objects' appearance in order to discriminate between the actions. Materzynska~\etal~\cite{materzynska2020something}, propose two data splits according to the objects’ distribution at training and test time. The data is divided such that the models encounter distinct objects during training and testing. In the first, compositional generalization split, the training and validation set contains $\sim$55k and $\sim$58k videos respectively, with 174 action categories. This data split is explicitly aimed at testing the compositional generalization of the models. In the few-shot split, there are $\sim$112k pre-training videos (with 88 base actions), 5 $\times$ 86 and 10 $\times$ 86 videos in the 5-shot and 10-shot setup respectively for fine-tuning, and $\sim$49k testing videos (with 86 novel actions).
% Furthermore, Materzynska~\etal~\cite{materzynska2020something} and a subsequent work of ours \cite{radevski2021revisiting} demonstrate that a standard RGB-based model \cite{carreira2017quo} exhibits significantly lower performance on the Something-Else split compared to the standard Something-Something split. To improve the generalization ability of standard RGB-based models, the work of Materzynska~\etal~\cite{materzynska2020something}, as well as subsequent works \cite{radevski2021revisiting, herzig2022object}, propose using object detections as input to the model \cite{ren2015faster}, as they are agnostic to the appearance of individual objects.

\subsubsection{Action Genome}
The Action Genome dataset \cite{ji2020action} is built on top of Charades \cite{sigurdsson2016hollywood}, and has $\sim$10k videos of people doing daily activities. Multiple object-specific actions simultaneously occur in each video out of 157 unique ones. Out of these 157 actions, 144 are human-object activities: e.g., ``carrying a blanket'', ``eating a sandwich'', etc. The frames where the action occurs (i.e., the person interacts with the objects), are annotated with bounding boxes and categories.

Importantly, in the Action Genome dataset, the majority of the objects present in the video are unrelated to the action (i.e., the person does not interact with these objects). Therefore, such background clutter increases the complexity of the action recognition task, where the models need to isolate the objects the person interacts with (i.e., the ones specific to the action).

\subsubsection{Epic-Kitchens}
EPIC-Kitchens is a large-scale benchmark dataset consisting of 700 videos recorded by 32 participants \cite{damen2018scaling, damen2020rescaling}, totaling 100 hours of egocentric videos capturing daily activities in kitchen environments.
% In our experiments, we use the annotations for the action recognition task.
It features actions that can be broken down into the noun (the active object participating in the action: e.g. ``carrot'', ``pan'', etc.) and the verb (the activity itself: e.g. ``cutting'', ``washing''). In total, there are 331 noun and 125 verb categories, while the training and validation set contains $\sim$68k and $\sim$10k videos respectively. The modalities we use in our experiments in \textbf{Part~\ref{part:fusion-transference}} include the RGB frames, the audio, and the optical flow. We use the audio directly from the mp4 files and the optical flow as released by Damen~\etal~\cite{damen2020rescaling}.

\subsubsection{Epic-Kitchens Unseen Participants}\label{iccv2023:appendix:datasets_epic_unseen}
This particular subset of the Epic-Kitchens validation split contains 1065 action sequences from two participants that were not observed in the training dataset (i.e., videos recorded by them are not included in the training data). Similar to the Something-Else dataset, we use this data split to more explicitly measure the compositional generalization performance of the models. Specifically, standard RGB models often incorporate undesirable biases to discriminate between different actions, such as objects or environmental cues unrelated to the action. Using the Epic-Kitchens Unseen Participants split, we verify the extent to which video understanding models are robust w.r.t. this type of distribution shift.

\subsection{Properties of Egocentric Vision}\label{background:sec:egovision-properties}
While both the Something-Something dataset and the Epic-Kitchens dataset consist of videos of people performing actions with their hands, it does not imply that both datasets are egocentric. Namely, in the Something-Something dataset, the person performing the action is holding the camera with their hands, while in the Epic-Kitchens dataset, the person performing the action has a head-mounted camera. This creates an important distinction between these data types.

Firstly, in the case of Something-Something, the camera's viewpoint does not directly align with the wearer's line of sight. Even though the wearers' hand(s) and the objects they manipulate are visible, the camera movement might be detached from the natural head movements. This results in variable spatial relations, where the spatial relationship between the camera and the objects might be unstable; i.e., the camera movement depends on how the wearer holds and maneuvers it, leading to potential variability in the perspective of the actions and objects.

In the case of Epic-Kitchens, the camera is mounted on the wearer's head---capturing a first-person perspective---where the viewpoint aligns with the wearers' line of sight. The wearer's hands and the objects they manipulate are featured in the frame, reflecting a first-person viewpoint as they perform tasks. The spatial relationship between the camera and the objects remains consistent with the wearer's viewpoint; i.e., as the wearer moves their head, the perspective on the objects and actions remains relatively stable.

\cleardoublepage%

% Insert here your own chapters
% Chapters are expected to be in a tex-file with the given name dot
% tex and in a directory with the given name in the chapters
% directory.

\part{Multimodal Alignment and Translation}\label{part:alignment-translation}
% \includechapter{preamble_part_i}
%
  \graphicspath{{\chaptersdir/emnlp2020/image/}}%
  %\cleardoublepage%
  \chapter{Translating Text-Based Spatial Relations to 2D Spatial Arrangements}\label{ch:emnlp2020}
In this chapter, we focus on the problem of multimodal translation from a set of language-expressed spatial relations (given by scene descriptions), to a set of 2D spatial arrangements (given by clip-arts arranged on a 2D canvas) in a multi-object and multi-relationship setting. The goal of such a setup is to study the problem of multimodal spatial understanding in a controlled environment, where we frame the task as arranging a scene of clip-arts given a textual description. 

To study this problem, we propose a simple and effective model architecture \textsc{Spatial-Reasoning Bert} (\blackgls{sr-bert}), trained to decode, or translate, text to 2D spatial arrangements in a non-autoregressive manner. We find that \blackgls{sr-bert} can understand and translate both explicit and implicit natural language to 2D spatial arrangements, can generalize to out-of-sample data to a reasonable extent, and can generate complete abstract scenes if paired with a clip-art predictor. Finally, we evaluate the quality of the generated abstract scenes with a user study, validating that our generated spatial arrangements between the clip-arts align with human expectations.

The content of this chapter is based on the following publication:

\begin{mdframed}[backgroundcolor=gray!30,linewidth=0pt]
\underline{Radevski, G.}, Collell, G., Moens, M-F., and Tuytelaars, T., (2020). Decoding Language Spatial Relations to 2D Spatial Arrangements. Findings of The Conference on Empirical Methods in Natural Language Processing (EMNLP). \cite{radevski2020decoding}
\end{mdframed}

\section{Introduction}\label{emnlp2020:sec:introduction}
\begin{figure}[t]
\centering
\includegraphics[width=0.9\textwidth]{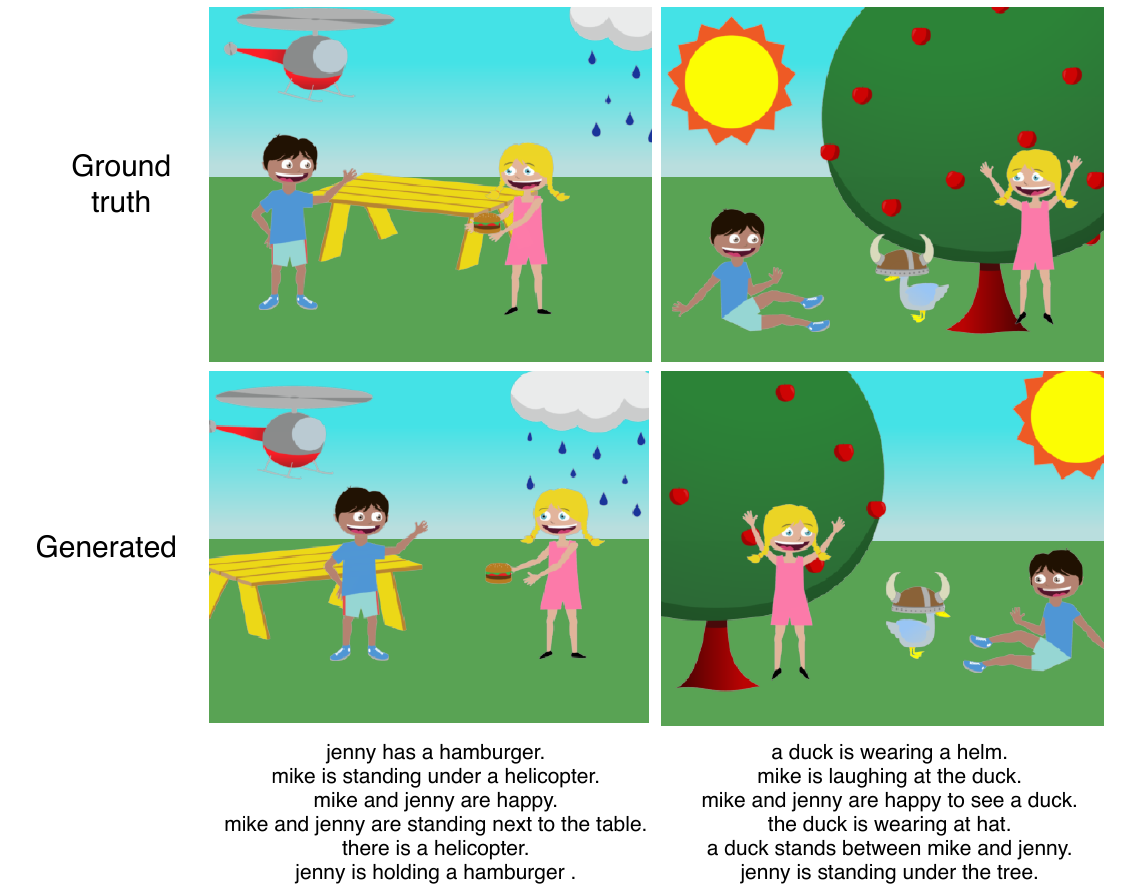}
\caption[Generated scene by \textsc{Sr-Bert}]{Given a set of clip-arts and a textual description of a scene, including both implicit and explicit spatial relations, our method automatically generates a spatial arrangement of the clip-arts that satisfies the spatial constraints.}
\label{emnlp2020:fig:generated_scenes}
\end{figure}

Multimodal spatial understanding is a problem of paramount importance to both the vision and the language community. For a machine learning model to be able to reason about the spatial domain w.r.t. another modality (e.g., language), it should incorporate common-sense spatial knowledge, which is often obvious to humans, yet hard to grasp by machines. If the spatial relations are expressed in language, such common sense knowledge can be implicit: e.g., ``Mike and Jenny see a duck'' (Figure \ref{emnlp2020:fig:generated_scenes}), implies that both ``Mike'' and ``Jenny'' should be facing the ``duck'', even though this is not explicitly stated. The complexity of the problem increases tremendously when there is no a priori-defined limit on the number/type of objects and relationships---the de facto setting in practical applications like computer graphics, video games, 3D modeling, etc. In such domains, it would be intuitive for the designers of content to communicate spatial relations via language, while arranging objects in a multidimensional space remains to be tedious and labor-intensive. For that reason, building models that exhibit spatial understanding is a key step towards automation: aiding humans in repetitive time-consuming tasks.

Up to 2020 (when this work came to fruition), multiple methods had explicitly investigated the spatial reasoning ability of machine learning models in a multidimensional space. However, at that time, the main limitations of these methods were:
\begin{inparaenum}[(i)]
    \item scene environments with strong priors on (relative) object placements (e.g., indoor home environments) \cite{chang2017sceneseer, fisher2012example, choi2013understanding, xu2013sketch2scene, jiang2012learning, chang2014learning, kermani2016learning};
    \item modeling only pairwise relationships (i.e., two objects and a single relationship) \cite{dan2020understanding, collell2018acquiring};
    \item not using natural language descriptions of scenes, but only structured language \cite{collell2018acquiring, dan2020spatial}.
\end{inparaenum}

To address the aforementioned limitations, in this chapter, we attempt to gain insights into the following research question:

\researchquestion{Research Question 1}{How effectively can we translate language-expressed spatial relations into 2D spatial arrangements between objects without constraints on their number and type?}

We further delve into the complexities of the spatial language by distinguishing explicit spatial relations (e.g., ``above'', ``left-of'') from implicit ones (e.g., ``wearing'', ``holding''). This exploration is important for enhancing the models' versatility in real-world applications, such as robotics, where commands can be either direct (i.e., explicit) to interpretive (i.e., implicit). Hence, we pose the following secondary research question:

\subresearchquestion{Research Question 1.1}{Does the nature of spatial language relations (explicit like ``above'', ``left-of'' or implicit like ``wearing'', ``holding'') significantly influence the correctness of the translation between the language and the 2D spatial arrangements?}

We next examine the models' robustness w.r.t. out-of-distribution (OOD) spatial relations not encountered during training. The ability to generalize to unseen data is critical for real-world applications where models often face novel descriptions. Thus, pose the following research question:

\subresearchquestion{Research Question 1.2}{How robust is the translation model we learn in dealing with spatial relations which are out-of-distribution (i.e., not encountered during training)?}

% We formally frame our research problem as: \textit{Given a set of discretely encoded clip-arts (people, objects, etc.), and a textual description of a scene, what is the best positioning of the clip-arts that corresponds to the spatial relations implied by the text?}.
\textbf{Contributions.} In response to these research questions, our contributions are as follows:

\circled{1} We propose \blackgls{sr-bert}, a method developed to jointly process multi-modal data with distinct positional encoding for each modality, which directly focuses on the spatial relations---decoding the language cues (both implicit and explicit spatial relations)---to 2D spatial arrangements.
% we propose a model---based on \blackgls{bert} \cite{devlin2018bert}, adapted to jointly process multi-modal data with distinct positional encoding. 

% achieved by masking the information related to the spatial arrangements during training.
\circled{2} We build on top of methods initially proposed for non-autoregressive text decoding \cite{ghazvininejad2019constant, kasai2020parallel, wang2019bert, lee2018deterministic, lawrence2019attending}, which, in our case, we re-purpose to iteratively mask-out and predict the spatial arrangements of the objects of interest.
% \circled{3} Inspired by Lawrence~\etal~\cite{lawrence2019attending} we develop distinct ways of imposing a decoding order, tailored for generating spatial arrangements from language.

\circled{3} We develop a complete abstract scene generation pipeline---predicting the clip-arts that appear in the scene and arranging them on the scene canvas---which outperforms competitive baselines on the same task. 

% We perform ad-hoc experiments to gain insights into three main research questions:
% \begin{inparaenum}[(RQ1)]
%     \item \label{emnlp2020:question:1} Can we decode a set of language spatial relations to the 2D space without imposing constraints on the number and type of objects and relationships?
%     \item \label{emnlp2020:question:2} 
%     Does the model merely exploit dataset bias to generate arrangements or does it acquire an understanding of the language and the spatial domain?
%     \item \label{emnlp2020:question:3} Is the model able to interpret only explicit spatial relationships (e.g., on, above) 
%     or can it cope with implicit ones as well (e.g., wearing, eating, etc.)?
% \end{inparaenum}
We release the code, data, and trained models at {\url{https://github.com/gorjanradevski/sr-bert}}.
\section{Related Work}\label{emnlp2020:sec:related-work}
Since this work came to fruition in early 2020, in this section we discuss \textit{only} the related and concurrent works on spatial language understanding at the time of publication.

\textbf{Spatial understanding in vision} is mainly limited to generating scenes that have strong priors on the relative locations of the objects (e.g., indoor home environments). The model proposed by Choi~\etal~\cite{choi2013understanding}, encodes the object interactions in 3D and performs object detection, layout estimation, and scene classification, limited to images of indoor home environments. Xu~\etal~\cite{xu2013sketch2scene}'s method maps indoor scene sketches in 3D space. Similar to our approach, they do not limit the number of objects. Kermani~\etal~\cite{kermani2016learning} synthesize 3D indoor home environments from RGB-D. Fisher~\etal~\cite{fisher2012example} also address the problem of generating spatial arrangements using 3D indoor home environments and propose a probabilistic model that can sample diverse 3D scenes.  The work of Jiang~\etal~\cite{jiang2012learning} explores human-object relationships in 3D indoor scenes. Given a scene, Zhao~\etal~\cite{zhao2016relationship} synthesize new scenes that preserve the nature of the original spatial relations by replacing the objects. On the other hand, our work is not limited in terms of the types of relationships, i.e., explores human-object, object-object, and human-human relationships in clip-arts scenes.

\textbf{Spatial understanding in language} often suffers from processing structured language or again, using scenes that have strong priors on the relative object's locations. Chang~\etal~\cite{chang2014learning} infer spatial relationships not explicitly stated in natural language and generate 3D indoor home environments. Contrary to our work, the 3D indoor home environments are limited in terms of implicit spatial language (e.g., wearing, holding, etc.). Chang~\etal~\cite{chang2017sceneseer} decode language spatial relations to 3D indoor home environment layouts, by first selecting the objects and then arranging them.
% However, we loosen the object selection part, and provide the objects and the language spatial relations as input.
In contrast with our work, they consider only object-object explicit relations. Collell~\etal~\cite{collell2018acquiring} introduced the notion of implicit spatial relations, expanding on prior research limited to explicit relations, yet they restrict to two objects and a single relationship in a structured format.
% To our knowledge, no work has decoded language-expressed spatial relationships to spatial arrangements without limiting the type of relations.
Although not related to spatial understanding of scenes, Dan~\etal~\cite{dan2020spatial} create a spatial representation language to describe spatial configurations, while some works \cite{kordjamshidi2010spatial, kordjamshidi2011spatial} tackle spatial role labeling with a relational learning framework.

\textbf{Multimodal spatial understanding} mostly consists of works that employ a spatial reasoning module in their pipeline, yet proper spatial understanding is not their main goal but rather a secondary sub-problem, hence less emphasis is placed on spatial correctness and evaluation. Several works \cite{johnson2018image, herzig2019learning} generate images from text descriptions with an intermediate scene layout generation step. Others \cite{lee2019neural, li2019seq, li2019object, hong2018inferring} explicitly focus on generating high-quality multi-object and multi-relationship 2D scene layouts from natural language, without limiting the type or number of relationships. However, their layout module does not aim for a precise depiction of the scene arrangements but rather provides a rough outline for the subsequently generated image. The closest works to ours \cite{tan2018text2scene, zitnick2013learning} use the same dataset \cite{zitnick2013bringing} and generate realistically-looking scenes, given a language description of the scene's spatial arrangements. In this chapter, we show that our method is superior to the method of Tan~\etal~\cite{tan2018text2scene}, and shed light on the the property of completing partial scenes.
% and is more extensively evaluated on the Abstract Scenes dataset.
% We, conversely, aim to provide a complete description of the arrangements as input and generate the spatial arrangements in 2D space.
Dan~\etal~\cite{dan2020understanding} predict the relationship word given the image, a bounding box, and the subject and object words by using a spatial model to filter the predictions of a fine-tuned \textsc{Bert} model. Their model does not decode language to 2D spatial arrangements while considering the position of the objects. Finally, Ghanimifard~\etal~\cite{ghanimifard-dobnik-2019-goes} propose a model that can generate spatial image descriptions to investigate what kind of spatial bottom-up knowledge, benefits the top-down methods the most.
\section{Methods}
The main building block for all our models is \blackgls{bert} \cite{devlin2018bert} (see \S\ref{background:sec:transformers} and \S\ref{background:sec:transfer} for details on \blackgls{bert}). In particular, we propose \blackgls{sr-bert}, a \blackgls{bert} variant based on a pre-trained \textsc{Bert\textsubscript{Base}}. Compared with existing \textsc{Bert}-based models \cite{sun2019videobert, tan2019lxmert, chen2019uniter, su2019vl, lu2019vilbert, li2019visualbert}, with \blackgls{sr-bert} our contributions are two-fold:
\begin{inparaenum}[(i)]
\item We alter the input-embedding module to process two discrete modalities with different positional encoding---sequential and spatial;
\item We design a novel training method---Masked Position Modelling---where we iteratively mask and predict the positional encoding of the input objects.
\end{inparaenum}

% \subsection{\textsc{Bert} revisited}\label{emnlp2020:sec:background}
% In \blackgls{bert}, the input sequence is tokenized using WordPiece tokenization \cite{wu2016google} and encoded as token indices $\{w^1,...,w^N\}$ with a \texttt{[CLS]} token prepended at the start and \texttt{[SEP]} token appended at the end. Then, a token embedding vector, a token index and a token type embedding vector for each word are summed together, and subsequently, layer-normalization \cite{ba2016layer} and dropout \cite{srivastava2014dropout} are applied.
% % The rest of the architecture resembles the Transformer model of Vaswani~\etal~\cite{vaswani2017attention}.
% The essence of \textsc{Bert}'s bi-directionality is the pre-training method - Masked Language Modelling (MLM) which we explain in \S\ref{emnlp2020:sec:mpm}. For a detailed description, we refer to Devlin~\etal~\cite{devlin2018bert}.

\subsection{Spatial Reasoning Bert}

\begin{figure}[ht]
\centering
\includegraphics[width=1.0\textwidth]{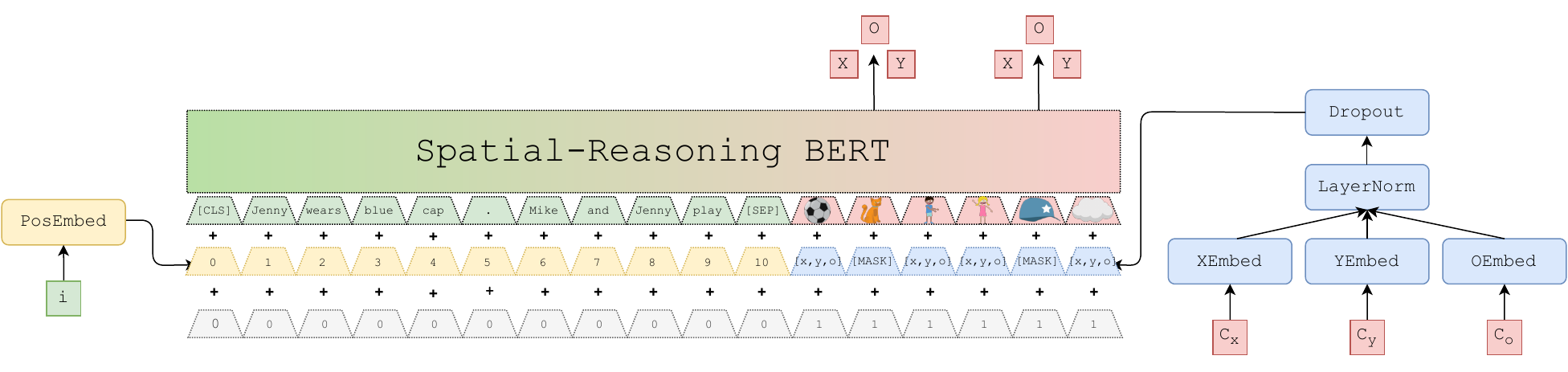}
\caption[The Spatial-Reasoning Bert Model]{The \blackgls{sr-bert} model with the text position embedding module as per \textsc{Bert}---Left (Yellow), clip-art spatial embedding module, which is novel in \blackgls{sr-bert}---Right (Blue). The blue \texttt{[MASK]} elements are the masked spatial positions, which the model learns to predict during training. During inference, all blue elements (the spatial encoding of the clip-arts) are masked, and the model non-autoregressively decodes them.}
\label{emnlp2020:fig:spatial_bert}
\end{figure}

In order to enable \textsc{Bert} to handle two modalities with different positional encoding, we keep the language embedding module as described in the work of Devlin~\etal~\cite{devlin-etal-2019-bert}, before the addition of the token, token position, and token type embeddings, and append the output of a specialized spatial embedding module. To represent a clip-art object on the 2D canvas, we store a unique clip-art object index $c_i$, a spatial encoding $[c_x,c_y]$ coordinate pair, and a binary indicator $c_o$ of the clip-art orientation (left or right). Therefore, we obtain a clip-art category embedding: $\mathbf{c_e} = \operatorname{CEmbed}(c_i)$; a clip-art position embedding: $\mathbf{x_e} = \operatorname{XEmbed}(c_x)$ and $\mathbf{y_e} = \operatorname{YEmbed}(c_y)$, applied for the $x$ and $y$ axis respectively; and embedding of the clip-art orientation: $\mathbf{o_e} = \operatorname{OEmbed}(c_o)$. Note that in practice, all four embedding layers ($\operatorname{CEmbed}$, $\operatorname{XEmbed}$, $\operatorname{YEmbed}$, and $\operatorname{OEmbed}$) are simple linear projection layers.
% Finally, we obtain a token type embedding $t^c_e$.
To obtain an embedding of the clip-art position in the scene we sum all positional embeddings together and apply layer-normalization and dropout: $\mathbf{s_e} = \operatorname{Dropout}(\operatorname{LayerNorm}(\mathbf{x_e} + \mathbf{y_e} + \mathbf{o_e}))$.
% \begin{equation}
% \begin{gathered}
% c_e = \operatorname{CEmbed}(c_i);~x_e = \operatorname{XEmbed}(c_x) \\
% y_e = \operatorname{YEmbed}(c_y);~o_e = \operatorname{OEmbed}(c_o) \\
% s_e = \operatorname{Drop}(\operatorname{LayerNorm}(\lambda(x_e + y_e + o_e)));~t^c_e = \operatorname{TokenTypeEmbed}(c_i)
% \end{gathered}
% \end{equation}

% where $\lambda$ is a scaling factor. Finally, we concatenate the word embeddings $w_e$ with the clip-art embeddings $c_e$, the word positional embeddings $p_e$ with the spatial embeddings $s_e$, the token type embeddings for the language $t^w_e$ and spatial parts $t^c_e$, and apply layer-normalization and dropout:
% \begin{equation}
% \begin{gathered}
%     wc = \operatorname{Concat}(w_e, c_e);~ps = \operatorname{Concat}(p_e, s_e);\\
%     tt = \operatorname{Concat}(t^w_e, t^c_e);~e = \operatorname{Drop}(\operatorname{LayerNorm}(wc + ps + tt))
% \end{gathered}
% \end{equation}
% where $e$ is the final input embedding.
We keep the rest of the model identical to the original model of Devlin~\etal~\cite{devlin2018bert} and re-use the pre-trained \textsc{Bert\textsubscript{Base}} modules. Figure \ref{emnlp2020:fig:spatial_bert} illustrates the \blackgls{sr-bert} model architecture. On top, we append modeling heads to predict the clip-art spatial position $[\hat{h}_x, \hat{h}_y]$ and orientation $\hat{h}_o$.
% that consist of two linear layers, with a GELU \cite{hendrycks2016gaussian} and a layer-normalization between them: $\mathbf{\hat{h}} = \operatorname{Linear}(\operatorname{LayerNorm}(\sigma(\operatorname{Linear}(\mathbf{x}))))$, where $\mathbf{x}$ is the hidden representation of a single clip-art. We create a continuous and a discrete model\footnote{Denoted according to the $[x,y]$ value they predict.} variant, where both output a discrete value for the object orientation $o$ (left or right). 

\textbf{Continuous Model.} With this model variant, for the clip-art position $[c_x,c_y]$ we directly regress the coordinates for the $x$ and $y$ axis for each clip-art: $\hat{h}_x$ and $\hat{h}_y$ respectively.
% apply sigmoid ($\sigma$) normalization and subsequently multiply them by $w$ and $h$ (the width and height of the scene canvas respectively) to project them in the range of the scene canvas: $\hat{x} = \sigma(\hat{h}_x) \odot w;~\hat{y} = \sigma(\hat{h}_y) \odot h$.
During training, we minimize a sum of all individual losses: $\mathcal{L} = \mathcal{L}_{absolute} + \mathcal{L}_{relative} + \mathcal{L}_{orientation}$. In \S\ref{emnlp2020:sec:quantitative}, we describe the absolute scene similarity and the relative scene similarity metrics, which we minimize with $\mathcal{L}_{absolute}$ and $\mathcal{L}_{relative}$, respectively.

\textbf{Discrete Model.} We additionally explore a discrete model that outputs a probability distribution over the quantized $x$ and $y$ axis (explained in \S\ref{emnlp2020:sec:mpm}): $\mathbf{\hat{h_x}} = \operatorname{softmax}(\mathbf{\hat{h}_x});~\mathbf{\hat{h}_y} = \operatorname{softmax}(\mathbf{\hat{h}_y})$, where $\mathbf{\hat{h_x}}$ and $\mathbf{\hat{h_y}}$ now represent confidence scores over the 2D canvas for the $x$ and $y$ axis respectively. We train this model by minimizing the sum of the individual per-axis cross-entropy losses together with the orientation loss: $\mathcal{L} = \mathcal{L}_x + \mathcal{L}_y + \mathcal{L}_{orientation}$. To obtain predictions we simply take the position with the highest probability.

\subsection{Masked Position Modelling}\label{emnlp2020:sec:mpm}
Originally, \textsc{Bert} is trained by reconstructing the ground truth sentence given a corrupted one as input, where 15\% of the tokens are replaced with a \texttt{[MASK]} token 80\% of the time, a random token 10\% of the time, or unchanged 10\% of the time. Then, \textsc{Bert} outputs a probability distribution for each token, while the loss is computed over the masked tokens. To conceptually 
preserve \textsc{Bert}’s input-embedding module, and explicitly encode a \textit{masked} clip-art position\footnote{In other words, to have a specialized embedding indicating that a clip-art is masked.}, we encode the spatial position with a discrete set of values. In our use-case, the ranges are $[0, 500]$ for $x$, $[0, 400]$ for $y$ and $[0, 1]$ for $o$. Due to the data size ($\sim$8000 scenes for training with a median of 6 clip-arts per scene), having $500 \times 400 \times 2$ spatial combinations renders the problem unfeasible. To overcome this issue, we quantize the values of $x$ and $y$ in equal $\text{bin\_size}$ intervals, yielding a range of $[0, \frac{500}{\text{bin\_size}}]$
 unique values for $x$ and $[0, \frac{400}{\text{bin\_size}}]$ for $y$. We empirically determine $\text{bin\_size} = 20$, which yields optimal results on the validation set.

In \textsc{SR-Bert}, we adjust the masking objective to decode spatial representations, i.e., \textit{instead of masking the clip-art tokens, we mask the spatial encoding tokens}. Namely, given a set of sentences and a set of clip-arts with their spatial position and orientation $[c_x, c_y, c_o]$, we train the model such that when spatial arrangements between the clip-arts are corrupted, the model learns to rely on the relations from the sentences to reconstruct the correct spatial arrangement. Thus, we use original masked language modeling ratios used in the work of Devlin~\etal~\cite{devlin2018bert}, and mask 15\% of the clip-arts spatial positions during training. Then, 80\% of the time each of the position elements $[x, y, o]$ are replaced with \texttt{[MASK]} token, 10\% of the time with a random $[x, y]$ position and random $o$ orientation, and 10\% of the time we keep them the same.

During training, we employ data augmentation techniques,
% (see Appendix~\ref{emnlp2020:appendix:sec:augmentation}),
specifically adapted to fit within the training objective we propose:
% \subsubsection{Data Augmentation}
% Due to the limited size of the dataset, we employ several data augmentation strategies to artificially increase the quantity of data, while preserving the meaning of each abstract scene. Furthermore, we want to impose greater importance on the relative positioning of the elements and obtain a model that is invariant on whether the scenes are mirrored across the y-axis and invariant on the order of the sentences from the scene descriptions.
% To this end, the data augmentations we use are:
\begin{inparaenum}[(i)]
     \item In the Abstract Scenes dataset, each scene is paired with a set of $\sim6$ sentences, where each sentence is entirely self-contained (e.g., ``The cat is on the bench''). We conclude that the order of the sentences does not affect the semantic meaning of the scenes, and randomly shuffle the sentences before providing them as input to the models;
    \item For each scene in the dataset, the mirrored scene across the $y$ axis is also valid. Therefore, we randomly mirror each scene with 50\% probability by inverting the $x$ coordinates and the orientation of the clip-arts;
    % - $\left| width - x \right|$, and reverting the orientation - $\left| 1 - o \right|$ of each object.
    \item With 50\% probability we move all elements in the scene up, down, left, or right, by a random number of quantization intervals, with 25\% probability for each movement. The number of quantization intervals is sampled uniformly from $[0, w]$ for $x$ and $[0, h]$ for $y$, where $w$ and $h$ are the scene width and height respectively.
\end{inparaenum}

\subsection{Non-Autoregressive Decoding of Spatial Arrangements}\label{emnlp2020:sec:decoding-strategies}
Because of the non-sequential nature of the 2D space, we have to impose a specific order when decoding the spatial arrangements. In contrast with the left-to-right sequential order in the English language, decoding a set of spatial relationships does not have a pre-defined order and one must consider all pairwise relative locations of the objects.
% Consequently, the model either has to learn the decoding order itself, or the decoding order has to be injected manually.
For that reason, we adopt non-autoregressive decoding based on mask-predict \cite{ghazvininejad2019mask}, to convert the language spatial relations into 2D spatial arrangements. By using a transformer model which can be effectively trained with mask-predict methods \cite{ghazvininejad2019mask}, we can derive decoding strategies where the model considers all relations between the objects.
% and learns the decoding order.
Specifically, we assess five decoding strategies inspired by Lawrence~\etal~\cite{lawrence2019attending}:

\begin{enumerate}[label=(\roman*)] % Use lowercase Roman numerals
    \item \textbf{Single-step (\blackgls{ss})}: The positions for \textit{all} clip-arts in the scene are generated in a single step, all at once.
    \item \textbf{Random-order (\blackgls{ro})}: We generate the positions of the clip-arts by randomly choosing the order.
    \item \textbf{Human-order (\blackgls{ho})}: We inject domain knowledge in the generation process and generate the clip-art positions following an order specified by the dataset itself: objects in the sky, large elements, people, animals, clothing, food and toys.
    \item \textbf{Highest-confidence (\blackgls{hc})}:
    We maintain a beam of 3 hypotheses for the clip-arts that exhibit the highest joint probability of the coordinates $c_x$, $c_y$, and orientation $c_o$. We repeat this process until all clip-arts are spatially arranged and select the hypothesis with the highest joint probability. We apply this decoding strategy only for the discrete model.
    \item \textbf{Lowest-entropy (\blackgls{le})}: Similarly, we perform beam search, however, we now use the lowest entropy of all hypotheses as a selection criterion. We apply this decoding strategy only for the discrete model.
\end{enumerate}

\subsubsection{Future work: Learning a decoding order}
Besides the mask-predict-based methods \cite{lawrence2019attending} we describe above, alternatively, we could train models that would learn a decoding order, facilitating an autoregressive formulation of the clip-art arrangement decoding problem. Namely, the sequence of clip-arts could be linearised in a consistent manner (e.g., following a left-to-right and top-to-bottom order), such that it would enable applying autoregressive methods to generate clip-art arrangements. Such methods would be more efficient to train, as the current \blackgls{sr-bert} models only learn from the portion of the mask clip-arts, and not the entire sequence. Furthermore, during inference, such models would follow a consistent decoding scheme that is aligned to how the models have been trained, which, in turn, could improve the models' performance.
\section{Evaluation and Discussion}\label{emnlp2020:sec:evaluation}
We use the Abstract Scenes dataset \cite{zitnick2013bringing}, which consists of 10.020 scenes of clip-arts, with an average of 6 sentences for each scene, describing which clip-art objects appear in the scene and the spatial relations between them. The clip-arts belong to 7 distinct groups: objects in the sky, large elements, people, animals, clothing, food, and toys. The scenes are organized in 1002 semantically different sets, where scenes within a particular set are generated from the same core-scene description. After removing empty scenes from each of the 1002 sets we allocate one scene for testing, one for model selection and we keep the rest for training.\footnote{We split the data this way in order to retain as much information as possible within the train-validation-test splits.}
% We also include an experiment with a random split in appendix \ref{emnlp2020:appendix:sec:add_quantitative}.}
That leaves us with 1002 scenes in the test and validation set respectively, and 7989 scenes in the training set. The maximum number of clip-arts in a scene is equal to 17 while the minimum number, and the median, are equal to 6. The total number of unique clip-arts in the dataset is 126.

\subsection{Quantitative Evaluation}\label{emnlp2020:sec:quantitative}
We propose two measures to evaluate how well we can translate language-expressed spatial relations (scene descriptions) to 2D spatial arrangements between objects (clip-arts). Both are similarity measures applied on normalized coordinates, hence the higher the better.

\textbf{Absolute spatial scene similarity} represents the average Euclidean distance between the ground truth and predicted position over all $N$ clip-arts in the scene, defined as Gaussian function:
\begin{gather}
    \mathcal{S}_{absolute} = \frac{1}{N} \sum_{i=1}^N \operatorname{exp}(-\frac{\sqrt{ (x_i - \hat{x}_i)^2 + (y_i - \hat{y}_i)^2}}{2\sigma})
\end{gather}
where $[x_i, y_i]$ and $[\hat{x}_i, \hat{y}_i]$ represent the ground truth and predicted position respectively, and $\sigma$ is set to 0.2 as per the concurrent work of Tan~\etal~\cite{tan2018text2scene}. We compute the similarity for both the original ground truth positions and the ground truth positions mirrored across the $y$ axis, and subsequently take the maximum as the absolute similarity for that scene.

\textbf{Relative spatial scene similarity} focuses on the relative positioning of the objects with respect to each other. We compute two separate square matrices for the ground truth and predicted relative discrepancies between the clip-art positions, $\mathbf{M}$ and $\mathbf{\hat{M}}$ respectively, where the $\mathbf{M}_{i,j}$ element is the Euclidean distance between the $i$-th and $j$-th object. Then, the similarity is the mean absolute difference between $\mathbf{M}$ and $\mathbf{\hat{M}}$, defined as a Gaussian function:
\begin{gather}
    \mathcal{S}_{relative} = \frac{1}{(N - 1)^2}\sum_{i=1}^N\sum_{j=1}^N\operatorname{exp}(-\frac{\left|\mathbf{M}_{i,j} - \mathbf{\hat{M}}_{i,j}\right|}{2\sigma})
\end{gather}
where $\sigma$ is again fixed to 0.2.

Due to ambiguity---some object groups are symmetrical across the y-axis---we evaluate the orientation accuracy only within the scene completion setup.

In all experiments, we test the statistical significance between the average similarities between any two methods with Welch's t-test $(p < 0.01)$ \cite{welch1947generalization} on the per scene similarities.

\textbf{Baselines.} We create two recurrent neural networks with attention baselines \cite{bahdanau2014neural}. The first baseline, dubbed as Attention, uses an attention-based decoder, while the second baseline, dubbed as Rnn + Attention, propagates contextual information regarding the spatial positions with a recurrent neural network.
% (see Appendix \ref{emnlp2020:appendix:sec:rnn_baseline} for details).
In both baselines, the clip-arts are ordered according to HO. We measure the models' performance in terms of the similarity metrics in five different scenarios.

\subsubsection{Translating Text-based Scene Descriptions to 2D Spatial Arrangements}
To evaluate the overall performance of \blackgls{sr-bert}, we measure the absolute and relative scene similarity metrics of the discrete and continuous variants on the full test set of 1002 semantically unique scenes. In Table \ref{emnlp2020:table:full-test-set}, we report results for all decoding strategies, when the model is, or is not, conditioned on the scene descriptions (dubbed as No-description). For these description-less methods, we remove the scene descriptions and use the concatenated \texttt{[CLS]} and \texttt{[SEP]} tokens to indicate to \blackgls{sr-bert} that the text input is excluded.

\begin{table*}[ht]
\centering
\resizebox{\textwidth}{!}{
\begin{tabular}{lcccc} \toprule
    {} & \multicolumn{2}{c}{Discrete Model} & \multicolumn{2}{c}{Continuous Model} \\
    \cmidrule(lr){2-3} \cmidrule(lr){4-5}
    {Method} & {Absolute similarity ($\uparrow$)} & {Relative similarity ($\uparrow$)}  & {Absolute similarity ($\uparrow$)} & {Relative similarity ($\uparrow$)} \\ \midrule
    \multicolumn{5}{l}{\textbf{\textit{\blackgls{sr-bert} methods}}} \\
    SS & 0.565 $\pm$ 0.002 & 0.769 $\pm$ 0.003 & 0.589 $\pm$ 0.002 & 0.814 $\pm$ 0.003  \\
    SS; No-description & 0.499 $\pm$ 0.002 & 0.695 $\pm$ 0.003 & 0.530 $\pm$ 0.002 & 0.779 $\pm$ 0.002 \\ \midrule
    RO & 0.585 $\pm$ 0.003 & 0.818 $\pm$ 0.003 & 0.603 $\pm$ 0.003 & 0.838 $\pm$ 0.003 \\
    RO; No-description & 0.523 $\pm$ 0.003 & 0.746 $\pm$ 0.003 & 0.546 $\pm$ 0.003 & 0.792 $\pm$ 0.002 \\ \midrule
    HO & 0.594 $\pm$ 0.003 & 0.823 $\pm$ 0.003 & \textbf{0.611 $\pm$ 0.003} & \textbf{0.846 $\pm$ 0.003} \\ 
    HO; No-description & 0.539 $\pm$ 0.003 & 0.745 $\pm$ 0.003 & 0.562 $\pm$ 0.003 & 0.794 $\pm$ 0.003 \\ \midrule
    HC & \textbf{0.598 $\pm$ 0.003} & \textbf{0.826 $\pm$ 0.003} & \NA & \NA  \\
    HC; No-description & 0.534 $\pm$ 0.003 & 0.750 $\pm$ 0.003 & \NA & \NA \\ \midrule
    LE & 0.592 $\pm$ 0.003 & 0.825 $\pm$ 0.003 & \NA & \NA \\
    LE; No-description & 0.536 $\pm$ 0.003 & 0.751 $\pm$ 0.003 & \NA & \NA \\ \midrule \midrule
    \multicolumn{5}{l}{\textbf{\textit{Baseline methods}}} \\
    Attention & 0.565 $\pm$ 0.002 & 0.746 $\pm$ 0.002 & 0.579 $\pm$ 0.002 & 0.812 $\pm$ 0.002 \\ 
    Rnn + Attention & 0.567 $\pm$ 0.002 & 0.746 $\pm$ 0.003 & 0.581 $\pm$ 0.002 & 0.813 $\pm$ 0.002 \\ 
    \bottomrule
\end{tabular}
}
\caption[Comparison between \textsc{Sr-Bert} and the baselines]{Absolute and relative similarity for \blackgls{sr-bert} and the baselines.}
\label{emnlp2020:table:full-test-set}
\end{table*}

For both metrics, we observe that the models that use a controlled way of decoding the spatial arrangements (\blackgls{hc}, \blackgls{le} and \blackgls{ho}) outperform the ones that are orderless (SS, RO). We conclude that decoding order matters, and orderless decoding (\blackgls{ss}, \blackgls{ro}) is undesirable. Regardless of the decoding strategy, we observe significant gains when the model is conditioned on the scene descriptions. This implies that our model is not trivially relying on dataset bias (e.g., Mike usually wears a blue cap) when decoding the spatial arrangements. Furthermore, we observe a significant increase in both absolute and relative similarity with \blackgls{sr-bert} compared to the baselines, in both the discrete and continuous variant, indicating the superiority of jointly attending on both all available clip-arts and spatial relations. Finally, the continuous model outperforms the discrete model and is less sensitive to the decoding strategy; which we hypothesize is due to the continuous model directly optimizing the evaluation metrics.

We conclude that we can train models to effectively translate text-based scene descriptions to 2D spatial arrangements without constraints on their number and types  \inlineresearchquestion{Research Question 1}.

\subsubsection{Completing Partial Scenes}
To get better insights into the performance of the models, we formulate an inference scenario where we decode spatial arrangements for each type of clip-arts separately (objects in the sky, large elements, people, animals, clothing, food, and toys), conditioned on both the scene descriptions and the remaining clip-arts. This can be interpreted as a scene completion setting; e.g., what is the position and orientation of Mike and Jenny in the 2D space w.r.t. the other clip-arts if: ``Mike is holding a football'', ``Mike wants to play football with Jenny'' and ``Jenny fell off the swing''.

\begin{table}[ht]
\centering
\resizebox{0.8\textwidth}{!}{
\begin{tabular}{lccc} \toprule
    {Clip-art group} & {Absolute similarity ($\uparrow$)} & {Relative similarity ($\uparrow$)} & {Orientation accuracy ($\uparrow$)} \\ \midrule
    Sky & 0.655 $\pm$ 0.011 & 0.782 $\pm$ 0.010 & 48.4 $\pm$ 1.54 \\
    Large & 0.671 $\pm$ 0.011 & 0.781 $\pm$ 0.012 & 57.8 $\pm$ 1.60 \\
    People & 0.796 $\pm$ 0.006 & 0.857 $\pm$ 0.005 & 86.7 $\pm$ 0.97 \\
    Animals & 0.661 $\pm$ 0.016 & 0.783 $\pm$ 0.017 & 70.5 $\pm$ 2.02 \\
    Clothing & 0.853 $\pm$ 0.021 & 0.900 $\pm$ 0.021 & 71.2 $\pm$ 2.13 \\
    Food & 0.772 $\pm$ 0.021 & 0.852 $\pm$ 0.023 & 46.9 $\pm$ 1.92 \\
    Toys & 0.664 $\pm$ 0.015 & 0.786 $\pm$ 0.017 & 51.0 $\pm$ 1.80 \\
    \bottomrule
\end{tabular}
}
\caption[Evaluation metrics reported per clip-art group]{Absolute and relative similarity for each clip-art group. Evaluation performed in a scene completion setting using the Discrete model.}
\label{emnlp2020:table:scene-completion}
\end{table}

In Table \ref{emnlp2020:table:scene-completion}, we see the highest absolute and relative similarity when we generate the arrangements for the ``clothing'' category conditioned on the other groups. On the contrary, the lowest reported absolute and relative similarity are for the ``animals'' category. Finally, we see almost random orientation accuracy for the symmetrical object categories (e.g., sky, food, etc.), with a major improvement for the object categories where their orientation matters.

\subsubsection{Coping with Explicit and Implicit Spatial Relations}
We split the test set into 4 different subsets, where each contains scenes with $[0\%, 25\%]$, $(25\%, 50\%]$, $(50\%, 75\%]$ and $(75\%, 100\%]$ ratio of sentences that consist of explicit relations exclusively (i.e., only concrete spatial relations: ``above'', ``next to'', etc.).
% We determine whether a relation between a pair of clip-arts is explicit if it  

\begin{figure}[ht]
\centering
\includegraphics[width=0.8\textwidth]{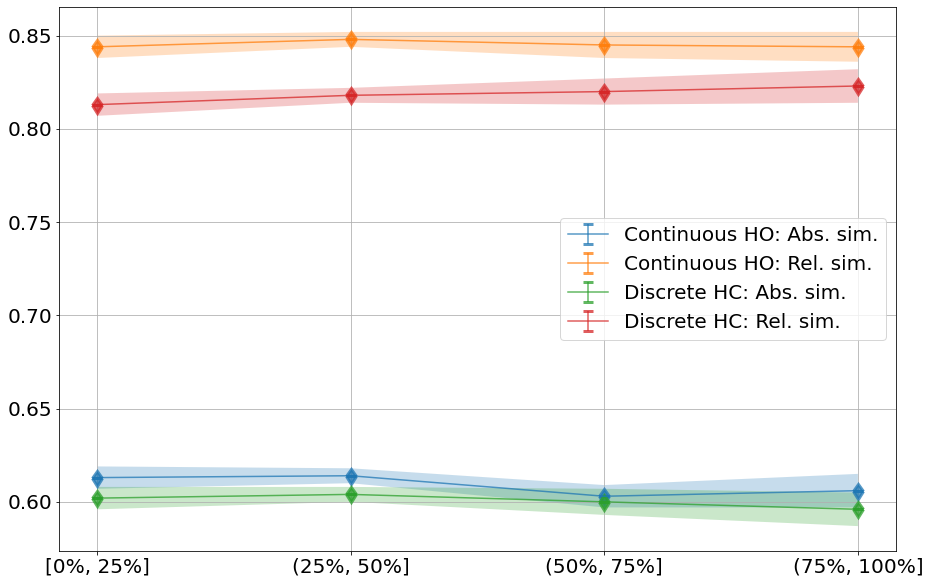}
\caption[Metrics on testing data as per the ratio of explicit spatial relations]{The Absolute and Relative scene similarity metrics reported on testing data splits obtained as per the ratio of explicit spatial relations (e.g., ``above'', ``below'', ``next to'', etc.)}
\label{emnlp2020:fig:explicit-implicit}
\end{figure}

In Figure \ref{emnlp2020:fig:explicit-implicit}, we report results with the discrete model with \blackgls{hc} decoding, and the continuous model using \blackgls{ho} decoding. We observe that both the continuous and discrete models obtain steadily similar absolute and relative similarities as the ratio of explicit relations increases. Note that if the models are not capable of coping with implicit spatial language, we would observe an increased performance as the ratio of explicit spatial relations increases. This shows the robustness of our method in successfully coping with both explicit and implicit spatial relations \inlinesubresearchquestion{Research Question 1.1}.

\subsubsection{Generalizing to Out-of-Distribution Data}
To gain insight into the out-of-distribution generalization ability, we split the test set by considering scenes that contain at least 2 sentences with a subject-relationship-object (S-R-O) combination\footnote{For each sentence there is a S-R-O triplet in the dataset.} that the model has not encountered during training. This yields 354 scenes from which we create 5 subsets:

\begin{inparaenum}[(i)]
    \item The scenes in their \textbf{Default} form, which contain sentences with (S-R-O) combinations that the model has encountered, and at least two combinations that it has \textit{not} encountered during training. 
    \item From each of the default scenes, we discard sentences consisting of (S-R-O) combinations that the model has encountered during training while keeping only the unseen ones. Hence, we are left with at least 2 sentences per scene, which are \textbf{Out-of-Distribution} (\blackgls{ood}).
    \item \textbf{In-Distribution} (\blackgls{id}) is the complement of the Out-of-Distribution and contains exclusively sentences consisting of (S-R-O) combinations encountered at training time.
    \item The scenes without the scene descriptions: \textbf{No-description}.
    \item A \textbf{Filtered} training set, from which we remove all sentences that have (S-R-O) combinations that appear in their default testing scenes. We use this set to train a new model.
\end{inparaenum}

\begin{table}[t]
\centering
\resizebox{0.7\textwidth}{!}{
\begin{tabular}{lcc} \toprule
    {Data type} & {Absolute similarity ($\uparrow$)} & {Relative similarity ($\uparrow$)} \\ \midrule
    \multicolumn{3}{l}{\textbf{\textit{Discrete model with HC decoding}}} \\
    Default & 0.599 $\pm$ 0.005 & 0.817 $\pm$ 0.005 \\
    Out-of-Distribution & 0.575 $\pm$ 0.005 & 0.799 $\pm$ 0.005 \\
    In-Distribution & 0.586 $\pm$ 0.005 & 0.807 $\pm$ 0.005 \\
    No-description & 0.505 $\pm$ 0.005 & 0.724 $\pm$ 0.004 \\
    Filtered & 0.581 $\pm$ 0.005 & 0.803 $\pm$ 0.005 \\
    \midrule
    \multicolumn{3}{l}{\textbf{\textit{Continuous model with HO decoding}}} \\
    Default & 0.608 $\pm$ 0.005 & 0.844 $\pm$ 0.004 \\
    Out-of-Distribution & 0.582 $\pm$ 0.005 & 0.817 $ \pm$ 0.004 \\
    In-Distribution & 0.593 $\pm$ 0.005 & 0.828 $\pm$ 0.004 \\ 
    No-description &  0.563 $\pm$ 0.004 & 0.795 $\pm$ 0.004 \\
    Filtered & 0.586 $\pm$ 0.005 & 0.813 $\pm$ 0.004 \\
    \bottomrule
\end{tabular}
}
\caption[Out-of-distribution generalization experiments]{Absolute and relative similarity with the Discrete and Continuous model on distinct data subsets derived for out-of-distribution generalization.}
\label{emnlp2020:table:compositional-generalization}
\end{table}

In Table \ref{emnlp2020:table:compositional-generalization}, we observe that despite the effective similarity between the No-description and the \blackgls{ood} setting (in both the model is exposed to unfamiliar descriptions or no descriptions at all), the difference in performance in favor of \blackgls{ood} is relatively big---especially with the discrete model. We also observe that the performance in the \blackgls{ood} setting degrades only moderately compared to the Default setting, and the \blackgls{ood} differs non-significantly from the ID setting. When comparing the Default and \blackgls{ood} settings, we must take into account that the model uses only partial scene descriptions in the \blackgls{ood} setting due to the held-out sentences, which might explain the moderate drop in performance. Finally, we observe that the Filtered setting fares remarkably close to the Default setting, even though in this case, the model encounters all spatial relations as out-of-distribution. This observation suggests potentially great value in real-life scenarios where one is frequently exposed to unfamiliar spatial relations \inlinesubresearchquestion{Research Question 1.2}.

\subsubsection{Generating Complete Abstract Scenes}
We extend our method to generate complete scenes, i.e., predicting the clip-arts and their spatial arrangements. We adopt a two-step approach where
\begin{inparaenum}[(i)]
    \item we fine-tune \textsc{Bert\textsubscript{Base}} as a backbone clip-art predictor, such that, given a description of the scene, the model outputs a vector of probabilities for the clip-arts in the scene: $\mathbf{\hat{y}} = \sigma(\operatorname{Linear}(\text{\textsc{Bert}}(\mathbf{x})))$, where $\mathbf{\hat{y}}$ are the output probabilities, and $\mathbf{x}$ is the tokenized scene description; and
    \item we use \blackgls{sr-bert} to arrange the predicted clip-arts w.r.t. the spatial relations. 
\end{inparaenum}
The clip-art predictor projects the text embedding from \textsc{Bert}'s hidden space to 126 (number of unique clip-arts), and $\sigma$ is the sigmoid non-linearity, thresholded at 0.35 during inference.

We compute the precision and recall of each clip-art, classification accuracy for their poses and expressions, and absolute similarity for the object positions\footnote{The \texttt{U-obj} coord. metric of Tan~\etal~\cite{tan2018text2scene} is equivalent to our absolute similarity.}. When generating the scene arrangements, we first provide the predicted clip-arts to \blackgls{sr-bert} and then arrange them on the scene. We then find the common clip-arts between the predictions and the ground truths and measure absolute similarity, as per Tan~\etal~\cite{tan2018text2scene}\footnote{Contrary to the other experiments, here we measure absolute similarity only w.r.t. the default ground truth positions and not the mirrored ones for a pessimistic comparison.}. We train a new discrete \blackgls{sr-bert} model on Tan \etal~\cite{tan2018text2scene}'s data splits and perform inference using \blackgls{hc} decoding.

\begin{table}[ht]
\centering
\resizebox{1.0\textwidth}{!}{
\begin{tabular}{lccccc} \toprule
    {Method} & {Clip-art Precision ($\uparrow$)} & {Clip-art Recall ($\uparrow$)} & {Pose Accuracy ($\uparrow$)} & {Expression Accuracy ($\uparrow$)} & {Absolute similarity ($\uparrow$)} \\ \midrule
    Zitnick~\etal~\cite{zitnick2013learning} & 72.2 & 65.5 & 40.7 & 30.0 & 0.449 \\
    Tan~\etal~\cite{tan2018text2scene} & 76.0 & 69.8 & \textbf{41.8} & 37.5 & 0.409 \\
    Clip-art predictor + \blackgls{sr-bert} & \textbf{82.7} & \textbf{72.5} & 40.4 & \textbf{38.0} & \textbf{0.512} \\
    \bottomrule
\end{tabular}
}
\caption[Full scene generation results]{Per-object (clip-art) precision and recall, pose and expression accuracy, and absolute similarity on the test set by Tan~\etal~\cite{tan2018text2scene}.}
\label{emnlp2020:table:full-scene-generation}
\end{table}

In Table \ref{emnlp2020:table:full-scene-generation}, we observe that our pipeline outperforms alternative methods \cite{tan2018text2scene, zitnick2013learning} in terms of precision, recall and expression accuracy, while it falls slightly short in terms of pose accuracy. Moreover, \blackgls{sr-bert} outperforms the concurrent methods according to absolute similarity by a large margin, which measures spatial reasoning. We want to stress however, that because of the simplicity of \blackgls{sr-bert}, the model can be trivially plugged in within a more powerful abstract scene generation from a language pipeline. 

\subsection{Qualitative Evaluation}\label{emnlp2020:sec:qualitative}

\begin{figure}[ht]
\centering
\includegraphics[width=0.8\textwidth]{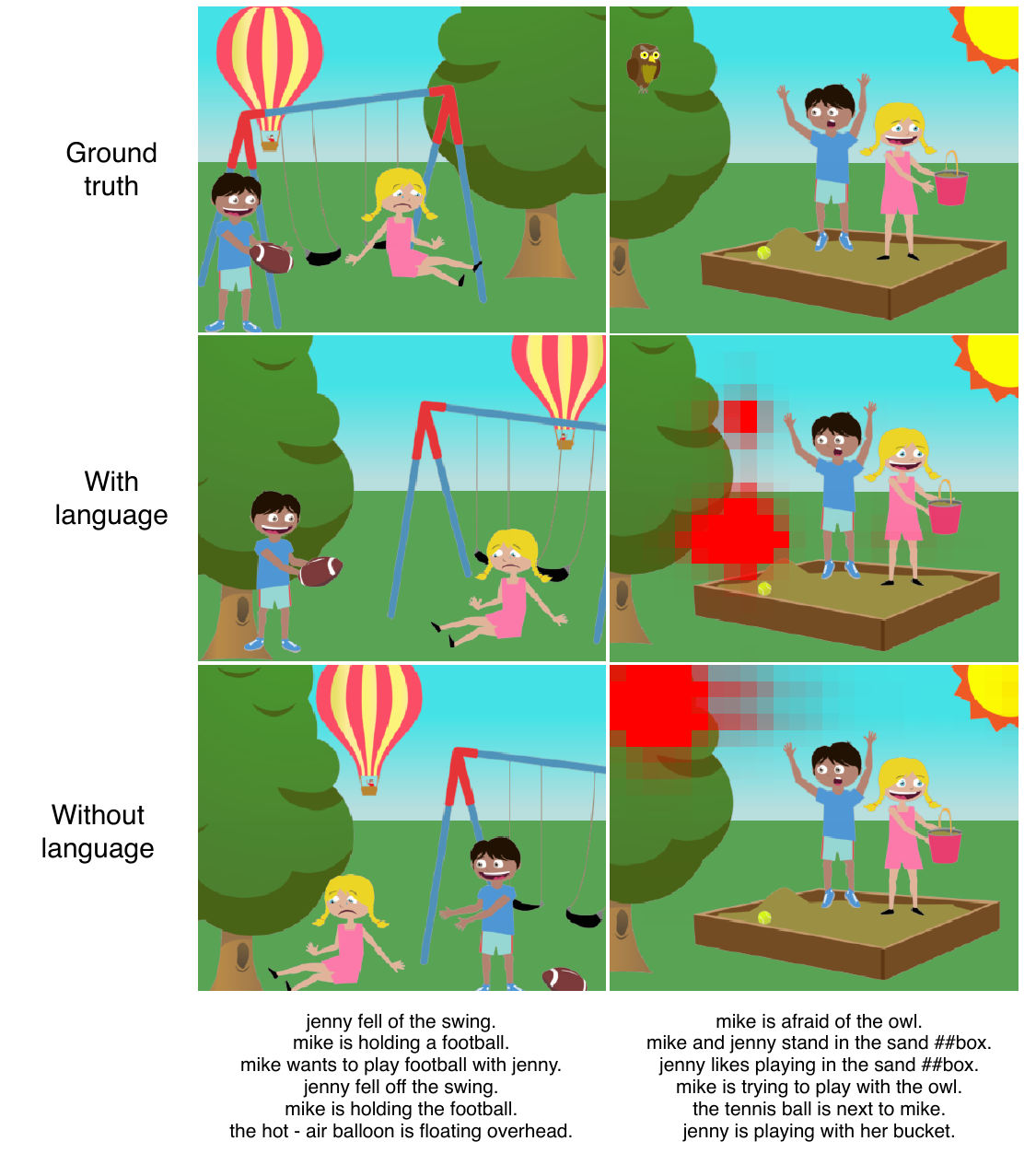}
\caption[Ground truth and generated scenes with a trained model]{Ground truth (top), generated scenes (middle-left, bottom-left), and heat-maps (middle-right, bottom-right) with and without conditioning on the language (i.e., scene descriptions).} % test-set: 1, 89
\label{emnlp2020:fig:with_without_language}
\end{figure}

In Figure \ref{emnlp2020:fig:with_without_language}, we compare the generated clip-art spatial arrangements, when conditioning the model on the full scene description, or no scene description at all. For example, if we exclude the descriptions, both the middle-left and the bottom-left are plausible scenes. However, when the description is present, ``Jenny'', the ``football'' and ``Mike'' take upon certain positions/orientations to satisfy the imposed spatial relations. In Figure \ref{emnlp2020:fig:with_without_language} (right), we showcase the scene completion feature of our method. We see that the most probable location of the \textit{owl} is in the \textit{tree}, which is intuitive. However, when conditioning on the sentences ``Mike is afraid of the owl'' and ``Mike is trying to play with the owl'', the distribution shifts to the \textit{owl} being in the \textit{sandbox}. This indicates that the model gains an understanding of the implicit spatial relationships.

\begin{figure}[ht]
\centering
\includegraphics[width=0.8\textwidth]{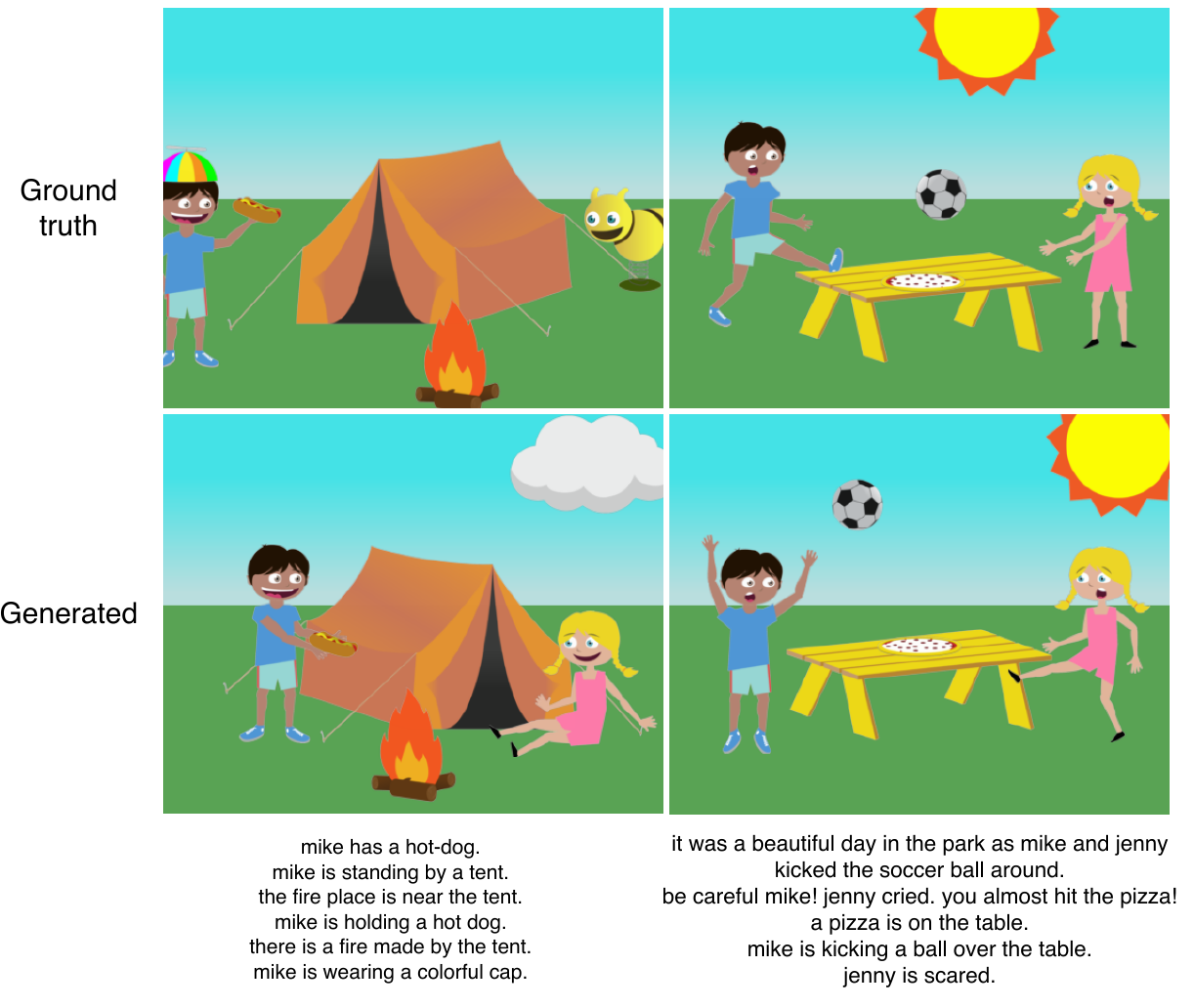}
\caption[Generated scenes: clip-art prediction and clip-art arrangements]{Ground truth (top) and generated scenes (bottom) by (1) predicting the clip-arts and (2) using \blackgls{sr-bert} to arrange them on the 2D canvas.}
\label{emnlp2020:fig:full_scene_generation}
\end{figure}

We also qualitatively evaluate the complete scene generation pipeline on the data split provided by Tan~\etal~\cite{tan2018text2scene}. In Figure \ref{emnlp2020:fig:full_scene_generation}, we observe scenes generated in a two-step process by 
\begin{inparaenum}[(i)]
    \item predicting the clip-arts, and
    \item arranging them using \blackgls{sr-bert}.
\end{inparaenum}
In both scenarios, we observe that the predicted clip-arts are relatively in line with the scene descriptions. Furthermore, despite providing clip-art predictions that do not perfectly resemble the ground truths, the scene arrangements generated with \blackgls{sr-bert} obtain consistent quality w.r.t. the cases when the ground truth clip-arts are provided as input and \blackgls{sr-bert} arranges them.

\subsection{User study}\label{emnlp2020:sec:user-study}
\begin{figure}[ht]
\centering
\includegraphics[width=0.8\textwidth]{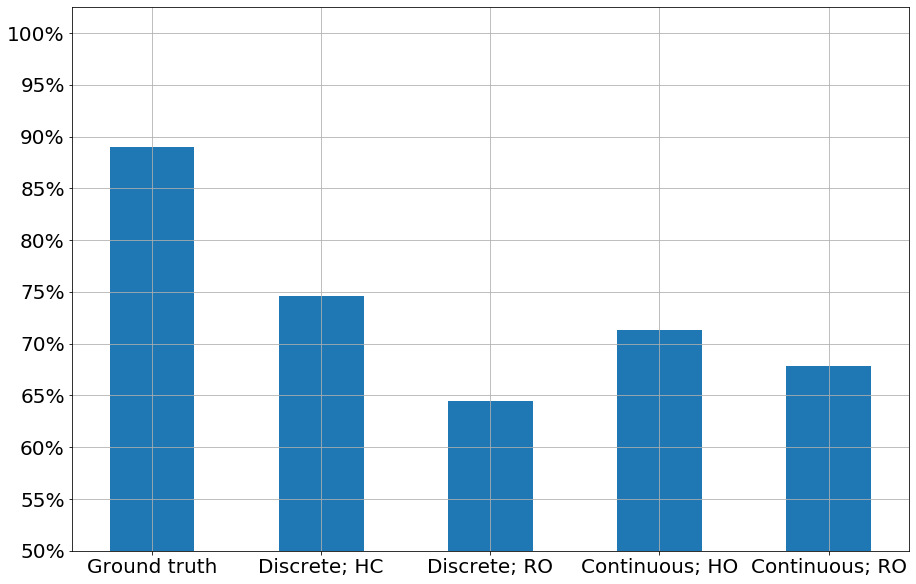}
\caption[User-study evaluation of the generated scenes]{Per-scene macro-average of the accepted spatial relation triplets (i.e., scene descriptions) from the participants in the user study.}
\label{emnlp2020:fig:rating-userstudy}
\end{figure}

We conduct a user study to evaluate to what extent the generated spatial arrangements align with human judgment. We randomly select 100 samples ($\sim$10\% of the full test set) and generate the spatial arrangements using the continuous model with RO and HO and the discrete model with RO and HC decoding. The participants are presented with a scene together with the corresponding sentences that imply spatial relatedness between the clip-arts\footnote{To avoid an optimistic estimate, we remove sentences that are always correct; i.e., sentences that do not imply any spatial connection between the clip-arts e.g., ``Mike is smiling''.}, and are asked to select the sentences which are spatially true for the scene. in Figure \ref{emnlp2020:fig:rating-userstudy}, we report the macro-average results per scene for each model including the ground truth scenes.

Irrespective of the model and decoding strategy, all models perform well compared to the ground truth, and users found that at least 64\% of the sentences are spatially true for the generated scenes while the ground truth scenes are at the 88\% mark, with a fair agreement (k=0.29) between raters as per Fleiss' Kappa score \cite{fleiss1973equivalence}.
% Despite the non-significant difference between the RO and HO decoding with the continuous model and RO and HC with the discrete model in Table \ref{emnlp2020:table:full-test-set},
We observe that the continuous model with HO decoding, and the discrete model with HC decoding, outperform the continuous and discrete model with RO decoding respectively by a large margin. We also observe a less amplified difference of 4\% between the HO and RO decoding with the continuous model compared to the discrete model which reports a 10\% difference between the HC and RO decoding. This is due to the continuous model being less sensitive to the decoding order. This strengthens our hypothesis that the decoding order is important, which comes automatically with the discrete model and HC decoding, while it needs to be injected manually in the continuous model with HO decoding. Consequently, the best performance is achieved with the discrete model with HC decoding---above 74.6\% accepted spatial relations and the continuous model with HO decoding---71.3\% accepted spatial relations. 
\section{Conclusion}
In this chapter, we focused on the capability of machine learning models to translate language-expressed spatial relations into 2D spatial arrangements between objects without constraints on their number and types.
% We empirically demonstrated that \blackgls{sr-bert} can effectively translate between linguistic descriptions and spatial configurations.

Concerning \inlineresearchquestion{Research Question 1}, our findings demonstrate that \blackgls{sr-bert} can effectively translate language-expressed spatial relations into 2D spatial arrangements. This capability extends across a diverse set of objects and relationship types, showcasing flexibility and robustness in handling scenarios without predefined constraints on the number of object or categories.

Regarding \inlinesubresearchquestion{Research Question 1.1}, we reveal that the nature of the spatial language---whether explicit (e.g., ``above'', ``left-of'') or implicit (e.g., ``wearing'', ``holding'')---does not significantly hinder the model's ability to accurately translate these relations into spatial arrangements.
% This underscores \gls{sr-bert}'s sophisticated understanding of spatial language, enabling it to decipher and implement both overt and subtle spatial cues.

Concerning \inlinesubresearchquestion{Research Question 1.2}, our experiments indicate that \blackgls{sr-bert} is robust w.r.t. spatial relations that are out-of-distribution (i.e., not encountered during training). This further attests to the potential of such models for real-world applications, where the models encounter out-of-distribution data at inference time.

\textbf{Limitations.} The first limitation of our approach is that it is restricted to 2D spatial understanding despite computer graphics data being frequently 3D. However, our approach can trivially be extended to work with 3D data. Second, we assume a fixed number of 126 clip-art categories and canvas of $500 \times 400$, which is a rather idealized setting considering that actual 2D/3D modelling is performed with a larger number of categories and larger scenes. Finally, scene layout generation is an ill-posed problem: i.e., there are multiple valid layouts conditioned on a scene description. Therefore, an ideal approach should not generate a deterministic layout, but rather address the uncertainty in both the models' output and the evaluation.

\textbf{What has happened since?} The majority of the work from this chapter dates back to early 2020, and therefore, numerous works since then have reported findings that corroborate the ones reported in the scope of this chapter. Specifically, works on text-to-image and -video generation using diffusion models have demonstrated that pre-trained \blackgls{llm}s encoders can sufficiently well encode the spatial knowledge present in language.

The work of Ramesh~\etal~\cite{ramesh2022hierarchical} (DALL-E) and Saharia~\etal~\cite{saharia2022photorealistic} (Imagen) have shown that for the task of text-to-image generation, large pre-trained language models (based on CLIP \cite{radford2021learning}) can encode the spatial knowledge from text. Namely, for a model to be successful in the task of image generation, it should generate an image that satisfies the spatial relations between the objects in the image (e.g., horse riding an astronaut). Finally, recent work even extends such an approach for video generation \cite{ho2022imagen}, a task requiring not only modelling both the spatial relations between the objects, but also maintaining and updating them temporally. 

%%%%%%%%%%%%%%%%%%%%%%%%%%%%%%%%%%%%%%%%%%%%%%%%%%
% Keep the following \cleardoublepage at the end of this file, 
% otherwise \includeonly includes empty pages.
\cleardoublepage

% vim: tw=70 nocindent expandtab foldmethod=marker foldmarker={{{}{,}{}}}
%

%
  \graphicspath{{\chaptersdir/conll2020/image/}}%
  %\cleardoublepage%
  \chapter{Translating Medical Text into Locations in a 3D Human Atlas}\label{ch:conll2020}
In this chapter, we focus on the problem of multimodal translation of medical texts to 3D locations in an interpretable space corresponding to a human atlas. To address the aforementioned problem, we
% develop a method for grounding medical text to the human atlas,
% We build on text embedding architectures such as \gls{bert}
% and
propose a novel loss function that allows the models to generate embeddings that capture the semantic and spatial relatedness of medical texts, by learning a projection of the text into a 3D space representing the human body. We quantitatively and qualitatively demonstrate that our proposed method learns context-sensitive and spatially-aware mapping in both the inter-organ and intra-organ sense.
% using large-scale medical text datasets.
% from the ``Large-scale online biomedical semantic indexing'' track of the 2020 BioASQ challenge.
Finally, we extend our approach to a self-supervised setting, and find it to be competitive with a fully supervised variant of itself.

The content of this chapter is based on the following publication:

\begin{mdframed}[backgroundcolor=gray!30,linewidth=0pt]
Grujicic*, D., \underline{Radevski*, G.}, Tuytelaars, T., Blaschko, M., (2020). Learning to ground medical text in a 3D human atlas. The Conference on Computational Natural Language Learning (CoNLL). \cite{grujicic-etal-2020-learning} (*authors contributed equally\footnote{Dusan started the project and outlined the task of translating medical text to locations in the 3D human atlas. Gorjan joined the project soon after, and the work was equally distributed: data processing, designing experiments, paper writing, etc.})
\end{mdframed}

% \begin{itemize}
%     \item Grujicic*, D., \underline{Radevski*, G.}, Tuytelaars, T., Blaschko, M., (2020). Learning to ground medical text in a 3D human atlas. The Conference on Computational Natural Language Learning (CoNLL). \cite{grujicic-etal-2020-learning} (*authors contributed equally)
% \end{itemize}

\section{Introduction}
The quantity of available medical literature increases daily \cite{wang2020cord, tsatsaronis2015overview}, however, it is often provided in a non-systematized, free form. The development of \textsc{Bert} \cite{devlin2018bert}, and the increased popularity of transfer learning in natural language processing (\blackgls{nlp}), prompted notable works that aim to leverage publicly available medical and scientific articles to develop (domain-specific) pre-trained language models \cite{lee2019biobert, alsentzer2019publicly, beltagy2019scibert, jin2019probing}. High-quality sentence representations that capture the semantics and structure of the text can be obtained by training models to solve the Natural Language Inference task on open-domain datasets \cite{bowman2015large, williams2017broad}, and predict whether two pieces of text entail, contradict, or are neutral to each other \cite{conneau2017supervised}. The aforementioned \textsc{Bert}-based models can serve as sentence encoders for such approaches \cite{reimers2019sentence}, and the setup can be trivially extended to enable searching through and retrieving relevant documents from large datasets.

\begin{figure}[t]
\centering
  \begin{minipage}[c]{0.6\columnwidth}
    \includegraphics[width=1.0\textwidth]{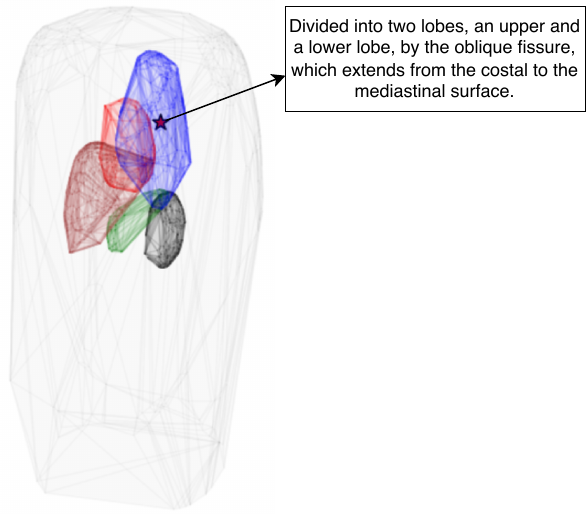}
  \end{minipage}\ \ 
  \begin{minipage}[c]{0.3\columnwidth}
    \caption[Medical text grounding]{Given the text with implicit reference to lungs \cite{drake2009gray}, a model trained with our method learns the grounding indicated by the \textit{star} symbol.}
    \label{conll2020:fig:sentence_grounding}
  \end{minipage}
\end{figure}

Despite proving useful in a variety of settings, these works suffer from the following limitations:
\begin{inparaenum}[(i)]
    \item the medical text is embedded in a non-interpretable space;
    \item there is no clear visually intuitive indication of how similar two retrieved pieces of text are;
    \item visualizing the embeddings requires dimensionality reduction techniques \cite{hotelling1933analysis, maaten2008visualizing}.
\end{inparaenum}

In contrast, in this chapter, we strive to answer the following research question:

\researchquestion{Research Question 2}{Can we train models that translate medical texts to 3D locations within a human anatomical atlas, leveraging the spatial co-occurrence of medical terms to create meaningful grounding?}

We further assess the model's generalization in terms of handling out-of-distribution data: i.e., novel locations in the 3D human atlas outside of the training dataset. This capability is important for models to remain applicable and support medical professionals without frequent retraining. To gain insights, we explore:

\subresearchquestion{Research Question 2.1}{How does the model perform in grounding medical texts to locations within the human body that were not part of the training dataset? In other words, human organs that are never referred to in the training data.}

In order to reduce the reliance on data annotated by medical practitioners, next, we explore whether such models can be trained in a self-supervised manner. Since annotated medical datasets are expensive to create, self-supervised training significantly lowers the barriers to developing advanced medical visualization tools. Consequently, we ask:

\subresearchquestion{Research Question 2.2}{Is it feasible to train medical text grounding models in a self-supervised manner, thereby eliminating the dependency on human-annotated data?}

\textbf{Contributions.} We propose a method that utilizes the natural tendency of functionally similar organs to be located near each other in the human body.
%, making it easier to navigate and interpret.
By taking advantage of this inductive bias, we can train models which able to generate compact 3D medical text representations (Figure~\ref{conll2020:fig:sentence_grounding}), that are as effective as the traditional, higher dimensional text embeddings.
% Furthermore, our method enables the searching and retrieval of documents that are contextually linked to specific locations within the human body's physical space.
More concretely, in response to the research questions at hand, we contribute the following:

\circled{1} We propose the task of grounding medical text in the physical space of the human body, where anatomically related substructures tend to be close to one another.

\circled{2} We design a novel loss function that allows us to reason about the semantic relatedness of medical texts.

\circled{3} We explore a specific use case for medical text retrieval, where our proposed method outperforms several competitive baselines.
% We perform an extensive evaluation to measure the performance of our method in two scenarios, namely, grounding in the human atlas (relevant for visualization and navigation), and medical text-to-text retrieval (directly assessing the performance of our model in an information retrieval setting). Furthermore, we set up an experimental setting explicitly tailored to measure the spatial reasoning ability of our model within an organ, a setting never directly imposed during training. We empirically demonstrate that our method is highly successful in all aforementioned experimental settings, effectively addressing the previously stated limitations.

We release code, data, and trained models at: \url{www.github.com/gorjanradevski/text2atlas}.
\section{Related Work}
Since this work came to fruition in early 2020, in this section we discuss \textit{only} the related and concurrent works at the time of publication.

\textbf{(Medical) text embeddings.} Before the development of \textsc{Bert}, a common approach to obtain text embeddings was to leverage a pre-trained recurrent neural network (RNN) language model (LM) \cite{peters2018deep, kiros2015skip}. An extension of such LM for the biomedical domain is BioELMO \cite{jin2019probing}. Despite being successful in a transfer learning setting, the usefulness of the generated embeddings for medical text navigation and retrieval is arguably limited. Furthermore, \textsc{Bert}-based medical language representation models such as \textsc{BioBert} \cite{lee2019biobert} and \textsc{ClinicalBert} \cite{alsentzer2019publicly}, despite outperforming RNN-based LMs on a variety of downstream tasks, also make embedded text navigation impractical. On the other hand, we directly focus on learning embeddings that are rich with visual information; i.e., medical text embeddings that are by default represented in a physically meaningful space of the human body.

\textbf{(Medical) text grounding.} There has been a variety of approaches \cite{krishnamurthy2013jointly, kong2014you, rohrbach2016grounding, hu2016natural, wang2018learning} and datasets, such as ReferIt \cite{kazemzadeh2014referitgame} and RefCOCO \cite{yu2016modeling}, focusing on general-purpose visual grounding of natural language. However, the applications dealing with medical text grounding have been limited, and to the best of our knowledge, there are no works that ground medical text in the human body. The main differences between works performing visual grounding and our work are:
\begin{inparaenum}[(i)]
    % \item we perform a grounding which is universal, and not specific to a single environment (e.g. the image),
    \item their models are trained with expensive bounding box annotations for the desired grounding location,
    \item the methods rely on explicit annotations of every concept referred to in the text; i.e., these models can only ground text referring to objects from the training set.
\end{inparaenum}
Finally, our method is designed to reduce labeling costs, as it relies on high-level annotations of the referred organs in a paragraph, which, in a self-supervised setting---as we show in our experiments---can be inferred from the text itself.

\textbf{(Medical) document retrieval.} Retrieving a set of relevant documents given a query requires that both the query and the documents are embedded in a joint latent space. A straightforward approach to obtaining a single text representation is to use the \texttt{[CLS]} token representation concatenated with the mean-pooled and max-pooled representations of the remaining tokens from a pre-trained \textsc{Bert} model. However, it is shown that this often leads to a worse representation than averaging GloVe embeddings \cite{pennington2014glove, reimers2019sentence}. Recently, such embeddings are obtained using a pre-trained \textsc{Bert} model which is subsequently fine-tuned
% as a Siamese model
\cite{reimers2019sentence} on the NLI task using general-purpose natural language datasets \cite{bowman2015large, williams2017broad}. The proxy task is proven to be effective as models trained this way generate embeddings in which documents that share similar semantics close in the representation space. Despite the useful feature, without inspecting the documents' content, it is not immediately obvious why a set of documents is clustered together in the representation space, and why they are considered to be semantically similar. We address this issue by translating the documents into a 3D physical space corresponding to the human body, where the document similarity is expressed in terms of physical proximities. This leads to an intuitive interpretation of why a set of documents are considered to be similar.
\section{Data Collection}

\subsection{Human Body Atlas}
We leverage the \textbf{S}egmented \textbf{I}nner \textbf{O}rgans (SIO) \cite{pommert2001creating}
% (see Appendix, Figure~\ref{conll2020:appendix:fig:voxelman}),
though our approach can be readily extended to other models of the human anatomy. We refer to this 3D model as the \textit{atlas}. We base the 3D atlas on the segmentation labels of the tissues in the human body provided in SIO, which come in the form of image slices that form a 3D voxel model of the male torso when stacked on top of one another. The stacked images from the torso represent a volume of $573 \times 330 \times 774$ voxels, with a 1-millimeter resolution along each axis. The value of each voxel represents the segmentation label of its corresponding organ or tissue. Therefore, an organ can be represented as the set of indices of voxels in the aforementioned volume, which contain the value corresponding to the organ's segmentation label.
% Details about the creation of the 3D human atlas can be found in the Appendix~\S\ref{conll2020:appendix:sec:human_atlas}.

\subsection{Dataset}
The dataset used in this work is built upon the training set of Task 8a: ``Large-scale online biomedical semantic indexing'' of the 2020 BioASQ challenge \cite{tsatsaronis2015overview}. Originally, it consisted of 14,913,939 samples, where each sample pertains to one medical article, and contains the abstract text and the Medical Subject Headings (\blackgls{mesh}) \cite{lipscomb2000medical} vocabulary terms of the organs. We consider the grounding of article abstracts to the locations in the atlas that correspond to the article's MeSH terms. Therefore, we use articles that contain one or more MeSH terms that match the names or the alias terms of the organs in the atlas glossary. To accommodate the maximal sequence length of \textsc{Bert\textsubscript{Base}}, we keep the articles whose abstracts have fewer than 512 WordPiece \cite{wu2016google} tokens. For each organ in the atlas glossary, we take 500 articles that mention it individually, and another 500 articles that mention it in addition to other organs. Subsequent removal of duplicates resulted in the final dataset of 25,552 abstracts annotated with organ MeSH terms, of which we use 70\% for training, 15\% for validation, and 15\% for testing.
% \footnote{Additional details about the dataset found in the Appendix Section~\ref{conll2020:appendix:sec:dataset}.}
\section{Proposed Task and Methods}
\subsection{Text-to-Atlas Grounding Objective}
Our goal is to ground, or translate, medical texts into 3D locations within a human body atlas. To achieve this, we learn a projection of the medical text, referring to one or more atlas organs, into the 3D volume in the atlas that corresponds to the mentioned organs. The volume of each organ is characterized by a set of voxels in the atlas, which capture its position, size, and shape. The voxels of one organ can, in turn, be represented by a point cloud in 3D space, where each point represents the coordinate indices of one voxel\footnote{For brevity, we also use the term voxel or organ point for a vector of indices of the actual voxel in the 3D volume.}.

The most straightforward way to associate texts with predefined regions of a physical space is to train a model that minimizes the cross-entropy between the predicted probability distribution over the set of all organs (e.g., each indexing their predefined location), and the target vector with 1's at the indices corresponding to the target organs, and 0's elsewhere. This approach is expected to yield a high accuracy of selecting the right organ, however, it has a critical downside of not providing any meaningful within organ reasoning; i.e., during inference, it grounds all articles pertaining to a single organ to either a random location within the organ, or a single predefined one. On the other hand, framing the task as a minimization of the mean squared error between the predicted 3D location and an average of the ground truth organ positions would result in grounding to some mid-way location, potentially belonging to some other, unrelated organ.

To overcome both of these issues and retain as much of the predictive power as possible, we frame the task as regressing a set of 3D coordinates within the human body, while forcing the prediction to snap to the most nearby ground truth organ. Namely, we design a loss function---Soft Organ Distance loss, henceforth abbreviated as \blackgls{sod} (\S\ref{conll2020:sec:loss})---which gives the model freedom to choose the most relevant organ in case there are multiple organ annotations for a particular medical document. 

\subsection{Model}
We use \textsc{Bert} \cite{devlin2018bert} as our model backbone due to its applicability in a wide range of domains. As per Devlin~\etal~\cite{devlin2018bert}, we tokenize the input text using WordPiece \cite{wu2016google}, and select the representation of the \texttt{[CLS]} token as the sequence representation. Finally, to obtain the 3D atlas grounding for a piece of medical text, we project \textsc{Bert}'s output with a linear layer, mapping from \textsc{Bert}'s hidden space to the 3D human-atlas space: $\mathbf{\hat{y}} = \operatorname{Linear}(\text{\textsc{Bert}}(\mathbf{x}))$, where $\mathbf{x}$ is a vector of tokens representing the medical text and $\mathbf{\hat{y}}$ is the 3D grounding in the human body. During training, we normalize $\mathbf{\hat{y}}$ by applying $\tanh$, which is subsequently re-scaled to the dimensions of the atlas during inference.

\subsection{Soft Organ Distance Loss}\label{conll2020:sec:loss}
The proposed loss function---\blackgls{sod}---allows us to sacrifice the least amount of predictive power and in turn, achieve within organ contextual reasoning; i.e., not only grounding the medical article to the right organ but also to the appropriate location within the organ without any explicit annotations at that level of granularity. Furthermore, a medical text may simultaneously refer to a single or multiple organs in the human body. In the former setting, we would like to have an approach based on mean squared error minimization, while in the latter, we would like to relax the target and pull the model's prediction to the location of the closest ground truth organ. Finally, the organs themselves are distributed in nature, and their volumes are characterized by a set of points in 3D space, rather than just one.\par

\begin{figure}[t]
\centering
\includegraphics[width=0.7\textwidth]{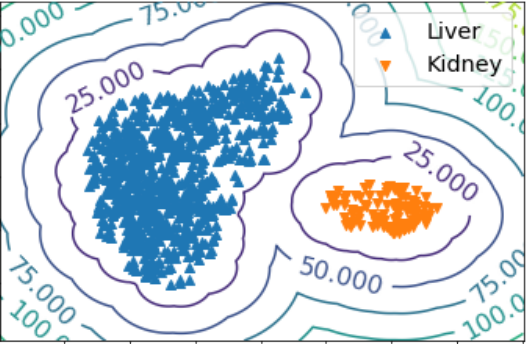}
\caption[Soft Organ Distance loss function isocurves]{Soft Organ Distance loss function isocurves around ``liver'' and ``kidney'' point clouds projected into 2D with PCA \cite{hotelling1933analysis}.}
\label{conll2020:fig:loss_isocurves_kidney_lung}
\end{figure}

In Figure~\ref{conll2020:fig:loss_isocurves_kidney_lung}, we observe our desired scenario when there are two ground truth organs: ``liver'' and ``kidney''. As the grounding approaches the ``liver'', we observe that the loss contribution from the ``kidney'' organ voxels diminishes, and vice versa. This effect extends to the loss contributions of individual voxel points. Namely, as the grounding approaches a particular region in the organ, the loss contribution from the other voxel points diminishes, thus allowing the model to ground the input text within the most appropriate organ substructure.

To account for the distributed nature of the organs and take a step towards the desired within organ semantic reasoning, for each sample during training, we randomly sample a set of $N$ points from the point cloud of each of its organs. Then, we calculate
\begin{inparaenum}[(1)]
    \item the Euclidean distances between the prediction and each sampled organ point, and
    \item the soft-min\footnote{Soft-max on the inputs reversed in sign, used to emphasize smaller quantities - in this case, shortest distances.} across these distances as weights for the contributions of individual points.
\end{inparaenum} 
The loss contribution $\mathcal{L}_{p}$ of a 3D organ point $\mathbf{y}$ is the product of its distance from the predicted point $\mathbf{\hat{y}}$ and its corresponding weight produced by the soft-min: 

\begin{equation}
    \mathcal{L}_{p} = \|\mathbf{\hat{y}} - \mathbf{y}\|_2 \frac{\exp(-\|\mathbf{\hat{y}} - \mathbf{y}\|_2/\gamma_p)}{\sum_{i=1}^N \exp(-\|\mathbf{\hat{y}} - \mathbf{y_i}\|_2/\gamma_p )} ,
\end{equation}

where $N$ is the number of points sampled from the organ point cloud and $\gamma_p$ is a temperature term. We calculate the loss for one organ $\mathcal{L}_{o}$ as the sum of contributions of its points: 
$    \mathcal{L}_{o} = \sum_{i=1}^N{\mathcal{L}_{p}^i}    $.

We calculate the loss for each target organ in the way described above. Then, we compute the soft-min over the set of such loss terms as contribution weights for each organ. The total loss is the sum of soft-min-weighted losses over each organ:
\begin{equation}
    \mathcal{L}_{t} = \sum_{i=1}^M \mathcal{L}_{o}^i \frac{\exp(-{\mathcal{L}_{o}^i}/\gamma_o)}{\sum_{j=1}^M \exp(-{\mathcal{L}_{o}^j}/\gamma_o)} ,
\end{equation}
where $M$ is the total number of target organs, $\mathcal{L}_{o}^i$ is the organ loss for the $i$-th organ, and $\gamma_o$ is a temperature term.
\section{Evaluation and Discussion}

\textbf{Experiments overview.} We quantitatively evaluate our trained models in two scenarios:
\begin{inparaenum}[(i)]
    \item Grounding to the human atlas: measuring to what extent models trained with our proposed loss function can ground medical articles to the correct locations.
    \item Medical information retrieval: directly assessing the quality of the medical document embeddings; i.e., evaluate to what extent medical articles characterized by a certain set of MeSH terms are grouped in the physical space of the human body.
\end{inparaenum}

\subsection{Grounding to the Human Atlas}
To evaluate the quality of the grounding/translation, we measure the performance of each of the models using three evaluation metrics.
% (more details in Appendix~\ref{conll2020:appendix:sec:metrics}):

\begin{inparaenum}[(i)]
\item Rate at which the texts are grounded within, or sufficiently close\footnote{As some organs are hollow (small intestine, colon, etc.), we record a ``hit'', when the grounding is less than 1cm away from the most nearby voxel.}, to the volume of the correct organ. The hit rate, which we denote as Inside Organ Ratio (\blackgls{ior}) and expressed as percentage points.
\item Distance to the nearest voxel of the nearest correct organ, denoted as Nearest Voxel Distance (\blackgls{nvd}), expressed in centimeters.
\item Distance to the nearest voxel of the nearest correct organ, calculated only on samples for which the projection is outside the organ volume, denoted as Nearest Voxel Distance Outside (\blackgls{nvdo}) expressed in centimeters.
\end{inparaenum}

We compute the aforementioned metrics in four distinct evaluation scenarios specifically tailored to measure the grounding ability of our models. In the following experiments, we show that our approach has an advantage over multiple baselines, and we demonstrate its ability to reason within the substructures of the organ and generalize to out-of-atlas organs.
% in addition to its other desirable properties that we demonstrate qualitatively.
The error bars represent the standard error.

\subsubsection{Notes on the metrics}
For the IOR, the predicted 3D point lies inside the organ volume (\textit{hit}) when its coordinates, rounded to the nearest integer to represent voxel indices, are within the set of indices of voxels that make up the corresponding organ.  However, in the case of hollow organs, such as the intestines and the stomach, scoring a \textit{hit} would require predicting a point that lies exactly within a (usually very thin) organ wall, as the region of the organ's lumen is typically not included. Therefore, we use a more relaxed criterion, and record a \textit{hit} when the predicted 3D point is inside or sufficiently close to the organ volume, which we consider to be the case when its coordinates are less than 1cm away from the nearest voxel of the target organ. In cases of multiple target organs, we measure a \textit{hit} when the predicted coordinates lie within or sufficiently close to any one of the given organs.

When the projection is exactly inside the volume of the organ, the NVD is zero, and otherwise, it is measured as the distance to the surface of the nearest organ in the text. The NVD-O metric complements the NVD metric, such that it gives insight into how far off the prediction is when it misses the correct organ.

\subsubsection{General Human Atlas Grounding}\label{conll2020:sec:full-test-set}
We generate a 3D grounding for each of the articles in the test set and measure the performance of our proposed loss function---\blackgls{sod}---against the following baselines:

\begin{inparaenum}[(i)]
    \item \textbf{Random}: We predict a randomly sampled point within a randomly chosen organ for each sample.
    \item \textbf{Center}: We use the center of the 3D atlas as the prediction.
    \item \textbf{Frequency}: We measure the frequency of the organ terms in the training set, and always predict the point within the most frequent organ.
    \item \textbf{Mean Squared Error Regressor}: We train a regression baseline model by minimizing the mean squared error between the prediction and the average of a set of randomly sampled points from all the target organs.
    \item \textbf{Classifier}: We frame the task as classification and train a model to predict an organ index. The model is trained to minimize the cross-entropy between the output probability distribution and the target vector, with 1's at the positions corresponding to the indices of organs present in the text and 0's elsewhere. During evaluation, the prediction is considered to be correct when it corresponds to any one of the target \blackgls{mesh} terms. When measuring \blackgls{nvd} and NVD-O, we select the average voxel point from the predicted organ as a 3D grounding.\footnote{We do not use the organ voxels centroids as prediction, as they can be outside of the organ volume for non-convex organs and yield a non-zero distance even when the correct organ is predicted, unfairly penalizing the classification baseline.}
\end{inparaenum}

\begin{table}[t]
\centering
\resizebox{0.7\textwidth}{!}{
\begin{tabular}{lccc} \toprule
{Method} & {IOR ($\uparrow$)} & {NVD ($\downarrow$)} & {NVD-O ($\downarrow$)} \\ \midrule
Random & 8.9 $\pm$ 0.5 & 17.9 $\pm$ 0.3 & 19.0 $\pm$ 0.3 \\
Center & 6.7 $\pm$ 0.4 & 13.3 $\pm$ 0.2 & 13.3 $\pm$ 0.2 \\
Frequency & 10.9 $\pm$ 0.5 & 13.9 $\pm$ 0.2 & 15.4 $\pm$ 0.2 \\
Mean Squared Error Regressor & 9.9 $\pm$ 0.5 & 6.8 $\pm$ 0.1 & 7.0 $\pm$ 0.1 \\ 
Classifier & \textbf{90.8 $\pm$ 0.5} & 0.9 $\pm$ 0.1 & 8.2 $\pm$ 0.5 \\
Soft Organ Distance (SOD) & 89.4 $\pm$ 0.5 & \textbf{0.8 $\pm$ 0.1} & \textbf{2.5 $\pm$ 0.1} \\
\bottomrule
\end{tabular}
}
\caption[Comparison against baseline for grounding to the human atlas]{IOR, NVD, and NVD-O, of our proposed method---Soft Organ Distance---and the baselines for grounding to the human atlas.}
\label{conll2020:table:full-test-set}
\end{table}

In Table \ref{conll2020:table:full-test-set}, we observe that \blackgls{sod} outperforms all baselines, and achieves nearly the same IOR as the Classifier baseline. Furthermore, \blackgls{sod} significantly outperforms the Classifier baseline according to the NVD and NVD-O metrics, which give a strong indication of the overall grounding performance, as per the one-sided Wilcoxon signed-rank test \cite{wilcoxon1945individual} ($p \approx 0$). We conclude that despite framing the task as soft regression, we sacrificed the least amount of predictive power (as per IOR), and exploited the atlas's inductive bias to achieve successful grounding (as per NVD and NVD-O).

\subsubsection{Within Organ Reasoning}\label{conll2020:sec:within-organ}
We perform a simulation to demonstrate that models trained with SOD can translate medical text to anatomical substructures not present at the granularity of labeling in a specific atlas. For that reason, we perform experiments in which we merge the voxels of two different organs---effectively treating them as a single organ---and keep only instances from the training set that contain these ``super-organs''\footnote{These super-organs may occur individually or co-occur with other organs.}. Then, we train a new model on each of these subsets and subsequently generate 3D groundings for each of the test set samples that only contain the individual occurrences of the two merged organs. The merged organ pairs are:
\begin{inparaenum}[(i)]
    \item the ``lung'' and the ``stomach'', functionally different organs that belong to different groups, respiratory and digestive, respectively; and
    \item the ``duodenum''\footnote{The duodenum is the first section of the small intestine in most higher vertebrates, including mammals.
    } and the ``small intestine'', organs which are functionally related and frequently jointly referred to as ``small intestine'' in the medical literature.
\end{inparaenum}
 
Then, we train three different models for each merger:
\begin{inparaenum}[(i)]
    \item \textbf{Classifier}: We train a classification baseline on each of the subsets. During inference, we substitute the organ index with the average voxel point within the predicted organ.
    \item \textbf{Soft Organ Distance; Unmerged}: We train a model using \blackgls{sod} with the individual organs from the filtered training set.
    \item \textbf{Soft Organ Distance; Merged}: We train a model with the \blackgls{sod} loss function, where the two organs are merged into one ``super-organ'', effectively training without the per-organ annotations.
\end{inparaenum}

% \begin{table}[t]
% \centering
% \resizebox{\textwidth}{!}{
% \begin{tabular}{lcccccc} \toprule
%     {Method} & {IOR} & {NVD} & {IOR} & {NVD} & {IOR} & {NVD} \\ \midrule
%     {} & \multicolumn{2}{c}{Lung} & \multicolumn{2}{c}{Stomach} & \multicolumn{2}{c}{Total} \\ \midrule
%     \blackgls{sod} w/ & 100.0 $\pm$ 0.0 & 0.0 $\pm$ 0.0 & 95.4 $\pm$ 2.6 & 0.2 $\pm$ 0.1 & 97.8 $\pm$ 1.3 & 0.1 $\pm$ 0.0 \\ \midrule
%     SMP & 46.5 $\pm$ 6.0 & 2.5 $\pm$ 0.3 & 44.6 $\pm$ 6.2 & 4.2 $\pm$ 0.6 & 45.6 $\pm$ 4.3 & 3.3 $\pm$ 0.3 \\
%     \blackgls{sod} w/o & \textbf{94.4 $\pm$ 2.8} & \textbf{0.2 $\pm$ 0.1} & \textbf{81.5 $\pm$ 4.8} & \textbf{0.5 $\pm$ 0.1} & \textbf{88.2 $\pm$ 2.8} & \textbf{0.3 $\pm$ 0.1} \\
%     \midrule
%     {} & \multicolumn{2}{c}{Small intestine} & \multicolumn{2}{c}{Duodenum} & \multicolumn{2}{c}{Total} \\ \midrule
%     \blackgls{sod} w/ & 97.4 $\pm$ 1.8 & 0.1 $\pm$ 0.0 & 90.9 $\pm$ 3.6 & 0.2 $\pm$ 0.1 & 94.4 $\pm$ 1.9 & 0.2 $\pm$ 0.0 \\ \midrule
%     SMP & \textbf{71.1 $\pm$ 5.2} & \textbf{1.0 $\pm$ 0.2} & 33.3 $\pm$ 5.8 & 5.1 $\pm$ 0.6 & 53.5 $\pm$ 4.2 & 2.9 $\pm$ 0.3 \\
%     \blackgls{sod} w/o & 50.0 $\pm$ 5.8 & 1.1 $\pm$ 0.1 & \textbf{93.9 $\pm$ 3.0} & \textbf{0.2 $\pm$ 0.0} & \textbf{70.4 $\pm$ 3.8} & \textbf{0.7 $\pm$ 0.1} \\ \bottomrule
% \end{tabular}
% }
% \caption{Within organ reasoning evaluated on test set subsets obtained according to choice of organs merged.}
% \label{conll2020:table:merged-organs}
% \end{table}

\begin{table}[t]
\centering
\resizebox{\textwidth}{!}{
\begin{tabular}{lcccccc} 
\toprule
Method & IOR ($\uparrow$) & NVD ($\downarrow$) & IOR ($\uparrow$) & NVD ($\downarrow$) & IOR ($\uparrow$) & NVD ($\downarrow$) \\ \midrule
{} & \multicolumn{2}{c}{Lung} & \multicolumn{2}{c}{Stomach} & \multicolumn{2}{c}{Micro-average} \\ 
\cmidrule(r){2-3} \cmidrule(lr){4-5} \cmidrule(l){6-7}
Soft Organ Distance (SOD); Unmerged & 100.0 $\pm$ 0.0 & 0.0 $\pm$ 0.0 & 95.4 $\pm$ 2.6 & 0.2 $\pm$ 0.1 & 97.8 $\pm$ 1.3 & 0.1 $\pm$ 0.0 \\
Classifier & 46.5 $\pm$ 6.0 & 2.5 $\pm$ 0.3 & 44.6 $\pm$ 6.2 & 4.2 $\pm$ 0.6 & 45.6 $\pm$ 4.3 & 3.3 $\pm$ 0.3 \\
Soft Organ Distance (SOD); Merged & \textbf{94.4 $\pm$ 2.8} & \textbf{0.2 $\pm$ 0.1} & \textbf{81.5 $\pm$ 4.8} & \textbf{0.5 $\pm$ 0.1} & \textbf{88.2 $\pm$ 2.8} & \textbf{0.3 $\pm$ 0.1} \\ \midrule
{} & \multicolumn{2}{c}{Small intestine} & \multicolumn{2}{c}{Duodenum} & \multicolumn{2}{c}{Micro-average} \\ 
\cmidrule(r){2-3} \cmidrule(lr){4-5} \cmidrule(l){6-7}
Soft Organ Distance (SOD); Unmerged & 97.4 $\pm$ 1.8 & 0.1 $\pm$ 0.0 & 90.9 $\pm$ 3.6 & 0.2 $\pm$ 0.1 & 94.4 $\pm$ 1.9 & 0.2 $\pm$ 0.0 \\
Classifier & \textbf{71.1 $\pm$ 5.2} & \textbf{1.0 $\pm$ 0.2} & 33.3 $\pm$ 5.8 & 5.1 $\pm$ 0.6 & 53.5 $\pm$ 4.2 & 2.9 $\pm$ 0.3 \\
Soft Organ Distance (SOD); Merged & 50.0 $\pm$ 5.8 & 1.1 $\pm$ 0.1 & \textbf{93.9 $\pm$ 3.0} & \textbf{0.2 $\pm$ 0.0} & \textbf{70.4 $\pm$ 3.8} & \textbf{0.7 $\pm$ 0.1} \\
\bottomrule
\end{tabular}
}
\caption[Within organ reasoning experiments]{IOR and NVD reported for within organ reasoning on subsets of the test set, which we obtain according to the choice of organs we merge.}
\label{conll2020:table:merged-organs}
\end{table}

\begin{figure}[t]
\centering
\includegraphics[width=0.7\textwidth]{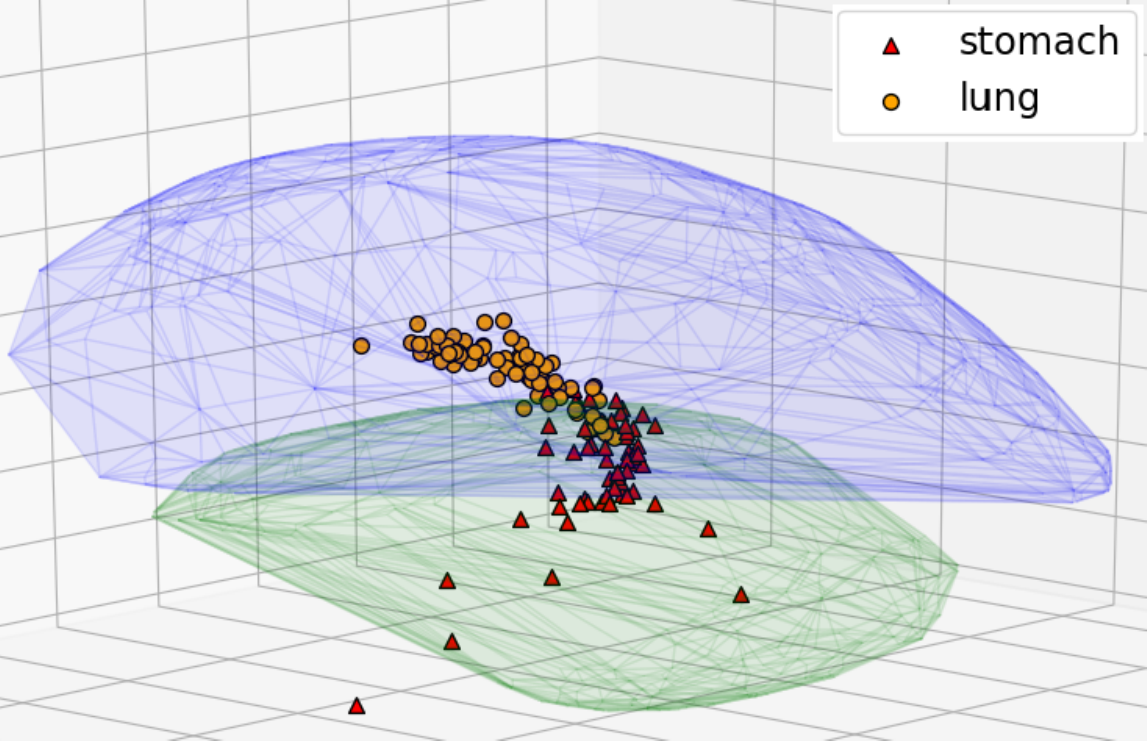}
\caption[Lung and stomach article grounding from a model trained on merged organs]{Grounding of medical articles referring to either the ``lung'' or the ``stomach''; obtained from a model trained with the two organs merged into one.}
\label{conll2020:fig:superorgan_lung_stomach}
\end{figure}

With the functionally different ``lung'' and ``stomach'' merged, in Table \ref{conll2020:table:merged-organs}, we observe that models trained with \blackgls{sod} significantly outperform the Classifier baseline, which can only predict the coarse label corresponding to the super-organ, but is unable to reason about the organ's subregions. We also observe that the performance of \blackgls{sod}; Merged, is relatively close to \blackgls{sod}; Unmerged, which is trained with the separated organs. Additionally, in Figure~\ref{conll2020:fig:superorgan_lung_stomach}, we observe the grounding of 136 articles related to the ``lung'' and the ``stomach'' generated with \blackgls{sod}; Merged. Despite being trained without per-organ annotations, the model infers the appropriate substructure where the medical articles should be grounded.

A notably harder problem is the ``small intestine''-``duodenum'' merger, which involves functionally related organs. We again observe that \blackgls{sod}; Merged significantly outperforms the Classifier baseline in both the micro-averaged performance and the grounding within the ``duodenum''. The Classifier baseline achieves higher IOR on the articles that belong to the ``small intestine,'' which is a result of the roughly 3 times larger number of ``small intestine'' voxels compared to the ``duodenum'' making the Classifier performance skewed.

Finally, we examine the approximate locations of anatomical structures that co-occur most frequently with a given organ. For the organs co-occurring with the ``lung'', the frequency-weighed arithmetic mean of their centroids lies roughly 13.8 centimeters above that of the organs that co-occur with the ``stomach''. Similarly, such mean location of organs co-occurring with the ``duodenum'' lies 6.5 centimeters to the upper-left of the ``small intestine''.

We conclude that despite the lack of explicit within organ annotations, with \blackgls{sod} we can train models that spatially reason about substructures within an organ based on the target organs' co-occurrences. In particular, the model learns to disambiguate between the organ regions because terms associated with different sub-regions tend to co-occur with different organs, typically the ones to which they are closer.
% (see Appendix \S\ref{conll2020:appendix:sec:coocurrences}).
This is an important observation from the following aspects:
\begin{inparaenum}[(i)]
    \item Medical articles would get mapped to the appropriate organ regions they refer to, even though they were never explicitly annotated as such during training;
    \item given an atlas with increased granularity, our method would, to a large degree, accommodate the newly added sub-regions without the need for re-training.
\end{inparaenum}

Altogether, we shed light on \inlineresearchquestion{Research Question 2}: we can effectively train models that leverage the spatial co-occurrence of medical terms to accurately translate medical text to specific 3D locations in a human atlas.

\subsubsection{Generalization to Unseen Organs}\label{conll2020:sec:out-of-sample-organs}
% We next experiment whether our approach the locations of organs that are absent in the atlas segmentation labels, we evaluate the generalization ability of our method to organs unseen during training.
Next, we perform experiments to test whether our approach can generalize to out-of-distribution data; that is, regions in the human atlas that are never referred to by the medical documents from the training data. To test this phenomenon, for every organ, we remove its annotation from the training set, train a separate model, and perform inference on the test set documents referring only to that corresponding organ. Finally, we report the metrics averaged over all held-out organs.\footnote{Samples referring solely to the held-out organ are removed, and the ones referring to it in addition to some other organs retain only the annotations of the other organs.} In addition to NVD, we measure the rate at which the prediction is within the convex hull enveloping the organs of the same functional group as the held-out organ, denoted as Inside Group Ratio---\blackgls{igr}. We compare \blackgls{sod}'s performance against Random, Center, and Classifier, defined in \S\ref{conll2020:table:full-test-set}.

\begin{table}[t]
\centering
\resizebox{0.7\textwidth}{!}{
\begin{tabular}{lcc} \toprule
Method & IGR ($\uparrow$) & NVD ($\downarrow$) \\ \midrule
Random & 34.5 $\pm$ 4.0 & 21.1 $\pm$ 1.5 \\
Center & 37.0 $\pm$ 9.5 & 15.8 $\pm$ 2.0 \\
Classifier & 72.1 $\pm$ 5.8 (84.21) & 8.3 $\pm$ 1.0 (6.9)  \\
Soft Organ Distance (SOD) & \textbf{76.5 $\pm$ 5.3 (86.15)} & \textbf{7.5 $\pm$ 1.0 (5.9)}  \\
\bottomrule
\end{tabular}
}
\caption[Experiments for grounding to organs held-out during training]{IGR and NVD for grounding of medical articles which refer to organs held out during training. Median values in parentheses.}
\label{conll2020:table:missing-organs}
\end{table}

In Table \ref{conll2020:table:missing-organs}, we observe that \blackgls{sod} significantly outperforms the other baselines. We also confirm a significant advantage of \blackgls{sod} over the Classifier baseline by performing a Wilcoxon signed-rank test (IGR: $p=0.0063$; NVD: $p=0.0014$). Therefore, we conclude that besides grounding texts regarding organs present in the atlas, models trained with \blackgls{sod} can reason about unannotated structures. More concretely, the model learns to leverage the shared context between the held-out organ and the functionally similar organs in its vicinity. Consequently, we conclude that models trained with \blackgls{sod} learn to relate the articles' context with the spatial domain of the human body, and exploit this knowledge to improve generalization in an out-of-distribution setting \inlinesubresearchquestion{Research Question 2.1}.
% . This suggests that our approach is robust to the granularity of the atlas used in training \inlinesubresearchquestion{Research Question 2.1}.

\subsubsection{Self-Supervised Human Atlas Grounding}\label{conll2020:sec:self-supervised}
We additionally evaluate our method in a self-supervised setting. Specifically, we ground medical abstracts in the atlas using only self-supervision in the form of (co-)occurrences of organ-related terms. For that, we aggregate a list of all organ names corresponding to the \blackgls{mesh} terms, together with their Unified Medical Language System synonyms \cite{bodenreider2004unified}. During training, instead of providing the ground truth MeSH term annotation as target organs, we provide the target organs that correspond to the elements of the aggregated list of organ terms that appear in the abstract. Then, we train the regular variant---\blackgls{sod}---of our loss function, and an additional one---\blackgls{sod} + Masking---where stochastically (with 50\% probability) substitute the occurrences of the organ names or their synonyms in the text with a \texttt{[MASK]}\footnote{The \texttt{[MASK]} token is included in \blackgls{bert}'s vocabulary.} token.
% \begin{inparaenum}[(i)]
%     \item \textbf{Soft Organ Din}: A model trained with our regular \blackgls{sod} loss function.
%     \item \textbf{SOD + M}: During training, we 
% \end{inparaenum}

We evaluate the performance of our method against the following baselines:
\begin{inparaenum}[(i)]
    \item \textbf{Random}: A naive model that predicts one of the organ names that appear in the text at random. When there is no explicit organ occurrence, it predicts the center of the atlas.
    \item \textbf{Classifier}: A classification baseline, trained to predict one of the organ name occurrences from the text.
    \item \textbf{Classifier + Masking}: A classification baseline boosted with the ``masking'' extension.
\end{inparaenum}
Finally, we perform inference on the annotated test set and measure the IOR, NVD, and NVD-O.

\begin{table}
\centering
\resizebox{0.8\textwidth}{!}{
\begin{tabular}{lccc} \toprule
Method & IOR ($\uparrow$) & NVD ($\downarrow$) & NVD-O ($\downarrow$) \\ \midrule
Random & 68.7 $\pm$ 0.7 & 3.2 $\pm$ 0.1 & 9.7 $\pm$ 0.3 \\ 
Classifier & 74.1 $\pm$ 0.7 & 2.5 $\pm$ 0.1 & 8.4 $\pm$ 0.3 \\
Classifier + Masking & 80.4 $\pm$ 0.6 & 1.7 $\pm$ 0.1 & 7.3 $\pm$ 0.3 \\
Soft Organ Distance (SOD) & 79.7 $\pm$ 0.7 & 1.6 $\pm$  0.1 & 5.1 $\pm$ 0.2 \\
Soft Organ Distance (SOD) + Masking & \textbf{83.2 $\pm$ 0.6} & \textbf{1.2 $\pm$ 0.1} & \textbf{3.9 $\pm$ 0.2} \\
\bottomrule
\end{tabular}
}
\caption[Self-supervised medical text grounding experiments]{IOR, NVD and NVD-O with models that are trained in a self-supervised fashion for medical text grounding.}
\label{conll2020:table:self-supervised}
\end{table}

In Table \ref{conll2020:table:self-supervised}, we observe that our method outperforms all baselines when trained both without (SOD), and with the masking extension (SOD + Masking). Since masking the organ names and their synonyms puts additional emphasis on their surrounding context, it allows the model to generalize better to the semantically annotated test set, yielding a considerable improvement for all metrics. It is noteworthy that despite training the model using organ names + synonym occurrences that appear within the medical articles as ground truth targets, we obtain performance competitive to the fully supervised training, included in Table \ref{conll2020:table:full-test-set}. This data efficiency feature of our method is especially important since obtaining annotated data for medically relevant NLP tasks requires the time and effort of medical experts \inlinesubresearchquestion{Research Question 2.2}.

\subsection{Medical Information Retrieval}

We formulate a text-to-text retrieval setting where each test set article serves as a query and the remaining articles as the database from which we retrieve the relevant ones. We measure the retrieval quality using the standard Recall@K metric; i.e., the fraction of queries for which the correct article is retrieved among the top $K$ articles. A retrieved article is considered correct when it has an identical set of MeSH term annotations as the query article. We fix $K$ to 1, 5, or 10. We evaluate the performance of our method against the following fully-supervised and self-supervised baselines:

\begin{inparaenum}[(i)]
    \item \textbf{Siamese-3D}: We train a Siamese \textsc{Bert} to group articles by optimizing the triplet loss \cite{oord2018representation}, enforcing the embedding of articles with matching sets of MeSH annotations to nearby locations, and the non-matching ones to distant locations in the embedding space. We set the embedding space dimension to 3, and use the Euclidean distance measure and online triplet mining to obtain the positives and negatives for each sample during training \cite{hermans2017defense}.
    \item \textbf{Siamese-768D}: We follow the same procedure as Siamese-3D, however, we extend the embedding space dimension to 768.
    \item \textbf{BaseBert}: We use a general domain pre-trained \textsc{Bert} and concatenate the mean-pooled, max-pooled, and \texttt{[CLS]} representations into a 2304 dimensional vector for each of the test articles.
    \item \textbf{BioBert}: We use \textsc{Bert} pre-trained on PubMed abstracts and follow the same procedure as with \textsc{BaseBert}.
    \item \textbf{SciBert-NLI}: We use the mean-pooled embeddings from \textsc{SciBert}, which is fine-tuned on datasets for the NLI task \cite{bowman2015large, williams2017broad}.
\end{inparaenum}

In all baselines, we perform retrieval by taking the top $K$ elements from the list of articles ranked by the Euclidean distance between their representation vectors and that of the query. The distance is computed in the representation space for the models trained on the retrieval task and the pre-trained sentence representation models, while for the \blackgls{sod} models we consider the physical distance in the 3D atlas.

\begin{table}[ht]
\centering
\resizebox{0.8\columnwidth}{!}{
\begin{tabular}{lcccc} \toprule
{Method} & {Embedding dimensionality} & {Recall@1 ($\uparrow$)} & {Recall@5 ($\uparrow$)} & {Recall@10 ($\uparrow$)} \\ \midrule
\multicolumn{5}{l}{\textbf{\textit{Fully-supervised methods}}} \\
3D-Sms & 3 & 34.6 $\pm$ 0.8 & 61.4 $\pm$ 0.8 & 69.7 $\pm$ 0.7 \\
Large-Sms & 768 & \textbf{42.9 $\pm$ 0.8} & \textbf{68.7 $\pm$ 0.7} & \textbf{75.4 $\pm$ 0.7} \\
SOD & 3 & 37.4 $\pm$ 0.8 & 64.3 $\pm$ 0.8 & 71.3 $\pm$ 0.7 \\\midrule
\multicolumn{5}{l}{\textbf{\textit{Self-supervised methods}}} \\
BaseBert & 2304 & 9.8 $\pm$ 0.5 & 26.1 $\pm$ 0.7 & 37.5 $\pm$ 0.8\\
BioBert & 2304 & 13.6 $\pm$ 0.6 & 35.2 $\pm$ 0.8 & 48.5 $\pm$ 0.8 \\
SciBert-NLI & 768 & 16.9 $\pm$ 0.6 & 41.8 $\pm$ 0.8 & 55.4 $\pm$ 0.8 \\
SOD+M & 3 & \textbf{26.7 $\pm$ 0.7} & \textbf{56.8 $\pm$ 0.8} & \textbf{65.9 $\pm$ 0.8} \\
\bottomrule
\end{tabular}
}
\caption[Medical text retrieval experiments]{Recall@k retrieval of medical texts. SOD-based methods perform retrieval by first grounding medical text to the human atlas.}
\label{conll2020:table:retrieval}
\end{table}

In Table \ref{conll2020:table:retrieval} (upper), we compare our method with fully-supervised siamese models trained to group documents based on their MeSH term annotations. An interesting observation is that despite being trained to choose between the target organs when there are multiple, \blackgls{sod} outperforms Siamese-3D, which is explicitly trained to group articles in 3D based on the whole set of MeSH annotations, without being limited to organizing the article embeddings in the rather constrained 3D human atlas. It is worth noting that \blackgls{sod} falls slightly short compared to Siamese-768D, most likely because it is trained to embed text in 768 dimensions, thus having a higher representational power.

In Table \ref{conll2020:table:retrieval} (lower), we evaluate the retrieval performance of our self-supervised method SOD+M (trained on occurrences of atlas glossary terms and their synonyms, see \S\ref{conll2020:sec:self-supervised}) against the pre-trained \textsc{Bert} baselines, in a setting that does not rely on ground truth MeSH term annotations. We observe that SOD+M significantly outperforms all of them, including SciBert-NLI\footnote{We report SciBert-NLI as it outperformed BioBert-NLI.}.

We finally acknowledge a performance gap between self-supervised SOD+M and the fully-supervised SOD.
% as well as the methods that are explicitly trained to optimize the retrieval performance (3D-Sms, Large-Sms).
However, in a (realistic) scenario of having large quantities of unannotated medical texts that require systematization, we argue that such fully-supervised approaches would not be feasible.

\subsection{Qualitative Evaluation and Use-Cases}
\begin{figure}
\centering
\begin{subfigure}[t]{0.4\columnwidth}
  \centering
  \includegraphics[width=0.5\columnwidth]{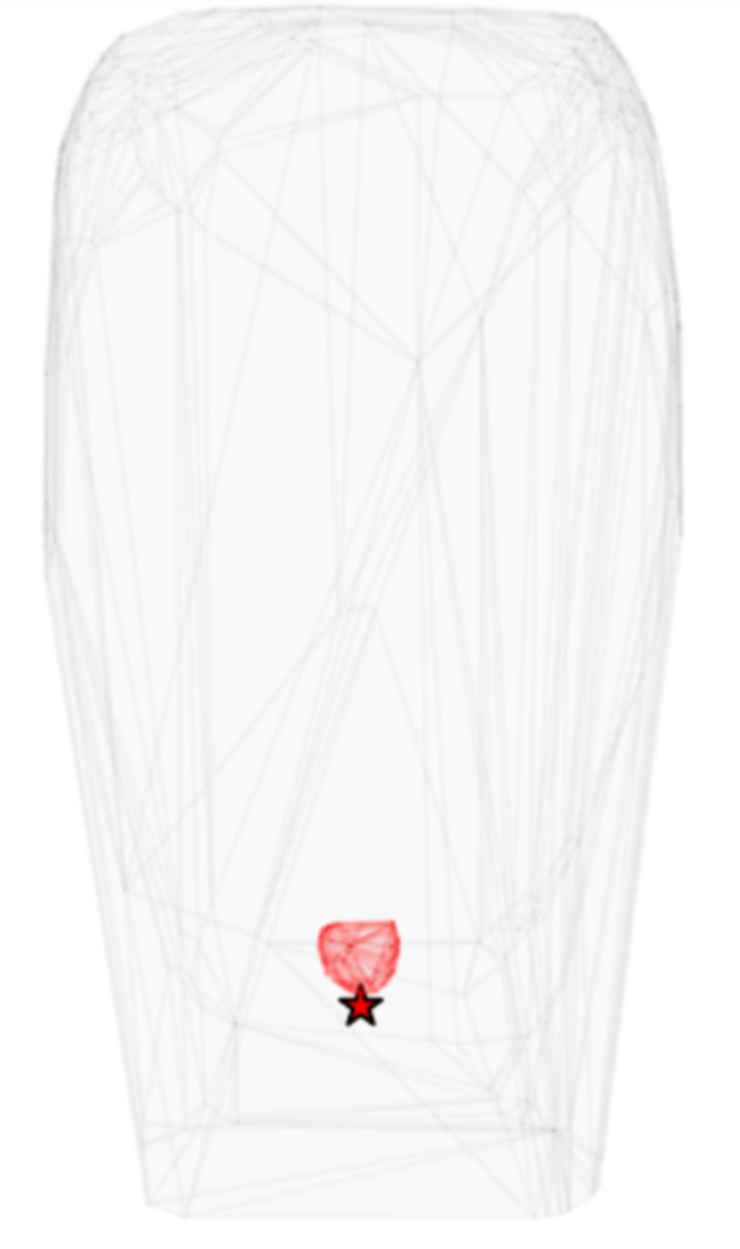}
  \caption{Ovaries}
  \label{conll2020:fig:ovaries_grounding}
\end{subfigure}%
\begin{subfigure}[t]{0.4\columnwidth}
  \centering
  \includegraphics[width=0.5\columnwidth]{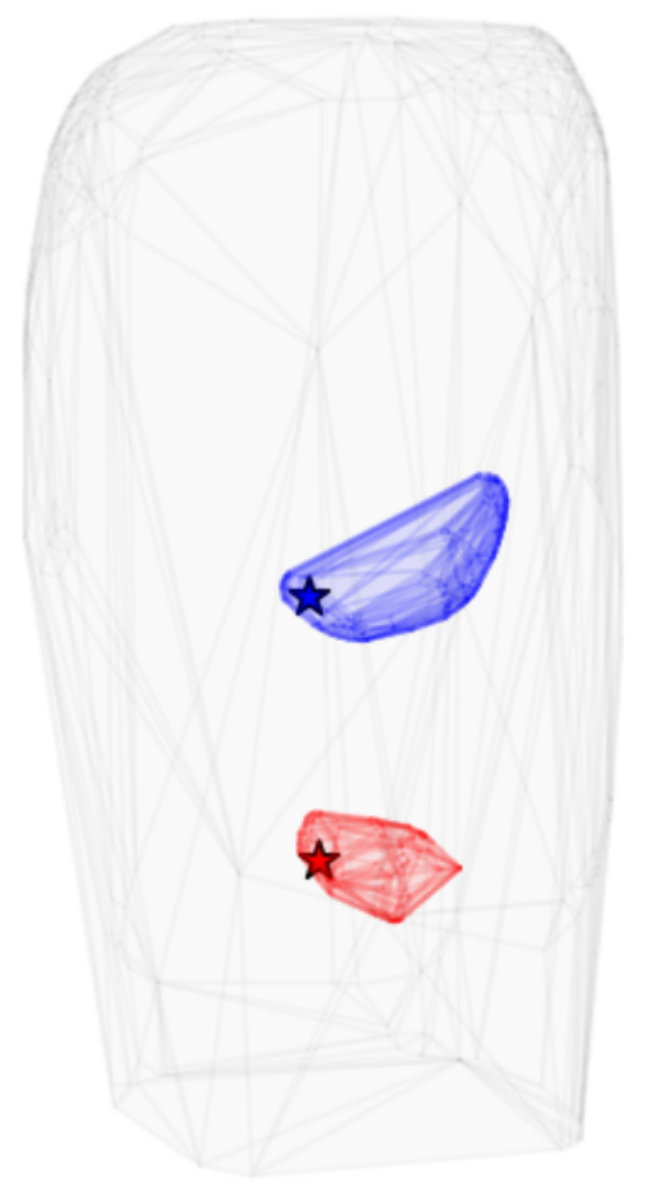}
  \caption{Colon segments}
  \label{conll2020:fig:colon_segments}
\end{subfigure}
\caption[Medical text grounding qualitative examples]{\textbf{Left}: Grounding of a paragraph about ``ovaries''.
% (Appendix \S\ref{conll2020:appendix:sec:qualitative}).
The red structure is ``urinary bladder'', which serves as a location reference;
\textbf{Right:} Grounding of Wikipedia articles describing ``transverse colon'' (upper) and ``sigmoid colon'' (lower), both contained within a common label ``colon'' during training.}
\label{conll2020:fig:qualitative_examples}
\end{figure}

We further qualitatively demonstrate several desirable properties of our approach. Although we perform training using a single male atlas, in Figure~\ref{conll2020:fig:ovaries_grounding}, we observe the grounding of a paragraph describing the ``ovaries''
% (See Appendix \S\ref{conll2020:appendix:sec:qualitative})
to a reasonably close vicinity of their actual location. We additionally qualitatively evaluate the results of \S\ref{conll2020:sec:within-organ} by mapping Wikipedia articles referring to the ``transverse colon'' and the ``sigmoid colon'' to the 3D atlas. In Figure~\ref{conll2020:fig:colon_segments}, we observe that the articles are mapped to the actual locations of the colon segments, despite that the terms shared a common label (``colon'') during training.

The low-dimensional text embeddings in the 3D atlas space can be put to use in multiple real-world applications. Integrated with a speech recognition system, they could be used to provide real-time localization of the steps taken during medical procedures based on the narrative operative reports. Additionally, the translation to a 3D atlas can be used as a way to systematize large corpora of unannotated medical text while being able to observe the relationship between embedded texts in an intuitively meaningful setting. Another advantage of medical text retrieval in the physical 3D space is the ability to retrieve information by directly specifying an observable location in the human atlas space, as opposed to using textual queries. To demonstrate this, we built a tool that accepts a query in the form of 3D coordinates and matches articles related to COVID-19,  based on the proximity of their embeddings in 3D space \cite{grujicic2020self}.
% The tool for visual-based retrieval of Covid-19 related articles can be accessed at: \url{www.github.com/dusangrujicic/cord19-visualizer}

\subsection{Ablation Study: Varying Temperatures and Sampled Voxels}\label{conll2020:appendix:sec:hyperparams}
Table \ref{conll2020:appendix:table:temperatures-100-points} and \ref{conll2020:appendix:table:temperatures-1000-points} showcase how varying the temperature terms $\gamma_p$ and $\gamma_o$ affect the results measured by \blackgls{ior}, \blackgls{nvd} and \blackgls{nvdo} on the full test set of medical articles. In particular, Table \ref{conll2020:appendix:table:temperatures-100-points} shows the results when the model is trained with 100 voxel points sampled for each organ during training, while table \ref{conll2020:appendix:table:temperatures-1000-points} shows the results when the model is trained by sampling 1000 voxel points during training.

\begin{table}[t]
\centering
\resizebox{0.7\textwidth}{!}{
\begin{tabular}{lccc} \toprule
{Method} & {IOR ($\uparrow$)} & {NVD ($\downarrow$)} & {NVD-O ($\downarrow$)} \\ \midrule
$\gamma_p$ = 0.1, $\gamma_o$ = 0.1 & 89.2 $\pm$ 0.5 & 0.9 $\pm$ 0.1 & 2.8 $\pm$ 0.1 \\
$\gamma_p$ = 0.1, $\gamma_o$ = 0.5 & 88.8 $\pm$ 0.5 & \textbf{0.8 $\pm$ 0.1} & 2.7 $\pm$ 0.2 \\
$\gamma_p$ = 0.1, $\gamma_o$ = 1.0 & \textbf{89.4 $\pm$ 0.5} & \textbf{0.8 $\pm$ 0.1} & 2.5 $\pm$ 0.2 \\
$\gamma_p$ = 0.5, $\gamma_o$ = 0.1 & 82.5 $\pm$ 0.6 & 1.0 $\pm$ 0.1 & 2.3 $\pm$ 0.1 \\
$\gamma_p$ = 0.5, $\gamma_o$ = 0.5 & 86.7 $\pm$ 0.5 & 0.9 $\pm$ 0.1 & 2.1 $\pm$ 0.1 \\
$\gamma_p$ = 0.5, $\gamma_o$ = 1.0 & 85.0 $\pm$ 0.6 & 1.0 $\pm$ 0.1 & 2.3 $\pm$ 0.1 \\
$\gamma_p$ = 1.0, $\gamma_o$ = 0.1 & 83.6 $\pm$ 0.6 & 1.0 $\pm$ 0.1 & 2.4 $\pm$ 0.1 \\
$\gamma_p$ = 1.0, $\gamma_o$ = 0.5 & 82.2 $\pm$ 0.6 & 1.0 $\pm$ 0.1 & 2.3 $\pm$ 0.1 \\
$\gamma_p$ = 1.0, $\gamma_o$ = 1.0 & 82.2 $\pm$ 0.6 & 1.0 $\pm$ 0.1 & \textbf{1.9 $\pm$ 0.1} \\ \bottomrule
\end{tabular}
}
\caption[Experiments with 100 sampled voxels during training]{IOR, NVD and NVD-O with SOD-based methods trained with varying inference while randomly sampling 100 voxels during training.}
\label{conll2020:appendix:table:temperatures-100-points}
\end{table}

\begin{table}[t]
\centering
\resizebox{0.7\textwidth}{!}{
\begin{tabular}{lccc} \toprule
{Method} & {IOR ($\uparrow$)} & {NVD ($\downarrow$)} & {NVD-O ($\downarrow$)} \\ \midrule
$\gamma_p$ = 0.1, $\gamma_o$ = 0.1 & 89.1 $\pm$ 0.5 & 1.0 $\pm$ 0.1 & 3.6 $\pm$ 0.2 \\
$\gamma_p$ = 0.1, $\gamma_o$ = 0.5 & \textbf{89.3 $\pm$ 0.5} & \textbf{0.8 $\pm$ 0.1} & 2.5 $\pm$ 0.2 \\
$\gamma_p$ = 0.1, $\gamma_o$ = 1.0 & 89.2 $\pm$ 0.5 & \textbf{0.8 $\pm$ 0.1} & 2.4 $\pm$ 0.2 \\
$\gamma_p$ = 0.5, $\gamma_o$ = 0.1 & 84.0 $\pm$ 0.6 & 1.0 $\pm$ 0.1 & 2.6 $\pm$ 0.2 \\
$\gamma_p$ = 0.5, $\gamma_o$ = 0.5 & 86.8 $\pm$ 0.5 & 0.9 $\pm$ 0.1 & \textbf{2.0 $\pm$ 0.1} \\
$\gamma_p$ = 0.5, $\gamma_o$ = 1.0 & 84.0 $\pm$ 0.6 & 0.9 $\pm$ 0.1 & \textbf{2.0 $\pm$ 0.1} \\
$\gamma_p$ = 1.0, $\gamma_o$ = 0.1 & 84.2 $\pm$ 0.6 & 1.1 $\pm$ 0.1 & 2.8 $\pm$ 0.2 \\
$\gamma_p$ = 1.0, $\gamma_o$ = 0.5 & 81.4 $\pm$ 0.6 & 1.0 $\pm$ 0.1 & 2.1 $\pm$ 0.1 \\
$\gamma_p$ = 1.0, $\gamma_o$ = 1.0 & 83.1 $\pm$ 0.6 & 1.0 $\pm$ 0.1 & 2.1 $\pm$ 0.1 \\ \bottomrule
\end{tabular}
}
\caption[Experiments with 1000 sampled voxels during training]{IOR, NVD and NVD-O with SOD-based methods trained with varying inference while randomly sampling 1000 voxels during training.}
\label{conll2020:appendix:table:temperatures-1000-points}
\end{table}
\section{Conclusion}
% One limitation of our method is that it does not explicitly take into account spatial descriptions and other modifier expressions. Rather, it uses abstract-level annotation to ground whole abstracts to the most semantically relevant regions and uses the co-occurrences between terms (which also reflect their spatial relationships to a significant degree) to organize and distribute the grounding within the same organ or out-of-atlas organs. A natural extension of this work would be to move up from the entity level and explicitly address the spatial language and descriptions of relationships between anatomical structures.
In this chapter, we addressed the task of translating medical text into 3D spatial representations grounded in a human anatomical atlas. Specifically, we proposed the task of mapping medical descriptions to locations within a human atlas and introduced the Soft Organ Distance loss function to facilitate this process. Furthermore, we showcased how our approach effectively utilizes the spatial relationships between medical terms to train models that accurately associate medical text with precise regions in the human atlas.

Concerning \inlineresearchquestion{Research Question 2}, we reveal that the Soft Organ Distance loss function significantly enhances the models' ability to ground medical texts to the human body's atlas. The grounding capability goes beyond traditional text embeddings, and offers a more intuitive and spatially relevant method of understanding medical literature.

Regarding \inlinesubresearchquestion{Research Question 2.1}, we observe that the model is capable of grounding medical texts to anatomical (sub)regions not present in the training data. This indicates the model's robustness and adaptability to out-of-distribution locations for organ references.
%outside of the training data.

Finally, we demonstrated that we can indeed train such grounding models in a self-supervised manner, thereby addressing \inlinesubresearchquestion{Research Question 2.2}. This reduces reliance on labor-intensive annotated datasets, and presents a scalable method for model training that retains effectiveness in grounding medical texts.
\textbf{Limitations.} One limitation of our method is that it does not explicitly take into account spatial descriptions (``above'', ``next to'', etc.) and other modifier expressions. Rather, it uses abstract-level annotation to ground whole abstracts to the most semantically relevant regions, and uses the co-occurrences between terms (which also reflect their spatial relationships to a significant degree) as a proxy, to organize and distribute the grounding within the same organ or organs not part of the training data. This approach, while effective in many scenarios, might not capture the full complexity of spatial descriptions in medical texts, suggesting a direction for future enhancements.

\textbf{What has happened since?} Text embedding methods had improved significantly since 2020 \cite{reimers2019sentence, wang2022text}, when the majority of the work in this chapter was done. Most recent work \cite{wang2024multilingual} even shows that high-quality text embeddings can be learned for multiple languages (i.e., multilingual embeddings). Nevertheless, despite being more powerful, these methods still suffer from similar limitations as prior (medical) text embedding methods: the embeddings are non-interpretable for medical professionals, with no clear notion of similarity between the documents.

%%%%%%%%%%%%%%%%%%%%%%%%%%%%%%%%%%%%%%%%%%%%%%%%%%
% Keep the following \cleardoublepage at the end of this file, 
% otherwise \includeonly includes empty pages.
\cleardoublepage

% vim: tw=70 nocindent expandtab foldmethod=marker foldmarker={{{}{,}{}}}
%

%
  \graphicspath{{\chaptersdir/emnlp2023/image/}}%
  %\cleardoublepage%
  \chapter{Translating Structured Text to Canonical Facts in Large-Scale Knowledge Graphs}\label{ch:emnlp2023}
In this chapter, we focus on the problem of multimodal translation from structured text to canonical facts in large-scale Knowledge Graphs (\blackgls{kg}). Specifically, we approach this task by establishing connections between Open Information Extractions (\blackgls{oie})---representing text in a structured form---and Knowledge Graphs.

Open Information Extraction methods extract facts from natural language text in the form of (\textit{``subject''}; \textit{``relation''}; \textit{``object''}) triplets. These facts are, however, merely surface forms, the ambiguity of which impedes their downstream usage; e.g., the surface phrase \emph{``Michael Jordan''} may refer to either the former basketball player or the university professor. Knowledge Graphs, on the other hand, contain facts in a canonical (i.e., unambiguous) form, but their coverage is limited by a static schema (i.e., a fixed set of entities and predicates).

To bridge this gap, we need the best of both worlds: (i) high coverage of free-text \blackgls{oie}s, and (ii) semantic precision (i.e., monosemy) of KGs. In order to achieve this goal, we propose a new benchmark with novel evaluation protocols that can, for example, measure fact-linking performance on a granular triplet slot level, while also measuring if a system can recognize that a surface form has no match in the existing KG. Our comprehensive analysis of various baseline models reveals that identifying entities and predicates not present in the Knowledge Graph poses a greater challenge than correctly linking to existing ones, highlighting the need for increased research focus on this complex task.

The content of this chapter is based on the following publication\footnote{This work was performed during a research internship at NEC Labs Europe.}:

\begin{mdframed}[backgroundcolor=gray!30,linewidth=0pt]
\underline{Radevski, G.}, Gashteovski, K., Hung, C-C., Lawrence, C., Glavaš, G., (2023). Linking Surface Facts to Large-Scale Knowledge Graphs. The Conference on Empirical Methods in Natural Language Processing (EMNLP). \cite{radevski-etal-2023-linking}
\end{mdframed}

\section{Introduction}
Open Information Extraction (\blackgls{oie}) methods extract surface \emph{(``subject''; ``relation''; ``object'')}-triplets from natural language text in a schema-free manner \cite{Banko2007OpenIE}. For example, given the sentence \emph{``Michael Jordan, who grew up in Wilmington, played for Chicago Bulls''}, such method should extract the triplets (i.e., \emph{surface facts}): $t_1=$\emph{(``Michael Jordan''; ``played for''; ``Chicago Bulls'')} and $t_2=$\emph{(``Michael Jordan''; ``grew up in''; ``Wilmington'')}. The output of such systems is used in a diverse range of downstream tasks, including summarization \cite{Ribeiro2022FactGraphEF}, question answering \cite{wu2022triple}, event extraction \cite{dukic2023leveraging}, text clustering \cite{viswanathan2023large} and video grounding \cite{nan2021interventional}. These triplets, however, consist of surface-form entities and relations, which are frequently ambiguous; e.g., in $t_1$ and $t_2$, the entity mention \emph{``Michael Jordan''} may refer to several entities: the basketball player or the computer scientist. Resolving such ambiguities---by linking the OIE triplet slots to inventories of unambiguous concepts---improves their downstream utility; e.g., in QA \cite{saxena2020improving}, automatic medical diagnosing \cite{li2022neural} or dialogue \cite{joko2021conversational}.

Knowledge Graphs (\blackgls{kg})s, on the other hand, are inventories of \emph{canonical} facts in the form of (subject; predicate; object)-triplets, where each slot is a unique (i.e., unambiguous) concept \cite{vrandevcic2012wikidata}. \blackgls{kg}s are, however, limited by their static and often hand-crafted schema (i.e., fixed set of entities and predicates). As a consequence, methods that directly extract canonical \blackgls{kg} triplets from text \cite{distiawan2019neural, josifoski2021genie}, preemptively discard any information outside of the reference \blackgls{kg} schema.
% Acknowledging this, Ye~\etal~\cite{ye2023schema} recently proposed \emph{schema-adaptable \blackgls{kg} construction}, with the goal of extracting information for a \blackgls{kg} with an evolving schema. Ye~\etal~\cite{ye2023schema} conclude that OIE methods indeed extract meaningful new knowledge for such KGs, however, they point precisely to the ambiguity of the surface forms as the major obstacle.  

To combine the best of both worlds, we need to bridge the gap between the schema-free (but ambiguous) surface facts extracted from text and the schema-fixed (but unambiguous) \blackgls{kg} knowledge. Therefore, we pose the following research question:

\researchquestion{Research Question 3}{How effectively can we translate general-purpose textual data, represented as structured triplets, to corresponding facts in a knowledge graph?}

We further explore a more nuanced scenario where such translation might be infeasible due to the absence of the corresponding facts in the KG. Importantly, by identifying instances where text triplets cannot be linked to existing \blackgls{kg} facts, we can identify novel or incorrect information. This functionality is essential both for \blackgls{kg} population and fact-checking. Consequently, we ask:

\subresearchquestion{Research Question 3.1}{Can we accurately identify instances where a valid translation between the text triplets and knowledge graph facts does not exist? That is, the text represents knowledge that is not in the knowledge graph: i.e., factually incorrect or novel information.}

% However, existing benchmarks and models only partially address the problem. One line of work \cite{zhang2019openki, jiang2021cori, wood2021integrating} assumes a setting, which is arguably unrealistic because the OIE entity slots are \textit{a priori} linked to \blackgls{kg} entities and thus only addresses relation linking. A second line of research \cite{distiawan2019neural,cabot2021rebel,josifoski2021genie,sakor2020falcon,elsahar2018t} entails canonical \blackgls{kg} triplets directly from text paragraphs, bypassing OIE, which makes such models tied to the fixed \blackgls{kg} schema. Critically, \textit{all} these works conjecture that \emph{each} OIE slot \textit{has} a corresponding \blackgls{kg} entry. Gashteovski~\etal~\cite{gashteovski2019opiec} showed that this is hardly ever the case in practice.

\begin{figure*}
    \centering
    \includegraphics[width=1.0\textwidth]{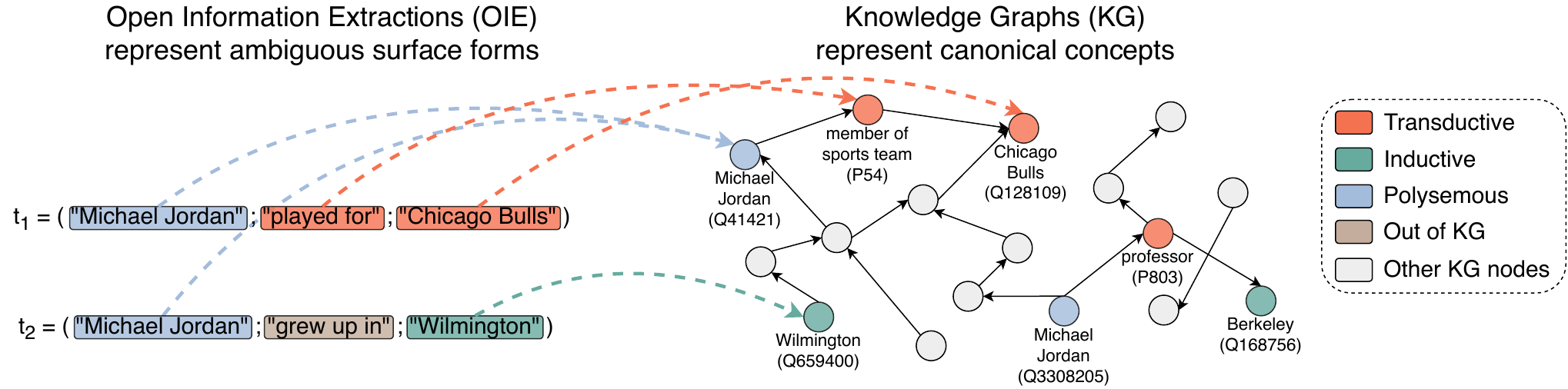}
    \caption[Problem statement: Linking Open Information Extractions to Knowledge Graphs]{Open Information Extractions represent free-text surface-form triplets, which may be ambiguous. Knowledge Graphs represent canonical, unambiguous concepts, yet are limited by a hand-crafted schema. By linking OIEs to \blackgls{kg} facts we bridge the gap between the schema-free (but ambiguous) surface facts extracted from text and the schema-fixed (but unambiguous)
    \blackgls{kg} knowledge. Our benchmark addresses \textit{all} aspects of the OIE-to-KG linking. In \textcolor{transductive}{\textbf{transductive}} evaluation, the \blackgls{kg} facts consist of entities \textit{seen} during training, while in \textcolor{inductive}{\textbf{inductive}} evaluation the \blackgls{kg} facts contain entities \textit{strictly outside} of the training data. \textcolor{polysemous}{\textbf{Polysemous}} \blackgls{oie}s match to \textit{multiple} \blackgls{kg} concepts (e.g., ``\emph{Michael Jordan}''). Finally, some OIEs might refer to \textcolor{outofkg}{\textbf{Out-of-KG}} entities or predicates.}
    \label{emnlp2023:fig:fig1}
\end{figure*}

\textbf{Contributions.} To gain insight into the aforementioned research questions we make the following contributions:

\circled{1} We automatically curate a novel large-scale benchmark for linking surface facts to KG, with multifaceted evaluation protocols that cover \textit{all} aspects of the linking (see Figure~\ref{emnlp2023:fig:fig1}).

\circled{2} We develop several strong baselines, inspired by cross-modal retrieval \cite{miech2021thinking, geigle2022retrieve} approaches.

\circled{3} We reveal that we can train successful fact-linking methods by training solely on synthetic data, while, on the other hand, detecting out-of-knowledge graph detection remains an open research problem.
% \circled{1}  and \circled{2} s\circled{3}
% Through our experiments, we found that the methods (i) perform well \textit{transductively} but (ii) their performance deteriorates in an \textit{inductive} evaluation.
% Further, we find that (iii) a dedicated OIE-to-KG fact re-ranking model improves the linking performance of both inductive and polysemous OIEs, and that (iv) we obtain high performance by training models solely on a synthetic variant of our dataset (i.e., with the \blackgls{kg} as the only human-annotated data). 
% Lastly, for Out-of-KG detection, we show that (v) it is possible to detect Out-of-KG entities to an extent, however, the same does not hold for predicates: a task that our experiments identify as a difficult open problem, which requires more research attention.

We release code, data, and trained models at: \url{https://github.com/nec-research/fact-linking}.

\section{Fact Linking: Problem Statement}\label{emnlp2023:sec:fl}
For a given surface-form \blackgls{oie} triplet $t_1=(``s";``r";``o")$, the goal is to link each slot to a canonical concept in a \blackgls{kg} (\textit{if} the corresponding concept exists in the KG): $``s" \rightarrow e_1 \in \mathcal{E}$; $``r" \rightarrow p \in \mathcal{P}$; $``o" \rightarrow e_2 \in \mathcal{E}$, with $\mathcal{E}$ and $\mathcal{P}$ as the (fixed) sets of \blackgls{kg} entities and predicates. Importantly, our problem definition (and consequently evaluation) focuses on linking at the \textit{fact level}, where each \blackgls{oie} slot is contextualized with the other two \blackgls{oie} slots.\footnote{Alternatively, additional context can be included, e.g., the provenance from which the \blackgls{oie} surface fact is obtained.}

Additionally and crucially, we want linking models that can assign an empty set to \blackgls{oie} surface forms (e.g., $``s"$ $\rightarrow \emptyset$) when they refer to concepts not present in the \blackgls{kg} (i.e., \textit{out-of-KG} concepts). To enable a realistic setup for linking free-text \blackgls{oie} triplet to a KG, we build a benchmark that considers four different facets (see Figure~\ref{emnlp2023:fig:fig1}).

\textbf{\textcolor{transductive}{Transductive Fact Linking.}}
In transductive linking, we measure how well the models link OIEs to \blackgls{kg} facts consisting of entities and predicates seen during training (as components of training \blackgls{kg} facts). Note that the testing \blackgls{kg} facts (as whole triplets) are not in the training data. Consider, for example, the extraction $t_1$ in Figure~\ref{emnlp2023:fig:fig1}. In the transductive linking task, the mentions \emph{``played for''} and \emph{``Chicago Bulls''} refer to the \blackgls{kg} predicate (P54) and entity (Q128109) respectively; both seen by the model as part of other training \blackgls{kg} facts. However, the whole triplet (Q41421; P54; Q128109) to which $t_1$ is linked was not part of the training data.

\textbf{\textcolor{inductive}{Inductive Fact Linking.}}
The inductive setup evaluates the linking to \blackgls{kg} facts that consist of entities that are not seen during the training of the models. 
 In other words, this setup tests the generalization of fact-linking models over entities. In Figure~\ref{emnlp2023:fig:fig1}, given $t_2$ as a test instance, the \blackgls{oie} entity \emph{``Wilmington''} is inductive as it refers to a \blackgls{kg} entity (Q659400) that is not part of any training \blackgls{kg} fact.

\textbf{\textcolor{polysemous}{Polysemous Fact Linking.}}
We focus on OIEs for which the \textit{``s''} and \textit{``o''} slots are ambiguous w.r.t. the KG, i.e., in isolation they refer to a set of \blackgls{kg} entities rather than a single entity. The mention \emph{``Michael Jordan''} from either $t_1$ and $t_2$ (Figure~\ref{emnlp2023:fig:fig1}), in isolation, refers to both the basketball player (Q41421) and the computer scientist (Q3308205). Here, the other two \blackgls{oie} slots offer the disambiguation signal that is necessary for successful linking.

\textbf{\textcolor{outofkg}{Out-of-KG Detection.}} 
We introduce a novel Outside-of-Knowledge Graph \blackgls{out-of-kg} detection task, in which the models are to recognize that an \blackgls{oie} triplet component (i.e., \textit{``s''}, \textit{``r''} or \textit{``o''}) cannot be linked because they do not have a corresponding \blackgls{kg} concept (e.g., the relation \emph{``grew up in''} from the triplet $t_2$ in Figure~\ref{emnlp2023:fig:fig1}).

\section{FaLB: Fact Linking Benchmark}\label{emnlp2023:sec:preparation}
We set up an automatic data processing pipeline to derive an OIE-to-KG fact-linking benchmark, which supports all four evaluation facets from \S\ref{emnlp2023:sec:fl}. We refer to both the process (i.e., pipeline) and the resulting benchmark as \textbf{\blackgls{falb}}. FaLB's input is a dataset with (gold) alignments between natural language sentences and KG facts entailed by the sentence; i.e., each data instance is \emph{(sentence, KG fact)} pair. Consequently, the creation of FaLB entails five design decisions: selection of (i) sentence-to-KG fact dataset(s), and (ii) a reference KG; (iii) generating OIE triplets, (iv) high-precision OIE-KG fact alignments, and (v) a data augmentation strategy to increase the diversity of the data. Below is an example instance of the FaLB dataset:

\begin{tcolorbox}[enhanced, fit to height = 4.2cm, colback = white, colframe = white, boxsep = 2pt, coltitle = black, left = 1pt, right = 1pt, top = 1pt, bottom = 0pt, title = \textbf{\underline{Example Data Instance from FaLB}}]
  \underline{\textbf{Sentence:}} \emph{``M.~J., who was born in Brooklyn, played for the Bulls.''} \vspace{2.5pt}
  
  \underline{\textbf{OIE surface facts:}} $t_1=$\emph{(``M.~J.''; ``played for''; ``the Bulls'')} and $t_2=$\emph{(``M.~J.''; ``was born in''; ``Brooklyn'')}\vspace{2.5pt}
  
  \underline{\textbf{KG canonical fact identifiers:}} (Q41421; P54; Q128109); (Q41421; P19; Q18419)\vspace{2.5pt}

  \underline{\textbf{KG text facts:}} (Michael Jordan; member of sports team; Chicago Bulls); (Michael Jordan; place of birth; Brooklyn)\vspace{2.5pt}
  
  \underline{\textbf{KG entity aliases:}} (Q41421: Air Jordan, Michael Jeffrey Jordan, His Airness); (Q128109: Bulls, The Bulls), ...
\end{tcolorbox}

\paragraph{Sentence-to-KG Fact Datasets.} We build FaLB benchmarks for two such existing datasets: REBEL \cite{cabot2021rebel} and SynthIE \cite{josifoski2023exploiting}. \textit{REBEL} is built from Wikipedia abstracts, where the manually hyperlinked entities in each sentence are linked to Wikidata entities, thus making them golden. To match the sentence with a KG fact, for each pair of KG entities $e_i$ and $e_j$ within the sentence, REBEL obtains all KG predicates $p_k$ such that $(e_i, p_k, e_j)$ or $(e_j, p_k, e_j)$ exist. However, two Wikidata entity nodes connected with a predicate (thus constituting a KG fact) do not guarantee the predicate validity in the sentence. For that reason, Cabot~\etal~\cite{cabot2021rebel} use a Natural Language Inference pre-trained \blackgls{roberta} \cite{liu2019roberta} to filter the predicates \textit{not entailed} by the Wikipedia sentence; see \cite{cabot2021rebel} for details. We also use \textit{SynthIE}, which is synthetically generated using a Large Language Model (\blackgls{llm}) \cite{brown2020language}. Given a set of KG facts, the \blackgls{llm} is prompted to generate a sentence that covers the KG facts.
% \footnote{Josifoski \etal \cite{josifoski2023exploiting} manually evaluated that SynthIE is of higher quality than REBEL (see Appendix~\ref{emnlp2023:appendix:rebel-vs-synthie} for details).}

\paragraph{Reference KG.}
We use Wikidata \cite{vrandevcic2012wikidata} as our reference KG, because REBEL and SynthIE align sentences to Wikidata facts. We select the subgraph of Wikidata that contains all entities which have a corresponding Wikipedia page, as per Wu~\etal~\cite{wu2019scalable}. Following Lerer~\etal~\cite{lerer2019pytorch}, we additionally filter out the most infrequent entities and predicates, appearing less than 5 times in the whole Wikidata dump. This results in a large reference \blackgls{kg} with $5,794,782$ unique entities and $4,153$ unique predicates. We further create two smaller, dataset-specific reference KGs; one for each of the two datasets for which we apply FaLB: REBEL and SynthIE. These Benchmark-Restricted KGs (\blackgls{brkg}s) contain only the Wikidata entities and predicates referenced by at least one OIE triplets extracted from their respective datasets. The \blackgls{brkg} for REBEL contains $625,125$ entities and $565$ predicates, while the one for SynthIE contains $702,334$ entities and $758$ predicates.

\paragraph{Generating OIE Triplets.}
We use four state-of-the-art OIE methods to obtain a set of \blackgls{oie} triplets for each of the sentences in the dataset. To increase diversity, we use two state-of-the-art rule-based OIE methods: MinIE \cite{gashteovski2017minie} and StanfordOIE \cite{angeli2015leveraging}; and two state-of-the-art neural OIE models: MilIE \cite{kotnis2022milie} and Multi$^2$OIE \cite{ro2020multi}.

\paragraph{High-precision OIE-KG Fact Alignments.}
Next, we need to match the extracted OIE triplets against the set of KG facts associated with the sentences. Crucially, this automatic step needs to have high precision in order to create a high-quality benchmark. As per the distant supervision assumption of Mintz~\etal~\cite{mintz2009distant}, we create an alignment $t\leftrightarrow f$ between any OIE triplet $t$ and any KG fact $f$ that exactly match (i.e., have identical text form) in subject and object, assuming that the relation of $t$ is aligned with the predicate of $f$. When multiple OIE triplets (excluding exact duplicates) $t_1$, $t_2$, \dots, $t_k$ (e.g., extracted with different OIE systems) match with the same KG fact $f$, we obtain all $k$ alignments: $t_1\leftrightarrow f$, \dots, $t_k\leftrightarrow f$. Finally, we remove all training alignments $t \leftrightarrow f$ that exist in the test portion. 
To verify that this strategy indeed has high precision, we randomly sample 100 test instances, and conduct a human study with two expert annotators. The annotators found 97\% correct pairings (inter-annotator agreement of 99\%; Kohen's kappa of 0.80)
% ; see Appendix~\ref{emnlp2023:appendix:data-quality} for details.
We therefore confirm the reliability of this design choice to produce high-quality data.

\paragraph{Data Augmentation to Increase Diversity.}
Lack of example diversity---most linked OIE entity mentions exactly match the text of the corresponding KG entities---is a common problem in existing linking benchmarks \cite{cabot2021rebel}. This stems from strict data curation procedures that aim for high alignment precision, representing a mismatch with practice, where often the entity mentions do not exactly match the canonical KG entity denotation. To train and evaluate fact-linking methods on more complex linking examples (e.g., linking from initials and abbreviations), we augment the data using the additional information available in the reference KG. For each alignment $t \leftrightarrow f$, we fetch the Wikidata aliases (i.e., non-canonical denotations) for the subject and object entities of $f$. We then create additional pairs $t' \leftrightarrow f$ with augmented OIE triplets $t'$ for all possible alias combinations. For example, for the original OIE triplet \emph{(``Michael Jordan''; ``played for''; ``Chicago Bulls'')}, this process results in augmented triplets such as \emph{(``Air Jordan''; ``played for''; ``Chicago Bulls'')}, \emph{(``M.J.''; ``played for''; ``The Bulls'')}, etc.

\section{OIE-to-KG Fact Linking Models}\label{emnlp2023:sec:methods}

\begin{figure*}[t]
\centering
\resizebox{1.0\textwidth}{!}{
\includegraphics[width=1.0\textwidth]{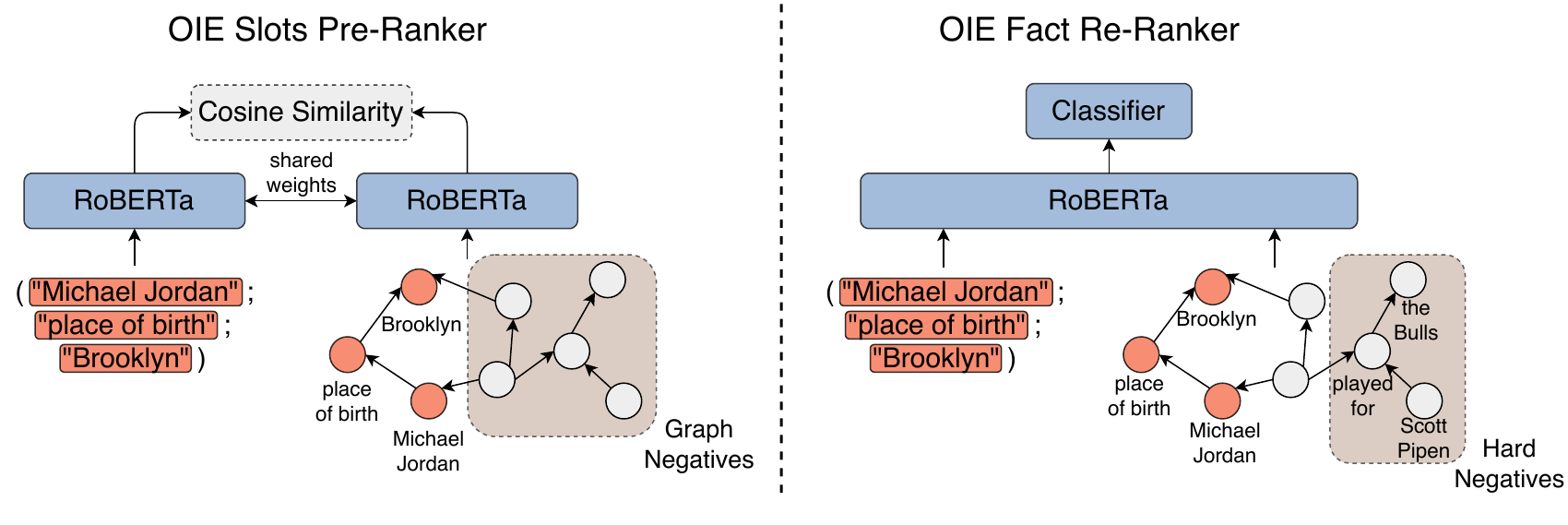}
}
\caption[Method(s) for linking Open Information Extractions to Knowledge Graphs]{\textbf{Left:} \preranker which performs pre-ranking of the \blackgls{oie} slots to \blackgls{kg} entries independently (i.e., no context between them is considered). Trained with negatives sampled from the whole KG; \textbf{Right:} \reranker which attends between the whole \blackgls{oie} triplet and \blackgls{kg} fact to output their similarity; Trained with hard-negatives.}
\label{emnlp2023:fig:models}
\end{figure*}

Our goal is to link surface-form $(``s";``r";``o")$ \blackgls{oie} triplets, to canonical Knowledge Graph facts $(e_1;p;e_2)$, where $e_1, e_2 \in \mathcal{E}$, and $p \in \mathcal{P}$. Each \blackgls{kg} entity or predicate is represented as its surface-form \blackgls{kg} label (e.g., ``Michael Jordan''), and its KG-provided description (e.g., ``American basketball player and businessman''). We henceforth refer to the entities and predicates as \blackgls{kg} entries. Intuitively, since both data streams---the \blackgls{oie} triplets and the \blackgls{kg} facts---are in natural language, we opt to obtain their representations (i.e., embeddings) with a pre-trained language model. We decouple the OIE-to-KG linking in two steps: pre-ranking and re-ranking (see \S\ref{emnlp2023:sec:pre-rank} and \S\ref{emnlp2023:sec:re-rank}, respectively),\footnote{We perform the linking in roughly $O(\vert\mathcal{E}\vert + \vert\mathcal{P}\vert + \vert\mathcal{E}\vert)$ time complexity, where $\vert\mathcal{E}\vert$ and $\vert\mathcal{P}\vert$ are the number of \blackgls{kg} entities and predicates respectively. Directly encoding whole \blackgls{kg} facts would have a prohibitive complexity: In the limit, we could have $\vert\mathcal{E}\vert \times \vert\mathcal{P}\vert \times \vert\mathcal{E}\vert$ \blackgls{kg} facts.} as is done commonly in entity retrieval \cite{wu2019scalable} and image-text matching \cite{geigle2022retrieve,li2021align}. See Figure~\ref{emnlp2023:fig:models} for an overview.
% and Appendix~\ref{emnlp2023:appendix:experimental} for implementation details.

\subsection{Pre-ranking OIE Slots to KG Entries}\label{emnlp2023:sec:pre-rank}
We denote this model as \preranker. It aims to generate \blackgls{oie} slot embeddings and \blackgls{kg} entry embeddings, such that an \blackgls{oie} slot embedding yields higher similarity with its aligned \blackgls{kg} entry's embedding compared to the other \blackgls{kg} entries. Therefore, during training, we contrast the positive pairs against a set of negatives, thereby training the model to generate embeddings for a matching \blackgls{oie} slot and \blackgls{kg} entry that lie close in the latent space. The motivation for such formulation is two-fold: (i) the number of entities is large, and could practically grow further, therefore computing the softmax over all \blackgls{kg} entities during training is prohibitive; (ii) there may be unseen \blackgls{kg} entries that we encounter during inference, therefore posing the problem as a standard classification inevitably leads to the model ignoring them. We use \blackgls{roberta} \cite{liu2019roberta} to encode the \blackgls{oie} slots and the \blackgls{kg} entries.

\paragraph{OIE Embeddings.}
We first add special tokens to indicate the start of each \blackgls{oie} slot: \texttt{<SUBJ>} for \emph{``subject''}, \texttt{<REL>} for \emph{``relation''}, \texttt{<OBJ>} for \emph{``object''}. Hence, $t_1$ = \emph{(``M. Jordan''; ``grew up in''; ``Wilmington''}) is represented as \emph{``\texttt{<SUBJ>} M. Jordan \texttt{<REL>} grew up in \texttt{<OBJ>} Wilmington''}. We then tokenize the \blackgls{oie} representation, denoted as $\mathbf{\hat{t}}$, and obtain \blackgls{roberta} token embeddings. Finally, we pool \textit{only} the special tokens' embeddings subsequently linearly projected in the desired latent space: $\mathbf{\hat{o}}_{i} = \text{Linear}(\text{RoBERTa}(\mathbf{\hat{t}}))$, where $\mathbf{\hat{o}}$ is the slot embedding, and $i \in\mathbb{R}^3$ is the \blackgls{oie} slot index.

\paragraph{KG Embeddings.}
We represent both the entities and predicates as their label followed by their description (if available in the KG); e.g., the entity ($e$) \textit{Michael Jordan} is represented as ``Michael Jordan \texttt{<DESC>} American basketball player and businessman...'', and the predicate ($p$) \textit{place of birth} is represented as ``place of birth \texttt{<DESC>} most specific known birth location of a person...''. The \texttt{<DESC>} special token indicates the start of the description. We then tokenize the representation,
% ($\mathbf{\hat{e}}$ for the entity, $\mathbf{\hat{p}}$ for the predicate),
and obtain embeddings using the same \blackgls{roberta} model. We pool the \texttt{<CLS>} token embedding and linearly project it in the \blackgls{oie} representation space as: $\mathbf{\hat{k}}_{j} = \text{Linear}(\text{RoBERTa}(\mathbf{b}_j))$, where $j \in \{entity, predicate\}$, $\mathbf{\hat{k}}$ is the \blackgls{kg} entry embedding, and $\mathbf{b}_j$ represents the sequence of tokens from the $j$-th \blackgls{kg} entry representation.

\paragraph{Linking OIEs to \blackgls{kg} Facts.} Given an \blackgls{oie} slot embedding $o_i$ and a \blackgls{kg} entry embedding $k_j$, we obtain their dot-product as: $\hat{s}_{pre} = \mathbf{o}_i^T \mathbf{k}_j$, where $\mathbf{o}_i$ and $\mathbf{k}_j$ are norm-scaled (i.e., $\hat{s}_{pre}$ represents the \preranker cosine similarity). During inference, we link an \blackgls{oie} to a \blackgls{kg} fact by selecting the most similar \blackgls{kg} entry for each of the \blackgls{oie} slots.

\paragraph{\preranker Training.}\label{emnlp2023:paragraph:preranker-training}
We train the model using standard contrastive loss: we sample $N-1$ in-batch negative \blackgls{kg} entries for each positive \blackgls{oie} slot $\leftrightarrow$ \blackgls{kg} entry pair, where $N$ is the batch size. As per standard practice \cite{oord2018representation}, we train the model using temperature-scaled InfoNCE contrastive loss:

\begin{equation}
  \begin{gathered}
\mathcal{L} = -\operatorname{log}\frac{e^{\mathbf{\hat{o}}^\text{T}\mathbf{\hat{k}}/\tau}}{e^{\mathbf{\hat{o}}^\text{T}\mathbf{\hat{k}}/\tau} + \sum_{n=1}^{\text{N}-1} e^{\mathbf{\hat{o}}^\text{T}\mathbf{\hat{k}}_n^{-}/\tau}},
\end{gathered}  
\end{equation}

where $\tau$ is the temperature. Note that during training, we only sample negative \blackgls{kg} entry embeddings for each \blackgls{oie} slot, but not the other way around. See \S\ref{background:sec:contrastive} for more information regarding multimodal alignment learning using contrastive loss.

Sampling only in-batch negatives presents an issue as the training data represents only a limited subset of the whole \blackgls{kg} (i.e., only the \blackgls{kg} entries with paired OIE). During inference, however, we contrast each \blackgls{oie} slot against the whole \blackgls{kg} to find the \blackgls{kg} entry with which it exhibits the highest similarity. Therefore, for each \blackgls{oie} slot, we additionally sample $e$ negative entities and $p$ negative predicates at random from the whole \blackgls{kg} (e.g., we would sample $\sum_{n=1}^{\text{N}-1 + p} \mathbf{\hat{k}}_n^{-}$ negative predicate embeddings).

\subsection{Re-ranking OIEs to KG Facts}\label{emnlp2023:sec:re-rank}
In certain scenarios (e.g., in the case of polysemous \blackgls{oie}s), matching whole \blackgls{oie}s with whole \blackgls{kg} facts (i.e., not decoupled per \blackgls{oie} slot) could resolve the ambiguity and thus improve performance. To that end, for each \blackgls{oie} slot, we re-rank the \preranker top-k most probable \blackgls{kg} links. We denote this model as \reranker. We perform self-attention between the \blackgls{oie} and the \blackgls{kg} fact (both provided as input, separated by a \texttt{<FACT>} special token) with a single \blackgls{roberta} transformer, and return their similarity as: $\hat{s}_{re} = \sigma(\text{Linear}(\text{RoBERTa}(\mathbf{\hat{c}})))$, where $\mathbf{\hat{c}}$ is the concatenated \blackgls{oie} and \blackgls{kg} fact representation, and $\hat{s}_{re}$ is the sigmoid ($\sigma$) normalized similarity.

\paragraph{\reranker Training.}
We train by sampling matching \blackgls{oie} $\leftrightarrow$ \blackgls{kg} fact pairs as positives, and negatives, where we replace some \blackgls{kg} fact slots (subject, predicate, object) with incorrect ones, generated as follows: We first obtain embeddings for each \blackgls{kg} entry using the \preranker, and find its top-k most similar candidates w.r.t. all other \blackgls{kg} entries. We then corrupt the ground truth \blackgls{kg} fact by randomly sampling \textit{only} from the top-k (hard) negative candidates. Lastly, with 50\% probability, we randomly mask (i.e., replace with a \texttt{<mask>} token) the description of the \blackgls{kg} fact entries.

\paragraph{\textbf{Why two-step procedure?}} We opt for a two-step approach where we apply the \preranker followed by the \reranker. While, in theory, we could directly apply the \reranker on all triplet combinations of two entities and a predicate between them, such linking would not be feasible in practice; we would need to perform $\vert\mathcal{E}\vert \times \vert\mathcal{P}\vert \times \vert\mathcal{E}\vert$ forward passes with the model to obtain all scores.
% Even though this number could be decreased significantly, as we can prune all invalid combinations given the KG ontology, still,
\section{Evaluation and Discussion}
We measure accuracy to evaluate both \blackgls{oie} slot linking to KGs (\S\ref{emnlp2023:sec:kg-population}), and Out-of-KG detection of \blackgls{oie} slots (\S\ref{emnlp2023:sec:out-of-kg-detection}). To measure \blackgls{oie} fact linking, we score a hit if \emph{all} \blackgls{oie} slots are linked correctly. The error bars represent the standard error of the mean.

\subsection{Linking OIEs to Knowledge Graphs}\label{emnlp2023:sec:kg-population}
\paragraph{Setup.} We explore the extent to which we can link \blackgls{oie} slots to a large-scale \blackgls{kg} (Wikidata). We address two main research questions concerning the OIE-to-KG fact-linking task: (i) To what extent do methods generalize to different \blackgls{kg} entity facets? We consider transductive, inductive, or polysemous entities (see \S\ref{emnlp2023:sec:fl} for detailed definition); (ii) What is the performance impact of the \blackgls{kg} size? We test two reference \blackgls{kg} sizes: Benchmark Restricted \blackgls{kg} ($\sim$650k entities, $\sim$0.6k predicates) and Large \blackgls{kg} ($\sim$5.9M entities, $\sim$4k predicates).

\paragraph{Methods.} We use the following methods to obtain results for the \blackgls{oie} slot linking task: (i) \textsc{Random:} for each \blackgls{oie} slot, we sample a random \blackgls{kg} entry; (ii) \textsc{Frequency:} based on the training data statistics, we link each \blackgls{oie} slot to the most frequent \blackgls{kg} entry (entity or predicate) in the training set; (iii) \textsc{SimCSE:} we use a pre-trained SimCSE model \cite{gao2021simcse}, where we represent each \blackgls{oie} slot and each \blackgls{kg} entry (optionally, its description as well---if available in the KG) in a natural language format, and obtain their embeddings. We finally obtain the cosine similarity between each \blackgls{oie} slot and \blackgls{kg} entry to perform the linking; (iv) \preranker: We train a pre-ranking model as per the setup described in \S\ref{emnlp2023:paragraph:preranker-training};
(v) $+$~Context: We append the context sentence to the OIE\footnote{We structure the input as: ``OIE \texttt{<SENT>} Sentence'', where \texttt{<SENT>} is a separator token.}; (vi) $+$\reranker: We additionally re-rank the top-k ($k=3$) pre-ranked \blackgls{oie} slot links as per the setup in \S\ref{emnlp2023:sec:re-rank}.

\begin{table}[t]
\centering
\resizebox{1.0\textwidth}{!}{
\begin{tabular}{lcccccccc} \toprule
    {} & \multicolumn{4}{c}{Benchmark-Restricted Knowledge Graph} & \multicolumn{4}{c}{Large Knowledge Graph} \\
    \cmidrule(lr){2-5} \cmidrule(lr){6-9}
    {Method} & {Subject ($\uparrow$)} & {Relation ($\uparrow$)} & {Object ($\uparrow$)} & {Fact ($\uparrow$)} & {Subject ($\uparrow$)} & {Relation ($\uparrow$)} & {Object ($\uparrow$)} & {Fact ($\uparrow$)} \\ \midrule
    \multicolumn{9}{l}{\textit{\textbf{Transductive data split}}} \\
    \preranker & 86.8 \error{0.1} & 93.5 \error{0.1} & 95.7 \error{0.0} & 79.1 \error{0.1} & 78.7 \error{0.2} & 93.5 \error{0.1} & 93.1 \error{0.1} & 70.7 \error{0.2} \\ \midrule
    \multicolumn{9}{l}{\textit{\textbf{Inductive data split}}} \\
    \preranker & 71.9 \error{0.4} & 69.8 \error{0.5} & 59.5 \error{0.5} & 34.9 \error{0.5} & 62.0 \error{0.5} & 69.8 \error{0.5} & 48.2 \error{0.5} & 25.4 \error{0.4} \\
    $+$ Context & 74.5 \error{0.4} & \textbf{73.8 \error{0.4}} & \textbf{62.3 \error{0.5}} & 38.9 \error{0.5} & \textbf{64.5 \error{0.5}} & \textbf{73.8 \error{0.4}} & 50.8 \error{0.5} & 29.2 \error{0.5} \\
    $+$ \reranker & \textbf{76.2 \error{0.4}} & 67.6 \error{0.5} & 60.8 \error{0.5} & \textbf{40.6 \error{0.5}} & \textbf{64.8 \error{0.5}} & 66.5 \error{0.5} & \textbf{54.8 \error{0.5}} & \textbf{32.9 \error{0.5}} \\ \midrule
    \multicolumn{9}{l}{\textit{\textbf{Polysemous data split}}} \\
    \preranker & 68.8 \error{0.4} & 93.0 \error{0.2} & 94.6 \error{0.2} & 62.5 \error{0.4} & 58.2 \error{0.4} &  93.0 \error{0.2} & 92.4 \error{0.2} & 51.7 \error{0.4} \\
    $+$ Context & \textbf{77.9 \error{0.3}} & \textbf{93.3 \error{0.2}} & \textbf{95.8 \error{0.2}} & \textbf{71.6 \error{0.4}} & 69.5 \error{0.4} & \textbf{93.3 \error{0.2}} & 94.0 \error{0.2} & 63.5 \error{0.4} \\
    $+$ \reranker & 75.3 \error{0.4} & 90.6 \error{0.2} & 95.3 \error{0.2} & 69.5 \error{0.4} & \textbf{74.1 \error{0.4}} & 90.6 \error{0.2} & \textbf{95.6 \error{0.2}} & \textbf{69.3 \error{0.4}} \\ \bottomrule
\end{tabular}
}
\caption[Experiments for linking Open Information Extractions to regular and large-scale Knowledge Graphs]{OIE slot and fact-linking accuracy with models trained and evaluated on REBEL. Evaluation on Benchmark-Restricted KG and Large KG.}
\label{emnlp2023:table:all-splits-new}
\end{table}

\paragraph{Results.}
In Table~\ref{emnlp2023:table:all-splits-new}, we report the OIE-to-KG linking performance for each \blackgls{oie} slot, as well as linking on the fact level. Overall, across the data splits, we observe that the unsupervised baselines perform poorly compared to models proposed in this chapter, indicating that \blackgls{oie} linking is not trivial, hence off-the-shelf zero-shot models fail.\footnote{Note that simply linking the \blackgls{oie} relation to the most frequent \blackgls{kg} predicate yields high accuracy, due to the \blackgls{kg} predicates imbalance in REBEL.} Additionally, there is a significant performance drop on the inductive and polysemous split compared to the transductive split, suggesting that the models are neither robust w.r.t.~entities unseen during training, nor can successfully cope with polysemous entities. Expectedly, leveraging extra context (via the sentence from where the \blackgls{oie} is obtained) aids the linking process in the inductive and polysemous split, as it helps generalization and disambiguation. Similarly, the \reranker brings a significant performance gain, especially prominent for linking complete facts. Finally, we observe that the \blackgls{kg} size affects the performance: across all splits, OIE-to-KG linking is more challenging using large-scale \blackgls{kg} compared to the smaller Benchmark-Restricted KG.

Altogether, we shed light on \inlineresearchquestion{Research Question 3}: we can train models that translate structured text to canonical knowledge graph facts, while the effectiveness is tightly coupled with the type of data (i.e., transductive, inductive, or polysemous) and knowledge graph size.

\paragraph{Training on Synthetic Data Improves Performance.}
We explore the extent to which we can learn fact-linking models using synthetic data. SynthIE \cite{josifoski2023exploiting} is a dataset that features natural language sentences paired with \blackgls{kg} facts, where the sentences are obtained using a \blackgls{llm}. Namely, given a set of \blackgls{kg} facts, Josifoski~\etal~\cite{josifoski2023exploiting} prompt the \blackgls{llm} to generate a sentence which mentions (i.e., entails) all of the \blackgls{kg} facts. Notably, if we can link \blackgls{oie} slots to \blackgls{kg} facts by training on such synthetic dataset, then the \blackgls{kg} remains the only human-annotated component for learning the OIE-to-KG fact linking task.

To measure to what extent we can link \blackgls{oie}s to \blackgls{kg} facts using only synthetic data, we train \preranker models on both REBEL and SynthIE, and report results (in Table~\ref{emnlp2023:table:across-splits}) on inductive testing splits w.r.t. each dataset.\footnote{To preserve the inductive property, SynthIE's inductive test split contains only entities found in SynthIE that are not present in the REBEL train data.}
\begin{table}[t]
\centering
\resizebox{0.8\textwidth}{!}{
\begin{tabular}{llcccc} \toprule
    {Train Data} & {Test Data} & {Subject ($\uparrow$)} & {Relation ($\uparrow$)} & {Object ($\uparrow$)} & {Fact ($\uparrow$)} \\ \midrule
    REBEL & REBEL & 62.0 \error{0.5} & 69.8 \error{0.5} & 48.2 \error{0.5} & 25.4 \error{0.4} \\
    REBEL & SynthIE & 53.6 \error{0.4} & 57.2 \error{0.4} & 44.9 \error{0.4} & 17.4 \error{0.3} \\
    \multicolumn{2}{l}{\large\textbf{\textit{Macro Score}}} & 57.8 & 63.5 & 46.6 & 21.4 \\ \midrule
    SynthIE & REBEL & 69.7 \error{0.3} & 63.2 \error{0.3} & 41.6 \error{0.3} & 22.4 \error{0.2} \\
    SynthIE & SynthIE & 64.1 \error{0.6} & 81.2 \error{0.5} & 57.6 \error{0.6} & 34.5 \error{0.6} \\
    \multicolumn{2}{l}{\large\textbf{\textit{Macro Score}}} & \textbf{66.9} & \textbf{72.2} & \textbf{49.6} & \textbf{28.5} \\
    \bottomrule
\end{tabular}
}
\caption[Experiments for linking Open Information Extraction using synthetic data]{OIE slot and fact-linking accuracy on human-created (REBEL) and synthetically generated (SynthIE) datasets. \blackgls{oie} linking to a Large \blackgls{kg} variant on inductive splits with respect to each dataset.}
% (see Appendix~\ref{emnlp2023:appendix:inductive-splits} for details).}
\label{emnlp2023:table:across-splits}
\end{table}
We observe that models trained on SynthIE are overall better OIE-to-KG fact linkers than models trained on REBEL (i.e., higher macro accuracy across datasets). This indicates that learning to link \blackgls{oie}s to KGs is possible using only synthetic data, thus the only human-annotated requirement remains to be a reference KG.

\paragraph{Ablation Study: Importance of Entity Alias Augmentation.} We observe that in current datasets most surface-form entities (in the natural language sentences) appear only with their canonical label.\footnote{REBEL is constructed from Wikipedia abstracts, where the references use the canonical form name; SynthIE provides the \blackgls{kg} fact (in text format) as is to the LLM, so naturally, the sentence generated does not feature the entity aliases. Therefore, \emph{Michael Jordan}'s synonyms such as \emph{M.J., Air Jordan} and \emph{His Airness,} rarely appear in the data.} Since the \blackgls{oie}s represent surface-form facts, such lack of diversity prevents the models from learning more complex linking patterns. To overcome this, we perform entity alias augmentation in \textbf{FaLB} by adding the surface-form aliases of the entities---available in Wikidata and manually curated---and ablate its impact on the \blackgls{oie} linking task. Besides \preranker models trained on REBEL and SynthIE \textit{with} entity alias augmentation, we train additional models \textit{without} the alias augmented samples. We report results in Table~\ref{emnlp2023:table:alias-augment} on inductive REBEL and SynthIE testing data, which \textit{does} and \textit{does not} feature entity aliases.

\begin{table}[t]
\centering
\resizebox{0.7\textwidth}{!}{
\begin{tabular}{cccccc} \toprule
    {Training} & {Testing} & {Subject ($\uparrow$) } & {Relation ($\uparrow$)} & {Object ($\uparrow$)} & {Fact ($\uparrow$)} \\
    \multicolumn{2}{c}{Augmentation} & \multicolumn{4}{c}{} \\ \midrule
    \multicolumn{2}{c}{} & \multicolumn{4}{c}{\textbf{\textit{Models trained and evaluated on REBEL}}} \\
    \cmidrule(lr){3-6}
    \tikzxmark & \tikzxmark & 90.7 \error{0.1} & 91.9 \error{0.1} & 87.1 \error{0.2} & 74.0 \error{0.2} \\
    \tikzxmark & \tikzcmark & 61.6 \error{0.2} & 64.8 \error{0.2} & 41.5 \error{0.2} & 25.2 \error{0.2} \\ 
    \multicolumn{2}{l}{\textbf{\textit{Macro Score}}} & 76.2 & 78.4 & 64.3 & 49.6 \\ \midrule
    \tikzcmark & \tikzxmark & 89.0 \error{0.1} & 92.8 \error{0.1} & 85.1 \error{0.2} & 72.4 \error{0.2} \\ 
    \tikzcmark & \tikzcmark & 79.0 \error{0.2} & 92.2 \error{0.1} & 89.0 \error{0.1} & 67.9 \error{0.2} \\
    \multicolumn{2}{l}{\textbf{\textit{Macro Score}}} & \textbf{84.0} & \textbf{92.5} & \textbf{87.1} & \textbf{70.15} \\ \midrule
    \multicolumn{2}{c}{} & \multicolumn{4}{c}{\textbf{\textit{Models trained and evaluated on SynthIE}}} \\
    \cmidrule(lr){3-6}
    \tikzxmark & \tikzxmark & 91.8 \error{0.2} & 86.3 \error{0.2} & 90.7 \error{0.2} & 72.8 \error{0.3} \\
    \tikzxmark & \tikzcmark & 61.2 \error{0.2} & 70.1 \error{0.2} & 40.7 \error{0.2} & 22.6 \error{0.2} \\
    \multicolumn{2}{l}{\textbf{\textit{Macro Score}}} & 76.5 & 78.2 & 65.7 & 47.7 \\ \midrule
    \tikzcmark & \tikzxmark & 91.1 \error{0.2} & 88.9 \error{0.2} & 90.5 \error{0.2} & 74.3 \error{0.3} \\
    \tikzcmark & \tikzcmark & 80.1 \error{0.2} & 89.9 \error{0.1} & 86.3 \error{0.2} & 64.8 \error{0.2} \\
    \multicolumn{2}{l}{\textbf{\textit{Macro Score}}} & \textbf{85.6} & \textbf{89.4} & \textbf{88.4} & \textbf{69.6} \\ \bottomrule
\end{tabular}
}
\caption[Experiments to study the impact of data augmentation]{OIE slot and fact-linking accuracy to study the impact of the \textbf{FaLB} entity alias augmentation step on REBEL and SynthIE. Inference is done with \preranker on the full test set (i.e., containing transductive, inductive, and polysemous OIE triplets).}
\label{emnlp2023:table:alias-augment}
\end{table}

Expectedly, we observe that training with entity aliases allows us to link such \blackgls{oie} entity mentions more successfully than training without them. This was reflected in models trained on both REBEL and SynthIE. We further observe that this step hurts the linking of specific \blackgls{oie} entity mentions only moderately, suggesting that \blackgls{oie} linking methods could be trained to be robust w.r.t.~entity synsets. Finally, across all \blackgls{oie} slots and fact linking, the macro scores are significantly in favor of the model trained with entity aliases, on both datasets.
% \begin{figure}[t]
% \centering
% \includegraphics[width=0.6\textwidth]{chapters/emnlp2023/images/reranker-k.png}
% \caption{Ablating the impact of $k$ in the \reranker.}
% \label{emnlp2023:fig:reranker-k}
% \end{figure}
% \paragraph{Ablation Study: Importance of Fact-reranking.} We ablate the number of \blackgls{kg} facts we rerank ($k$) with \reranker and report results in Figure~\ref{emnlp2023:fig:reranker-k}. We observe the highest fact linking accuracy when performing reranking using the top-2 highest scoring \blackgls{kg} entries for each \blackgls{oie} slot, although performance is within 1 standard deviation for $k = 2,3,4$. If reranking $k=2$ facts, effectively, the \reranker constructs $2^3$ OIE-KG fact pairs, reranks the list, and returns the highest scoring \blackgls{kg} fact.

\subsection{Detecting Out-of-Knowledge Graph OIEs}\label{emnlp2023:sec:out-of-kg-detection}
\begin{table*}[t]
\centering
\resizebox{1.0\textwidth}{!}{
\begin{tabular}{lccccccccc} \toprule
    {} & \multicolumn{4}{c}{Benchmark-Restricted Knowledge Graph} & \multicolumn{4}{c}{Large Knowledge Graph} \\
    \cmidrule(lr){2-5} \cmidrule(lr){6-9}
    {Method} & {Subject ($\uparrow$)} & {Relation ($\uparrow$)} & {Object ($\uparrow$)} & {Fact ($\uparrow$)} & {Subject ($\uparrow$)} & {Relation ($\uparrow$)} & {Object ($\uparrow$)} & {Fact ($\uparrow$)} \\ \midrule
    \multicolumn{9}{l}{\textit{\textbf{Testing data with entity alias augmentation}}} \\
    Random & 50.0 \error{1.3} & \textbf{50.0 \error{1.3}} & 50.0 \error{1.3} & 12.5 \error{0.8} & 50.0 \error{1.3} & \textbf{50.0 \error{1.3}} & 50.0 \error{1.3} & 12.5 \error{0.8} \\
    Entropy-based Heuristic & \textbf{67.7 \error{1.2}} & \textbf{49.0 \error{1.2}} & \textbf{68.6 \error{1.1}} & \textbf{22.8 \error{0.4}} & \textbf{63.1 \error{1.1}} & \textbf{49.0 \error{1.2}} & \textbf{64.5 \error{1.1}} & \textbf{23.0 \error{0.4}} \\
    Query-Key-Value Cross-Attention & 62.4 \error{1.2} & \textbf{48.8 \error{1.0}} & 63.8 \error{1.2} & 17.2 \error{0.2} & 55.3 \error{1.2} & \textbf{49.2 \error{1.2}} & 56.8 \error{1.2} & 16.6 \error{0.2} \\ \midrule
    \multicolumn{9}{l}{\textbf{\textit{Testing data without entity alias augmentation}}} \\
    Random & 50.0 \error{2.8} & \textbf{50.0 \error{2.8}} & 50.0 \error{2.8} & 12.5 \error{1.8} & 50.0 \error{1.3} & \textbf{50.0 \error{1.3}} & 50.0 \error{1.3} & 12.5 \error{1.8} \\
    Entropy-based Heuristic & \textbf{77.5 \error{2.3}} & \textbf{49.0 \error{2.7}} & \textbf{78.4 \small{$\pm$}} 2.2 & \textbf{29.9 \error{1.7}} & \textbf{73.3 \error{2.3}} & \textbf{49.1 \error{2.7}} & \textbf{72.8 \error{2.4}} & \textbf{27.0 \error{1.7}} \\
    Query-Key-Value Cross-Attention & 70.7 \error{2.5} & \textbf{49.6 \error{2.7}} & 70.9 \error{2.5} & 25.9 \error{1.0} & 59.8 \error{2.7} & \textbf{49.4 \error{2.7}} & 60.8 \error{2.5} & 19.7 \error{0.9} \\
    \bottomrule
\end{tabular}
}
\caption[Experiments for detecting Out-of-Knowledge Graph entities and predicates]{Detection accuracy of Out-of-Knowledge Graph entities and predicates on the Out-of-KG split (built on top of SynthIE). All models trained on REBEL with entity alias augmentation.}
\label{emnlp2023:table:out-of-kg}
\end{table*}

\paragraph{Setup.} Prior works which study the \blackgls{oie} to \blackgls{kg} linking problem \cite{zhang2019openki, wood2021integrating, jiang2021cori} assume that \emph{all} \blackgls{oie} slots that need to be linked are present in the KG, which is rarely the case in practice. Here, we study (i) whether \blackgls{oie} slot linking methods can be converted to Out-of-KG detectors; (ii) the performance impact of the \blackgls{kg} size on the Out-of-KG detection task; and (iii) to what extent is it more difficult to recognize the presence or absence of aliased \blackgls{oie} entity mentions.

\paragraph{Datasets.} We create the Out-of-KG test split by imposing constraints to mimic an out-of-distribution setting. The constraints are: (i) both the \blackgls{kg} entities and predicates referred by \blackgls{oie}s are unseen during training (i.e., not in the KG); (ii) the testing OIE-to-KG pairs do not come from the training data distribution: we use models trained on REBEL, but evaluate on SynthIE (which we use to create the Out-of-KG split). We measure Out-of-KG detection performance on the original OIEs, and \blackgls{oie}s with alias augmented entity slots.

\paragraph{Evaluation Protocol.} We evaluate to what extent we can detect if an \blackgls{oie} slot refers to a concept outside of the KG. For each \blackgls{oie} slot, we add its corresponding concept (KG entity or predicate) to the KG, and score a hit if the model outputs a positive score for that slot. Conversely, for each \blackgls{oie} slot, we remove its corresponding entity or predicate from the KG, and score a hit if the model outputs a negative score. We report the averaged accuracy over the two scenarios for each \blackgls{oie} slot.

\paragraph{Methods.} We create three methods on top of \preranker trained on REBEL:
\begin{inparaenum}[(i)]
\item \textsc{Confidence@1-based heuristic}: We get the cosine similarities of the top-5 predictions from \preranker, compute the softmax, and threshold the top-1 confidence;
\item \textsc{Entropy-based heuristic:} We also obtain the top-5 cosine similarities, however, after softmax normalization compute their entropy. Finally, we threshold the entropy to obtain the prediction;\footnote{For each method, we find the optimal threshold on a hold-out set which we build on top of the REBEL validation set.}
\item \textsc{Query-Key-Value Cross-Attention}: We train a lightweight query-key-value cross-attention module on top of the frozen \preranker embeddings. Given a query \blackgls{oie} slot embedding, the model attends over the \blackgls{kg} entry embeddings representing the keys and values, and outputs a probability indicating the presence of the \blackgls{oie} slot in the \blackgls{kg}
% (See Appendix~\ref{emnlp2023:appendix:out-of-kg-models} for details).
\end{inparaenum}

\paragraph{Results.} We report the Out-of-KG detection performance in Table~\ref{emnlp2023:table:out-of-kg}. First, we observe that none of the models we evaluate can recognize whether an \blackgls{oie} relation has a corresponding \blackgls{kg} predicate, indicating that models do not cope with zero-shot relations. Intuitively, the number of entities is orders of magnitude more than the relations, thus the models learn features that generalize to unseen data. On the other hand, the number of relations is limited ($\sim$600 during training) and therefore, the models overfit on this limited set. Second, similar to our observations on the \blackgls{oie} linking task, detection of out-of-KG slots is significantly more difficult on data that features entity aliases, and even more difficult using larger KGs. Lastly, the best-performing model is an entropy-based heuristic on top of the \preranker output scores.

Overall, our experiments clarify \inlinesubresearchquestion{Research Question 3.1}: We can identify entities expressed in structured text as being outside of the knowledge graph, but we cannot do the same for the relations, suggesting a potential avenue for future work.
\section{Conclusion}
% \researchquestion{Research Question 3}{How effectively can we link general-purpose textual data, represented as structured triplets, to corresponding facts in a knowledge graph?}

% A key consideration is that not every triple can be directly associated with a knowledge graph fact. Specifically, triples that mention entities not present in the Knowledge Graph, such as new or erroneous information, pose a challenge. This leads us to our subsequent research question:

% \textit{Research Question 3.1:} Can we accurately identify instances where a valid translation between the text triplets and knowledge graph facts does not exist? That is, the text represents knowledge which is not in the knowledge graph (i.e., factually incorrect or novel information).
In this chapter, we explored the task of translating structured text---given by Open Information Extractions---to canonical facts within large-scale knowledge graphs, aiming to unify the broad coverage of the Open Information Extractions with the precise, unambiguous property of knowledge graphs.

Regarding \inlineresearchquestion{Research Question 3}, we developed a novel benchmark designed for a comprehensive evaluation across various facets. We observed that translating between the modalities is most challenging using out-of-distribution data, while, on a positive note, we can train multimodal translation models for this task using only synthetic data (e.g., data generated by Large Language Models).

Concerning \inlinesubresearchquestion{Research Question 3.1}, we revealed that despite the models' capability to translate \blackgls{oie}s to \blackgls{kg} facts, detecting out-of-KG entities and predicates (i.e., entities and predicates for which the translation is not feasible) remains a challenging task. We thus emphasize a critical area where current methods fall short, highlighting the need for novel approaches.

\textbf{Limitations and future work.} Even though in this chapter we have made significant progress toward addressing the aforementioned research questions, we acknowledge several limitations. Namely, in our methods, we do not utilize the \blackgls{kg} structural information. Future work could explore methods which leverage the \blackgls{kg} structure to enhance the models' ability to generalize to unseen entities or relations. Furthermore, we only focus on the English language, while future work could explore similar approaches which are also multilingual.
% We shed light on the OIE to \blackgls{kg} linking problem, allowing us \textit{to fuse the surface-form and open-ended knowledge found in OIEs, with the canonical real-world \blackgls{kg} facts}. We introduced a novel multifaceted benchmark that fixes prior work deficiencies and proposed a set of task-specific baselines. Our experiments uncover that (i) linking inductive or polysemous OIEs to large KGs is challenging; (ii) we can learn OIE linking methods using only synthetic data; and (iii) detecting whether OIEs are out-of-KG is an open research problem.

% \textbf{Limitations}
% Notably, the set of models we explore ignore the \blackgls{kg} structure to obtain \blackgls{kg} entry embeddings. Leveraging the underlying graph structure should, in theory, yield representations that generalize better to zero-shot samples (e.g., as is the case with detecting out-of-KG relations). Such \blackgls{kg} entry embeddings could even be trained offline (i.e., as a separate step) with standard \blackgls{kg} embedding methods \cite{bordes2013translating}. 

% Last but not least, all data, resources, and models used in this work are specific to the English language. Notice, however, that our approach can be readily extended to languages other than English, while Wikipedia and Wikidata have versions in other languages -- which we leave for future work.

% \textbf{Ethical Impact} We are not aware of any direct ethical impact generated by our work. However, in general, care should be taken when applying our technology to sensitive use cases in high-risk domains, such as healthcare.

\cleardoublepage%

% \includechapter{epilogue_part_i}
\part{Multimodal Fusion and Transference}\label{part:fusion-transference}
% \includechapter{preamble_part_ii}
%
  \graphicspath{{\chaptersdir/bmvc2021/image/}}%
  %\cleardoublepage%
  \chapter{Multimodal Fusion for Improving Compositional Action Recognition}\label{ch:bmvc2021}
In this chapter, we address the challenge of multimodal fusion, specifically focusing on the task of recognizing human actions from videos. At its core, the task of identifying human actions requires understanding the spatial and temporal interactions between the relevant video objects, which often necessitates abstracting away some of the visual characteristics of the video (i.e., its appearance).

Based on this hypothesis, we first propose a model that leverages only object detections as input (represented as an abstract modality), and emphasizes the spatial and temporal interactions between the video objects, without being influenced by the video appearance. We then delve into the intricacies of multimodal fusion, that is, combining the representation spaces of the object detections and the video appearance. Using the Something-Else \cite{materzynska2020something} and the Action Genome \cite{ji2020action} datasets, we  
\begin{inparaenum}[(i)]
\item explore how to design transformer-based models for spatio-temporal object detections-based action recognition,
\item experiment with various multimodal fusion strategies of the representation spaces from an object detections-based model and a video appearance-based model.
\end{inparaenum}

The content of this chapter is based on the following publication (Oral presentation):

\begin{mdframed}[backgroundcolor=gray!30,linewidth=0pt]
\underline{Radevski, G.}, Moens, M-F., and Tuytelaars, T. (2021). Revisiting Spatio-Temporal Layouts for Compositional Action Recognition. The British Machine Vision Conference (BMVC).
\end{mdframed}

\section{Introduction}
Whether a person is ``taking an apple out of a box'' or  ``taking a screwdriver out of a box'', we (humans) can recognize the action performed with ease. In fact, even if we have never seen the object before, we are still able to recognize the action that occurred. Moreover, for us, it makes no difference where the action takes place (indoors, outdoors, etc.), as long as the spatio-temporal interactions between the objects involved in the action are visible. This suggests that in certain settings (e.g., compositional generalization), action recognition should be, to a degree, invariant to the appearance of the objects, as well as the environment where the action takes place.

In spite of the aforementioned, most state-of-the-art action recognition methods are appearance-based 3D convolutional neural networks (\blackgls{3d-cnn})~\cite{lin2019tsm, yang2020temporal, xie2018rethinking, carreira2017quo} or video transformers \cite{liu2021swin, tong2022videomae}. These methods are indeed powerful, albeit heavily reliant on large-scale (pre-)training datasets \cite{kataoka2020would}. Unfortunately, despite the great pre-training efforts taken, their performance rapidly deteriorates on compositional action recognition: i.e., when the objects encountered at inference time are novel \cite{materzynska2020something}.

For this reason, several works \cite{baradel2018object, wang2018videos, sun2018actor, girdhar2019video, zhang2019structured} have advocated for an object-centric approach for video action recognition, reporting improved robustness, interpretability, and overall performance. To achieve object-centric video understanding, on top of the video appearance (RGB frames), these works either use a region proposal network \cite{wang2018videos, girdhar2019video, baradel2018object, sun2018actor}, or leverage an independently trained object detector (e.g., Faster R-CNN \cite{ren2015faster}), to obtain object detections for each video frame \cite{materzynska2020something, yan2020interactive}. Within these methods, one model is specialized for encoding the object detections, while the RGB-based model (e.g., 3D CNN) specializes in encoding the video appearance. Then, some works \cite{wang2018videos, girdhar2019video, yan2020interactive} perform an early fusion of the two representation spaces, while others \cite{materzynska2020something} perform only late fusion without allowing for fine-grained knowledge transfer in the earlier stages of the fusion process.

In contrast to these works, we seek to answer the following research question:

% In a \textit{two-branch model} the layout and appearance input follow separate pathways and are fused late, while in a \textit{one-branch model} they follow a single pathway (fused early in the model). Some of the limitations include:
% \begin{inparaenum}[(i)]
% \item With a two-branch model, fusion is performed by concatenation, not fully exploiting the complementarity of the spatio-temporal layouts and the video appearance \cite{baradel2018object, materzynska2020something, wang2018videos}, and
% \item the layout module is treated as a peripheral component \cite{wang2018videos, ji2020action}, so it remains unclear to what extent in different evaluation settings (compositional, few-shot, background cluttered videos), a well-assembled layout-based model can recognize human actions.
% \end{inparaenum}

\researchquestion{Research Question 4}{What is the most effective way to utilize multimodal fusion of two abstract modalities---representations of video frames and object detections---in order to enhance task-specific performance in video-based action recognition?}

The core of multimodal fusion lies in effectively integrating diverse modalities to leverage their unique strengths. Therefore, understanding how to maintain this complementarity is crucial for ensuring that the integrated representation is richer and more informative than the representation of any single modality alone. This is particularly important in video action recognition, where different modalities---like video frames and object detections---contribute unique information. Consequently, we ask:

\subresearchquestion{Research Question 4.1}{How do we preserve the complementarity of the modality representations prior to their fusion, which is the most crucial property that needs to be maintained to achieve optimal task performance?}

% At the same time, the transformer model \cite{vaswani2017attention} has been demonstrated to be a powerful common-sense reasoning tool over sets of spatially distributed objects in images for visual question-answering \cite{tan2019lxmert, lu2019vilbert}, abstract scenes generation \cite{radevski2020decoding}, etc. By performing beyond-pairwise spatial reasoning, it encapsulates the scene's global spatial context, which is indicative of its semantics, to a certain extent. Just as importantly, a variety of works specifically examine the problem of multimodal fusion \cite{neverova2015moddrop,vielzeuf2018centralnet,perez2019mfas}, attempting to determine how and where to fuse the different modalities.

\textbf{Contributions.} In this chapter we focus on the task of compositional and few-shot action recognition, and present the following key contributions:

\circled{1} We propose a transformer-based method, which we apply purely over highly abstract concepts (i.e., object detections).
% successful action recognizer even in isolation;

\circled{2} We perform an exhaustive experimental study to reveal how to optimally fuse the multimodal representations of the object detections-based model and video 
appearance-based model, to ultimately improve the action recognition performance.

\circled{3} We perform experiments on background cluttered video data such as Action Genome \cite{ji2020action}, and find that multimodal fusion by ensembling the predictions of the object detections-based model and the video appearance-based model significantly improves the action recognition performance.

We release code, data, and trained models at: \url{https://github.com/gorjanradevski/revisiting-spatial-temporal-layouts}.
\section{Related Work}
Since this work came to fruition in mid-2021, in this section we discuss \textit{only} the related and concurrent works at the time of publication. When we conclude this chapter in \S\ref{bmvc2021:sec:conclusion}, we discuss some of the future research published concurrently or soon after our work.

Action recognition methods are mostly video-appearance-based \blackgls{3d-cnn} \cite{carreira2017quo, xie2018rethinking, lin2019tsm, simonyan2014two, kataoka2020would, wang2016temporal, simonyan2014two, feichtenhofer2019slowfast, yang2020temporal} or video transformers \cite{liu2021swin,bertasius2021space}. These methods often use a 2D CNN/Transformer pre-trained on ImageNet \cite{deng2009imagenet}, subsequently inflated to 3D \cite{carreira2017quo}. Other works explore (pre)training 3D CNNs \cite{kataoka2020would} on large-scale curated datasets \cite{carreira2019short, monfort2019moments}, as well as the best practices for doing so \cite{wang2016temporal}. Finally, several works explore efficient video understanding methods \cite{lin2019tsm, xie2018rethinking, tran2018closer}.
%, or propose plug-in components to improve the temporal reasoning \cite{zhou2018temporal}

\textbf{Applications of Multi-Head Attention (MHA) in computer vision.}
% The applications of MHA in computer vision are rapidly expanding \cite{khan2021transformers}.
Up to mid-2021, MHA has been applied in conjunction with a CNN \cite{carion2020end, sun2019videobert, girdhar2019video, meng2019interpretable}, as a stand-alone MHA over low-level, raw image pixels \cite{dosovitskiy2021an, wang2020axial, Cordonnier2020On, bertasius2021space}, for vision and text tasks \cite{lu2019vilbert, tan2019lxmert}, or tasks involving spatial understanding \cite{radevski2020decoding} to name a few. In contrast, in this work, we apply MHA:
\begin{inparaenum}[(i)]
\item over high-level, abstract, spatio-temporal object detections, and
\item to perform multimodal fusion of the representation spaces of two distinct modalities (object detections and video appearance).
\end{inparaenum}

\textbf{Object-centric methods for action recognition (with attention).} The issue with appearance-based action recognition methods is their inherent tendency to overfit on the appearance of the environment and the objects, deteriorating the performance on fine-grained \cite{baradel2018object} or compositional datasets \cite{materzynska2020something}. Recently, multiple works that address this issue emerged \cite{baradel2018object, materzynska2020something, girdhar2019video, ji2020action, wang2018videos, sun2018actor, perez2020knowing, xu2020spatio, materzynska2020something, yan2020interactive}. Object Relation Network \cite{baradel2018object} models the spatio-temporal interactions of the detected video objects with a \blackgls{gru} recurrent neural network \cite{cho-etal-2014-learning}. STAR \cite{xu2020spatio} regresses the bounding box coordinates where the action occurs and classifies the action. STRG \cite{wang2018videos} uses a graph CNN \cite{kipf2017semi} and a region proposal network (RPN) \cite{ren2015faster}, applied over I3D features \cite{carreira2017quo}, with a non-local neural network temporal module \cite{wang2018non}. Actor Centric Relation Network \cite{sun2018actor} fuses the cropped feature map of the actors' regions and the global video feature map. In parallel, multiple works leverage attention, to augment existing methods or propose new ones. LFB \cite{wu2019long} uses attention as a non-local block \cite{wang2018non} to accumulate video features. SINet's \cite{ma2018attend} coarse- and fine-grained branches are attention-based, subsequently fused for action recognition. Compared to our work, SINet's fine-grained branch applies attention over the region of interest (\blackgls{roi}) pooled features from an RPN for each frame, subsequently fed to LSTM \cite{hochreiter1997long}. In turn, we apply MHA over the object detections (category and bounding box) to encode the spatio-temporal interactions in the videos. W3 \cite{perez2020knowing} is an attention-based plug-in module on top of appearance models, while SGFB \cite{ji2020action} utilizes attention through LFB \cite{wu2019long} over per-frame scene graphs, combined with I3D \cite{carreira2017quo} and a non-local neural network\cite{wang2018non}.

The Video Action Transformer \cite{girdhar2019video} uses I3D \cite{carreira2017quo} in conjunction with RPN \cite{ren2015faster} and a transformer \cite{vaswani2017attention}. It
\begin{inparaenum}[(i)]
\item obtains an I3D feature map around a center frame,
\item generates region proposals for the center frame,
\item applies MHA where the I3D feature map is the memory and the center frame RoI pooled features are the query.
\end{inparaenum}
In our work, we also evaluate the performance of a VAT \cite{girdhar2019video}-inspired fusion scheme between the appearance and object-detections branch. Lastly, STIN \cite{materzynska2020something} and SFI \cite{yan2020interactive} demonstrate that a graph neural network model \cite{kipf2017semi} applied on object detections can surpass I3D's \cite{carreira2017quo} performance for compositional/few-shot action recognition with ground truth object detections, while multimodal fusion with I3D video features embeddings improves the performance. Unlike these works, inspired by MHA-based methods for spatial understanding, we
\begin{inparaenum}[(i)]
\item develop a specifically tailored model for object detections-based action recognition which applies attention over high-level, spatio-temporal layouts,
\item empirically evaluate diverse multimodal fusion approaches, to uncover how and when the appearance- and layout-based models should be fused.
\end{inparaenum}

\textbf{Multimodal fusion} attempts to extract the relevant, complementary information from multimodal inputs, resulting in a better joint model, compared to training separate models on the individual modalities. The literature is extensive \cite{lahat2015multimodal, vielzeuf2018centralnet, neverova2015moddrop, arevalo2017gated}, without a universal approach that generalizes across different modalities and tasks. Many works \cite{tan2019lxmert, lu2019vilbert} that use attention have demonstrated remarkable results on multimodal tasks; e.g., VQA \cite{antol2015vqa} when fusing image and text features. In action recognition, late fusion by concatenation works well \cite{materzynska2020something}, also confirmed in our work. We demonstrate that multimodal fusion with cross-attention \cite{tan2019lxmert}, based on the CentralNet approach \cite{vielzeuf2018centralnet}, preserves the complementarity of the modalities and achieves state-of-the-art performance.
\section{Methodology}
In this section, we introduce the necessary Transformer model background (\S\ref{bmvc2021:sec:revisit}), describe how we extend it for modeling spatio-temporal layouts\footnote{Note that we often refer to the video object detections as spatio-temporal layouts, or in some occurrences just ``layouts''.}, i.e., object detections (\S\ref{bmvc2021:sec:spatial-temporal}), discuss the models we use to encode the video appearance and schemes to fuse them with the object detections model (\S\ref{bmvc2021:sec:appearance-fusion}).

\subsection{Transformer Revisited}\label{bmvc2021:sec:revisit}
The core Transformer \cite{vaswani2017attention} model component is the multi-head attention module, defined as: $\operatorname{Attention}(\mathbf{Q}, \mathbf{K}, \mathbf{V}) = \operatorname{Softmax}(\frac{\mathbf{Q}\mathbf{K}^T}{\sqrt{d_k}})\mathbf{V}$, where $\mathbf{Q}$, $\mathbf{K}$, $\mathbf{V}$ are the queries, keys and values respectively, and $d_k$ is the per-attention head hidden size. With self-attention, $\mathbf{Q}$, $\mathbf{K}$, and $\mathbf{V}$ come from the same and only modality (in our case, the layout modality), and with cross-attention, $\mathbf{Q}$ originates from the target modality, while $\mathbf{K}$ and $\mathbf{V}$ from the source modality we attend on. Due to the permutation invariance of the multi-head attention, the input should include positional information. In this work, we rely on
\begin{inparaenum}[(i)]
\item bidirectional and causal attention,
\item self- and cross-attention, and
\item different variants of position embeddings.
\end{inparaenum}
Refer to \S\ref{background:fig:transformer} and Vaswani~\etal~\cite{vaswani2017attention} for details.

\begin{figure*}[t]
\centering
\includegraphics[width=1.0\textwidth]{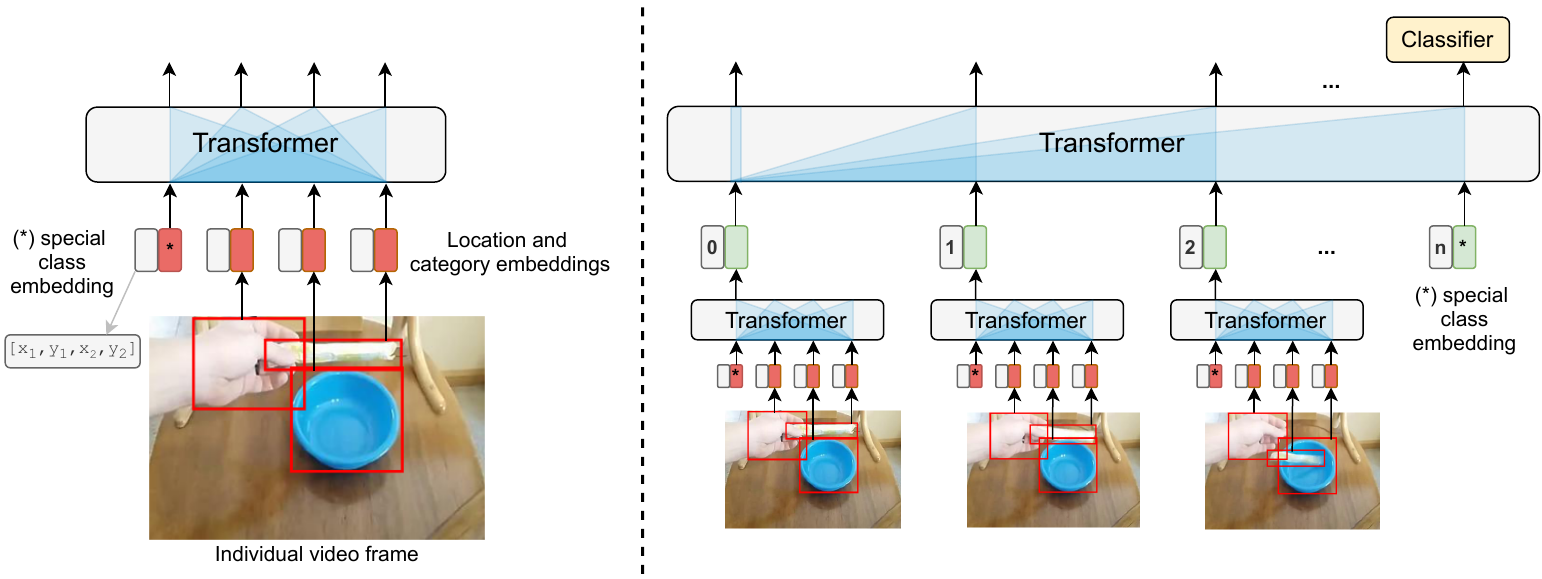}
\caption[Overview of the Spatial-Temporal Layout Transformer]{\blackgls{stlt} overview. \textbf{Left: Spatial-Transformer}. The inputs are the object categories and their [$x_1, y_1, x_2, y_2$] frame location normalized by the frame size. We select the special \texttt{class} embedding as the module output. \textbf{Right: Temporal-Transformer}. The inputs are the Spatial-Transformer outputs, summed with trainable position embeddings. We select the temporal \texttt{class} embedding as the module output, and add a classifier for action recognition.}
\label{bmvc2021:fig:main-model}
\end{figure*}

\subsection{Layout Branch: Spatial-Temporal Transformer}\label{bmvc2021:sec:spatial-temporal}
The input to the layout model (Figure~\ref{bmvc2021:fig:main-model}) is sequence of $n$ frames: $f_1, f_2,...,f_n$. Each frame $f_i$ is composed of $m$ objects: $o_1, o_2,...,o_m$, where $o_j$ consists of the object category $c_j$ and location in the frame $\mathbf{l}_j = [x_1, y_1, x_2, y_2]$. Note that this is all the information required and used by the layout branch: no appearance information, just bounding boxes and category labels. Two separate fully-connected layers yield the category embedding $\mathbf{\hat{c}}_j$ and the frame location embedding $\mathbf{\hat{l}}_j$, which we subsequently sum together and apply layer-normalization~\cite{ba2016layer} and dropout \cite{srivastava2014dropout} to obtain the final object embedding: $\mathbf{\hat{o}}_j = \operatorname{Dropout(LayerNorm(\mathbf{\hat{c}}_j + \mathbf{\hat{l}}_j))}$.

With the layout model, dubbed as Spatial-Temporal Layout Transformer, henceforth abbreviated as \blackgls{stlt}, we decouple the spatial (per-frame) reasoning from the temporal (across-video) reasoning. To that end, to model the per-frame spatial relations, given a set of frame objects: $o_1, o_2, ..., o_{m}$, we firstly prepend an $o_{\text{\texttt{class}}}$ object, with $c_{\text{\texttt{class}}}$ (special \texttt{class} category) and $l_{\text{\texttt{class}}}$ equal to the frame size. Then, we obtain the embedding for each frame object: $\mathbf{\hat{F}}_i = [\mathbf{\hat{o}}_{\text{\texttt{class}}}, \mathbf{\hat{o}}_1, \mathbf{\hat{o}}_2, ..., \mathbf{\hat{o}}_{m}]$, and use a bidirectional transformer (each object embedding can attend on all others). We denote this module as $\operatorname{Spatial-Transformer}$ (Figure~\ref{bmvc2021:fig:main-model}, left), which we apply on the set of object embeddings for each frame $\mathbf{\hat{F}}_i$ separately. Subsequently, we select the output hidden state corresponding to the \texttt{class} category as a global representation of a single frame: $\mathbf{\hat{s}}_i = \operatorname{Spatial-Transformer}(\mathbf{\hat{F}}_i)$.

To model the temporal evolution of the interactions between the video objects, we use a causal transformer (each frame reprensentation can only attend on the past ones), denoted as $\operatorname{Temporal-Transformer}$ (Figure~\ref{bmvc2021:fig:main-model}, right). We append another special \texttt{class} embedding $\mathbf{\hat{s}}_{\text{\texttt{class}}}$ to the outputs of the Spatial-Transformer. Then, for each frame, with a fully-connected layer, we obtain frame-position-in-the-video embedding $\mathbf{\hat{p}}_i$. This is summed with the frame spatial embedding $\mathbf{\hat{s}}_i$, followed by layer-normalization and dropout: $\mathbf{\hat{t}}_i = \operatorname{Dropout(\operatorname{LayerNorm(\mathbf{\hat{s}}_i + \mathbf{\hat{p}}_i)})}$. 
Finally, we forward propagate the sequence of frame embeddings $\mathbf{\hat{T}} = [\mathbf{\hat{t}}_1, \mathbf{\hat{t}}_2, ... \mathbf{\hat{t}}_{n}, \mathbf{\hat{t}}_{\text{\texttt{class}}}]$ through the Temporal-Transformer: $\mathbf{\hat{H}} = \operatorname{Temporal-Transformer}(\mathbf{\hat{T}})$, where $\mathbf{\hat{H}}$ are the output hidden states. Suppose we use \blackgls{stlt} as a standalone action recognizer, i.e., given spatio-temporal layouts as input we want to infer the action; in that case, we select the hidden state corresponding to the \texttt{class} embedding, and add a classifier on top: $\mathbf{\hat{y}} = \operatorname{Linear}(\mathbf{\hat{h}}_{\text{\texttt{class}}})$, where $\mathbf{\hat{y}}$ are the logits.

\subsubsection{Spatial \& Temporal Layout Transformer}
We also explore another model variant, where instead of having decoupled spatial and temporal reasoning over the spatio-temporal layouts, we perform it jointly. We name this model as: Spatial \& Temporal Layout Transformer, abbreviated as \blackgls{s_and_tlt}. Similar to \blackgls{stlt}, we obtain the category embedding $\mathbf{\hat{c}}_j$ and the location in the frame embedding $\mathbf{\hat{l}}_j$ with two separate fully-connected layers. However, in this case, for each frame object $o_j$, we additionally obtain its frame position index, where 
%, e.g., if the object is from the first frame the index would be $0$, while if it is from the last frame, it would be $n$, where
$n$ is the number of sampled video frames.
Therefore, for a single object, we obtain its embedding as $\mathbf{\hat{o}}_j = \operatorname{Dropout}(\operatorname{LayerNorm}(\mathbf{\hat{c}}_j + \mathbf{\hat{l}}_j + \mathbf{\hat{p}}_j))$, where an additional fully-connected layer yields the $\mathbf{\hat{p}}_j$ embedding. Finally, we append a special \texttt{class} object embedding $\mathbf{\hat{o}}_{\text{\texttt{class}}}$ at the sequence end, and forward-propagate the sequence of object embeddings through a transformer model. We control the attention pattern using an attention mask, such that we perform bidirectional attention for objects spanning a single frame, while we perform causal attention across the video\footnote{An object from frame $o_i$, can only attend on the objects from frames $o_1, o_2, ..., o_{i}$.}.

\subsection{Appearance Branch and Multimodal Fusion}\label{bmvc2021:sec:appearance-fusion}
\begin{figure}[t]
\centering
\includegraphics[width=1.0\textwidth]{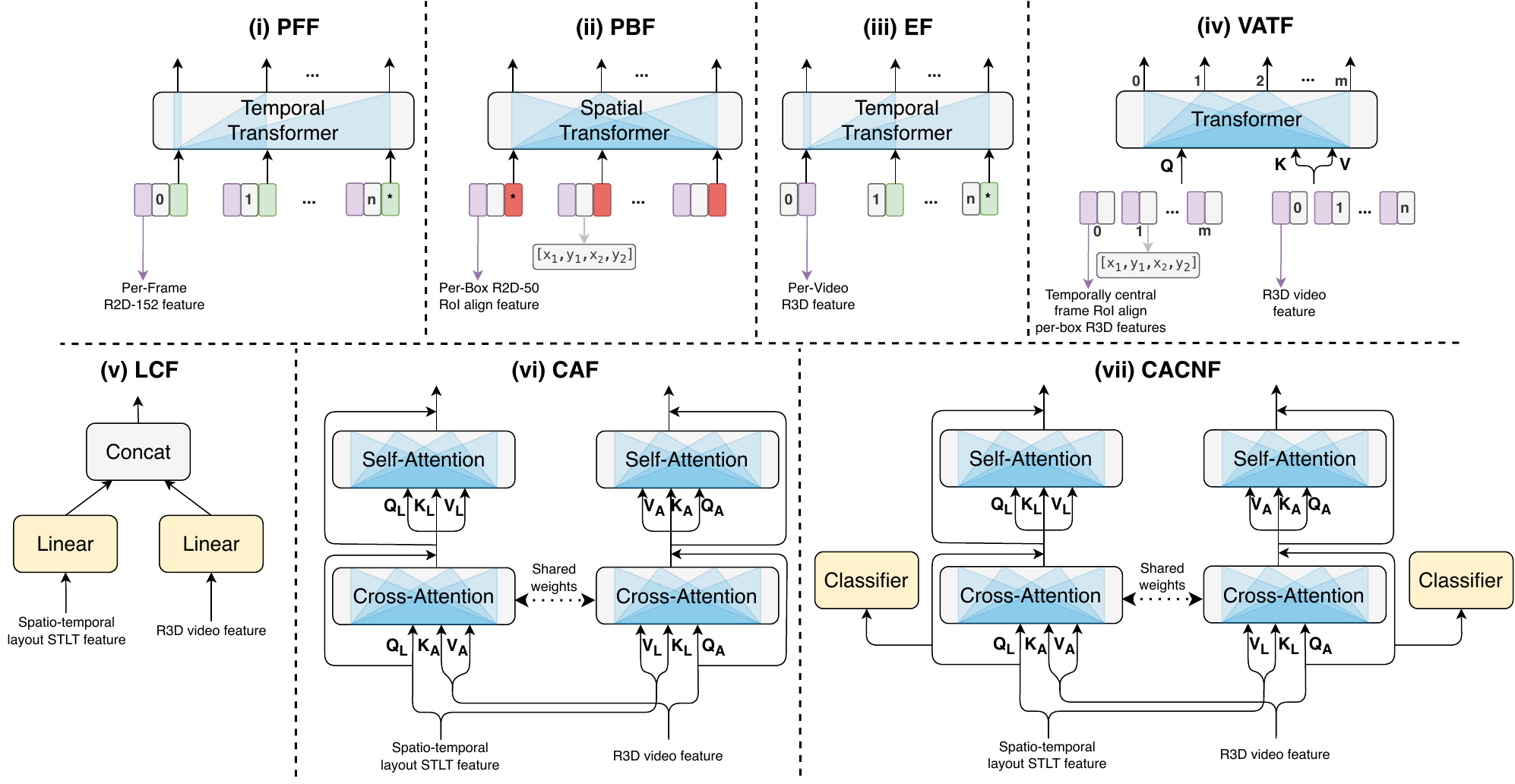}
\caption[Overview of the multimodal fusion schemes]{Fusion schemes. \textbf{Top:} One-branch. \textbf{Bottom:} Two-branch fusion methods.}
\label{bmvc2021:fig:fusion-overview}
\end{figure}
As an appearance model, we deem a neural network (e.g., a CNN \cite{lecun1998gradient} or a Vision Transformer \cite{bertasius2021space}), applied over pixels of video frames. In this work, we use four different types of appearance models, each tightly coupled with the corresponding fusion approaches:
\begin{inparaenum}[(i)]
\item 2D ResNet152 (R2D-152) \cite{he2016deep}, pre-trained on ImageNet \cite{deng2009imagenet}, which we apply over individual frames.
\item 2D ResNet50 (R2D-50) backbone, which we obtain from a COCO \cite{lin2014microsoft} pre-trained Faster R-CNN \cite{ren2015faster}. Given class-agnostic bounding boxes from a video frame, we extract \blackgls{roialign} \cite{he2017mask} features from each.
\item An inflated 3D Resnet50 (\blackgls{r3d}) \cite{kataoka2020would}, pre-trained on \cite{carreira2019short, monfort2019moments, yoshikawa2018stair}.
\item R3D, same as (iii), with a Transformer encoder on top to enable multimodal fusion with cross-attention.
\end{inparaenum}

Due to the specific nature of the appearance features, we devise different ways to fuse them with the layout model (Figure~\ref{bmvc2021:fig:fusion-overview}). We evaluate different configurations where each has its weaknesses, while we gradually build the ultimate, empirically superior fusion approach

\subsubsection{Per-frame Fusion}
With the Per-Frame Fusion (\blackgls{pff}) method---Figure~\ref{bmvc2021:fig:fusion-overview} (i)---our goal is to inject the R2D-152 frame embeddings in \blackgls{stlt}. As we have a separate embedding for each frame, we feed the embeddings directly into STLT's Temporal Transformer module. Note that the output of the Spatial Transformer module for a single frame $\mathbf{\hat{s}}_i$ is summed with a positional embedding $\mathbf{\hat{s}}_i + \mathbf{\hat{p}}_i$, and is provided as input to the Temporal Transformer module. Here, with PFF, we sum the R2D-152 frame embedding $\mathbf{\hat{a}}^f_i$ together with $\mathbf{\hat{s}}_i$ and $\mathbf{\hat{p}}_i$, apply layer-normalization \cite{ba2016layer} and dropout \cite{srivastava2014dropout}: $\mathbf{\hat{t}}_i = \operatorname{Dropout}(\operatorname{LayerNorm}(\mathbf{\hat{s}}_i + \mathbf{\hat{p}}_i + \mathbf{\hat{a}}^f_i))$. Then, as in STLT, we provide $\mathbf{\hat{t}}_i$ as input to the Temporal Transformer. The remaining part of the model follows STLT.

\subsubsection{Per-Box Fusion}
The goal with the Per-Box Fusion (\blackgls{pbf}) method---Figure~\ref{bmvc2021:fig:fusion-overview} (ii)---is to inject appearance information at a low, object-specific level. To that end, we utilize the appearance video features obtained from R2D-50. With STLT, given a frame object $o_j$, we obtain its embedding as $\mathbf{\hat{o}}_j = \operatorname{Dropout}(\operatorname{LayerNorm}(\mathbf{\hat{c}}_j + \mathbf{\hat{l}}_j))$. With PBF, the embedding for a single object region in a single video frame is denoted as $\mathbf{\hat{a}}^r_j$, which we fuse with the object embedding as $\mathbf{\hat{o}}_j = \operatorname{Dropout}(\operatorname{LayerNorm}(\mathbf{\hat{c}}_j + \mathbf{\hat{l}}_j + \mathbf{\hat{a}}^r_j))$. Then, same as \blackgls{stlt}, we provide the object embeddings $\mathbf{\hat{f}}_i$ for a single frame $f_i$ to the Spatial Transformer.

\subsubsection{Early Fusion}
With the Early Fusion (\blackgls{ef}) method---Figure~\ref{bmvc2021:fig:fusion-overview} (iii)---we inject the R3D video appearance features into STLT. First, we obtain $\mathbf{\hat{A}} \in \mathbb{R}^{2048 \times T_d \times H_d \times W_d}$ from R3D, perform average pooling across both the temporal ($T_d$) and the spatial dimensions ($H_d$ and $W_d$), and project the embedding in the corresponding hidden size using a fully-connected layer. Then, we prepend the global video appearance embedding $\mathbf{\hat{a}}_{global}$ as the first element in the sequence of layout-based frame representations obtained from the Spatial Transformer. By doing so, each of the layout-based frame embedding can attend on the R3D video embedding.

\subsubsection{Video Action Transformer Fusion}
We take inspiration from the Video Action Transformer \cite{girdhar2019video}, and attempt to conceptually adapt it to perform multimodal fusion. We abbreviate this fusion method as \blackgls{vatf} and display it in Figure~\ref{bmvc2021:fig:fusion-overview} (iv). To achieve such fusion, given the R3D appearance embedding $\mathbf{\hat{A}}$, we firstly perform average pooling across the time dimension, such that $\mathbf{\hat{A}} \in \mathbb{R}^{2048 \times H_d  \times W_d}$. Then, given a set of bounding boxes corresponding to the temporally central frame, we perform RoI Align and obtain region of interest embeddings (\blackgls{roialign}). Finally, the non-temporally pooled R3D video embedding represents the memory, while the regional embeddings represent the query as input to a transformer model. We select the output hidden state corresponding to the bounding box of the whole frame ($[0,0,w,h]$) and add a classifier on top. For details refer to Video Action Transformer \cite{girdhar2019video}.

\subsubsection{Late Concatenation Fusion}
Late Concatenation Fusion (\blackgls{lcf}) is always used as a strong baseline for multimodal fusion. We visualize the method in Figure~\ref{bmvc2021:fig:fusion-overview} (v). It follows the general paradigm for late fusion, where we obtain the STLT and R3D outputs, linearly project them with separate fully-connected layers so that both embeddings have the same dimensionality, and concatenate the representations before the classifier.

\subsubsection{Cross-Attention Fusion}
The main building block of Cross-Attention Fusion, abbreviated as \blackgls{caf}---Figure~\ref{bmvc2021:fig:fusion-overview} (vi)---is the cross-attention module. We draw inspiration from successful methods for the fusion of text and image embeddings \cite{tan2019lxmert}, mainly used for visual question answering, image-text matching, etc. To be specific, we get \blackgls{stlt}'s and R3D-Transformer's hidden states, $\mathbf{\hat{H}} \in \mathbb{R}^{Z \times T}$ and $\mathbf{\hat{A}} \in \mathbb{R}^{Z \times T_d * H_d * W_d}$ respectively, where $Z$ is the hidden size. $T$ does not need to be equal to $T_d * H_d * W_d$; i.e., we can increase or decrease the number of sampled layout-based or appearance-based frames.
However, they both have to have the same hidden size $Z$. Then, the module consists of cross-attention, self-attention, and feed-forward modules. The cross-attention is applied on each branch separately, such that for the layout branch, the multi-head attention queries are $\mathbf{\hat{H}}$, while the keys and values are $\mathbf{\hat{A}}$. When the cross-attention is applied on the appearance branch, the queries are $\mathbf{\hat{A}}$, while the keys and values are $\mathbf{\hat{H}}$. The remaining modules (self-attention, feed-forward module) closely follow the standard transformer block (refer to \cite{vaswani2017attention} for details). Finally, we select the output hidden state corresponding to the special \texttt{class} embedding from both the layout $\mathbf{\hat{h}}^{\text{\texttt{caf}}}_{\text{\texttt{class}}}$ and the appearance $\mathbf{\hat{a}}^{\text{\texttt{caf}}}_{\text{\texttt{class}}}$ branch, concatenate the embeddings, and add a classifier on top.

\subsubsection{Cross-Attention CentralNet Fusion}
Our motivation for performing multimodal fusion of the layout and appearance branch representations is their complementarity. But even though \blackgls{caf} follows state-of-the-art practices for multimodal fusion of image and text embeddings, we observe that \blackgls{caf} performance is often weaker than \blackgls{lcf} (Table~\ref{bmvc2021:table:compositional} and Table~\ref{bmvc2021:table:few-shot}). We hypothesize that the reason is the layout and appearance branch not preserving their individual capabilities (Figure~\ref{bmvc2021:fig:caf-cacnf}). To overcome this issue, we develop a multimodal fusion method based on CentralNet \cite{vielzeuf2018centralnet}. Namely, \blackgls{caf} remains unchanged, however, before forward-propagating $\mathbf{\hat{H}}$ and $\mathbf{\hat{A}}$ through \blackgls{caf}, we select the hidden states corresponding to \texttt{class}, namely $\mathbf{\hat{h}}_{\text{\texttt{class}}}$ and $\mathbf{\hat{a}}_{\text{\texttt{class}}}$, and add classifiers (implemented as linear layers) on top of each of them. See Figure~\ref{bmvc2021:fig:fusion-overview} (vii).

During training, we minimize three cross-entropy losses between the model predictions and the ground truth action classes:
\begin{inparaenum}[(i)]
\item \blackgls{stlt} loss,
\item 3D-Resnet Transformer loss,
\item \blackgls{caf} loss.
\end{inparaenum}

During inference, we average the logits of the three classifiers and take the action with the maximal probability as the model's prediction.
\section{Evaluation and Discussion}\label{bmvc2021:sec:eval}
We perform experiments on the Something-Something \cite{goyal2017something}, Something-Else \cite{materzynska2020something} (compositional and few-shot split) and the Action Genome datasets \cite{ji2020action}. During training, we randomly sample 16 frames (each represented as spatio-temporal layouts) for \blackgls{stlt}, and uniformly sample 32 RGB frames for models using \blackgls{r3d}, subsequently rescaled to 112 $\times$ 112. We measure performance using mean average precision (\blackgls{map}) and perform training with binary cross-entropy loss.

On the Something-Else compositional and few-shot splits, we perform experiments with Faster R-CNN \cite{ren2015faster} object detections (object predictions setting), and with ground truth object detections (oracle setting), both released by Materzynska~\etal~\cite{materzynska2020something}. The input object categories in STLT are either ``hand'' or ``object'' (i.e., the detections are object-agnostic). We measure performance using top-1 and top-5 accuracy and train all models with the cross-entropy loss.

On the Action Genome dataset, we train a Faster R-CNN \cite{ren2015faster} on these ground truth annotated frames and obtain object detections for the entire videos. We perform experiments using our object detections (obj. predictions setting), as well as the ground truth object detections (oracle setting), released by Ji~\etal~\cite{ji2020action}.

See \S\ref{background:sec:videodata} for details on the datasets.

\subsection{Ablation Studies and Main Findings}
% We ablate our models to gain insights in \textit{why} one should leverage spatio-temporal layouts for action recognition, and \textit{how} to come up with the best approach for it. We do an ablation study on the Something-Else compositional dataset, while we also verify our findings on the Something-Something (non-compositional) validation set, albeit only in an oracle setting.

\begin{table}[t]
\begin{center}
\resizebox{0.85\textwidth}{!}{
\begin{tabular}{lcccccc} \toprule
    {} & \multicolumn{4}{c}{Something-Else: Compositional generalization} & \multicolumn{2}{c}{Something-Something} \\ 
    \cmidrule(lr){2-5} \cmidrule(lr){6-7}
    {} & \multicolumn{2}{c}{Obj. predictions} & \multicolumn{2}{c}{Oracle} & \multicolumn{2}{c}{Oracle}  \\
    \cmidrule(lr){2-3} \cmidrule(lr){4-5} \cmidrule(lr){6-7}
    {Method} & {Action@1 ($\uparrow$)} & {Action@5 ($\uparrow$)}  & {Action@1 ($\uparrow$)} & {Action@5 ($\uparrow$)} & {Action@1 ($\uparrow$)} & {Action@5 ($\uparrow$)} \\ \midrule
    \multicolumn{7}{l}{\textbf{\textit{Object detections-based methods}}} \\
    GNN-NL & 33.3 & 58.9 & 50.7 & 78.6 & 47.1 & 76.3 \\
    S\&TLT & 40.6 & 66.9 & 57.7 & 84.7 & 55.6 & 84.3 \\
    STLT & \textbf{41.6} & \textbf{67.7} & \textbf{59.4} & \textbf{85.8} & \textbf{57.0} & \textbf{85.2} \\ \midrule
     \multicolumn{7}{l}{\textbf{\textit{RGB-based methods}}} \\
    \blackgls{r3d} & 51.3 & 78.6 & 51.3 & 78.6 & 52.2 & 80.7 \\ \midrule
    \multicolumn{7}{l}{\textbf{\textit{One-branch fusion methods}}} \\
    PFF & 47.3 & 73.7 & 62.5 & 87.5 & 62.9 & 88.0 \\
    PBF & 48.5 & 73.2 & 62.9 & 86.5 & \textbf{64.5} & 89.1 \\
    EF & \textbf{52.8 }& \textbf{79.3} & \textbf{63.8} & \textbf{88.1} & 64.4 & \textbf{89.4} \\
    VATF & 49.1 & 78.0 & 53.0 & 79.6 & 54.9 & 82.8 \\ \midrule
    \multicolumn{7}{l}{\textbf{\textit{Two-branch fusion methods}}} \\
    LCF & 54.1 & 79.8 & 66.1 & 88.8 & 64.4 & 89.3 \\
    CAF & 52.3 & 78.9 & 64.4 & 88.6 & 64.5 & 89.1 \\
    CACNF & \textbf{56.9} & \textbf{82.5} & \textbf{67.1} & \textbf{90.4} & \textbf{66.8} & \textbf{90.6} \\ \bottomrule
\end{tabular}
}
\end{center}
\caption[Experiments comparing the performance of different multimodal fusion configurations]{Accuracy comparison between the different model configurations: unimodal methods and multimodal fusion-based methods (one-branch and two-branch). The best method within the group is in bold.}
\label{bmvc2021:table:ablation-studies}
\end{table}

\textbf{Layout branch: How to encode the spatio-temporal layouts?} In Table~\ref{bmvc2021:table:ablation-studies} (Top), we measure STLT's performance against:
\begin{inparaenum}[(i)]
    \item A baseline model \cite{materzynska2020something} with a spatial reasoning graph neural network (GNN) and temporal reasoning non-local block (NL);
    \item An STLT variant performing joint spatial-temporal reasoning---Spatial \& Temporal Layout Transformer, abbreviated as S\&TLT---on unrolled frames' bounding boxes.
% (for details see Appendix~\ref{bmvc2021:appendix:sec:layout}).
\end{inparaenum}

Across different settings and layout types (obj. predictions or oracle), \blackgls{stlt} and S\&TLT significantly outperform GNN-NL, suggesting the appropriateness of multi-head attention-based methods for modeling spatio-temporal layouts. Furthermore, we observe that decoupled spatio-temporal reasoning is preferable: i.e., STLT outperforms S\&TLT which applies attention jointly over all video objects. Additionally, despite decoupled spatial and temporal reasoning over the layouts (STLT) being empirically superior compared to the joint one (S\&TLT), it is also computationally more efficient; i.e., with STLT, each object can attend only on objects within its frame, while with S\&TLT, each object can attend on all objects throughout the video. In Big-O notation, STLT has a time complexity of $O(n \cdot m^2) + O(n^2)$, while S\&TLT has a time complexity of $O(n^2 \cdot m^2)$.

\textbf{Multimodal fusion: How and where to fuse?} In Table~\ref{bmvc2021:table:ablation-studies}, we report action recognition results with the fusion methods we explore in this manuscript. We compare among the fusion methods plus a fine-tuned \blackgls{r3d} \cite{kataoka2020would}, and summarize our empirical findings as:
\begin{inparaenum}[(i)]
\item 2D fusion methods (\blackgls{pff}, \blackgls{pbf}) exhibit good performance on non-compositional datasets (Something-Something), where object appearance matters, while their performance deteriorates datasets which measure the compositional generalization ability of the methods (Something-Else);
\item Early Fusion (EF) yields a competitive performance across different datasets and is superior to the other one-branch fusion methods;
\item The Video Action Transformer \cite{girdhar2019video} fusion type (VATF), does not fully exploit the spatial-temporal layout and video-context specific to the action (it only performs \blackgls{roialign} on the temporally central frame); thus it yields consistently lower results on these types of data;
\item one-branch methods only marginally outperform \blackgls{r3d} without oracle object detections;
\item Late Fusion by Concatenation (LCF) remains a strong baseline, as reported by others \cite{vielzeuf2018centralnet, neverova2015moddrop};
\item Cross-Attention Fusion (\blackgls{caf}) is consistently weaker than \blackgls{lcf} and \blackgls{cacnf}, while \blackgls{cacnf} (CAF with CentralNet \cite{vielzeuf2018centralnet}) outperforms other methods regardless of the data type (compositional, non-compositional) and type of layouts (obj. predictions or oracle) \inlineresearchquestion{Research Question 4}.
\end{inparaenum}
% These observations are related to \inlineresearchquestion{Research Question 4}: certain fusion methods are inferior regardless of the type of data and object detections, while others (i.e., CACNF) yield superior performance in all our experiments.

\begin{figure}[t]
\centering
\includegraphics[width=0.8\textwidth]{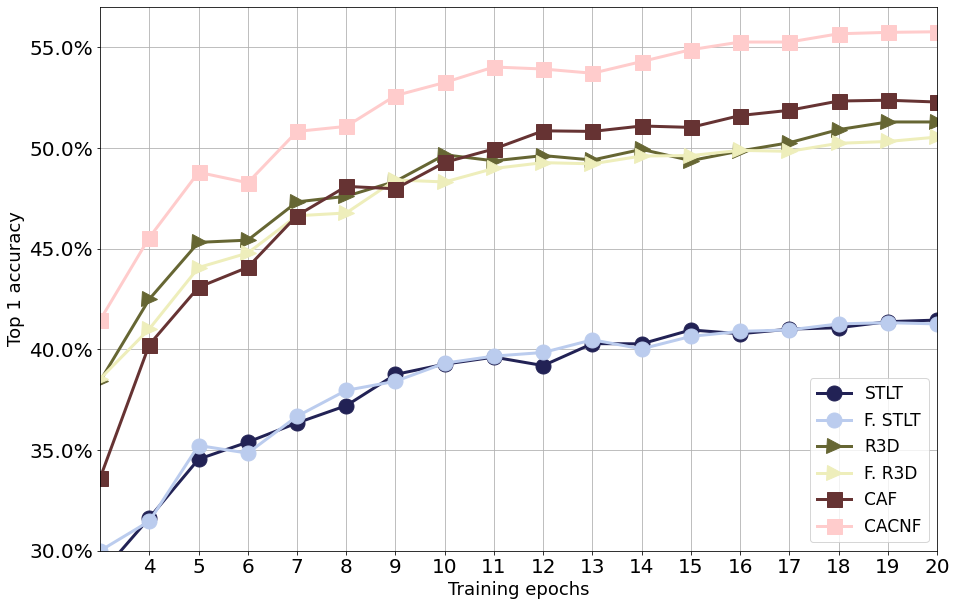}
\caption[Action recognition performance benefit of CACNF]{Top-1 validation accuracy (epoch 3 to 20) of \blackgls{stlt} and \blackgls{r3d} trained individually, trained in conjunction with \blackgls{cacnf}---F. STLT and F. R3D respectively---\blackgls{caf} and \blackgls{cacnf}.}
\label{bmvc2021:fig:caf-cacnf}
\end{figure}

\textbf{What makes CACNF superior to CAF?} To gain better insights into the underlying reason behind the superiority of CACNF w.r.t. CAF, in Figure~\ref{bmvc2021:fig:caf-cacnf}, we report the top 1 accuracy on the Something-Else compositional split, in the obj. predictions setting of:
\begin{inparaenum}[(i)]
\item \blackgls{stlt} trained individually;
\item \blackgls{stlt} trained within the CACNF model (F. STLT);
\item \blackgls{r3d} trained individually;
\item \blackgls{r3d} trained within the CACNF model (F. R3D);
\item \blackgls{caf};
\item \blackgls{cacnf}.
\end{inparaenum}

We see that the performance of \blackgls{stlt} and F. STLT, and, \blackgls{r3d} and F. R3D, is remarkably similar, indicating that with CACNF the layout and appearance branches remain complementary to one another and also preserve their individual abilities (i.e., yield the same performance as if they are trained separately). This phenomenon results in better overall performance of the Cross-Attention Fusion module (\blackgls{cacnf}), compared to training it without CentralNet \cite{vielzeuf2018centralnet} (CAF) \inlinesubresearchquestion{Research Question 4.1}.

% How do we preserve the complementarity of the modality representations before
% their fusion, which is the most crucial property that needs to be maintained
% to achieve optimal task performance?
\textbf{How does it look visually?} We are interested in visually inspecting three error types:
\begin{inparaenum}[(i)]
\item \blackgls{r3d} predicts wrong, STLT predicts correct action;
\item STLT predicts wrong, \blackgls{r3d} predicts correct action;
\item STLT and \blackgls{r3d} predict wrong, CACNF predicts correct action.
\end{inparaenum}

\begin{figure}[t]
\centering
\includegraphics[width=0.7\textwidth]{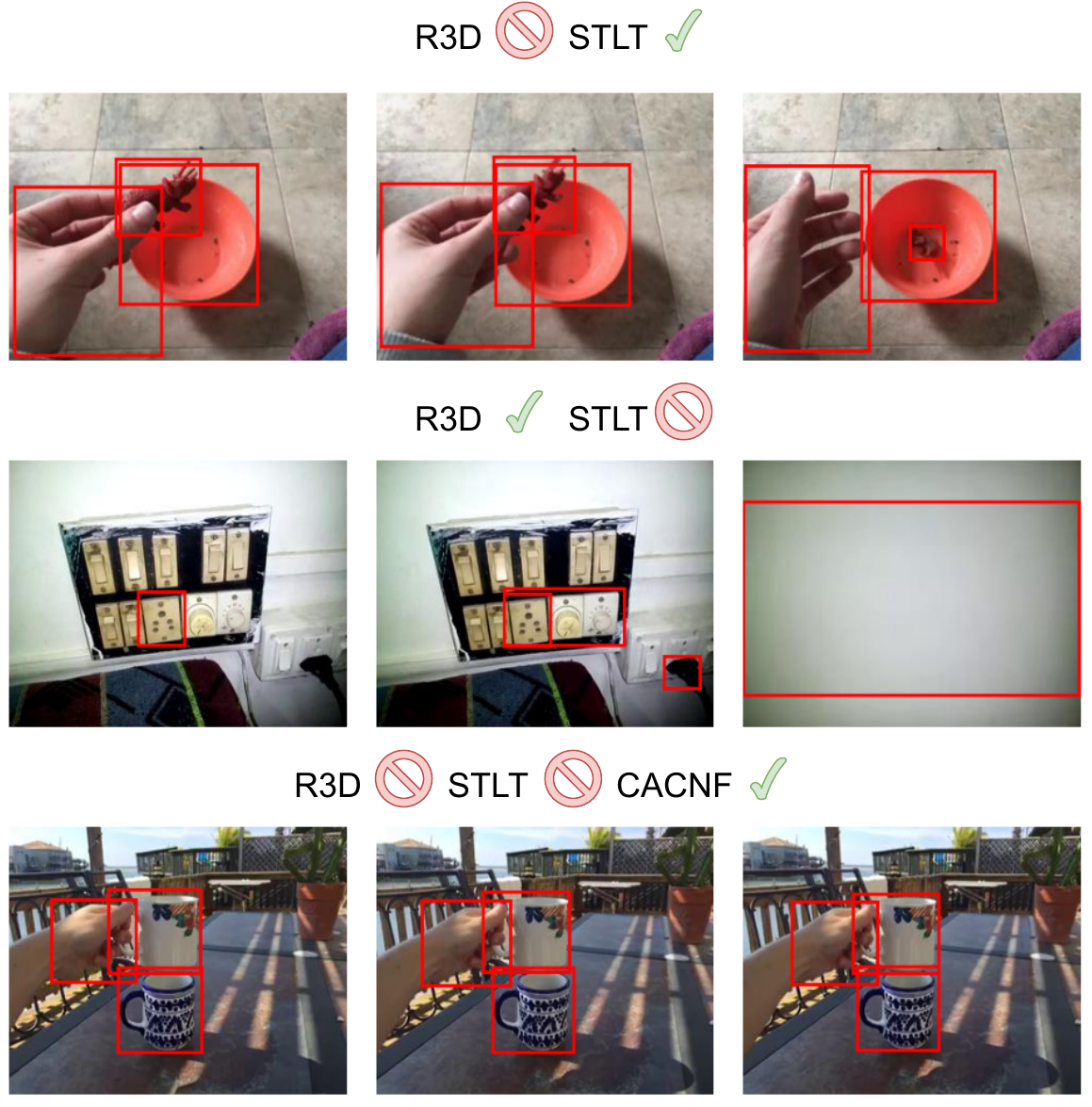}
\caption[Qualitative examples with STLT, R3D and CACNF]{\textbf{Top:} \blackgls{r3d} mispredicts, STLT predicts correctly. \textbf{Middle:} STLT mispredicts, \blackgls{r3d} predicts correctly. \textbf{Bottom:} STLT and \blackgls{r3d} mispredict, \blackgls{cacnf} predicts correctly.} 
\label{bmvc2021:fig:qualitative}
\end{figure}

In Figure~\ref{bmvc2021:fig:qualitative} (top), we observe the action ``Dropping smth. into smth.'', suitable for layout-based methods (e.g., STLT), as they directly model the spatio-temporal properties of the objects (location, movement, size, etc.). On the contrary, STLT is unable to recognize the action ``Turning the camera upwards while filming smth.'', Figure~\ref{bmvc2021:fig:qualitative} (middle), while \blackgls{r3d} recognizes the change in appearance which is indicative of the action. Lastly, the action ``Holding smth. over smth.'' in Figure~\ref{bmvc2021:fig:qualitative} (bottom), requires modeling both the temporal consistency of the layout and appearance, which STLT and \blackgls{r3d} individually fail, while CACNF recognizes the correct action. Note that these examples are selected to illustrate the particular phenomenon we are trying to illustrate.
% Furthermore, in Figure~\ref{bmvc2021:fig:qualitative-barplot}, we compare the performance of \blackgls{r3d} and STLT with CACNF, on five actions from Something-Else, where the difference between the averaged \blackgls{r3d} and STLT accuracy, and CACNF accuracy for each action is most prominent. We observe a consistent pattern across all five actions, i.e., CACNF successfully fuses the appearance (R3D) and layout branch (STLT), and it yields superior performance compared to the unimodal (layout or appearance) methods.

\subsection{Something-Else: State-of-the-art Comparisons}\label{bmvc2021:sec:state-of-the-art}
% \begin{figure}[t]
% \centering
% \includegraphics[width=0.6\textwidth]{chapters/bmvc2021/images/barplot.png}
% \caption[Action classes where the performance between CACNF, and \blackgls{r3d} and STLT is most prominent]{Five Something-Else actions where the performance difference between \blackgls{r3d} and STLT (averaged) with CACNF is most prominent.} 
% \label{bmvc2021:fig:qualitative-barplot}
% \end{figure}
We compare \blackgls{stlt} and \blackgls{cacnf} against the following methods:
\begin{inparaenum}[(i)]
\item \textbf{I3D} \cite{carreira2017quo}: An inflated Resnet50 \cite{he2016deep}-based 3D-CNN as in Wang~\etal~\cite{wang2018videos}, pre-trained on ImageNet \cite{deng2009imagenet}, subsequently fine-tuned on the Something-Else dataset;
\item \textbf{STRG} \cite{wang2018videos}: A multimodal method, with a GNN
\cite{kipf2017semi} applied over region proposals, combined with I3D using late fusion;
\item \textbf{STIN} \cite{materzynska2020something}: A GNN \cite{kipf2017semi} for spatial reasoning and a non-local neural network \cite{wang2018non} for temporal reasoning;
\item \textbf{STIN + I3D}: STIN combined with I3D in a late-fusion manner;
\item \textbf{SFI} \cite{yan2020interactive}: A multimodal fusion method, trained with an auxiliary task of predicting the future state of the video objects.
\end{inparaenum}

\textbf{Compositional action recognition}. We report results in Table~\ref{bmvc2021:table:compositional}. In both the obj. predictions and oracle setting, we observe that \blackgls{stlt} outperforms STIN and the other methods, with a more prominent difference in performance in the oracle setting. We also observe that STLT's performance is remarkably close to the best multimodal method in the oracle setting---SFI.
% an indication that, if perfect object detections are available, a multi-head attention method captures finer interactions between the objects compared to a (convolutional) GNN, 1D convolution, etc.
In the multimodal section---Table~\ref{bmvc2021:table:compositional} and Table~\ref{bmvc2021:table:few-shot} (bottom)---in the obj. predictions setting, \blackgls{cacnf} outperforms the other methods significantly, suggesting a notable improvement in robustness w.r.t. compositional data: a likely real-life scenario.

\begin{table}[t]
\begin{center}
\resizebox{0.8\textwidth}{!}{
\begin{tabular}{lcccc} \toprule
    {} & \multicolumn{2}{c}{Obj. predictions} & \multicolumn{2}{c}{Oracle} \\
    \cmidrule(lr){2-3} \cmidrule(lr){4-5}
    {Method} & {Action@1 ($\uparrow$)} & {Action@5 ($\uparrow$)} & {Action@1 ($\uparrow$)} & {Action@5 ($\uparrow$)} \\ \midrule
    \multicolumn{5}{l}{\textit{\textbf{Object detections-based methods}}} \\
    STIN \cite{materzynska2020something} & 37.2 & 62.4 & 51.4 & 79.3 \\
    SFI \cite{yan2020interactive} & \NA & \NA & 44.1 & 74.0  \\
    STLT (Ours) & \textbf{41.6} & \textbf{67.9} & \textbf{59.0} & \textbf{86.0} \\ \midrule
    \multicolumn{5}{l}{\textit{\textbf{RGB-based and multimodal fusion-based methods}}} \\
    I3D \cite{carreira2017quo} & 46.8 & 72.2 & 46.8 & 72.2 \\
   STIN \cite{materzynska2020something} + I3D \cite{carreira2017quo} & 48.2 & 72.6 & 54.6 & 79.4 \\
   STRG \cite{wang2018videos} & 52.3 & 78.3 & \NA & \NA \\
    SFI \cite{yan2020interactive} & \NA & \NA & 59.6 & 85.8 \\
    CACNF (Ours) & \textbf{56.9} & \textbf{82.5} & \textbf{67.1} & \textbf{90.4} \\ \bottomrule
\end{tabular}
}
\end{center}
\caption[Experiments for state-of-the-art comparisons on Something-Else (compositional generalization)]{Something-Else state-of-the-art action recognition accuracy comparisons in a compositional generalization setting.}
\label{bmvc2021:table:compositional}
\end{table}

\textbf{Few-shot action recognition.} We report results in Table~\ref{bmvc2021:table:few-shot}. We observe that in the obj. predictions setting, \blackgls{stlt} outperforms STIN in both the 5-shot and 10-shot setup. In the oracle setting, \blackgls{stlt} significantly outperforms the other methods as per the top-1 accuracy, allowing a significant gap for improvement as object detectors continue to improve \cite{dai2021dynamic}. Furthermore, \blackgls{cacnf} outperforms the other multimodal methods in the obj. predictions setting, with a bigger difference in top-1 accuracy in the 10-shot setup. We interpret the performance improvement as evidence that \blackgls{stlt} and \blackgls{cacnf} can successfully generalize in a low-data regime, a valuable observation considering the cost of acquiring curated video data.

\begin{table}[t]
\centering
\resizebox{1.0\textwidth}{!}{
\begin{tabular}{lcccc} \toprule
    {} & \multicolumn{2}{c}{Obj. predictions} & \multicolumn{2}{c}{Oracle} \\
    \cmidrule(lr){2-3} \cmidrule(lr){4-5}
    {Method} & {Action@1: 5-shot ($\uparrow$)} & {Action@1: 10-shot ($\uparrow$)} & {Action@1: 5-shot ($\uparrow$)} & {Action@1: 10-shot ($\uparrow$)} \\ \midrule
    \multicolumn{5}{l}{\textit{\textbf{Object detections-based methods}}} \\
    STIN \cite{materzynska2020something} & 17.7 & 20.8 & 27.7 & 33.5 \\
    SFI \cite{yan2020interactive} & \NA & \NA & 24.3 & 29.8 \\
    STLT (Ours) & \textbf{18.8} & \textbf{24.8} & \textbf{31.4} & \textbf{38.6} \\ \midrule
    \multicolumn{5}{l}{\textit{\textbf{RGB-based and multimodal fusion-based methods}}} \\
    I3D \cite{carreira2017quo} & 21.8 & 26.7 & 21.8 & 26.7 \\
   STIN \cite{materzynska2020something} + I3D \cite{carreira2017quo} & 23.7 & 27.0 & 28.1 & 33.6 \\
   STRG \cite{wang2018videos} & 24.8 & 29.9 & \NA & \NA \\
    SFI \cite{yan2020interactive} & \NA & \NA & 30.7 & 36.2 \\
    CACNF (Ours) & \textbf{27.1} & \textbf{33.9} & \textbf{37.1} & \textbf{45.5} \\ \bottomrule
\end{tabular}
}
\caption[Experiments for state-of-the-art comparisons on Something-Else (few-shot)]{Something-Else state-of-the-art action recognition accuracy comparisons in a few-shot setting.}
\label{bmvc2021:table:few-shot}
\end{table}

\subsection{Action Genome: Coping with Background-Clutter}
Multimodal fusion of appearance-based models with object-centric layout-based models appears to be unsuited for dealing with background-cluttered videos. To investigate whether this is the case, we use the Action Genome \cite{ji2020action} dataset, as it consists of videos
\begin{inparaenum}[(i)]
\item of people performing actions at home, where it is hard to isolate the objects specific to the action,
\item with objects presence overlapping between training and testing.
\end{inparaenum}
See \S\ref{background:sec:videodata} for details on the Action Genome dataset.

We measure action recognition \blackgls{map}, as well as \blackgls{map} relative improvement on top of trained \blackgls{i3d} \cite{carreira2017quo}, by ensembling it with \blackgls{stlt}. Note that different than the experiments on Something-Else, in this case
% We conduct experiments in a setup where we replace all object-specific categories, e.g., book, broom, phone, etc., except for person, with a generic ``object'' category. Therefore, we have only 2 object categories (person, object).
we utilize all 38 object categories as input to \blackgls{stlt}. We compare our methods against:
\begin{inparaenum}[(i)]
\item \textbf{LFB} \cite{wu2019long}: Long-term feature bank model, which performs well when video features are aggregated over time;
\item \textbf{SGFB} \cite{ji2020action}: Method which predicts (or uses ground truth in oracle setting) symbolic scene graph, further encoded and combined with LFB \cite{wu2019long}.
\end{inparaenum}

% \begin{table}[t]
% \begin{center}
% \resizebox{0.8\textwidth}{!}{
% \begin{tabular}{lcccc} \toprule
%     {Method} & {Input modalities} & Num. categories & {Obj. predictions mAP} & {Oracle mAP} \\ \midrule
%     \multicolumn{5}{l}{\textit{\textbf{Baseline methods}}} \\
%     LFB \cite{wu2019long} & RGB frames & \NA & 42.5 & 42.5 \\
%     SGFB \cite{ji2020action} & Scene Graphs \& RGB frames & 38 & 44.3 (1.8) & 60.3 (17.8) \\
%     I3D (Ours) \cite{carreira2017quo} & RGB frames & \NA & 33.5 & 33.5 \\ \midrule
%     STLT (Ours) & Obj. detections & 2 & 16.1 & 19.9 \\
%     STLT + I3D (Ours) & Obj. detections \& RGB frames & 2 & 33.8 (0.3) & 36.5 (3.0) \\ \midrule
%     STLT (Ours) & Obj. detections & 38 & 30.2 & 60.6 \\
%     STLT + I3D (Ours) & Obj. detections \& RGB frames & 38 & 38.5 (5.0) & 61.63 (28.1) \\  \bottomrule
% \end{tabular}
% }
% \end{center}
% \caption[Action recognition results on the Action Genome dataset]{Action Genome results. \textbf{From top to bottom:} Baselines, I3D, STLT and STLT + I3D ensemble with 2 generic obj.~categories (person, obj.), STLT and STLT + I3D ensemble with all 38 obj.~categories (person, book, phone, etc.). Relative mAP improvement over appearance method in parenthesis.}
% \label{bmvc2021:table:charades}
% \end{table}
\begin{table}[t]
\begin{center}
\resizebox{0.9\textwidth}{!}{
\begin{tabular}{lccc} \toprule
    {Method} & {Input modalities} & {Obj. predictions mAP ($\uparrow$)} & {Oracle mAP ($\uparrow$)} \\ \midrule
    \multicolumn{4}{l}{\textit{\textbf{Baseline methods}}} \\
    LFB \cite{wu2019long} & RGB frames & 42.5 & 42.5 \\
    SGFB \cite{ji2020action} & Scene Graphs \& RGB frames & 44.3\textsubscript{\textcolor{redred}{\textbf{+1.8}}} & 60.3\textsubscript{\textcolor{redred}{\textbf{+17.8}}} \\ \midrule
    % STLT (Ours) & Obj. detections & 16.1 & 19.9 \\
    % STLT + I3D (Ours) & Obj. detections \& RGB frames & 33.8 (0.3) & 36.5 (3.0) \\ \midrule
    \multicolumn{4}{l}{\textit{\textbf{Methods we propose}}} \\
    I3D \cite{carreira2017quo} & RGB frames & 33.5 & 33.5 \\
    STLT & Obj. detections & 30.2 & 60.6 \\
    STLT + I3D & Obj. detections \& RGB frames & 38.5\textsubscript{\textcolor{redred}{\textbf{+5.0}}} & 61.63\textsubscript{\textcolor{redred}{\textbf{+28.1}}} \\  \bottomrule
\end{tabular}
}
\end{center}
\caption[Background cluterred action recognition on Action]{Experiments on Action Genome. Relative \blackgls{map} improvement over appearance method in \textbf{\textcolor{redred}{red}}.}
\label{bmvc2021:table:charades}
\end{table}

In Table~\ref{bmvc2021:table:charades},
% we observe that \blackgls{stlt} yields significantly weaker performance compared to a standard appearance method, e.g., \blackgls{i3d}.
% Interestingly, despite the negatively biased setup, i.e., the actions in Action Genome are object-specific, e.g., opening a book, while the input categories are object-agnostic -- person/object, STLT still boosts I3D's performance by 3.0 \blackgls{map} points in the oracle setting.
we observe that \blackgls{stlt} performs well even in the obj. predictions setting, considering the Faster R-CNN's $\sim$ 11.5 average precision on the validation set. In the oracle setting the performance increases drastically, being on par with SGFB (which relies on ground truth scene graph), indicating a high upper bound, considering that object detection is merely a subset of scene graph generation. In the obj. predictions setting, we observe a solid relative improvement over \blackgls{i3d} when we employ multimodal fusion by ensembling \blackgls{stlt} with \blackgls{i3d}; even larger compared to SGFB and LFB, concluding that despite the complex setup (i.e., background cluttered videos) multimodal fusion of the layout-based and appearance-based methods yields improved performance.
\section{Conclusion}\label{bmvc2021:sec:conclusion}
% \researchquestion{Research Question 4}{To what extent does multimodal fusion of two abstract modalities -- representations of video frames and object detections -- enhances task-specific performance in video-based action recognition?}
 %Consequently, we ask another subsequent research question: \textit{Research Question 4.1:} How do we preserve the complementarity of the modalities representations prior to their fusion, which is the most crucial property that needs to be maintained to achieve optimal task performance?
In this chapter, we focused on multimodal fusion, as an approach for enhancing the object-centric understanding abilities of action recognition models. In particular, we developed models that leverage an abstract modality as input---object detections---and then, using multimodal fusion, we combined their representations with the representations obtained as output from standard video appearance-based models (which receive RGB frames as input).

We performed an extensive experimental study which helped us gain insight into \inlineresearchquestion{Research Question 4}. Across datasets, we observed that multimodal fusion (early, late, etc.) between the representation spaces of the object detections and the video appearance generally improves the action recognition performance. We also observed that attention-based fusion methods, despite being more expressive (i.e., they enable fine-grained interaction between the representations of the object detections and the video appearance), perform worse than simple late-fusion-based methods.

Consequently, we explored \inlinesubresearchquestion{Research Question 4.1} and developed a method which preserves the complementarity between the representation spaces before their fusion, yielding a method superior to other multimodal fusion schemes across various types of data.

% In this paper we shed light on the problem of compositional and few-show action recognition. We advocated the use of multi-head attention over spatio-temporal layouts and attempted to conclude how layout- and appearance-based models should be fused. Our main empirical findings suggest that
% \begin{inparaenum}[(i)]
% \item a layout-based model is robust w.r.t. compositional data, and generalizes from a few samples,
% \item when fusing a layout- and appearance-based model, it is crucial for the individual models to preserve their capabilities,
% \item Even on non-compositional, background-cluttered video datasets, a layout-based model can reasonably recognize human actions, and boost the performance of appearance-based models.
% \end{inparaenum}\par
\textbf{Limitations.} While multimodal fusion between the representations of the object detections and video appearance yields improved action recognition performance, it makes two key assumptions: (i) all modalities involved in the fusion during training the models have to be available during inference, and (ii) the compute budget at inference time is sufficient to perform multimodal fusion. However, for certain applications (e.g., egocentric video understanding), this might be problematic since the available compute budget is usually limited.

\textbf{What has happened since?} Many studies published since the start of 2021 (when the majority of this work took place)---either simultaneously or shortly thereafter---have reported findings similar to those we present in this chapter. Some works have leveraged a multimodal fusion-based approach with better image encoders (e.g., Video Transformers) \cite{herzig2022object}. Interestingly, other works \cite{zhang2022object} have shown that such object-centric representations, obtained with multimodal fusion between the video and object-detections representations, perform better in transfer learning settings (i.e., report higher linear probing performance) than the RGB video embeddings. Despite these advancements, these works still suffer from the two aforementioned limitations. To mitigate these challenges, the subsequent, \textbf{Chapter~\ref{ch:iccv2023}}, delves into multimodal knowledge transference as a method to enable multimodal knowledge transfer during training; thereby leveraging the strengths of multimodal fusion while relaxing its constraints.

\cleardoublepage%

  \graphicspath{{\chaptersdir/iccv2023/image/}}%
  %\cleardoublepage%
  \chapter{Multimodal Knowledge Transference for Improving Egocentric Action Recognition}\label{ch:iccv2023}
In this chapter, we focus on the challenge of multimodal knowledge transference, which we tackle through the lens of egocentric video understanding. The focal point of egocentric video understanding is modeling hand-object interactions. In \textbf{Chapter \ref{ch:bmvc2021}},
% and prior works \cite{carreira2017quo, kazakos2021little},
we observe that the performance of state-of-the-art models (e.g., 3D Resnets, Vision Transformers, etc.)---which receive RGB frames as input---can be improved further by employing multimodal fusion with \textit{additional} input modalities (e.g., object detections, optical flow, audio, etc.). These modalities provide cues complementary to the RGB modality; however, the added complexity stemming from the multimodal fusion makes these models impractical for deployment.

The goal of this chapter is to retain the high performance of a multimodal fusion approach while using \textit{only} the RGB frames as input at inference time. To achieve this, we perform multimodal knowledge transference by applying multimodal knowledge distillation during model training. We demonstrate that for egocentric action recognition using the Epic-Kitchens dataset, and the Something-Something datasets\footnote{See \S\ref{background:sec:egovision-properties} for a comparison between these datasets, in particular, focusing on the unique properties of egocentric vision.}, students which are taught by multimodal teachers tend to be more accurate and better calibrated than architecturally equivalent models trained on ground truth labels in a unimodal or multimodal fashion. We further develop a multimodal knowledge distillation framework, allowing us to deal with issues that occur when a noisy modality is added to the distillation process. Lastly, we demonstrate the reduction in computational complexity we achieve and show that our approach maintains higher performance with the reduction of the number of input views.

The content of this chapter is based on the following publications:

\begin{mdframed}[backgroundcolor=gray!30,linewidth=0pt]
\underline{Radevski*, G.}, Grujicic*, D., Blaschko, M., Moens, M-F., Tuytelaars, T., (2022). Students taught by multimodal teachers are superior action recognizers. The 2nd International Ego4D Workshop @ European Conference on Computer Vision (ECCV). \cite{radevski2022students} (*authors contributed equally\footnote{Gorjan had the initial observations that multimodal knowledge distillation works in this setup, and Dusan joined the project soon after. From that moment on, all tasks were equally divided between the authors without any specific arrangement.})
\end{mdframed}

\begin{mdframed}[backgroundcolor=gray!30,linewidth=0pt]
\underline{Radevski*, G.}, Grujicic*, D., Blaschko, M., Moens, M-F., Tuytelaars, T., (2023). Multimodal Distillation for Egocentric Action Recognition. The International Conference on Computer Vision (ICCV). \cite{radevski2023multimodal} (*authors contributed equally)\footnote{Both Gorjan and Dusan worked jointly on all tasks: preparing and processing datasets, designing experiments, paper writing, etc.}

\end{mdframed}

\section{Introduction}\label{iccv2023:sec:introduction}
The purpose of egocentric vision, as a field, is to enable machines to interpret real-world data taken from a human's perspective. The applications are numerous, ranging from recognizing \cite{zhou2015temporal} or anticipating \cite{furnari2019would} actions, to more complex tasks such as recognizing egocentric object-state changes, localizing action instances of a particular video moment \cite{grauman2022ego4d}, etc. One of the main focal points of egocentric vision is the understanding and interpretation of hand-object interactions. Usually, these hand-object interactions take place in cluttered environments, where the object of interest is often occluded or occurs only during a short time period. Furthermore, egocentric vision often suffers from motion blur---due to the movement of the scene objects or the camera itself---and thus, understanding video content from RGB frames alone may be challenging.

\begin{figure}[t]
\centering
\includegraphics[width=\columnwidth]{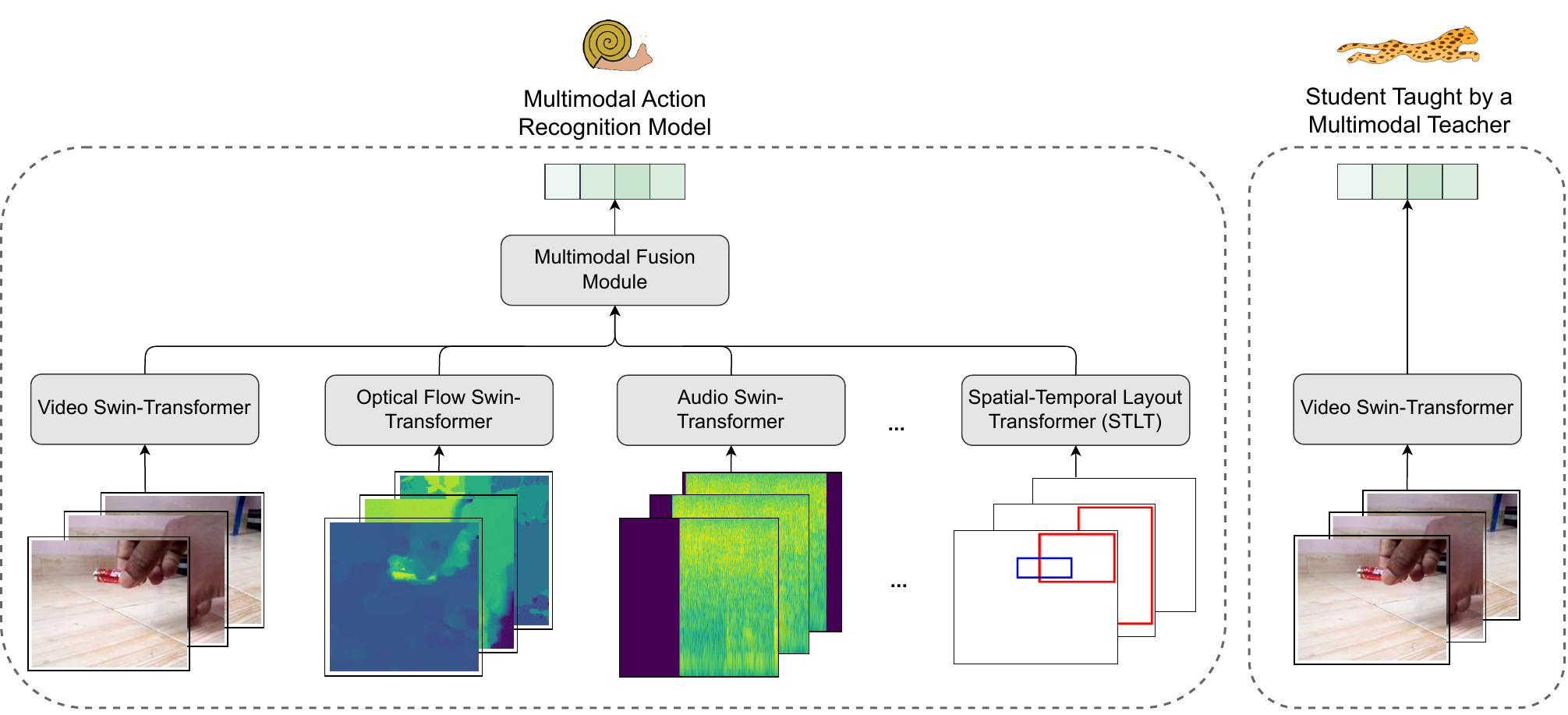}
\caption[Motivation for using multimodal knowledge distillation]{\textbf{Motivation:} The multimodal action recognition model is powerful, but too computationally inefficient to be used in practice. The distilled student is significantly faster yet achieves competitive performance.}
\label{iccv2023:fig:teaser}
\end{figure}

To cope with these challenges, various egocentric action recognition methods \cite{radevski2021revisiting, herzig2022object, zhang2022object, materzynska2020something, wang2018videos, yan2020interactive, kim2021motion} demonstrate that explicitly modeling hand-object interactions (usually represented via bounding boxes and object categories), by employing multimodal fusion, significantly improves the action recognition performance. Such performance gains are most notable in compositional generalization setup \cite{radevski2021revisiting}. Similarly, other works show that multimodal fusion with other diverse modalities (optical flow, audio, etc.) improves the performance even further \cite{xiong2022m, gabeur2020multi, kazakos2021little, nagrani2021attention}.

The main assumption these methods make is that multimodal fusion remains feasible at inference time. Namely, they assume: (i) that all modalities used during training are also available at inference time, and (ii) the compute budget at inference time would be sufficient to obtain and process these additional modalities. Such assumptions make these methods cumbersome or even impossible to use in practice: e.g., on a limited compute budget such as in the case of embedded devices. Specifically, using dedicated models for each additional modality (e.g., object detector, object tracker, and a transformer when using bounding boxes \& object categories as input \cite{radevski2021revisiting}), increases both the memory footprint as well as the inference time. 
Ideally, video understanding models would leverage additional modalities during training, while the resulting model would use only RGB frames at inference time: i.e., when deployed in practice. Such an approach---conducting multimodal knowledge transfer during training---is referred to as multimodal knowledge transference \cite{liang2022foundations}.

Therefore, we seek to gain insights into the following research question:

\researchquestion{Research Question 5}{Can we come up with a method which enables successful multimodal knowledge transference during training, to emulate the capabilities of a multimodal fusion-based model during inference?}

One way to perform multimodal knowledge transference is training Omnivorous models, i.e., models trained jointly on multiple modalities, which have been shown to generalize better than unimodal counterparts \cite{girdhar2022omnivore}. In this work, we take a different route and perform multimodal knowledge transfer by employing multimodal knowledge distillation.
% transfer multimodal knowledge to models subsequently used in an unimodal setting.
Namely, we distill the knowledge from a multimodal ensemble---exhibiting superior performance, but unviable for deployment---to a standard RGB-based action recognition model \cite{bertasius2021space} (see Figure~\ref{iccv2023:fig:teaser}).

In practical settings, the data quality across different modalities often varies due to sensor quality, environmental conditions, and operational limitations. Thus, we investigate methods to refine the multimodal knowledge transference process, aiming to understand the impact of lower-quality modalities on model learning and performance. This leads us to the following secondary research question:
% Importantly, in real-world applications, the data quality from different modalities varies significantly due to factors like sensor quality, environmental conditions, or operational constraints. Therefore, we explore strategies for optimizing the knowledge transference process, and helping to determine how lower-quality modalities can contribute to the models' learning and performance. Consequently, we ask:
% Practical Applicability: By addressing the challenges posed by noisier modalities, this research could significantly broaden the applicability of multimodal learning models, making them more versatile and useful across a wider range of conditions and applications.

\subresearchquestion{Research Question 5.1}{What is the impact of incorporating a new modality that is less reliable, or noisier, compared to other modalities in the multimodal transference process?}

Finally, we measure the trade-offs between computational efficiency and task effectiveness for multimodal transference vs. multimodal fusion. The ultimate goal could be, for example, to provide insights for practitioners on choosing multimodal fusion or multimodal transference according to their specific performance and computational needs. This leads us to pose the final secondary research question:

\subresearchquestion{Research Question 5.2}{How does multimodal knowledge transference compare to multimodal fusion in terms of downstream task performance and computational complexity?}

\textbf{Contributions.} In this research we employ state-of-the-art knowledge distillation practices \cite{beyer2022knowledge} and contribute the following:

\circled{1} We show that a student taught by a multimodal teacher, is both more accurate and better calibrated than the same model trained from scratch or in an omnivorous fashion.% (\S\ref{iccv2023:sec:general-results});

\circled{2} We provide motivation and establish a simple but reliable multimodal distillation approach, which overcomes the issue of potentially suboptimal modality-specific teachers.% (\S\ref{iccv2023:sec:weighting});

\circled{3} We demonstrate that the distilled student performs on par with significantly larger models, and maintains performance in computationally cheaper inference setups.%(\S\ref{iccv2023:sec:speed}).
\section{Related Work}\label{iccv2023:sec:related_work}
\textbf{Knowledge distillation.} Originally introduced by Hinton~\etal~\cite{hinton2015distilling}, knowledge distillation is used to transfer the knowledge from one model (i.e., a teacher), to another model (i.e., a student), by training the student to match the teacher's (intermediate) outputs on a certain dataset. Shown to be useful in a variety of contexts, the primary goal is model compression: transferring the knowledge from a larger, cumbersome teacher model, or a teacher exhibiting a different inductive bias \cite{touvron2021training, chen2022dearkd}, to a typically lightweight student model \cite{wang2021knowledge, cho2019efficacy, mishra2017apprentice}. Another line of work \cite{shen2020meal, asif2019ensemble} proposes to distill from ensembles of large teacher models to lightweight student models, obtaining promising results. Compared to these works, we focus on knowledge distillation from a multimodal teacher ensemble, i.e., a set of models where each is trained on a distinct modality.

\textbf{Multimodal knowledge distillation} has been previously used mainly in a cross-modal fashion, where the teacher and the student receive different modalities as input. In some works, the teacher receives RGB images while the student receives depth or optical flow images \cite{gupta2016cross}, while in others, the teacher receives RGB images as input, while the student receives audio as input \cite{aytar2016soundnet}. In contrast, other works explore a multimodal knowledge expansion scenario, where a multimodal student learns from pseudo-labels of an unimodal teacher \cite{xue2021multimodal}. We, on the other hand, focus on scenarios where obtaining additional modalities (optical flow, object detections, audio, etc.) during inference is prohibitive due to a limited compute budget, and therefore multimodal data is only used during training time. Multimodal knowledge distillation for action recognition has been previously explored in the works of Gracia~\etal~\cite{garcia2018modality, garcia2019dmcl}. They propose to train a model on aligned data from two modalities, where a model which receives data from modality A is trained to imitate the intermediate features of a model which receives training data from modality B. This approach is shown to yield performance improvements compared to an RGB baseline when using RGB and depth data. Moreover, the Mars \cite{crasto2019mars} and D3d \cite{stroud2020d3d} methods, similarly to us, leverage RGB frames and optical flow during training to improve test-time performance of a model that performs inference using RGB frames alone. This is achieved by matching the corresponding features or probabilistic outputs of the modality-specific models during training. Compared to these works, we consider a more general knowledge distillation setting \cite{hinton2015distilling, beyer2022knowledge}, with a multimodal teacher ensemble, used to provide a better approximation of the true posterior. Lastly, we consider a more diverse and broader set of modalities (RGB frames, optical flow, audio, and object detections) compared to the aforementioned works.

\textbf{Multimodal (egocentric) video understanding.} In the context of (egocentric) video understanding, several works have shown that using additional modalities at inference time significantly improves performance \cite{radevski2021revisiting, zhang2022object, herzig2022object, materzynska2020something, kim2021motion, nagrani2021attention, wang2018videos, tan2023egodistill}. The hypothesis is intuitive: certain actions are more easily understood from specific modalities; e.g., to recognize that a person is ``pushing something from left to right'' the bounding boxes alone are sufficient \cite{radevski2021revisiting, materzynska2020something}. Nevertheless, the assumption these works make is that all modalities used during training are available during inference, and that the compute budget allows for processing additional modalities other than the RGB frames. To that end, multiple works \cite{radevski2021revisiting, materzynska2020something, herzig2022object, wang2018videos, kim2021motion, zhang2022object} effectively use a Faster R-CNN \cite{ren2015faster}, multiple object tracker (MOT), and object detection-specific models at inference time. In contrast, we posit that for egocentric video understanding, computing additional modalities on the fly may be prohibitive. Therefore, we propose a distillation approach which uses multimodal data \textit{only} during training, while the resulting model is dependent on RGB frames alone during inference.

\textbf{Models robust to missing modalities during inference.} A parallel route to our goal---retaining the performance of a multimodal fusion approach, while using only the RGB frames at inference time---is to explicitly train models to be robust to missing modalities during inference \cite{zhao2021missing, parthasarathy2020training, neverova2015moddrop}; or more recently, to process different modalities altogether interchangeably---Omnivorous models \cite{girdhar2022omnivore, dai2022one, girdhar2022omnimae}. These models have been shown to generalize better than models trained on unimodal data. In this work, we train an Omnivorous model using the same architecture as our student, and show that the student distilled from a multimodal teacher generalizes better than its Omnivorous variant.
\section{Methodology} \label{iccv2023:sec:methods}

\subsection{Egocentric Action Recognition}
We assume we are given an input $\mathbf{x} \in \mathbb{R}^{T \times D_1 \times D_2 ... \times D_L}$, which describes an egocentric action sequence, where $T$ is the number of time-steps, while $D_1 \times D_2 ... \times D_L$ represent other dimensions of the input data, e.g., the height, width and the number of channels of a video frame. The goal of the model $f$ is to produce a discrete probability distribution over a predefined set of $C$ classes, i.e., $\mathbf{\hat{y}} = \sigma \left( f \left( \mathbf{x} \right) \right) \in \mathbb{R}_+^C$, where $\sigma$ is the softmax operator. The classes represent the actions, or alternatively, the nouns and verbs constituting the actions (e.g., the active video object and the activity). 

Given a dataset $\mathbb{D} = \{ (\mathbf{x}_1, \mathbf{y}_1), \dots, (\mathbf{x}_{N}, \mathbf{y}_{N}) \}$ of $N$ egocentric action sequences $\mathbf{x}_i$ paired with labels $\mathbf{y}_i \in \mathbb{R}_+^C$, the model is trained by minimizing the standard cross-entropy objective $\mathcal{L}_{\text{CE}} = - \frac{1}{N}\sum_{i=1}^{N} \mathbf{y}_i \cdot \log \sigma \left( f \left( \mathbf{x}_i \right) \right)$, where $\cdot$ represents the scalar product. In the case of actions characterized by separate nouns and verbs, and accompanied by their respective labels $\mathbf{y}^n_i$ and $\mathbf{y}^v_i$, separate prediction heads produce $f^n(\mathbf{x}_i)$ and $f^v(\mathbf{x}_i)$. Finally, the model is trained by minimizing the sum of the loss terms corresponding to the nouns and verbs, $\mathcal{L}^n_{\text{CE}}$ and $\mathcal{L}^v_{\text{CE}}$ respectively.

\subsection{Multimodal Knowledge Distillation}\label{iccv2023:sec:mmkd}
In egocentric vision, there often exists more than one input modality that characterizes the same actions. The action recognition task may thus be performed via the use of multiple modalities, leveraged only during training \cite{girdhar2022omnivore, radevski2022students}, or both during training and inference \cite{xiong2022m, gabeur2020multi, kazakos2021little, nagrani2021attention} by ensembling \cite{xiong2022m} models, multimodal-fusion \cite{radevski2021revisiting, kazakos2021little}, etc. However, in the case of the latter, processing multimodal data may be computationally prohibitive at inference time (e.g., due to a limited compute budget).

The fundamental concept our method builds on is knowledge distillation \cite{hinton2015distilling}, featuring a teacher (usually a larger model, exhibiting strong performance, but cumbersome to use in practice) and a student (typically a smaller model, trained to mimic the teacher \cite{beyer2022knowledge}).
Focusing on the most accessible data modality---RGB video frames (e.g., obtained using a single monocular video camera)---we opt for distilling the knowledge of a multimodal ensemble to a single model that relies on RGB inputs alone. We modify the standard knowledge distillation approach, by altering the teacher such that (i) it is not a single model, but rather an ensemble of models, and (ii) the constituting models receive different modalities as input.

\textbf{Teacher ensemble.} Given $M$ datasets $\mathbb{D}^m = \{ (\mathbf{x}^m_1, \mathbf{y}_1), \dots, (\mathbf{x}^m_{N}, \mathbf{y}_{N}) \}$ of different modalities, we train a separate model $f^m$ by minimizing the learning objective $\mathcal{L}^m_{\text{CE}} = - \frac{1}{N} \sum_{i=1}^{N} \mathbf{y}_i \cdot \log \sigma \left( f^m \left( \mathbf{x}^m_i \right) \right)$ for each modality. Finally, the output of the ensemble can be obtained by averaging the outputs of the individual teachers:
\begin{equation}
\begin{gathered}
\mathbf{\hat{y}}^t_i = \sigma \left( \frac{1}{M}\sum_{m=1}^{M} f^m (\mathbf{x}_i^m) \right).
\end{gathered}
\label{iccv2023:eq:ensemble-output}
\end{equation}

Intuitively, under-performing modality-specific models could negatively affect the performance of the ensemble. We thus consider assigning different weights to the output logits of each model in the ensemble, before aggregating their predictions. Ideally, we would want to perform a Bayesian prediction: $p( \mathbf{y} | \mathbf{x}, \mathbb{D}) = \int_{f \in F} p( \mathbf{y} | \mathbf{x} , f) p( f | \mathbb{D}) df$. For a given finite ensemble of $M$ diverse predictors, we replace the integral with a sum over the individual models: $p( \mathbf{y} | \mathbf{x}, \mathbb{D}) \approx \sum_{m=1}^{M} p( \mathbf{y} | \mathbf{x} , f^m) p( f^m | \mathbb{D})$.
We further approximate $p( f^m | \mathbb{D})$ via its proportionality to the data likelihood $p( \mathbb{D} | f^m)$ under the Bayes rule: $p( f^m | \mathbb{D}) \propto p( \mathbb{D} | f^m)$; which itself can be expressed in terms of the cross-entropy that the model $f^m$ exhibits on the dataset $\mathbb{D}$. 

The cross-entropy $e^m$ of each modality-specific model in the ensemble can be estimated on a holdout set: 

\begin{equation}
\begin{gathered}
e^m = -\frac{1}{Z}\sum_{i=1}^{Z} \mathbf{y}_i \cdot \log \sigma \left( f^m \left( \mathbf{x}^m_i \right) \right),
\end{gathered}
\end{equation}

where $Z$ is the number of held-out samples used to estimate the weights. Then, we can obtain the weights for the modality-specific models via softmax normalization of the negative cross-entropy terms: 

\begin{equation}
\begin{gathered}
w^m \propto \text{exp}(-e^m/\gamma),
\end{gathered}
\label{iccv2023:eq:weighting}
\end{equation}

where $\gamma$ is a temperature term which controls the entropy of the model weights, e.g., if $\gamma \rightarrow \infty$, equal weights would be given to each teacher---resulting in an arithmetic mean. 

We finally compute the weighted average of the predictions of $M$ modality-specific models as the teacher output:

\begin{equation}
\begin{gathered}
\mathbf{\hat{y}}^t_i = \sigma \left( \sum_{m=1}^{M} w^m f^m (\mathbf{x}_i^m) \right).
\end{gathered}
\end{equation}

\begin{figure}[t]
\centering
\includegraphics[width=1.0\columnwidth]{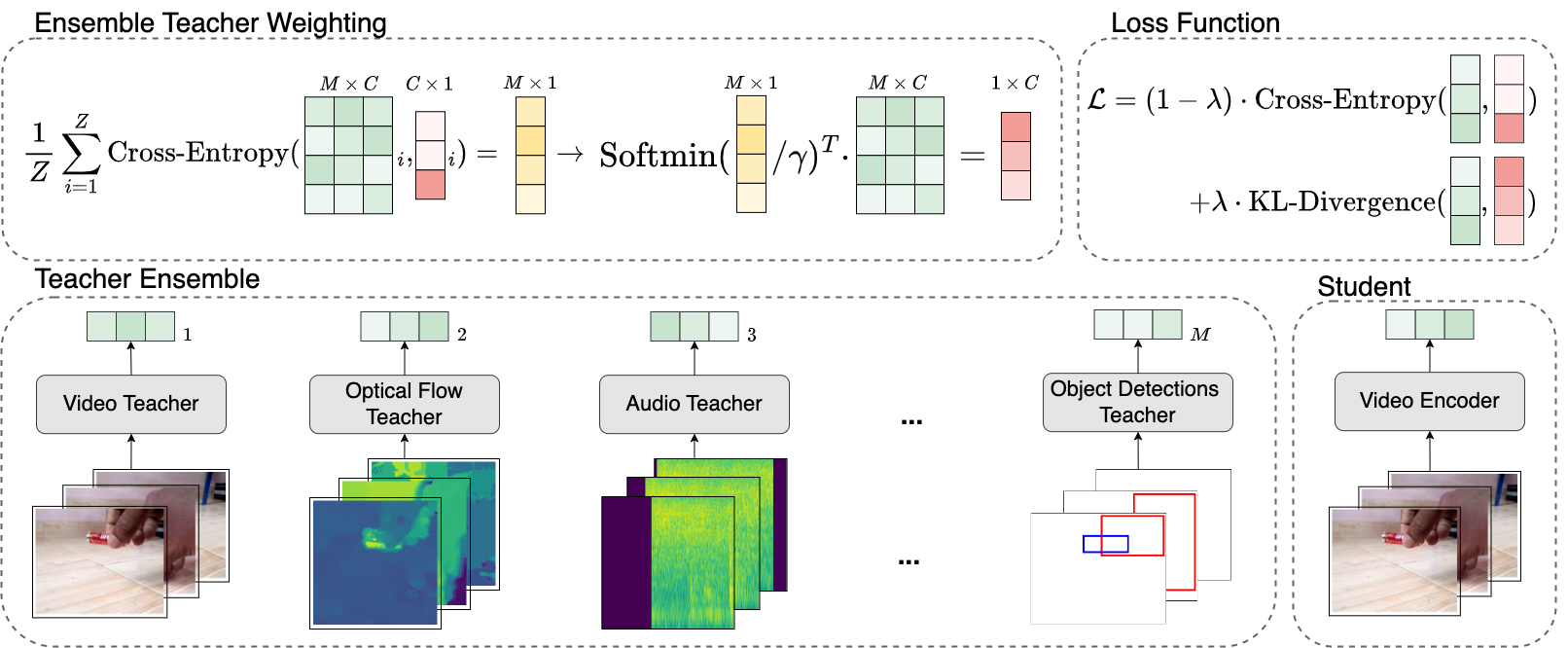}
\caption[Overview of the main components of our approach. ]{Overview of the main components of our approach.}
\label{iccv2023:fig:mainfigure}
\end{figure}

Figure \ref{iccv2023:fig:mainfigure} presents a high-level overview of our approach. In summary, our student is taught by a multimodal teacher which is itself an ensemble of multiple modality-specific models, trained separately on each modality.

\textbf{Training objective.} During training, we perform multimodal knowledge distillation, as originally proposed by Hinton~\etal~\cite{hinton2015distilling}. Specifically, we minimize the KL-divergence $\mathcal{L}_{\text{KL}}$ between the class probabilities predicted by the teacher $\mathbf{\hat{y}}^t_i = [\hat{y}^t_{i,1}, \dots, \hat{y}^t_{i,C}] \in \mathbb{R}_+^C$ and the class probabilities of the student $\mathbf{\hat{y}}^s_i = [\hat{y}^s_{i,1}, \dots, \hat{y}^s_{i, C}] \in \mathbb{R}_+^C$ as $\mathcal{L}_{\text{KL}} = \frac{1}{N}\sum_{i=1}^{N} (- \mathbf{\hat{y}}^t_i \cdot \log \mathbf{\hat{y}}^s_i + \mathbf{\hat{y}}^t_i\cdot \log \mathbf{\hat{y}}^t_i)$.

Additionally, we use a temperature parameter $\tau$ to control the entropy of the predicted probability scores while preserving their ranking, i.e., $\hat{y}^s_j \propto \text{exp}(\hat{y}^s_j/\tau)$. As per standard practice \cite{hinton2015distilling}, we use the temperature parameter $\tau$ to also rescale the KL-divergence loss, i.e., $\mathcal{L}_{\text{KL}} = \mathcal{L}_{\text{KL}} \cdot \tau^2$. We further use the standard cross-entropy action recognition loss $\mathcal{L}_{\text{CE}}$, and compute the final loss:

\begin{equation}
\begin{gathered}
 \mathcal{L} = \lambda \cdot \mathcal{L}_{\text{KL}} + (1 - \lambda) \cdot \mathcal{L}_{\text{CE}},
\end{gathered}
\label{iccv2023:eq:final_loss}
\end{equation}

where $\lambda$ balances the distillation loss $\mathcal{L}_{\text{KL}}$ and the action recognition loss $\mathcal{L}_{\text{CE}}$. For example, with $\lambda = 0.0$ we would effectively train the modality-specific RGB model, while with $\lambda = 1.0$, we would perform solely multimodal knowledge distillation.

\textbf{Optimality of a multimodal ensemble teacher.} The work of Menon~\etal~\cite{menon2021statistical} further demonstrates that in the context of distillation, a teacher that predicts the Bayes class-probability distribution over the labels $\mathbf{p}^*(\mathbf{x}) = [\mathbb{P}(y|\mathbf{x})]_{y \in [C]}$ exhibits the lowest possible variance of the student objective for any convex loss function.

The student trained to minimize the KL divergence between its output and the Bayes class probabilities would thus generalize the best. In a preliminary experiment shown in Table~\ref{iccv2023:table:calibration-error-teacher}, we demonstrate that the multimodal ensemble (both for $\gamma=1$ and $\gamma=30.0$) achieves significantly higher accuracy and lower calibration error (see \S\ref{iccv2023:sec:calibration} for details), and thus represents a better approximation of the Bayes class probabilities than a single modality-specific model. This may lead to a lower-variance objective for the student and improved generalization during knowledge distillation \cite{menon2021statistical}.

\begin{table}[t]
\centering
\resizebox{1.0\textwidth}{!}{
\begin{tabular}{lcccccc} \toprule
    {Method} & {Noun Acc. ($\uparrow$)} & {Noun ECE ($\downarrow$)} & {Verb Acc. ($\uparrow$)} & {Verb ECE ($\downarrow$)} & {Action Acc. ($\uparrow$)} & {Action ECE ($\downarrow$)} \\ \midrule
    {} & \multicolumn{4}{c}{Epic-Kitchens Regular} & \multicolumn{2}{c}{Something-Something} \\
    \cmidrule(lr){2-5} \cmidrule(lr){6-7}
    RGB Baseline & 52.0 & 19.9 & 61.7 & 17.1 & 59.3 & 21.2 \\
    Teacher ($\gamma = 30.0$) & 52.3 & 2.1 & 66.8 & 3.4 & 66.6 & 8.1 \\
    Teacher ($\gamma = 1.0$) & 53.9 & 8.9 & 66.9 & 4.2 & 66.7 & 9.0 \\ \midrule
    {} & \multicolumn{4}{c}{Epic-Kitchens Unseen Environments} & \multicolumn{2}{c}{Something-Else} \\
    \cmidrule(lr){2-5} \cmidrule(lr){6-7}
    RGB Baseline & 38.3 & 26.9 & 51.7 & 24.0 & 51.8 & 21.4 \\
    Teacher ($\gamma = 30.0$) & 42.9 & 6.0 & 58.0 & 9.9 & 63.3 & 6.5 \\
    Teacher ($\gamma = 1.0$) & 43.8 & 9.5 & 57.7 & 9.7 & 63.5 & 8.1 \\ \bottomrule
\end{tabular}
}
\caption[Comparison between the RGB baseline and the Teacher ensemble]{Accuracy \& Expected Calibration Error (ECE) of the RGB baseline and Teacher ensemble on Epic-Kitchens and Something-Something (regular \& unseen splits). We use all available modalities for the respective dataset in the teacher ensemble during multimodal knowledge distillation.}
\label{iccv2023:table:calibration-error-teacher}
\end{table}
\section{Experiments and Discussion}\label{iccv2023:sec:experiments}
\textbf{Datasets.} We use the Something-Something \cite{goyal2017something}, Something-Else \cite{materzynska2020something}, and the Epic-Kitchens (100) \cite{damen2020rescaling} datasets. See \S\ref{background:sec:videodata} for details on the datasets, and \S\ref{background:sec:egovision-properties} for an extended discussion on the differences between Epic-Kitchens and Something-Something, focused on the unique properties of egocentric vision.

\textbf{Modalities.} In addition to the RGB frames (the modality of interest at inference time), in our experiments we consider the following modalities:
\begin{inparaenum}[(i)]
    \item \textit{Optical flow} (\blackgls{of}): First used by I3D \cite{carreira2017quo} for action recognition, it is also used by recent state-of-the-art (egocentric) video understanding methods \cite{xiong2022m}. We use optical flow obtained using TV-L\textsubscript{1}\cite{zach2007duality}.
    \item \textit{Object detections} (\blackgls{obj}): Shown to significantly improve the performance of standard (RGB) egocentric video understanding models across datasets \cite{radevski2021revisiting, herzig2022object, materzynska2020something, zhang2022object, wang2018videos}. 
    % The object detector is usually trained on object-agnostic annotations \cite{shan2020understanding}, i.e., it can recognize two ``hands'' and ``objects''. 
    As per \cite{shan2020understanding}, we use the object detector trained on object-agnostic annotations, i.e., with ``hands'' and ``objects'' object labels.
    \item \textit{Audio} (\blackgls{audio}): For certain datasets \cite{damen2018scaling, damen2020rescaling}, several works \cite{kazakos2021little, xiong2022m} have shown that using audio improves action recognition performance. The audio is obtained directly from the video.
\end{inparaenum}

Moreover, if available, one may include additional modalities: e.g., depth estimates, heat maps, etc. For Something-Something and Something-Else there is no audio available, while we use the object detections provided by \cite{herzig2022object, materzynska2020something}, and optical flow from \cite{wang2016temporal}. For Epic-Kitchens we use the modalities available and released with the dataset itself \cite{damen2020rescaling}: optical flow and audio.

\textbf{Models.} We use a Swin-Tiny (\blackgls{swin_t}) Transformer \cite{liu2021swin} to encode RGB frames, optical flow, and audio. Each optical flow frame is represented as a $224 \times224 \times 2$ tensor, where the two values at each spatial location represent the $x$ and $y$ velocity components. In the case of audio, we extract 1.116 second-audio segments (0.558s before its corresponding time-step and 0.558s after it) for each frame. We compute the mel-spectrogram of the audio segment,
% (see details in Appendix~\ref{iccv2023:appendix:modalities}),
which is subsequently resized to the desired width and height. We thus treat each modality as a sequence of $224 \times 224$ multi-channel images, which we provide as input to the vision transformer, as per the common practice in recent vision models \cite{girdhar2022omnimae, girdhar2022omnivore}. To encode the object detections, i.e., bounding boxes and object categories of the scene objects, we use a state-of-the-art model---\blackgls{stlt} \cite{radevski2021revisiting}. In \blackgls{stlt}, a spatial and temporal transformer separately encodes the spatial and temporal arrangement of the objects occurring in the video. The multimodal teacher we use during knowledge distillation is an ensemble of individual, modality-specific models. Unless noted otherwise, the student is a Swin-T model that receives RGB frames as input, both during training (distillation) and inference.
In addition to Swin-T, in \S\ref{iccv2023:sec:speed}, we also consider \blackgls{r3d} \cite{kataoka2020would} (18 and 50 layers deep) light-weight student models which receive video frames of size $112 \times 112$ as input.

\textbf{Metrics.} Besides accuracy, we measure Expected Calibration Error (\blackgls{ece}) \cite{guo2017calibration}. As per \cite{guo2017calibration, rousseau2021post}, we sort the predictions based on the per-class confidence scores and group them into $K$ bins $B_k$, each associated with a confidence interval $I_{B_k} = (\frac{k-1}{K} ,\frac{k}{K})$, where $K=15$. \blackgls{ece} represents the discrepancy between the average accuracy $acc(B_k)$ and the average confidence $conf(B_k)$ in each bin $B_k$:
\begin{equation}
\begin{gathered}
    \text{ECE} = \sum_{k=1}^{K} \frac{|B_k|}{N} |acc(B_k) - conf(B_k)|,
\end{gathered}
\end{equation}
where $N$ is the number of samples in the dataset.

\subsection{Multimodal Distillation for Egocentric Vision}\label{iccv2023:sec:general-results}
We first verify the overall effectiveness of multimodal knowledge distillation on the task of egocentric action recognition for both object-agnostic actions (Something-Something) and actions represented as noun-verb compositions (Epic-Kitchens). Across all experiments, we fix $\lambda$ to 1.0: i.e., we train solely with multimodal knowledge distillation. Similarly, we set $\gamma$ to a large value ($\gamma = 30.0$), where effectively each teacher equally contributes to the ensemble output. Note that in this setting, models trained on modalities such as optical flow and audio may underperform, and thus adversely affect the ensemble teacher's performance of recognizing active objects (i.e., nouns).

\subsubsection{Recognizing Egocentric Actions}
\begin{table}[t]
\centering
\resizebox{\textwidth}{!}{
\begin{tabular}{lccccc} \toprule
    {Method} & {Training Modalities} & {Inference Modalities} & {Noun@1 ($\uparrow$)} & {Verb@1 ($\uparrow$)} & {Action@1 ($\uparrow$)} \\ \midrule
    Most Frequent & \NA & \NA & 3.93 & 20.04 & 0.0 \\
    Baseline & RGB & RGB & 52.0 & 61.7 & 38.3 \\ \midrule
    Modality-specific & OF & OF & 34.1 & 59.0 & 25.9 \\
    Teacher & \NA & RGB \& OF & 51.9 & 65.3 & 39.5 \\
    Student & RGB \& OF & RGB & 52.2\textsubscript{\textcolor{redred}{\textbf{+0.2}}} & 65.6\textsubscript{\textcolor{redred}{\textbf{+3.9}}} & 39.9\textsubscript{\textcolor{redred}{\textbf{+1.6}}} \\ \midrule
    Modality-specific & A & A & 22.3 & 46.5 & 15.1 \\
    Teacher & \NA & RGB \& A & 52.7 & 64.4 & 39.8 \\
    Student & RGB \& A & RGB & 51.5\textsubscript{\textcolor{redred}{\textbf{-0.5}}} & 62.4\textsubscript{\textcolor{redred}{\textbf{+0.7}}} & 37.9\textsubscript{\textcolor{redred}{\textbf{-0.4}}} \\ \midrule
    Teacher & \NA & RGB \& OF \& A & 52.3 & 66.8 & 40.5 \\
    Student & RGB \& OF \& A & RGB & 51.7\textsubscript{\textcolor{redred}{\textbf{-0.3}}} & 65.4\textsubscript{\textcolor{redred}{\textbf{+3.7}}} & 39.3\textsubscript{\textcolor{redred}{\textbf{+1.0}}} \\ \bottomrule
\end{tabular}
}
\caption[Action recognition experiments on Epic-Kitchens]{Experiments on Epic-Kitchens: Accuracy of recognizing the active video object (noun) and the activity (verb). Improvement over RGB frames baseline \cite{liu2021swin} in \textbf{\textcolor{redred}{red}}.}
\label{iccv2023:table:epic-regular}
\end{table}

\begin{table}[t]
\centering
\resizebox{0.9\textwidth}{!}{
\begin{tabular}{lcccc} \toprule
    {Method} & {Training Modalities} & {Inference Modalities} & {Action@1 ($\uparrow$)} & {Action@5 ($\uparrow$)} \\ \midrule
    Baseline & RGB & RGB & 60.3 & 86.4 \\ \midrule
    Modality-specific & OF & OF & 49.3 & 79.0 \\
    Teacher & \NA & RGB \& OF & 64.3 & 88.9 \\
    Student & RGB \& OF & RGB & 62.8\textsubscript{\textcolor{redred}{\textbf{+2.5}}} & 88.9\textsubscript{\textcolor{redred}{\textbf{+2.5}}} \\ \midrule
    Modality-specific & OBJ & OBJ & 47.9 & 76.2 \\
    Teacher & \NA & RGB \& OBJ & 65.3 & 89.5 \\
    Student & RGB \& OBJ & RGB & 63.2\textsubscript{\textcolor{redred}{\textbf{+2.9}}} & 88.7\textsubscript{\textcolor{redred}{\textbf{+2.3}}} \\ \midrule
    Teacher & \NA & RGB \& OF \& OBJ & 66.6 & 90.5 \\
    Student & RGB \& OF \& OBJ & RGB & 63.0\textsubscript{\textcolor{redred}{\textbf{+2.7}}} & 88.9\textsubscript{\textcolor{redred}{\textbf{+2.5}}} \\
    \bottomrule
\end{tabular}
}
\caption[Multimodal knowledge transference experiments on Something-Something]{Experiments on Something-Something: Accuracy of recognizing object-agnostic actions. Improvement over \blackgls{rgb} frames baseline \cite{liu2021swin} in \textbf{\textcolor{redred}{red}}.}
\label{iccv2023:table:something-regular}
\end{table}

In Table~\ref{iccv2023:table:epic-regular} and Table~\ref{iccv2023:table:something-regular}, we report performance on Epic-Kitchens 100 \cite{damen2020rescaling} and Something-Something \cite{goyal2017something}. In line with the previous findings reported in \textbf{Chapter~\ref{ch:bmvc2021}} and in the literature \cite{herzig2022object, materzynska2020something}, we find that employing additional modalities at inference time significantly improves the performance compared to the RGB baseline model, for both the Epic-Kitchens and the Something-Something datasets.

A novel observation by our work is that multimodal knowledge distillation performs well in the context of egocentric video understanding, with student models often approaching the performance of the multimodal teacher ensemble. For Epic-Kitchens (Table~\ref{iccv2023:table:epic-regular}), we observe that when the student is distilled from an RGB \& \blackgls{of}, or \blackgls{rgb} \& \blackgls{of} \& \blackgls{audio} teacher, it is superior to the baseline model, as well as all modality-specific models, for recognizing actions. On the other hand, distilling from an RGB \& A teacher yields performance lower than the baseline, due to the low performance of the model trained only on audio data. Specifically, we observe that the audio-specific teacher lowers the active object recognition performance. In \S\ref{iccv2023:sec:weighting}, we propose a solution for this issue.
For Something-Something (Table~\ref{iccv2023:table:something-regular}), the student model is superior to the baseline for each combination of modalities. When distilling from all available modalities (RGB \& OF \& OBJ), the resulting model outperforms the baseline by \textcolor{redred}{\textbf{3.7\%}} in terms of the top-1 accuracy. In terms of the top-5 accuracy on Something-Something, the students achieve performance that is nearly on par with the multimodal teacher ensemble.

\subsubsection{Generalizing to Unseen Environments and Objects}
\begin{table}[t]
\centering
\resizebox{\textwidth}{!}{
\begin{tabular}{lccccc} \toprule
    {Method} & {Training Modalities} & {Inference Modalities} & {Noun@1 ($\uparrow$)} & {Verb@1 ($\uparrow$)} & {Action@1 ($\uparrow$)} \\ \midrule
    Most frequent & \NA & \NA & 3.7 & 15.9 & 0.0 \\
    Baseline & \blackgls{rgb} & \blackgls{rgb} & 38.3 & 51.7 & 25.4 \\ \midrule
    Modality-specific & OF & OF & 28.0 & 53.2 & 21.6 \\
    Teacher & \NA & RGB \& OF & 41.0 & 54.9 & 28.4 \\
    Student & RGB \& OF & RGB & 42.5\textsubscript{\textcolor{redred}{\textbf{+4.2}}} & 55.9\textsubscript{\textcolor{redred}{\textbf{+4.2}}} & 30.2\textsubscript{\textcolor{redred}{\textbf{+4.8}}} \\ \midrule
    Modality-specific & A & A & 15.0 & 41.5 & 9.1 \\
    Teacher & \NA & RGB \& A & 41.9 & 55.3 & 28.5 \\
    Student & RGB \& A & RGB & 41.8\textsubscript{\textcolor{redred}{\textbf{+3.5}}} & 51.8\textsubscript{\textcolor{redred}{\textbf{+0.1}}} & 27.5\textsubscript{\textcolor{redred}{\textbf{+2.1}}} \\ \midrule
  Teacher & \NA & RGB \& OF \& A & 42.9 & 58.0 & 30.3 \\
    Student & RGB \& OF \& A & RGB & 43.7\textsubscript{\textcolor{redred}{\textbf{+5.4}}} & 54.1\textsubscript{\textcolor{redred}{\textbf{+3.4}}} & 29.6\textsubscript{\textcolor{redred}{\textbf{+4.2}}} \\ \bottomrule
\end{tabular}
}
\caption[Multimodal knowledge transference on Epic-Kitchens (Unseen environments)]{Experiments on Epic-Kitchens (Unseen environments): Accuracy of recognizing the active video object (noun) and the activity (verb). Improvement over RGB frames baseline \cite{liu2021swin} in \textbf{\textcolor{redred}{red}}.}
\label{iccv2023:table:unseen-epic}
\end{table}

\begin{table}[t]
\centering
\resizebox{0.9\textwidth}{!}{
\begin{tabular}{lcccc} \toprule
    {Method} & {Training Modalities} & {Inference Modalities} & {Action@1 ($\uparrow$)} & {Action@5 ($\uparrow$)} \\ \midrule
    Baseline & RGB & RGB & 51.8 & 79.5 \\ \midrule
    Modality-specific & OF & OF & 49.0 & 77.4 \\
    Teacher & \NA & RGB \& OF & 61.0 & 86.4 \\
    Student & RGB \& OF & RGB & 58.2\textsubscript{\textcolor{redred}{\textbf{+6.4}}} & 85.1\textsubscript{\textcolor{redred}{\textbf{+5.6}}} \\ \midrule
    Modality-specific & OBJ & OBJ & 41.4 & 67.3 \\
    Teacher & \NA & RGB \& OBJ & 59.4 & 84.5 \\
    Student & RGB \& OBJ & RGB & 57.5\textsubscript{\textcolor{redred}{\textbf{+5.7}}} & 84.1\textsubscript{\textcolor{redred}{\textbf{+4.6}}} \\ \midrule
    Teacher & \NA & RGB \& OF \& OBJ & 63.6 & 87.7 \\
    Student & RGB \& OF \& OBJ & RGB & 59.1\textsubscript{\textcolor{redred}{\textbf{+7.3}}} & 86.1\textsubscript{\textcolor{redred}{\textbf{+6.6}}} \\
    \bottomrule
\end{tabular}
}
\caption[Multimodal knowledge transference experiments on Something-Else]{Experiments on Something-Else: Accuracy of recognizing object-agnostic actions featuring objects unseen during training. Improvement over \blackgls{rgb} frames baseline \cite{liu2021swin} in \textbf{\textcolor{redred}{red}}.}
\label{iccv2023:table:unseen-something}
\end{table}

We investigate to what extent our findings translate to the compositional generalization setting \footnote{Compositional generalization measures to what extent the model can generalize to novel combinations of concepts observed during training.}, in which the performance of standard video models deteriorates significantly \cite{materzynska2020something}. In the Something-Something dataset, the objects the people interact with might overlap between training and testing, potentially allowing models to pick up undesirable biases w.r.t. object appearance to discriminate between the actions. Therefore, Something-Else proposes a compositional generalization data split, on which the objects at training and testing time do not overlap, such that the models encounter strictly novel objects during testing. Similarly, the rich visual environments featured in the Epic-Kitchen training data may also provide the model with spurious cues for predicting the actions, and lead to significant performance drop on new, unobserved environments. The Epic-Kitchens Unseen split consists of a subset of videos in the validation set from participants whose videos were not present in the training dataset, giving insight into how our approach generalizes to new visual environments.

We report the performance of our approach on the Epic-Kitchens Unseen split and the Something-Else \cite{materzynska2020something} split in Table~\ref{iccv2023:table:unseen-epic} and Table~\ref{iccv2023:table:unseen-something} respectively. The general observation across datasets and modalities is that the distilled students significantly outperform the RGB baseline model, and are sometimes even competitive with their respective teacher. Notably, on Something-Else, the student distilled from an RGB \& \blackgls{of} \& \blackgls{obj} teacher outperforms the baseline by \textcolor{redred}{\textbf{7.3\%}} in terms of top-1 accuracy.

\subsubsection{Performance Breakdown on Seen and Unseen Participants}\label{iccv2023:sec:epic_seen_unseen}
To assess whether our approach yields improvements in terms of generalizing to new visual environments, we isolate the videos of participants in the Epic-Kitchens validation dataset that are not included in the Epic-Kitchens Unseen split. We dub this dataset split as Epic-Kitchens Seen, as it consists only of videos of participants that have been featured in the training data (and thus show visual environments already observed during training). In Table~\ref{iccv2023:table:epic-seen-unseen}, we contrast the results on the Unseen split with those on the Seen split. We notice that while the performance improves on the Seen split, the gain of our approach is even larger on the Unseen split (in line with optical flow and audio being more invariant w.r.t.\ environment appearance). Consequently, we observe a smaller performance drop going from Seen to Unseen dataset splits in the case of our approach, suggesting a lower degree of overfitting to the appearance of objects and environments.

\begin{table}[t]
\centering
\resizebox{0.85\columnwidth}{!}{
\begin{tabular}{lcccccc} \toprule
    {Dataset} & {Objective} & {$\lambda$} & {$\gamma$} & {Noun@1 ($\uparrow$)} & {Verb@1 ($\uparrow$)} & {Action@1 ($\uparrow$)} \\ \midrule
    \multirow{2}{*}{Epic Kitchens \textbf{Regular}} & $\mathcal{L}_{\text{CE}}$ & 0.0 & \NA & 52.0 & 61.7 & 38.3 \\
    & $\mathcal{L}_{\text{CE}} \wedge \mathcal{L}_{\text{KL}}$ & 0.8 & 1.0 & 53.5 & 65.4 & 41.2 \\ \midrule
    \multirow{2}{*}{Epic Kitchens \textbf{Unseen}} & $\mathcal{L}_{\text{CE}}$ & 0.0 & \NA & 38.3 & 51.7 & 25.4 \\
     & $\mathcal{L}_{\text{CE}} \wedge \mathcal{L}_{\text{KL}}$ & 0.8 & 1.0 & 42.5 & 54.7 & 30.4 \\ \midrule
    \multirow{2}{*}{Epic Kitchens \textbf{Seen}} & $\mathcal{L}_{\text{CE}}$ & 0.0 & \NA & 53.7 & 63.0 & 39.9 \\
     & $\mathcal{L}_{\text{CE}} \wedge \mathcal{L}_{\text{KL}}$ & 0.8 & 1.0 & 54.8 & 66.7 & 42.5 \\ \bottomrule
\end{tabular}
}
\caption[Experiments on Epic-Kitchens: Regular, seen, and unseen split]{Per-split action recognition accuracy breakdown on Epic-Kitchens: Regular split, environments \textit{unseen} and environments \textit{seen} during training).}
\label{iccv2023:table:epic-seen-unseen}
\end{table}

\subsubsection{Comparison with Omnivorous Models}
We explicitly compare multimodal knowledge distillation against Omnivorous models \cite{girdhar2022omnivore, girdhar2022omnimae}, which are trained jointly on all modalities featuring non-aligned data, e.g., with RGB frames obtained from one dataset and depth maps from another \cite{girdhar2022omnivore}. These models have been proven to generalize better compared to their unimodal counterparts. We train the Omnivore model using batches comprised of data from the same modality, where we randomly sample the modality for each training batch\footnote{Girdhar~\etal~\cite{girdhar2022omnivore} show that there is no performance difference if the batches contain mixed data from different modalities, however, we found that using homogeneous batches yields better performance.}. Additionally, we train the Omnivore models for $M \times$ more epochs ($M$ is the number of modalities), to account for the random sampling of modalities during training. We represent the object detections modality as bounding boxes with a line thickness of 2px on a white canvas where each category is colored uniquely, e.g., hand in blue and object in red.

\begin{table}[t]
\centering
\resizebox{0.9\textwidth}{!}{
\begin{tabular}{lccccc} \toprule
    {Method} & {Noun@1 ($\uparrow$)} & {Verb@1 ($\uparrow$)} & {Action@1 ($\uparrow$)} & {Action@1 ($\uparrow$)} & {Action@5 ($\uparrow$)} \\  \midrule
    {} & \multicolumn{3}{c}{Epic-Kitchens Regular split} & \multicolumn{2}{c}{Something-Something} \\
    \cmidrule(lr){2-4} \cmidrule(lr){5-6}
    Omnivore \cite{girdhar2022omnivore} & 47.8 & 62.8 & 35.9 & 58.4 & 86.2 \\
    Student & \textbf{51.7} & \textbf{65.4} & \textbf{39.3} & \textbf{63.0} & \textbf{88.9} \\ \midrule
    \cellcolor{white}{} & \multicolumn{3}{c}{Epic-Kitchens Unseen Participants} & \multicolumn{2}{c}{Something-Else} \\
    \cmidrule(lr){2-4} \cmidrule(lr){5-6}
    Omnivore \cite{girdhar2022omnivore} & 38.5 & \textbf{54.5} & 27.9 & 58.3 & 84.9 \\
    Student & \textbf{43.7} & 54.1 & \textbf{29.6} & \textbf{59.1} & \textbf{86.1} \\ \bottomrule
\end{tabular}
}
\caption[Experiments comparing multimodal distillation and Omnivore-type models]{Action recognition accuracy comparsion with Omnivorous models \cite{girdhar2022omnivore}. All models are \blackgls{swin_t} and perform inference using only RGB frames. Multimodal distillation using all modalities with $\lambda = 1.0$ and $\gamma = 30.0$.}
\label{iccv2023:table:omnivorous-models}
\end{table}

We report results in Table~\ref{iccv2023:table:omnivorous-models}. Even though on the Epic-Kitchens Unseen split and the Something-Else dataset the Omnivore model exhibits strong performance ($\sim$5.5\% improvement on top of the baseline on Something-Else), the multimodal distillation approach achieves even higher performance. Nevertheless, we conclude that both approaches represent viable options for training multimodal models that leverage unimodal inputs during inference.

\subsubsection{Effect on Model Calibration}\label{iccv2023:sec:calibration}
\begin{figure}[t]
\centering
\includegraphics[width=0.49\textwidth]{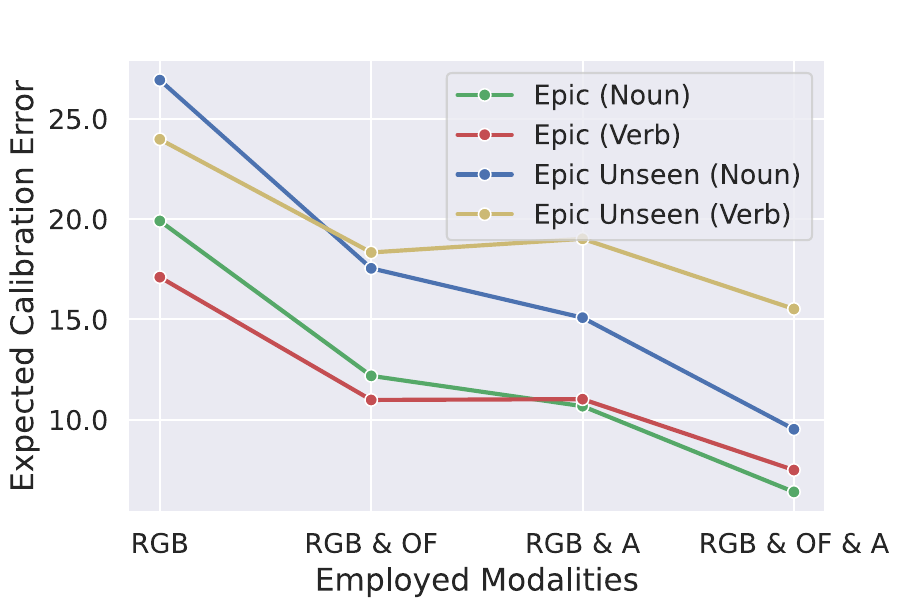}
\includegraphics[width=0.49\textwidth]{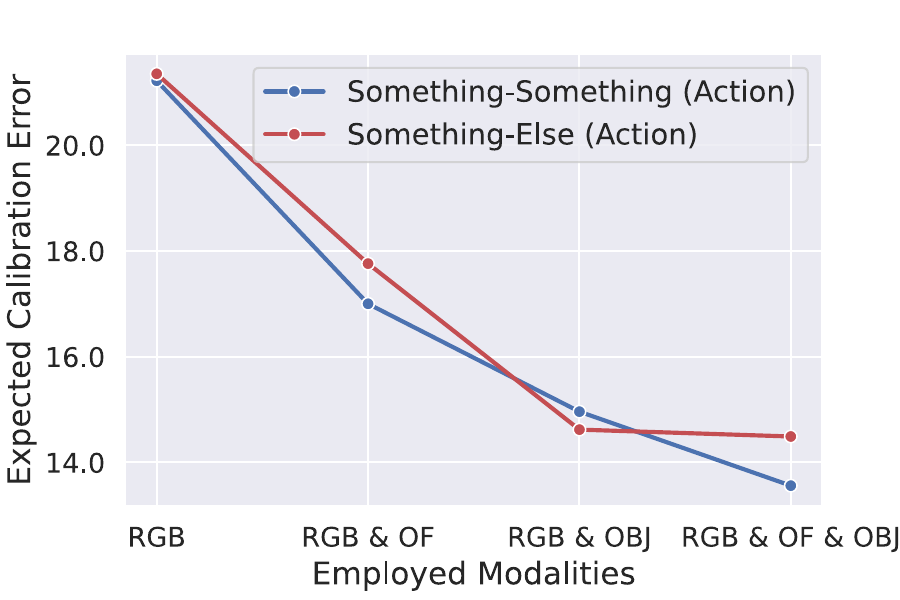}
\caption[Expected Calibration Error across datasets]{Expected Calibration Error across datasets. The student is Swin-T trained with $\lambda = 1.0$ and $\gamma = 30.0$.}
\label{iccv2023:fig:calibration-error}
\end{figure}

Next to the improvements in action recognition performance we observe, we also investigate how the distilled student fares against the baseline in terms of model calibration. That is, whether the probability score of the predicted class reflects the accuracy of predicting the said class \cite{guo2017calibration, popordanoska2022consistent}. More importantly, we verify whether distilling from additional modalities yields better-calibrated students. In Figure~\ref{iccv2023:fig:calibration-error}, we measure the Expected Calibration error \cite{guo2017calibration} (\blackgls{ece}) on the Epic-Kitchens (regular and unseen), and the Something-Something and Something-Else datasets. We report the ECE for an RGB model trained using the ground truth labels, as well as all distilled students (reported in Table~\ref{iccv2023:table:epic-regular}, \ref{iccv2023:table:something-regular}, \ref{iccv2023:table:unseen-epic}, and \ref{iccv2023:table:unseen-something}). The general observation is that \textit{across datasets, distillation improves the models' calibration}. Furthermore, we find that increasing the number of modalities used in the ensemble improves the model calibration.

\subsubsection{Per-Class Performance Breakdown}
\begin{figure}[t]
\centering
\includegraphics[valign=t, width=0.495\textwidth]{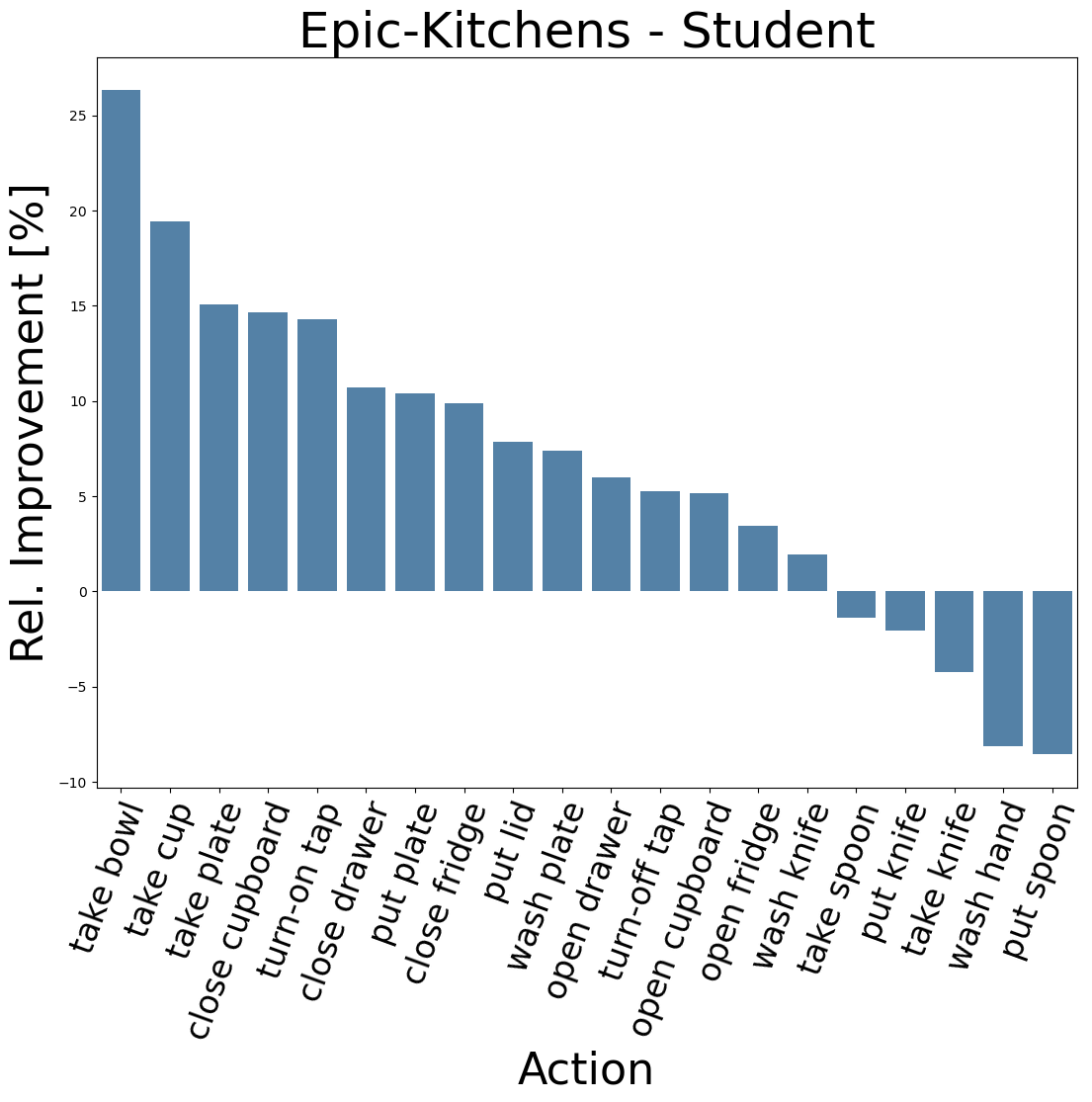}
\includegraphics[valign=t, width=0.495\textwidth]{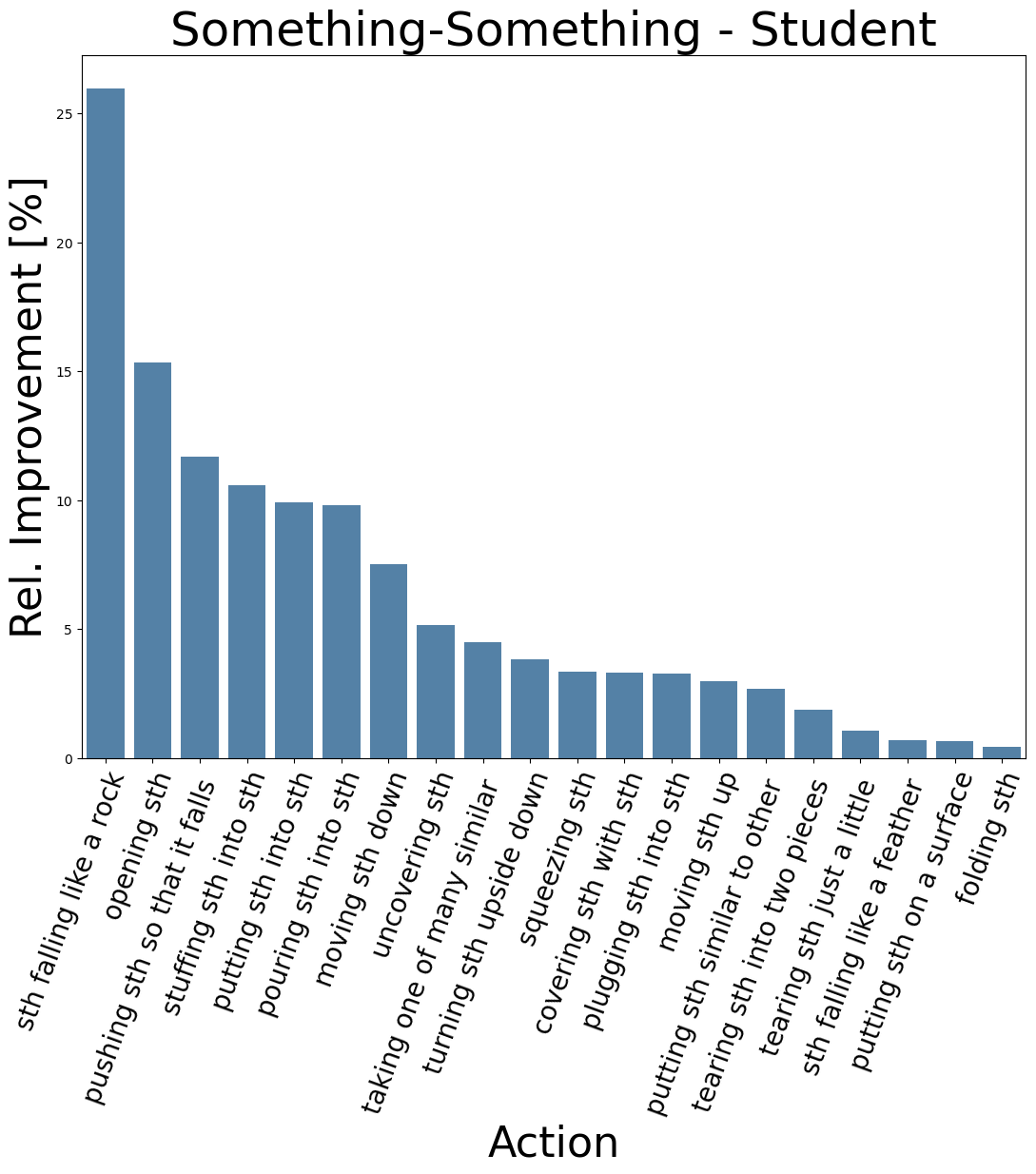}
\caption[Student per-class improvement over the baseline on the top-20 most frequent actions]{Per-class improvement of the student over the RGB baseline on the top-20 most frequent actions on Epic-Kitchens and Something-Something.}
\label{iccv2023:fig:per-class-student}
\end{figure}

In Fig~\ref{iccv2023:fig:per-class-student}, we present the relative change in action recognition accuracy of the student model obtained via multimodal distillation w.r.t.\ the architecturally equivalent baseline RGB model trained on ground truth labels, computed on the top-20 most frequent classes (actions) on Epic-Kitchens and Something-Something. Overall, we observe that multimodal distillation generally improves performance across different actions, particularly so on Something-Something, where we achieve improvements in terms of all of the top 20 most frequent action classes.

\subsection{Ensemble Teacher Weighting}\label{iccv2023:sec:weighting}
\begin{table}[t]
\centering
\resizebox{0.75\textwidth}{!}{
\begin{tabular}{lccccc} \toprule
    {Objective} & {$\lambda$} & {$\gamma$} & {Noun@1 ($\uparrow$)} & {Verb@1 ($\uparrow$)} & {Action@1 ($\uparrow$)} \\ \midrule
    $\mathcal{L}_{\text{CE}}$ & 0.0 & \NA & 52.0 & 61.7 & 38.3 \\
    $\mathcal{L}_{\text{KL}}$ & 1.0 & 30.0 & 51.7 & 65.4 & 39.3 \\
    $\mathcal{L}_{\text{CE}} \wedge \mathcal{L}_{\text{KL}}$  & 0.8 & 30.0 & 52.6 & 65.1 & 40.4 \\
    $\mathcal{L}_{\text{CE}} \wedge \mathcal{L}_{\text{KL}}$ & 0.8 & 3.0 & 53.0 & 66.9 & 41.0 \\
    $\mathcal{L}_{\text{KL}}$ & 1.0 & 1.0 & 53.1 & 65.5 & 40.5 \\
    $\mathcal{L}_{\text{CE}} \wedge \mathcal{L}_{\text{KL}}$ & 0.8 & 1.0 & 53.5 & 65.4 & 41.2 \\
    $\mathcal{L}_{\text{CE}} \wedge \mathcal{L}_{\text{KL}}$ & 0.8 & 0.33 & 53.6 & 64.7 & 40.5 \\ \bottomrule
\end{tabular}
}
\caption[Ablation study of ensemble teacher weighing module on Epic-Kitchens]{Ablation study on the Epic-Kitchens dataset (action recognition accuracy). $\lambda$: Distillation and Cross-Entropy loss balancing term; $\gamma$: Temperature of the Ensemble Teacher Weighting.}
\label{iccv2023:table:ablation-weighting}
\end{table}

In \S\ref{iccv2023:sec:general-results}, we observe that distilling from multimodal teachers yields students who are superior to models trained on the ground truth labels. Nevertheless, if a weak teacher is added to the ensemble, it negatively affects the knowledge distillation and yields a student which is inferior to using the ground truth labels: e.g., adding the audio-specific teacher in the ensemble for Epic-Kitchens. To cope with this issue, we weigh the logits of the teacher ensemble as discussed in \S\ref{iccv2023:sec:mmkd}. Namely, we use the two hyperparameters ($\lambda$ and $\gamma$) for
\begin{inparaenum}[(i)]
    \item balancing between the ground truth and the distillation loss: $\lambda$ (Equation~\ref{iccv2023:eq:final_loss}), and
    \item controlling how the predictions of modality-specific models are combined in the ensemble: $\gamma$ (Equation~\ref{iccv2023:eq:weighting}).
\end{inparaenum}
We report results for Epic-Kitchens in Table~\ref{iccv2023:table:ablation-weighting}, including the values for $\lambda$ and $\gamma$ as well as the objective we effectively minimize. We observe that the model trained only on the ground truth labels ($\lambda = 0.0$), is inferior to all other models. Using a large $\gamma$ (e.g., $30.0$)---effectively assigning equal weights to all models in the ensemble---we observe to perform well despite the simple setup (we use this model in the experiments in \S\ref{iccv2023:sec:general-results}). Additionally, training using the task loss in addition to the distillation loss ($\lambda = 0.8$) further improves the performance. The best-performing model uses both the task and the distillation loss ($\lambda = 0.8$), and assigns weights to each teacher in the ensemble based on its performance on $Z = 1000$ randomly sampled videos held out from the training dataset, with a normalization temperature $\gamma = 1.0$. Moreover, we find that lowering the normalization temperature $\gamma$ further ($\gamma = 0.33$)---giving higher weight to the best-performing model in the ensemble---yields lower performance. Similarly, increasing the normalization temperature to $\gamma = 3.0$, thus equalizing the model weights, also negatively affects performance.

\subsubsection{Ensemble Teacher Weighting: Something-Something and Something-Else}\label{iccv2023:sec:weighting-something}
In the vein of the ablation study reported in Table~\ref{iccv2023:table:ablation-weighting}, we conduct experiments on the Something-Something and the Something-Else datasets.
% Namely, the main findings from Table~\ref{iccv2023:table:ablation-weighting} suggest that (i) training with the ground-truth labels cross-entropy loss, in conjunction with the multimodal knowledge distillation loss, overcomes the issue of inferior modality-specific teachers, and (ii) weighting the teachers in the ensemble (such that their each individual cross-entropy loss on a holdout set of $Z = 1000$ samples are minimized) improves the performance further.
Different than Epic-Kitchens, in Table~\ref{iccv2023:table:ablation-weighting-something}, however, we observe that in the case of Something-Something and Something-Else, the addition of the loss term featuring ground-truth labels has a small effect on the performance of the student. Namely, on both the Something-Something and the Something-Else datasets, all modality-specific models perform well, and are complementary to each other, therefore, there is a lesser need for joint training using the ground truth labels and the distillation loss.

\begin{table}[t]
\centering
\resizebox{0.85\columnwidth}{!}{
\begin{tabular}{lcccccc} \toprule
    {} & {} & {} & \multicolumn{2}{c}{Something-Something} & \multicolumn{2}{c}{Something-Else} \\
    \cmidrule(lr){4-5} \cmidrule(lr){6-7}
    {Objective} & {$\lambda$} & {$\gamma$} & {Action@1 ($\uparrow$)} & {Action@5 ($\uparrow$)} & {Action@1 ($\uparrow$)} & {Action@5 ($\uparrow$)} \\ \midrule
    $\mathcal{L}_{\text{CE}}$ & 0.0 & \NA & 60.3 & 86.4 & 51.8 & 79.5 \\
    $\mathcal{L}_{\text{KL}}$ & 1.0 & 30.0 & 63.0 & 88.9 & 59.1 & 86.1 \\
    $\mathcal{L}_{\text{CE}} \wedge \mathcal{L}_{\text{KL}}$ & 0.8 & 30.0 & 63.1 & 88.3 & 59.3 & 86.3 \\ \bottomrule
\end{tabular}
}
\caption[Ablation study of ensemble teacher weighing module on Something-Something and Something-Else]{Ablation study on the Something-Something and Something-Else datasets (action recognition accuracy). $\lambda$: Distillation and Cross-Entropy loss balancing term; $\gamma$: Temperature of the Ensemble Teacher Weighting.}
\label{iccv2023:table:ablation-weighting-something}
\end{table}

\subsubsection{Discussion on the Ensemble Teacher Weighing Method}
The current method for determining the per-modality ensemble teacher weights involves estimating them on a holdout set of $Z$ samples. These samples come from the validation set and have not been seen by the ensemble teacher components (i.e., the unimodal models) during training. This method yields a fixed weight for each unimodal model in the ensemble teacher.

Alternatively, one could estimate the ensemble teacher weights online during the multimodal knowledge distillation process. For each training sample with a ground truth label, the discrepancy between the output of each teacher component and the ground truth can be measured. These weights can then be used to re-weigh the predictions of the individual teacher models. However, this approach has at least two potential issues: (i) the weights are estimated on the same dataset used to train the individual models, and (ii) the weights are estimated separately for each sample, which may introduce noise.

While we have not empirically confirmed that our adopted approach is superior to the online estimation method, we hypothesize that it is more robust because of overcoming these two limitations.

\subsection{Efficiency Analysis}\label{iccv2023:sec:speed}
Despite the strong performance multimodal action recognition models exhibit, we argue that the high computational complexity makes them cumbersome for deployment, particularly compared to a model which uses only \blackgls{rgb} video during inference. The teacher ensemble used for Epic-Kitchens uses three modality-specific \blackgls{swin_t} models, where each has 28.22M parameters, and requires 140.33 GFLOPs for processing single-view 32 frame/spectrogram video. Assuming such an ensemble is deployed, the optical flow would have to be computed on-the-fly\footnote{The Duality-based TV-L1 \cite{zach2007duality, perez2013tv} can be efficiently computed on a GPU (5-10 frames per second) \cite{bao2014comparison}. Deep Learning-based approaches, e.g., RAFT \cite{teed2020raft}, require 163.37 GFLOPs, however, they achieve higher frames per second of 21.10, measured with 10 refinement iterations and resolution of $256 \times 456$.}. Using RAFT, we measure a total added computation of 163.37 GFLOPs for such a model. We consider the computation required to obtain the spectrograms of 1.116s audio segments to be negligible in comparison. Therefore, when using all three modalities, updating the input sequences for each newly observed frame and performing action recognition would require 584.36 GFLOPs. In contrast, the distilled student is a single RGB model, and in the case of \blackgls{swin_t} requires 140.33 GFLOPs---\textit{which represents a reduction of 75.98\%.}

\begin{figure}[t]
\centering
% \resizebox{2.0\textwidth}{!}{
\includegraphics[width=0.8\textwidth]{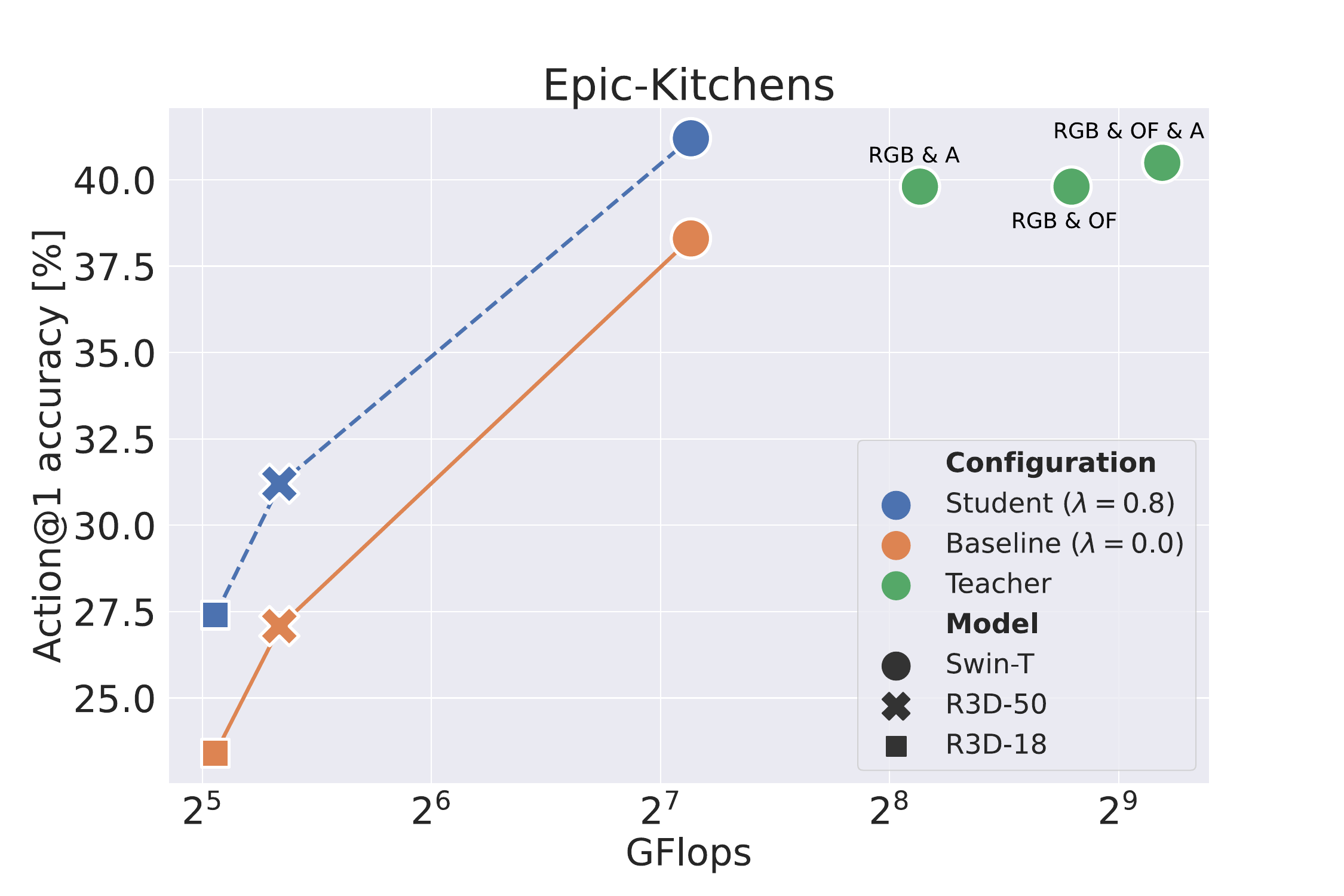}
\vspace{1cm}
\includegraphics[width=0.8\textwidth]{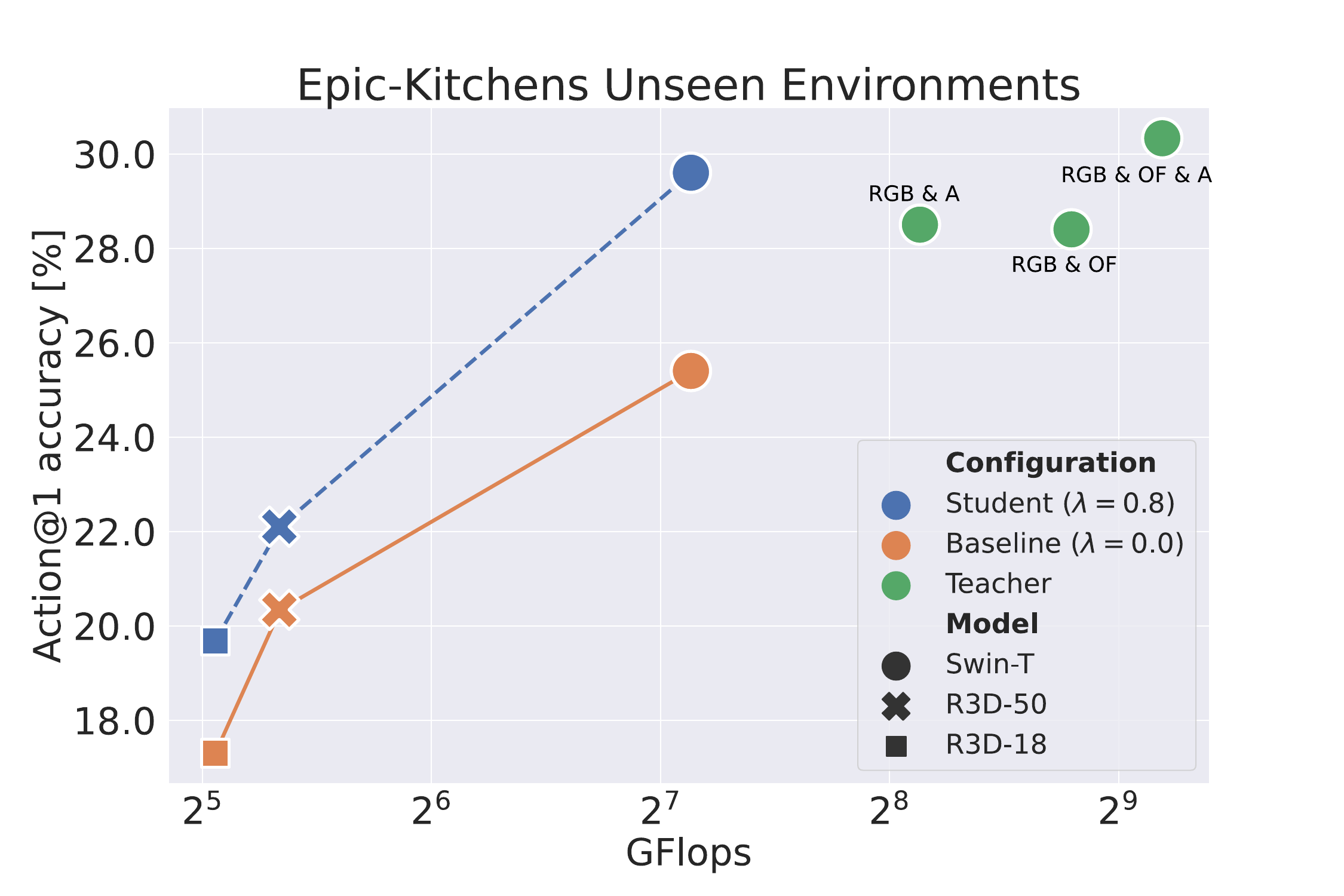}
% }
\caption[Analysis on the efficiency and action recognition ability]{Top-1 action recognition accuracy of \textcolor{teacher}{Teacher}, \textcolor{student}{Student} \& \textcolor{baseline}{Baseline} models and their associated computational cost in giga-FLOPs ($10^9$) required to update the input and predict the action. Note: Top-left corner is optimal; i.e., faster and most accurate models.}
\label{iccv2023:fig:model-families-complexity}
\end{figure}

We report results on the Epic-Kitchens dataset in Figure~\ref{iccv2023:fig:model-families-complexity} for:
\begin{inparaenum}[(i)]
    \item All variations of teacher models (RGB \& OF, RGB \& A, and RGB \& OF \& A);
    \item The \blackgls{swin_t} student model, distilled with $\lambda = 0.8$ and $\gamma = 1.0$;
    \item The \blackgls{swin_t} baseline model, trained with $\lambda = 0.0$;
    \item Two new ResNet3D (\blackgls{r3d}) models \cite{kataoka2020would} (with a depth of 50 and 18 layers), exhibiting fewer parameters and GFLOPs compared to the \blackgls{swin_t} model. The resolution size of the \blackgls{r3d} models is $112 \times 112$. For each \blackgls{r3d}, we report the action recognition performance of the baseline with $\lambda = 0.0$, and the performance of the distilled students with $\lambda = 0.8$ and $\gamma = 1.0$.
\end{inparaenum}\par
We observe a consistently higher performance of the student models compared to the same model architecture trained on ground truth labels alone. Notably, our best-performing student achieves comparable performance to the significantly more expensive RGB \& OF \& A teacher. Furthermore, the \blackgls{r3d}-18 and \blackgls{r3d}-50 students outperform their counterparts trained on ground-truth labels. Finally, we observe that the \blackgls{r3d}-18 student matches the performance of the larger, and computationally more expensive \blackgls{r3d}-50 trained on ground truth labels.

\begin{figure}[t]
\centering
\includegraphics[width=0.49\textwidth]{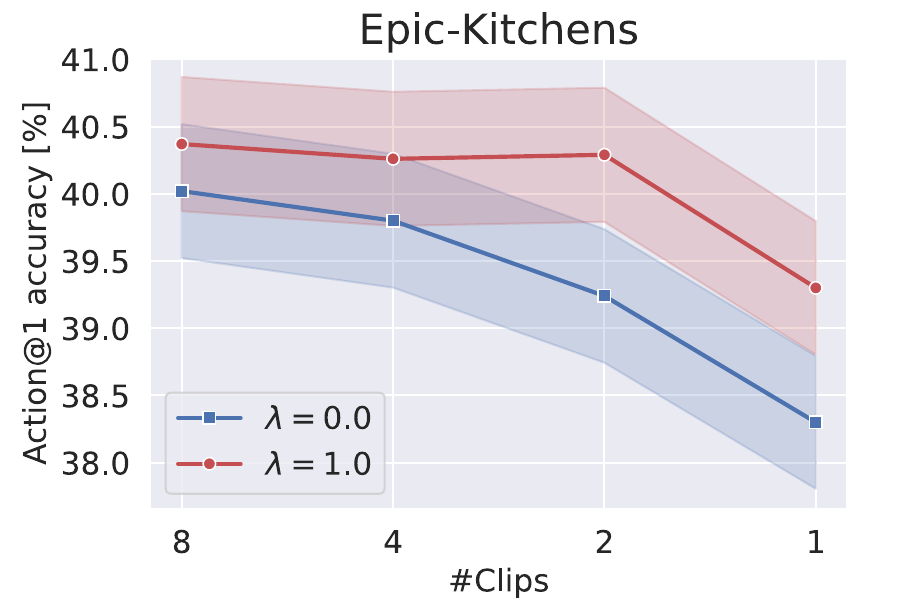}
\includegraphics[width=0.49\textwidth]{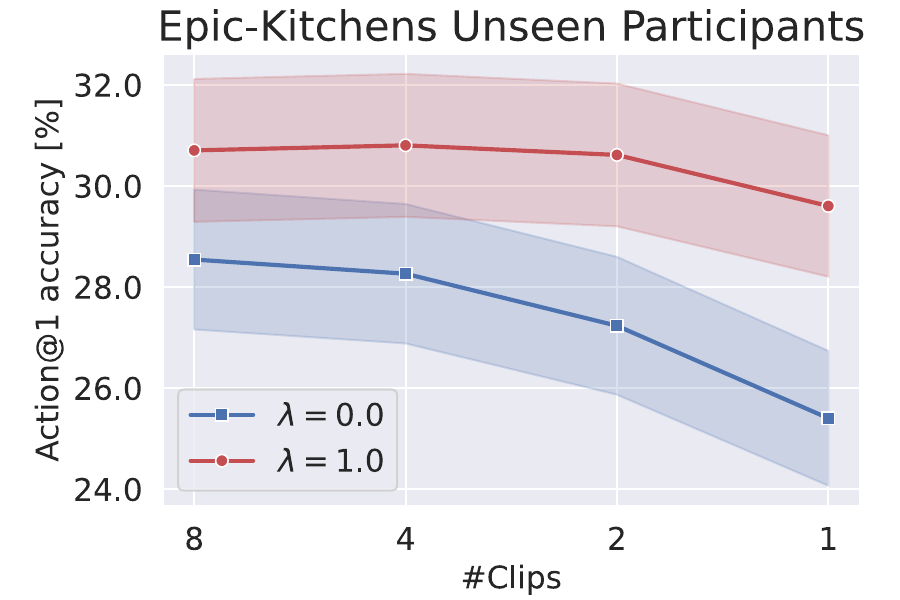}
\includegraphics[width=0.49\textwidth]{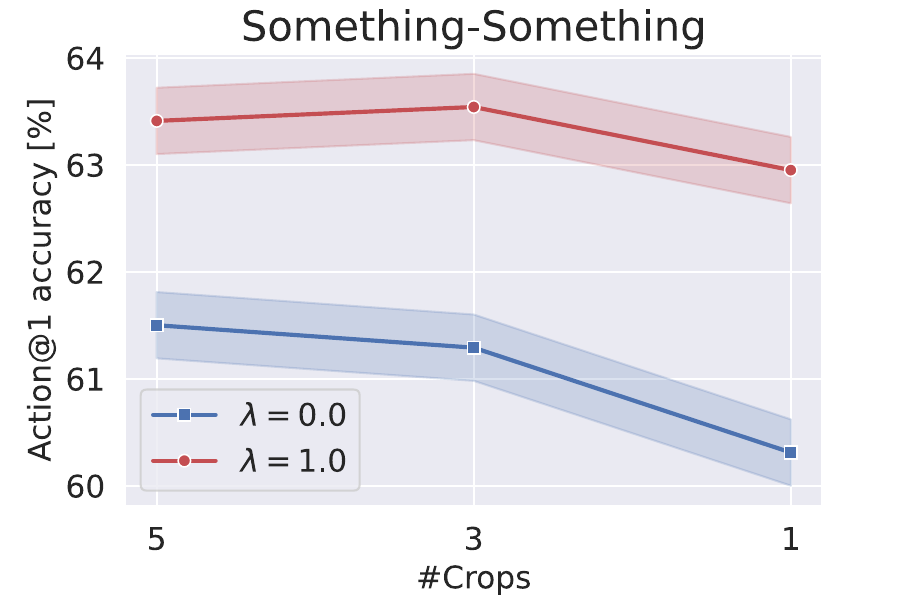}
\includegraphics[width=0.49\textwidth]{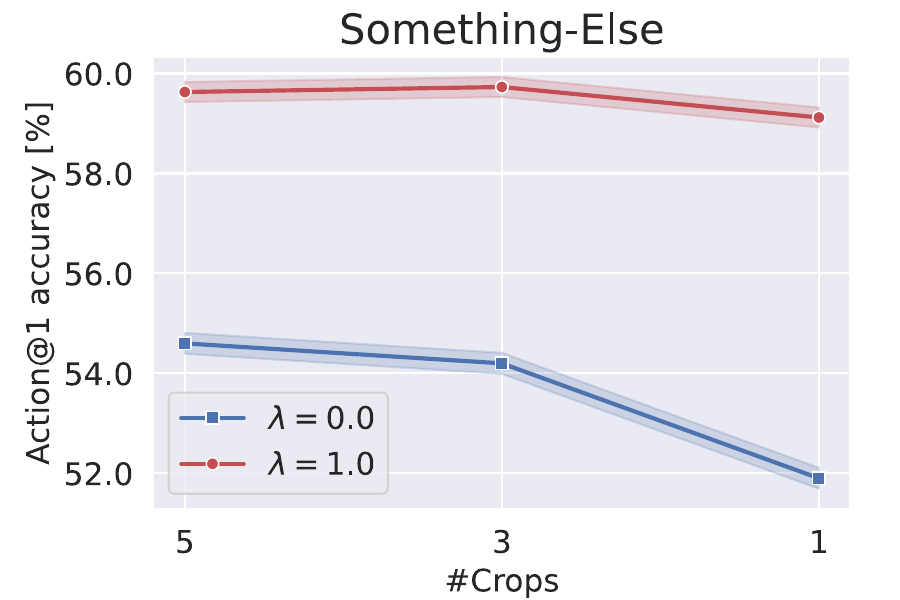}
\caption[Performance degradation when reducing the number of inference clips/crops]{Performance degradation when reducing the number of inference clips and crops on the Epic-Kitchens and Something-Something datasets.}
\label{iccv2023:fig:ablation_num_clips}
\end{figure}

\begin{figure}[t]
\centering
\includegraphics[width=0.63\textwidth]{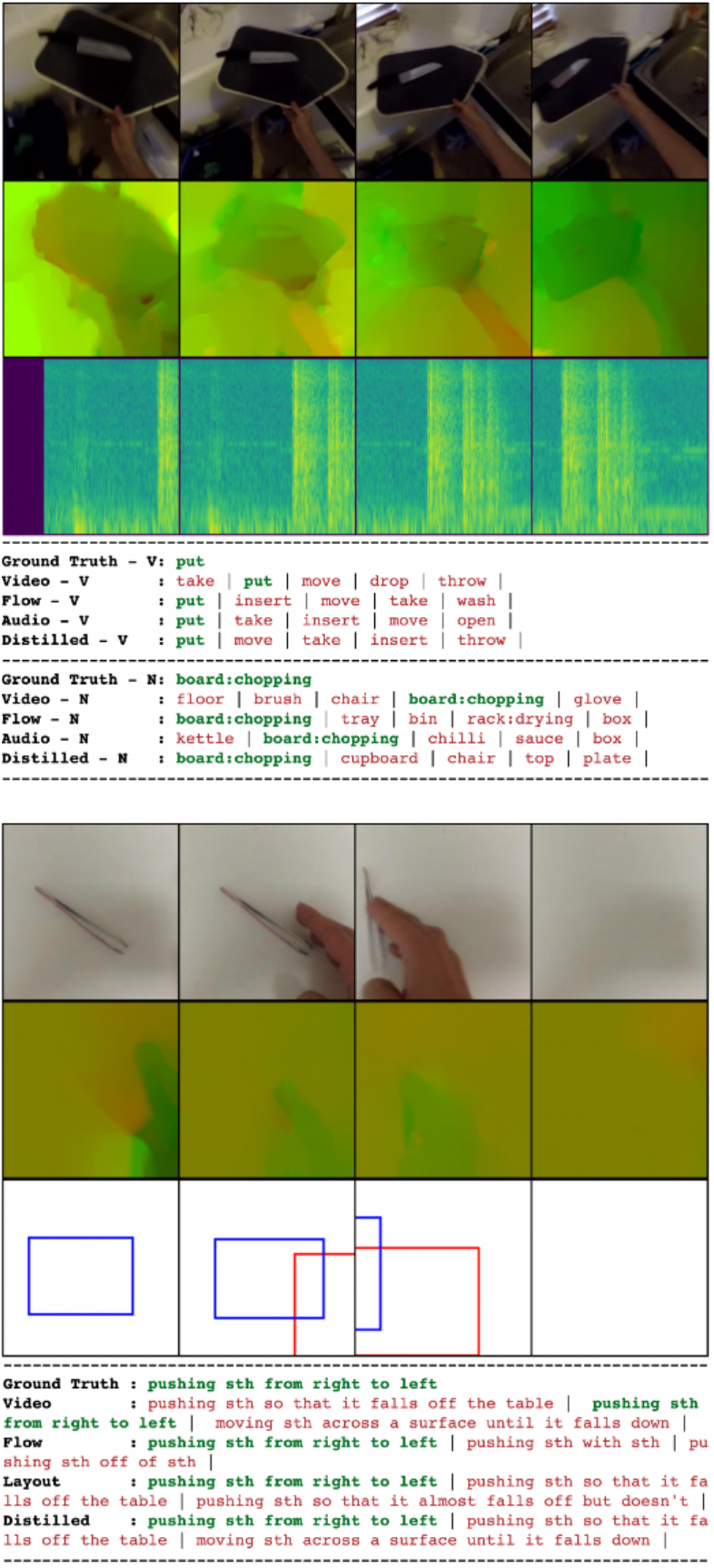}
\caption[Qualitative analysis on Epic-Kitchens and Something-Something]{Qualitative example for the Epic Kitchens (Top) and the Something-Something (Bottom) datasets.}
\label{iccv2023:fig:qualitative}
\end{figure}

\subsubsection{Effect of Test-Time Augmentation}
Lastly, we inspect the relationship between action recognition performance and the number of temporal clips (on Epic-Kitchens) and spatial crops (on Something-Something) sampled from the video during inference. Note that our goal is to reduce the computational complexity while maintaining the performance. Since standard video models \cite{bulat2021space, liu2021swin, li2022mvitv2, arnab2021vivit} use multiple temporal clips and spatial crops as a test-time augmentation to obtain further performance improvements, we explore the extent to which the distilled model is dependent on their availability during inference. We report results in Figure~\ref{iccv2023:fig:ablation_num_clips}, where we observe that the distilled model ($\lambda = 1.0$) is much less adversely affected by the reduction of both sampled temporal clips and spatial crops during inference, compared to the same model trained on the ground truth labels ($\lambda = 0.0$).

\subsection{Qualitative Examples}
In Figure~\ref{iccv2023:fig:qualitative}, we showcase the classes corresponding to the highest scores predicted by the student and the individual modality-specific models in the teacher ensemble, as well as the ground truth label (on Epic-Kitchens and Something-Something datasets). For both examples, we observe that the student picks up on relevant cues from each modality and accurately predicts the action of interest.

\section{Conclusion}\label{iccv2023:sec:conclusion}
In this chapter, we tackled the challenges associated with multimodal fusion in egocentric action recognition; specifically the assumptions of the availability of the different modalities during inference, as well as the computational demands of processing multiple modalities jointly, during inference time. We introduced a novel approach leveraging multimodal knowledge distillation, which allows models to benefit from rich multimodal data during training while operating solely on RGB frames at inference. Our method not only circumvents the limitations of multimodal fusion, but also significantly enhances the action recognition performance, effectively addressing \inlineresearchquestion{Research Question 5}.

Further, in response to \inlinesubresearchquestion{Research Question 5.1}, we propose a simple way to handle the inclusion of noisier modalities in the ensemble without compromising the overall performance. This ensures that our multimodal knowledge transference approach remains robust even when faced with less reliable modalities.

Regarding \inlinesubresearchquestion{Research Question 5.2}, our findings illustrate a beneficial trade-off between the high performance achieved through multimodal fusion and the efficiency of unimodal models. Notably, our method achieves this balance without heavily relying on expensive test-time augmentations, which are commonly employed to boost performance in egocentric action recognition tasks.

\textbf{Limitations.} The primary focus of this chapter has been on action recognition, leaving the exploration of multimodal distillation's applicability to other egocentric tasks for future work. Additionally, the integration of other diverse modalities (e.g., \blackgls{imu})---which are available in recent egocentric datasets (e.g., Ego4D \cite{grauman2022ego4d})---presents an avenue for future work, should we apply such multimodal knowledge transference methods using other modalities and tasks.

\textbf{Concurrent and future work with similar observations.} The concurrent works of Gong~\etal~\cite{gong2023mmg} and Ramazanova~\etal~\cite{ramazanova2024exploring} build models which are capable of dealing with missing modalities during inference, and report positive results on the Ego4D dataset, which we did not include in the experiments in this chapter. Specifically, Gong~\etal~\cite{gong2023mmg}, also leverage other diverse modalities such as \blackgls{imu} during the training process to perform multimodal knowledge transference. Finally, the recent work of Plizzari~\etal~\cite{plizzari2023outlook} discusses the overall future of egocentric vision, both real-life application and research problems.

\cleardoublepage%

% \includechapter{epilogue_part_ii}
%
  \graphicspath{{\chaptersdir/conclusion/image/}}%
  %\cleardoublepage%
  % !TeX root = ../../thesis.tex
\chapter{Conclusion}\label{ch:conclusion}
In the concluding chapter, we recap the main research questions and findings, summarize our research efforts, and provide an outlook on future research. Finally, we provide a short epilogue of the manuscript.

\section{Contributions Revisited}
With this manuscript, our goal is to explore and address the multifaceted challenges of multimodal machine learning, in particular: multimodal alignment, translation, fusion, and transference. Through a series of diverse experimental analyses, we shed light on how---by addressing the aforementioned challenges---we can enhance the machine learning models' ability to interpret and interact with the world around us. Our research is driven by two transversal research questions, each aiming to dissect specific aspects of multimodal learning and provide insights into bridging the semantic gap between heterogeneous data forms. Below, we revisit these guiding questions and outline how our findings contribute to their clarification. Note that, importantly, we do not attempt to answer these questions for \textit{any} combination of modalities, but rather illustrate the general feasibility with diverse modalities across the chapters of this manuscript.

\universalresearchquestion{Transversal Research Question I}{How effectively can we align and translate across different heterogeneous modalities, while accounting for the modality-specific nuances and challenges of out-of-distribution data?}

A central topic of \textbf{Part~\ref{part:alignment-translation}} of this manuscript is the exploration of multimodal alignment and translation, where we demonstrate the potential of machine learning models to effectively translate linguistic scene descriptions into spatial arrangements (\textbf{Chapter~\ref{ch:emnlp2020}}), medical texts into anatomical human atlas locations (\textbf{Chapter~\ref{ch:conll2020}}), and structured texts into knowledge graph facts (\textbf{Chapter~\ref{ch:emnlp2023}}).

In \textbf{Chapter~\ref{ch:emnlp2020}}, we develop a model architecture, training methodology, and translation (i.e., decoding) strategies, suitable for translating language-expressed spatial relations---given by scene descriptions---to spatial arrangements between objects in a 2D space. The model we propose is specifically designed for generating abstract scenes that satisfy the spatial constraints given by the scene descriptions. We further quantify to what extent the method we develop is capable of dealing with spatial relations that are out-of-distribution; i.e., spatial relations that the model has not encountered during training. Finally, we perform a human study to verify that the generated abstract scenes of clip-arts align with human expectations.

In \textbf{Chapter~\ref{ch:conll2020}}, we propose a novel loss function that enables translating medical texts into 3D locations in an anatomical human atlas. The loss function we design can be used as a plug-in component with any sentence embedding model and could yield a 3D dimensional medical text representation/embedding, which is grounded in a human body atlas. We further explore whether models trained by our proposed loss function can generalize to anatomical substructures not explicitly referred to in the training data, an ability that is crucial for such models to be used in practice. Lastly, we verify the usability of these low-dimensional embeddings in an information retrieval setting,
facilitating applications such as medical text retrieval and exploration, where the medical documents are grounded in a structure intuitive for humans to navigate.

In \textbf{Chapter~\ref{ch:emnlp2023}}, we contribute a novel benchmark dataset and methods for the task of translating structured general-purpose text to canonical knowledge graph facts. Furthermore, we introduce the task of Out-of-Knowledge Graph detection, which requires methods to not only translate between the modalities but also detect when such translation is not feasible. The method and resources we contribute enable translating natural language text to knowledge graph facts in various scenarios, which, in turn, facilitates tasks such as automatic knowledge graph population and automatic fact-checking.

Collectively, these chapters address the challenges of heterogeneity and connectivity among modalities, demonstrating that we can develop models capable of achieving multimodal understanding across diverse data forms.
% Notably, our contribution (e.g., model, loss function, etc.) in each chapter adheres to the modality-specific nuances and sheds light on the out-of-distribution generalization ability to out-of-distribution data to a certain extent.

\universalresearchquestion{Transversal Research Question II}{How can we harness the power of multimodal fusion and knowledge transference to create more capable and efficient video understanding machine learning models?}

In the latter \textbf{Part~\ref{part:fusion-transference}} of this manuscript, we delve into multimodal fusion and transference, focusing on how the complementary nature of different modalities enhances the action recognition performance of video understanding models (\textbf{Chapter~\ref{ch:bmvc2021}}). Additionally, we explore how we can use multimodal knowledge transference methods to yield (egocentric) video understanding models that operate under specific computational efficiency constraints (\textbf{Chapter~\ref{ch:iccv2023}}).

In \textbf{Chapter~\ref{ch:bmvc2021}}, we present a comprehensive study on multimodal fusion between two abstract modalities---representations of \blackgls{rgb} video frames and object detections---to enhance the action recognition performance of video understanding models. The main takeaway from our study is that maintaining the complementarity between these modalities is crucial for the successful fusion of multimodal inputs. Ultimately, the multimodal fusion method proves to be more robust to out-of-distribution data than unimodal methods, thereby increasing the models' applicability in real-life scenarios. However, we identify two significant issues with such methods during inference: the computational complexity which increases by orders of magnitude, and the strict assumption of the availability of all modalities involved in the fusion process.

In \textbf{Chapter~\ref{ch:iccv2023}}, we address the limitations of multimodal fusion by proposing a method for enabling multimodal knowledge transference during training through multimodal knowledge distillation. By adopting this method, we demonstrate that we can largely maintain the recognition performance and ability to handle out-of-distribution inputs of a multimodal fusion-based approach for egocentric video understanding. Importantly, the resulting model relies only on unimodal data during inference, thereby requiring the same computational budget as a model trained solely on unimodal data.

All together, in these two chapters, we highlight the interactivity property of multimodal data and illustrate how the integration and interaction of diverse multimodal data can lead to improved action recognition performance and robustness w.r.t. out-of-distribution inputs.

\textbf{Recap of the contributions.} This manuscript advances the field of multimodal machine learning in several key areas. We introduce novel methodologies and benchmarks for aligning and translating between modalities, enhancing the broader understanding of the semantic richness of multimodal data. Additionally, our work on multimodal fusion and knowledge transference opens new avenues for video understanding methods to efficiently leverage multimodal data, addressing practical limitations and paving the way for more robust and versatile machine learning applications.

\section{Future Work}
We highlight several crucial avenues for future research to address the complexities and current limitations of multimodal machine learning methods. Below, we outline some of these key areas for future exploration.

\subsubsection{Exploring Multilinguality} One limitation of our current research is the predominant focus on data restricted to the English language. This specificity raises questions about the universality and adaptability across languages of the findings we report. Exploring the property of multilinguality acknowledges the diversity of human languages and the need for models that can understand and process information not only in English. This is especially relevant in the current globalized world, where information and user interactions, span multiple languages. Research in this area could focus on developing models that seamlessly integrate and translate between modalities in different languages, leveraging multilingual datasets and cross-lingual transfer learning methods.

However, current methods cannot simply be repurposed in a multimodal context. We anticipate at least three challenges that will likely be present, should one decide to implement such methods---multimodal alignment and translation methods as the ones we propose here, or beyond that (e.g., text-to-image generation, text-to-video generation, etc.)---for low-resource languages other than English (e.g., Macedonian):

\begin{enumerate}[label=(\roman*)] % Use lowercase Roman numerals
    \item \textbf{Weaker pre-trained methods.} Notably, due to the lack of publicly available datasets, current pre-trained \blackgls{llm}s perform significantly better on tasks for high-resource languages. Even though recent state-of-the-art multilingual text embedding methods \cite{wang2024multilingual} are trained for a large number of languages, their performance remains to be significantly worse for under-represented languages.
    \item \textbf{Smaller task-specific datasets.} Importantly, in each chapter of \textbf{Part~\ref{part:alignment-translation}}, besides relying on high-quality pre-trained text embedding methods, we utilized fairly large-scale datasets to train our models. Similar to the previous point, training methods capable of translating text to representations from other modalities would pose significant challenges if conducted in a few-shot manner.
    % Importantly, besides relying on high-quality pre-trained text embedding methods, in each of the chapters in 
    % \textbf{Part~\ref{part:alignment-translation}}
    % , we leveraged a fairly large-scale dataset to train our models. Similar to (i), training methods that can translate text to a representation from another modality would become significantly more challenging should we perform training in a few-shot manner.
    \item \textbf{Adapting to cultural diversity.} Consider training a multimodal text-to-image translation model. In this context, what often might be considered a straightforward depiction in one culture, could have different connotations or even be inappropriate in another; e.g., certain symbols, animals, or colors often have specific cultural significance that multimodal translation model must account for to avoid misrepresentations. Furthermore, languages use idiomatic expressions which have interpretative meaning.
    % e.g., in Macedonian, we say: ``\textcyr{не умри, магаре, до зелена трева}'',
    % to indicate that a task is taking too long, and might be never finished. However, the literal translation in English would be: ``Do not die donkey, until the green grass''.
    Therefore, multimodal translation models trained primarily on English language might struggle to interpret such expressions in low-resource languages, leading to completely wrong outputs.
\end{enumerate}

% \subsubsection{Expanding the Scope of Evaluation Criteria}
% For almost all methods we developed, the performance is evaluated based on the correctness of the predictions on the downstream task w.r.t. the ground truth. While indeed, our primary interest is often only the correctness of the predictions, methods should also account for the uncertainty in their predictions. Importantly, reliable uncertainty estimation plays a pivotal role in predictive models, especially in safety-critical applications (e.g., machine learning applications in the medical domain), where decisions based on the neural network's predictions can have significant consequences~\cite{amodei2016concrete,kompa2021second}. Therefore, future work should at least report how calibrated the models are; i.e., \textit{calibration error} (\blackgls{ece})~\cite{guo2017calibration, popordanoska2022consistent} serves as a key indicator of the reliability and trustworthiness of a model's predictions, offering essential insights into its overall dependability.
% Informally, CE is a measure of discrepancy between predicted probabilities and empirically observed class frequencies. For instance, when a $calibrated$ model predicts an 80\% chance that a patient has the flu, we expect that out of 100 patients with similar symptoms, 80 indeed have the flu.
\subsubsection{Expanding the Scope of Evaluation Criteria}
In almost all the methods we have developed, performance evaluation is primarily based on the accuracy of the predictions in relation to the ground truth. While correctness of predictions is indeed our primary concern, methods should also incorporate uncertainty estimation in the evaluation. This is particularly important in predictive models, especially within safety-critical applications such as machine learning in the medical domain, where decisions based on neural network predictions can have significant consequences \cite{amodei2016concrete,kompa2021second}. Therefore, future research should report on the calibration of models, indicated by the \textit{calibration error} (\blackgls{ece}) \cite{guo2017calibration, popordanoska2022consistent}, which serves as a key indicator of a model's reliability and trustworthiness, offering insights into its overall dependability.

\subsubsection{Enhancing Generalization Performance w.r.t.~Out-of-Distribution Data}
% Our experiments reveal relative robustness in handling out-of-distribution data, yet there remains a noticeable dip in performance. This gap highlights the critical need for models to generalize beyond the training data distribution, which is crucial for the deployment of these models in real-world scenarios, where they will inevitably encounter inputs that differ from the training data distribution. Future work should, ideally, explore new methods for enhancing model robustness and generalization capabilities; e.g., domain adaptation methods---that can adapt to covariate shift in the input data distribution \cite{wang2007transferable}, or adapt to label-shift in the target distribution \cite{popordanoska2023estimating}---to identify and dynamically adapt to out-of-distribution samples.
Our experiments demonstrate relative robustness in handling out-of-distribution inputs, yet we observe a noticeable decline in performance. This disparity highlights the importance of models being able to generalize beyond the training data distribution. Such generalization is essential for deploying these models in real-world scenarios, where they will inevitably encounter inputs that deviate from the training data distribution. Ideally, future research should explore new methods to improve model robustness and generalization capabilities. For instance, domain adaptation methods---which can adjust to covariate shift in the input distribution \cite{wang2007transferable}, or adapt to label-shift in the target distribution \cite{popordanoska2023estimating}---offer promising avenues to identify and dynamically adapt to out-of-distribution samples.

\subsubsection{Integration of Multimodal and Multitask Learning}
Our experimental analysis is limited to a specific subset of modalities and tasks, overlooking the broader landscape of relevant modalities and tasks in real-world applications. Moreover, recent large-scale datasets such as Ego4D \cite{grauman2022ego4d}, encompass a much more diverse set of modalities: e.g., RGB, text narrations, audio, 3D scans, eye gaze, inertial measurement unit (\blackgls{imu}), and multi-camera inputs. These datasets feature culturally diverse data, underscoring the necessity for multimodal machine learning models capable of operating effectively in cross-cultural settings.

Furthermore, multitask learning methods can exploit the inherent relationships between different tasks to enhance learning efficiency and model performance \cite{crawshaw2020multi, vandenhende2021multi}. Developing models trained on multiple tasks concurrently can further benefit from the complementary information provided by different modalities. Hence, future research should explore methodologies that seamlessly integrate multimodal and multitask learning, with a focus on shared representation learning, task-specific adaptation, and effective knowledge transfer across tasks and modalities.

% \subsubsection{Exploring Multimodal and Multitask Learning}
% The experimental analysis we perform is focused on only a specific set of modalities and tasks. On the other hand, there are numerous modalities and tasks relevant to real-world applications. For example, for the task of egocentric video understanding, recent large-scale datasets---Ego4D \cite{grauman2022ego4d}---contain a significantly more diverse set of modalities: RGB, text narrations, audio, 3D scans, eye gaze, inertial measurement unit (\blackgls{imu}), multi-camera inputs, etc. To add on top of this, the dataset features culturally diverse data, emphasizing the requirement of multimodal machine learning models that operate in cross-cultural settings.

% What is more, multitask learning methods can leverage the inherent connections between different tasks to improve learning efficiency and model performance \cite{crawshaw2020multi, vandenhende2021multi}. Developing models that are trained on multiple tasks simultaneously can further enhance the benefits of the complementary information provided by different modalities. Therefore, future work could explore methods that effectively combine multimodal and multitask learning, focusing on shared representation learning, task-specific adaptation, and effective knowledge transfer across tasks and modalities.

\section{A Short Epilogue}
In conclusion, this manuscript represents a small step forward in the pursuit of truly multimodal machine learning models capable of understanding and integrating the complex, heterogeneous, and interconnected data that constitute (a part of) our experiences of the world. By addressing critical challenges in multimodal alignment, translation, fusion, and knowledge transference, we hope to have laid \textit{some} groundwork for future research in the field of multimodal machine learning. The ultimate goal for us is to deepen our understanding of these systems and move closer to the development of machine learning models with a more nuanced and comprehensive understanding of the multimodal world they are designed to navigate.
%%%%%%%%%%%%%%%%%%%%%%%%%%%%%%%%%%%%%%%%%%%%%%%%%%
% Keep the following \cleardoublepage at the end of this file, 
% otherwise \includeonly includes empty pages.
\cleardoublepage

% vim: tw=70 nocindent expandtab foldmethod=marker foldmarker={{{}{,}{}}}
%

%%%%%%%%%%%%%%%%%%%%%%%%%%%%%%%%%%%%%%%%%%%%%%%%%%%%%%%%%%%%%%%%%%%%%%

\appendix

% \includeappendix{emnlp2020_appendix}
% \includeappendix{conll2020_appendix}
% \includeappendix{emnlp2023_appendix}
% \includeappendix{bmvc2021_appendix}
% \includeappendix{iccv2023_appendix}

%%%%%%%%%%%%%%%%%%%%%%%%%%%%%%%%%%%%%%%%%%%%%%%%%%%%%%%%%%%%%%%%%%%%%%
\backmatter

\includebibliography
% BibTex
\bibliographystyle{acm}
\bibliography{allpapers}
% BibLatex (comment lines above and comment out biblatex lines in preamble)
%\printbibliography[title=\bibname]
\instructionsbibliography

\includecv{curriculum}

\includepublications{publications}

\makebackcoverXII

\end{document}